%% file: main.tex
\documentclass[lettersize,journal]{IEEEtran}
\usepackage{amsmath,amsfonts}
\usepackage{algorithmic}
\usepackage{algorithm}
\usepackage{array}
\usepackage[caption=false,font=normalsize,labelfont=sf,textfont=sf]{subfig}
\usepackage{textcomp}
\usepackage{stfloats}
\usepackage{url}
\usepackage{verbatim}
\usepackage{graphicx}
\usepackage{cite}
\usepackage{framed,multirow}  %
\usepackage[x11names]{xcolor} %
\usepackage{booktabs} %
\usepackage{arydshln} %

\usepackage[capitalize]{cleveref}
\crefname{section}{Sec.}{Secs.}
\Crefname{section}{Section}{Sections}
\Crefname{table}{Table}{Tables}
\crefname{table}{Tab.}{Tabs.}

\def\etal{\emph{et al.}} %

\begin{document}

\title{Blind Motion Deblurring with Pixel-Wise Kernel Estimation via Kernel Prediction Networks}

\author{Guillermo~Carbajal, Patricia Vitoria, José~Lezama, and Pablo~Musé~%
\thanks{G. Carbajal, J. Lezama and P. Musé are with the Department of Electrical Engineering, Universidad de la República, Uruguay. Patricia Vitoria is with the Image Processing Group, Universitat Pompeu Fabra, Spain.}
}

\maketitle

\begin{abstract}
In recent years, the removal of motion blur in photographs has seen impressive progress in the hands of deep learning-based methods, trained to map directly from blurry to sharp images. For this reason, approaches that explicitly use a forward degradation model received significantly less attention. However, a well-defined specification of the blur genesis, as an intermediate step, promotes the generalization and explainability of the method. Towards this goal, we propose a learning-based motion deblurring method based on dense non-uniform motion blur estimation followed by a non-blind deconvolution approach. Specifically, given a blurry image, a first network estimates the dense per-pixel motion blur kernels using a lightweight representation composed of a set of image-adaptive basis motion kernels and the corresponding mixing coefficients. Then, a second network trained jointly with the first one, unrolls a non-blind deconvolution method using the motion kernel field estimated by the first network. The model-driven aspect is further promoted by training the networks on sharp/blurry pairs synthesized according to a convolution-based, non-uniform motion blur degradation model. Qualitative and quantitative evaluation shows that the kernel prediction network produces accurate motion blur estimates, and that the deblurring pipeline leads to restorations of real blurred images that are competitive or superior to those obtained with existing end-to-end deep learning-based methods. 
Code and trained models are available at \texttt{\url{https://github.com/GuillermoCarbajal/J-MKPD/}}.

\end{abstract}

\begin{IEEEkeywords}
Non-uniform motion kernel estimation, kernel prediction networks, motion deblurring, deep learning.
\end{IEEEkeywords}

\section{Introduction}

\label{intro}

Motion blur is a major source of degradation in images. This effect, preeminent in low-light photography, occurs when the exposure time is long compared to the relative motion speed between the camera and the scene. As a result, the camera sensor at each pixel receives and accumulates light coming from different sources, producing different amounts of blur.
\begin{figure}[h]
    \centering   \includegraphics[width=0.48\textwidth]{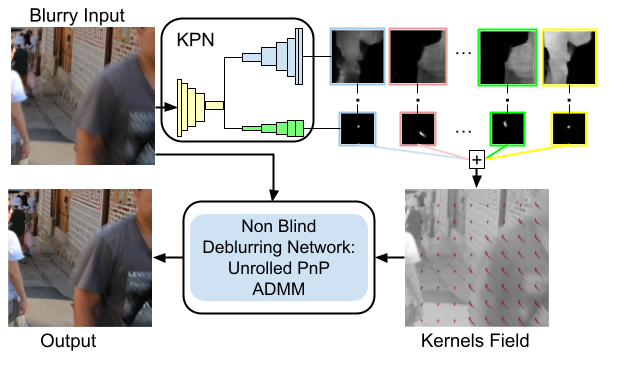}
    \caption{\textbf{Overview of the image deblurring pipeline.} Given a blurry image, a two-branch Kernel Prediction Network (KPN) predicts a set of image-adaptive basis motion kernels and corresponding pixel-wise mixing coefficients,
producing a per-pixel motion blur field. Then, a non-blind deconvolution network  \cite{laroche2022deep} fed with the blurry image and the motion kernel field, unrolls a linearized ADMM-based optimization, and generates the deblurred image.}
    \label{fig:pipeline2}
\end{figure}

\begin{figure*}[t!]
\setlength\tabcolsep{1.0pt} %
\centering
\begin{tabular}{c@{\hspace{1em}}c}
\multirow{2}{*}[0.26cm]{\includegraphics[width=0.07\textwidth]{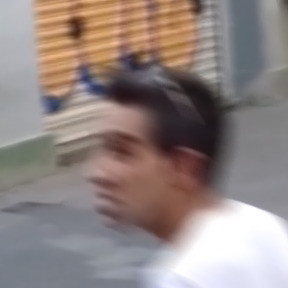}} & \includegraphics[width=0.9\textwidth]{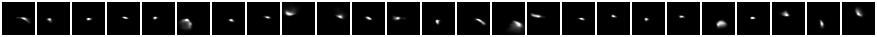}\\
& \includegraphics[width=0.896\textwidth]{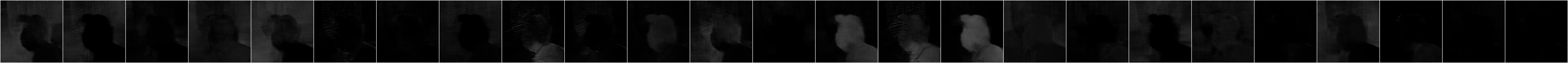}\\
\end{tabular}
\begin{tabular}{c@{\hspace{1em}}c}
\multirow{2}{*}[0.26cm]{\includegraphics[width=0.07\textwidth]{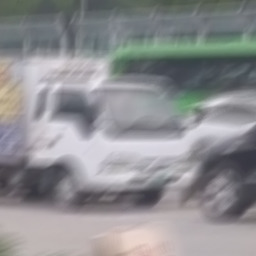}} & \includegraphics[width=0.9\textwidth]{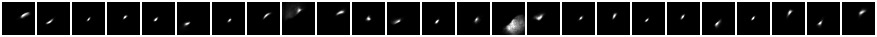}\\
& \includegraphics[width=0.896\textwidth]{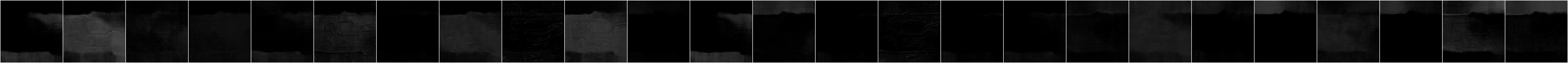}\\
\end{tabular}
\begin{tabular}{c@{\hspace{1em}}c}
\multirow{2}{*}[0.26cm]{\includegraphics[width=0.07\textwidth]{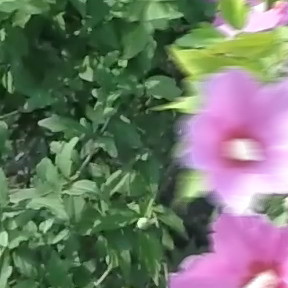}} & \includegraphics[width=0.9\textwidth]{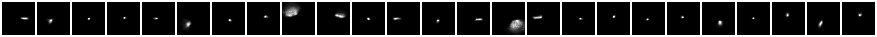}\\
& \includegraphics[width=0.896\textwidth]{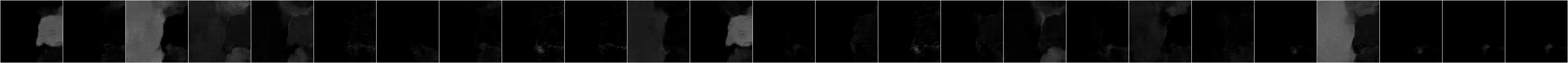}\\
\end{tabular}

\caption{\textbf{Examples of generated kernel basis $\{\mathbf{k}^b\}$  and corresponding mixing coefficients $\{\mathbf{m}^b\}$} predicted from the  blurry images shown on the left. The adaptation to the input is more notorious for the elements  that have significant weights. }
    \label{fig:KernelsAndMasks}
\end{figure*}
Motion deblurring from a single image is an ill-posed problem, as the same blurry image can be obtained from many pairs of blur kernels and sharp images. In model-based methods, an accurate blur kernel estimation is crucial for obtaining a high-quality sharp image. Existing non-uniform methods approximate the blur kernel by assuming a global parametric motion field model~\cite{tai2011richardson-lucy, gupta2010singleimage,hirsch2011fastremoval, whyte2010nonuniform}, or pixel-wise kernels parameterized by the length and orientation of the kernel support~\cite{gong2017motion,kim2014segmentation-free,dai2008motion,sun2015learning} or by quadratic curves~\cite{zhang2021exposure}.

While parametric methods significantly reduce the complexity and computational cost by solving for a simple approximate motion field, the approximation does not generalize to most real case scenarios, where kernel shapes are more complex, and the kernel values along the motion trajectory are not constant.  %

Motion blur kernel estimation from a single image proves useful to infer scene motion information and to solve related tasks such as tracking~\cite{guo2021exploring}, optical flow estimation~\cite{argaw2021optical}, motion segmentation~\cite{pan2016soft}, motion from blur~\cite{dai2008motion}, camera pose estimation and structure from motion blur~\cite{qiu2019world, zheng2011structure}, among others. However, until recently, its major motivation  has certainly been blind motion deblurring.
Nowadays, the unprecedented improvements in deblurring obtained by deep learning (DL) methods have contributed to a decreased interest in model-based approaches and motion blur kernel estimation. Indeed, DL deblurring methods skip the kernel estimation step arguing that~\cite{Nah_2017_CVPR}: {\em (i)} blur kernels models are too simplistic and unrealistic to be used in practice; {\em (ii)}  kernel estimation process is subtle and sensitive to noise and saturation; {\em (iii)} finding a spatially varying kernel for every pixel in a dynamic scene requires a huge amount of memory and computation. 
While these limitations are certainly true, current deblurring DL-based methods also suffer from a major limitation related to the way the training datasets are produced. To synthesize motion-blurred images without relying on a forward kernel-based model, the training dataset is obtained by averaging consecutive frames from a sequence of a dynamic scene captured with a high-speed camera~\cite{sun2015learning,Nah_2019_CVPR_Workshops_REDS,su2017deep}. Such datasets can only characterize the motion blur to a limited extent, and their end-to-end training schemes induce a mapping that might be specific to the camera used, capturing transformations other than deblurring. %
As a consequence, DL-based image deblurring methods sometimes fail to generalize to real blurred images \cite{tran2021explore}.

In this work, we propose a new method for image deblurring, combining a first network in charge of estimating a dense, spatially-varying motion blur field, together with a second network unrolling a deep plug-and-play deconvolution method, recently proposed in~\cite{laroche2022deep}.

More precisely, the main contributions are:

\begin{itemize}
\item A novel approach for non-parametric, dense, spatially-varying motion blur estimation. For each blurred image, a convolutional {\em kernel prediction network} estimates an image-specific set of kernel basis functions and a set of pixel-wise mixing coefficients, cf. Figures~\ref{fig:pipeline2} and \ref{fig:KernelsAndMasks}. The combination of the two results in a  unique motion blur kernel per pixel. We show that the estimated kernels obtained by our method manage to capture a wide variety of motion blurs, and are significantly more accurate than the ones obtained by the prior art~\cite{gong2017motion,sun2015learning, zhang2021exposure}.

\item A new blind image deblurring pipeline, obtained by jointly training our \textit{kernel prediction network} with an unrolled deep plug-and-play deconvolution network. We show that our approach compares favorably with state-of-the-art motion deblurring methods~\cite{kupyn2018deblurgan, kupyn2019deblurgan, tao2018scale, rim_2020_ECCV, Zhang_2019_CVPR} on real images.
\end{itemize}

Both contributions are validated by an extensive set of experiments on several benchmark datasets, including qualitative and quantitative comparisons with state-of-the-art methods. Cross-dataset experiments are considered to evaluate the capability of the proposed deblurring procedure to generalize to real motion-blurred photographs.

The remainder of the article is organized as follows. In~\cref{sec:related_work}, we present relevant previous work on spatially-varying motion blur estimation, kernel prediction networks, and blind motion deblurring. The motion blur degradation model, and the proposed kernel prediction network for spatially-varying motion blur estimation, are presented in Sections~\ref{sec:model} and~\ref{sec:kernel_estimation_method}.  Supportive experiments on  the proposed motion blur kernel estimation method are presented in~\cref{sec:kernel_estimation_experiments}. The blind motion deblurring pipeline is presented in~\cref{sec:deblurring_pipeline}. Supportive experiments on the blind image deblurring pipeline are presented in~\cref{sec:deblurring_experiments}. Possible extensions are presented in ~\cref{sec:extensions}. Last, conclusions and future work are summarized in~\cref{sec:conclusions}. Code and trained models are available at \texttt{\url{https://github.com/GuillermoCarbajal/J-MKPD/}}.

\section{Related Work}
\label{sec:related_work}

\subsection{Single Image Non-Uniform Motion Blur Estimation} 

Early methods attempting to estimate spatially-varying motion blur kernels consider that such non-uniformity is mainly caused by 3D camera tilts or rotations \cite{gupta2010singleimage,tai2011richardson-lucy,whyte2010nonuniform,hirsch2011fastremoval}. In this setting, the blurred image results from the integration of the intermediate images, which are varying perspective projections of the scene. By assuming that the scene is static and that the focal length is sufficiently long or the scene is far enough to be considered planar, the transformations are reduced to homographies. This leads to the so-called Projective Motion Blur Model (PMBM)~\cite{tai2011richardson-lucy}. While these methods achieve impressive results when the planar assumptions hold, most blur scenes are not planar or contain moving objects, leading to unsatisfactory results.

A different approach, more related to ours, that deals with both scene depth variations and moving objects, consists in predicting motion blur locally. Sun \etal~\cite{sun2015learning} and Gong \etal~\cite{gong2017motion} use neural networks to predict dense motion blur fields, with pixel-wise line-shaped kernels parameterized by their lengths and orientations. Both methods propose to deblur images using a non-blind deblurring method with the estimated kernels, regularized with the EPLL image prior~\cite{zoran2011learning}.
More recently, Zhang \etal ~\cite{zhang2021exposure} model the blurry image as pixel-wise displacements of the latent sharp image at continuous time points. The pixel-wise motion kernels are parameterized using a quadratic model.

\begin{figure*}[t]
    \centering
    \includegraphics[width=\textwidth]{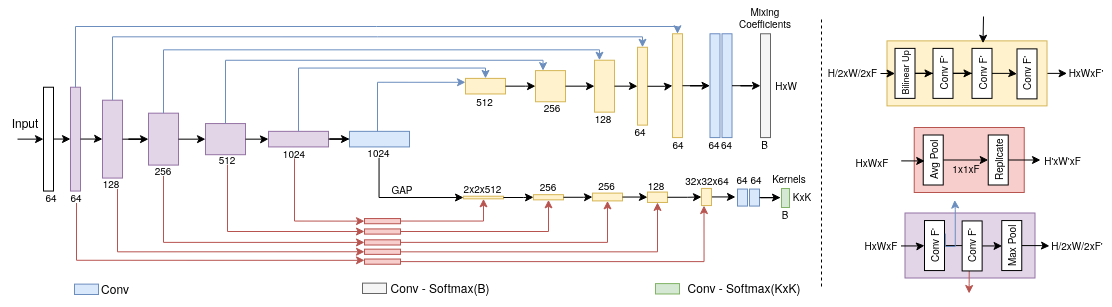}
    \caption{\textbf{Architecture Details} The proposed network is composed of an encoder and two decoders. The encoder takes as an input a blurry image. The decoders output the motion kernel basis and corresponding per-pixel mixing coefficients. \label{fig:architecture}}
    
\end{figure*}

\subsection{Kernel Prediction Networks} Recently, Kernel Prediction Networks (KPN) have been proposed for low-level vision tasks such as burst denoising~\cite{mildenhall2018burst,xia2019basis}, optical flow estimation, frame interpolation~\cite{niklaus2017video,niklaus2017video2}, stereo and video prediction~\cite{jia2016dynamic}. 
 Several works have used KPNs in the context of burst denoising. In~\cite{mildenhall2018burst}, denoised estimates are produced at each pixel as a weighted average of observed local noisy pixels across all the input frames. These averaging weights, or kernels, are predicted from the noisy input burst using a CNN.  
To improve the computational efficiency,~\cite{xia2019basis} proposes a basis prediction network that, given an input burst, predicts a set of
global basis kernels — shared within the image — and the corresponding mixing coefficients, which are specific to individual pixels.

\subsection{Blind Motion Deblurring from a Single Image}

A possible classification of blind image deblurring approaches is defined by whether or not they explicitly use a forward image degradation model. Traditional approaches model motion blur as the convolution of the sharp latent image with a motion blur kernel contaminated by additive noise. The other class of methods, known as {\em end-to-end} methods, emerged much more recently with the advent of deep learning. These methods learn to map blurry images to their corresponding latent sharp images, by training a CNN from a large dataset of sharp/blurry image pairs. 

\subsubsection{Blind motion deblurring based on kernel estimation}

In the classic approach, blind deblurring is formulated as a variational problem, with joint optimization on the kernels and latent images, usually using an alternate scheme. The problem is regularized using different priors on the latent sharp image and the kernels to deal with its ill-posed nature. Most of these methods assume a spatially uniform blur model~\cite{fergus2006removing, pan2014deblurring,cho2009fast,xu2010two}, or a PMBM kernel field~\cite{whyte2010nonuniform,xu2013unnatural}.

More recently, Kaufman and Fattal~\cite{kaufman2020deblurring} proposed to use an analysis-synthesis neural network for motion deblurring. The analysis network is pre-trained to predict a uniform blur kernel, while the synthesis network is pre-trained to deblur the images. Ground-truth kernels guide the synthesis network training during the pre-training phase, and the kernels predicted with the analysis network guide the synthesis network during joint training. Since there is no degradation model during the joint training, the quality of the kernels degrades severely. According to the authors, omitting the kernel loss during the joint training allows the networks to abstract its representation for better deblurring performance. 

\subsubsection{End-to-end motion deblurring}

Recently, DL-based end-to-end approaches have become the dominant trend for deblurring~\cite{Nah_2017_CVPR,tao2018scale, Zhang_2019_CVPR, Zamir2021MPRNet,cho2021rethinking, NAFNet}. One of the reasons for their success is the ability of neural networks to resolve the deconvolution and remove any artifacts of that process. 

Early DL methods \cite{Nah_2017_CVPR,tao2018scale, Zhang_2019_CVPR} sought to minimize the $L^2$-norm between the sharp image and the model output. This might introduce a blurring effect due to the regression-to-the-mean problem~\cite{bruna2015super,ledig2017photo}, motivating  the use of Generative Adversarial Networks (GANs) to obtain more realistically looking restorations~\cite{kupyn2018deblurgan,kupyn2019deblurgan}. However, GAN-based approaches introduce the potential pitfall of hallucinating image content~\cite{blau2018perception}. Another advantage of knowing the forward model is that it can be used to impose consistency between the restored and input blurry images~\cite{chen2018reblur2deblur}. Recent end-to-end DL methods produce outstanding restoration results on synthetic datasets using multi-scale~\cite{cho2021rethinking}, multi-stage~\cite{Zamir2021MPRNet} or carefully simplified~\cite{NAFNet} architectures. However, these models suffer from in-domain overfitting and may fail to generalize well to unseen data~\cite{tran2021explore}.

\subsection{Relation with the Proposed Approach} 

In this work, we first propose a method performing dense spatially-varying motion blur kernel estimation, based on an efficient computation of pixel-wise motion blur kernels. As in~\cite{hirsch2011fastremoval} the per-pixel kernels are modeled as linear combinations of base elements, but instead of limiting them to a single pre-computed basis of elements, we use a KPN to infer a non-parametric basis specific for each input image. Additionally, instead of learning the kernels to solve the image restoration problem (e.g. denoising in~\cite{xia2019basis}), we learn them to fit the forward degradation model. 

Then, we propose a motion deblurring pipeline that combines our KPN with a network unrolling a deep plug-and-play deconvolution method, recently proposed in~\cite{laroche2022deep}. Since both networks have been pre-trained independently, combining them directly for the blind image deblurring task tends to produce artifacts that degrade the quality of the restored images. These artifacts are effectively reduced by jointly fine-tuning the whole pipeline, as shown by the experiments presented in Section~\ref{sec:deblurring_experiments}. %
The quality of our restorations is demonstrated through cross-dataset quantitative evaluations that show the generalization capacity of our motion deblurring pipeline.

\section{Motion Blur Degradation Model} \label{sec:model}

Non-uniform motion blur can be modeled as the local convolution of a sharp image with a spatially varying filter, the \emph{motion blur kernel field}. This simple model represents the integration, at each pixel, of photons arriving from different sources due to relative motion between the camera and the scene. Given a sharp image $\mathbf{u}$ of size $H\times W$, and a set of per-pixel blur kernels $\mathbf{k}_i$ of size $K\times K$, we will assume that the observed blurry image $\mathbf{v}$ is generated as
\begin{equation}
    {v}_i = \langle \mathbf{u}_{nn(i)}, \mathbf{k}_i \rangle + n_i,
        \label{eq:model}
\end{equation}
where $\mathbf{u}_{nn(i)}$ is a window of size $K \times K$ around pixel $i$ in image $\mathbf{u}$, and $n_{i}$ is additive noise. We assume that kernels are non-negative (no negative light) and of unit area (conservation of energy).

Predicting the full motion field of per-pixel kernels $\mathbf{k}_i$ would lead to an estimation problem in a very high-dimensional space ($K^2 H W$), being computationally intractable for large images and kernels. We propose an efficient solution, based on the assumption that there exists significant redundancy between the kernels present in the image. We incorporate this assumption in our model by decomposing the per-pixel kernels as linear combinations of image-dependent kernel basis elements. Specifically, if the blur kernel basis has $B$ elements, then, only $B$ mixing coefficients ${m}_i^b$ are required per pixel instead of the original $K \times K$. Additionally, the $B$ basis elements need to be estimated leading to an estimation problem of dimension $B(K^2 + HW)$. A notorious gain is obtained when $B\ll K^2$, particularly relevant for large blur kernels. The mixing coefficients are normalized to sum to one at each pixel location. Thus, the per-pixel kernel $\mathbf{k}_i$ results from the convex combination of the basis kernels; conservation of energy is guaranteed, and the degradation model becomes:
\begin{equation}
    {v}_{i} = \langle \mathbf{u}_{nn(i)}, \sum_{b=1}^B\mathbf{k}^b m^b_{i} \rangle + n_{i}.
    \label{eq:modelKernelBasis}
\end{equation}
This approximation is well suited for motion blur due to moving objects and  proved useful to approximate motion fields due to camera shake as well.

Taking into account the sensor saturation and the gamma correction performed by the camera, a more accurate model is given by
\begin{equation}
    v_i = R \left( \right \langle \mathbf{u}^\gamma_{nn(i)}, \sum_{b=1}^B\mathbf{k}^b m^b_i \rangle + n_i )^{\frac{1}{\gamma}} ,
      \label{eq:model_sat}
\end{equation}
where $\gamma$ is the gamma correction coefficient and $R(\cdot)$ is the pixel saturation operator that clips image values $v_{i}$ which are larger than 1. To avoid the non-differentiability of this function at $v_{i} = 1$, a smooth approximation for the captor response function is considered~\cite{whyte2014deblurring}:
\begin{equation}
    R(v_i) = v_i - \frac{1}{a}\log (1 + e^{a(v_i-1)}).
      \label{eq:sat_func}
\end{equation}
The parameter $a$ controls the smoothness of the approximation and is set to $a=50$. As for the gamma correction factor, a typical value of $\gamma = 2.2$ is used.

\paragraph*{Limitations} 

The main limitation of our image degradation model is that the motion fields that can be captured are limited by the size of the kernel support $K$. This size is limited for computational reasons, and because a larger kernel dimension would require a larger number of base elements to capture the complexity of the motion field, i.e., for the low-rank approximation to be accurate.

\begin{figure}[t]
    \centering
    \includegraphics[width=0.45\textwidth]{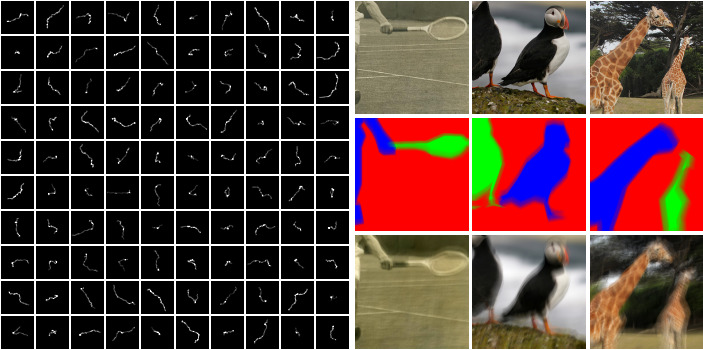}
    \caption{\textbf{Synthetic dataset for learning non-uniform motion kernel estimation.} Left: example random kernels. Right: different segments of the scene are convolved with different random kernels and are mixed with the (also convolved) segmentation masks.}
    \label{fig:dataset_gen}
\end{figure}

\section{Estimation of the Motion Blur Kernel Field}
\label{sec:kernel_estimation_method}

The architecture of our motion blur kernel prediction network is presented in Figure~\ref{fig:architecture}. A CNN is designed to estimate, from a given input blurry image, both the $B$ basis motion kernels $\{\mathbf{k}^b\}_{b=1,...,B}$ and the mixing coefficients $\{\mathbf{m}^b\}_{b=1,...,B}$. Building upon recent work in kernel prediction networks~\cite{xia2019basis}, the network is composed of a shared backbone and two generator heads. 
Normalization of the blur kernels and the mixing coefficients is achieved by using Softmax layers.
In our experiments, we set $K=33$ and $B=25$. %

To train the proposed motion blur kernel prediction network, we need to build a dataset of sharp/blurry image pairs, along with the corresponding motion blur kernel fields. 
Building such a dataset using real images requires a sophisticated setup. To our knowledge, the only benchmark dataset of this kind was proposed by K\"ohler \etal ~\cite{kohler2012recording}. The dataset is an excellent benchmark for quantitative performance evaluation of deblurring methods. However, because of its limited size and the planar geometry of the scene, this dataset is not well-suited to train motion kernel estimation or motion deblurring methods for real images\footnote{More details on K\"ohler's dataset generation procedure are provided in the supplementary material.}. Consequently, we propose to train our network on a synthetic dataset described below. As confirmed by the experiments presented in \cref{sec:kernel_estimation_experiments} and \cref{sec:deblurring_experiments}, training on this synthetic data generalizes remarkably well to real photographs with different types of scenes and motion.

\subsection{Synthetic Dataset Generation}\label{subsec:DatasetGen}

We build a synthetic dataset using 118,287 images from the COCO semantic segmentation dataset~\cite{lin2014microsoft}. To generate random motion kernels, we use a camera-shake kernel generator~\cite{gavant2011physiological,delbracio2015removing} based on physiological hand tremor data, and pre-compute 500,000 kernels with support smaller than $K\times K$, cf. Figure~\ref{fig:dataset_gen}.

More specifically, for a random sharp image $\mathbf{u}$, we perform a convolution of the image with a random kernel $\mathbf{k}$. Additionally, each segmented object (if any) and its corresponding mask is convolved with a different random kernel. This ensures a soft transition between different blurry regions. To simplify, a maximum of three segmentation masks are considered for each image. Finally, for each image, we obtain a tuple $\big(\mathbf{u}^{GT}, \mathbf{v}^{GT}, \{\mathbf{k}\}^{GT}, \{\mathbf{m}\}^{GT}\big)$ containing the sharp image, blurry image and the pairs of ground truth kernels and masks applied to generate the blurry image. %

To simulate latent scenes with saturating light sources, sharp images are first converted to the \textit{hsv} color space and then the histogram is transformed by multiplying the {\em v}-channel by a random value between $[0.5, 1.5]$.  The transformed image is converted back to the  \textit{rgb} color space, and the blurry image is generated as described above. Finally, the blurry image is clipped to the $[0,1]$ range. From now on we refer to this dataset as {\em SBDD} (Segmentation-Based Dataset for Deblurring).

\subsection{Objective Function}

To train the generation of basis kernels and mixing coefficients, we propose combining two reconstruction losses. It is worth emphasizing that the network is trained to predict a specific kernel basis (and mixing coefficients) for each image.  

\paragraph{Reblur Loss} Given a blurry image $\mathbf{v}^{GT}$, we aim to find the global  kernel basis  $\{\mathbf{k}^b\}$ and mixing coefficients $\{\mathbf{m}^b\}$ that minimize
\begin{equation} \label{eq:reblur_loss}
    \mathcal{L}_{reblur} = \sum_i  w_{i} ( v^\gamma_{i} - (v^{GT}_{i})^\gamma)^2,
\end{equation}
where the $v_i$ are computed using~\eqref{eq:model_sat}. The re-blurring of the sharp image can be performed efficiently by first convolving it with each of the kernels in the base, and then doing an element-wise blending of the $B$ resulting images with the corresponding mixing coefficients. To prevent a single kernel from dominating the losses \eqref{eq:reblur_loss} and \eqref{eq:kernel_loss}, weights $w_{i}$ are computed as the inverse of the number of pixels that belong to the same segmented object. \\ %

\paragraph{Kernel Loss}  Ground truth pixel-wise motion blur kernels are compared to the predicted per-pixel kernels. %
Given a ground truth per-pixel blur kernel $\{\mathbf{k}^{GT}_{i}\}$, the computed kernel basis $\{\mathbf{k}^b\}$ and the corresponding mixing coefficients $\{m^b_{i}\}$, the \emph{kernel loss} is defined as:
\begin{equation} \label{eq:kernel_loss}
    \mathcal{L}_{kernel} = \sum_i  w_{i} \left\Vert \sum_{b=1}^Bm^b_{i}\mathbf{k}^b-\mathbf{k}^{GT}_{i} \right\Vert_2^2,
\end{equation}
where the weights $w_{i}$ are the ones in the \emph{reblur loss}.

Note that there are infinitely many combinations of kernel basis and mixing coefficients that can generate the same motion kernel field. While it is possible to impose different constraints to promote a unique decomposition, we are not interested in any particular one, so we just let the network learn how to predict the decomposition.

\subsection{Training strategy}
We train the \textit{Kernel Prediction Network} using the sum of the \emph{reblur loss} \eqref{eq:reblur_loss} and the \emph{kernel loss} \eqref{eq:kernel_loss} with equal weights. %
We trained our model for $900$ epochs using Adam optimizer and image patches of $256\times256$ pixels. We started with a learning rate of $1e-4$ and halved it after $500$ and $750$ epochs. 

\input{section_experiments_kernels}

\section{Blind Motion Deblurring with Pixel-Wise Kernel Estimation}
\label{sec:deblurring_pipeline}

\begin{figure}[ht]
    \centering
    \includegraphics[width=0.45\textwidth]{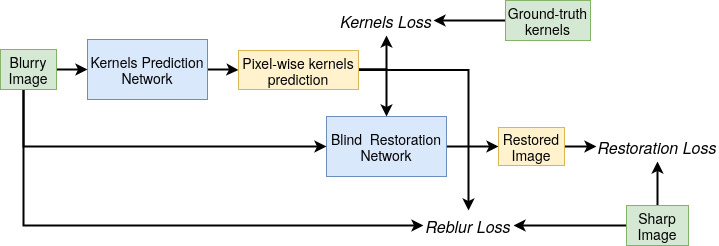}
    \caption{Scheme of the Joint Motion Kernels Prediction and Deblurring (J-MKPD) training procedure. A Blind Reconstruction Network \cite{laroche2022deep} receives the kernels estimated by the proposed Kernel Prediction Network (KPN) from the blurry input. The joint training is useful to remove artifacts produced by the mismatch between ideal and estimated kernels.}
    \label{fig:losses_diagram}
\end{figure}

The motion blur kernel field produced by the proposed KPN could be used to feed any non-blind image deconvolution algorithm, to solve the inverse problem of blind motion deblurring of real photographs. For that purpose, we used a \textit{deep plug-and-play} deconvolution method. In particular, we used a network that unrolls a linearized ADMM optimization method, recently proposed by Laroche \etal~\cite{laroche2022deep}, with a modified objective function to account for saturated pixels. More details on this method can be found in the Supplementary Material, however, it is worth mentioning here why we selected this particular deconvolution network. First, this method is efficient, as it does not require inverting the spatially-varying blur kernel operator, which would be computationally intractable. Second, this deconvolution method is a plug-and-play method that introduces a modern denoiser prior as trained on natural images as a regularizer. Lastly, since the linearized ADMM iteration is implemented as an unrolled feedforward network, it can be concatenated with the KPN, and both networks can be jointly trained or fine-tuned. This last point is important: indeed, it is often the case that, when directly plugging the kernels estimated by the proposed KPN into the restoration network, the restored images present several artifacts due to a mismatch between separately trained networks. Jointly fine-tuning the KPN and the restoration network allows us to remove artifacts and produce high-quality results, as shown in the experiments presented in Section~\ref{sec:deblurring_experiments}. To this end, fine-tuning of both networks is performed by introducing a {\em Restoration Loss}, which aims to minimize the distance between the target sharp images and the image predicted by the restoration network, i.e.: 
\begin{equation} \label{eq:restoration_loss}
    \mathcal{L}_{restoration} = \sum_i ( u_{i} - u^{GT}_{i})^2.
\end{equation}
The deblurring results we present in the next section are obtained by jointly fine-tuning both networks using a loss that combines the {\em restoration loss} with the {\em reblur loss} \eqref{eq:reblur_loss} and the {\em kernel loss} \eqref{eq:kernel_loss}. We refer to this model as J-MKPD (Joint Motion Kernels Prediction and Deblurring). A diagram illustrating the joint-training procedure with the involved losses is presented in \cref{fig:losses_diagram}. While only the {\em restoration loss}  could have been considered, we found that the other two losses add desirable regularization properties for the solution. 

\input{section_experiments_deblurring}

\input{section_extensions}

\section{Conclusions}
\label{sec:conclusions}

In this work, we proposed a method for predicting dense, spatially varying motion kernel fields from a single image. The method features an efficient image-dependent kernel basis decomposition, that results in a compact, non-parametric definition of the spatially-varying motion field. This KPN, trained jointly with a network unrolling a deep plug-and-play deconvolution method, leads to a new blind motion deblurring method that explicitly enforces the degradation model. Extensive experimental results validate the estimated motion blur kernel fields, and show that the blind motion deblurring based on KPN outperforms state-of-the-art end-to-end deep learning methods in terms of generalization to real blurry images, with the additional advantage of providing accurate motion blur kernel fields. The estimated motion kernels, besides other potential uses in computer vision, prove useful for interpretability and diagnosis of restoration failures. 

We consider that this work contributes to bridging a gap between model-based and data-driven methods for motion blur removal. Extending this method to other related image restoration tasks, and quantifying the impact of applying this method in conjunction with other algorithms to resolve other computer vision tasks, is a matter of future work.  

\section*{Acknowledgments}
This work was partially supported by Agencia Nacional de Investigación e Innovación (ANII, Uruguay) grant POS\_FCE\_2018\_1\_1007783. The experiments presented in this paper were carried out using ClusterUY (https://cluster.uy) and GPUs donated by NVIDIA Corporation.

\bibliography{refs}
\bibliographystyle{IEEEtran}

 \input{biography}

\appendices

\end{document}

\title{Supplementary Material for Blind Motion Deblurring with Pixel-Wise Kernel Estimation via Kernel Prediction Networks }

\maketitle

\section{Evaluation datasets}
\label{sec:evaluation_datasets}
\input{evaluation_datasets}

\section{Procedure to compute the distortion metrics \label{sec:quantification_distortion}}

The procedure to compute the PSNR and SSIM on K\"{o}ler's and RealBlur datasets is slightly different. As the K\"{o}ler's dataset has many possible \textit{ground-truth} images (each of the 192 frames in the sequence), the metrics are computed with respect to each frame, and the best one is chosen. To consider geometric misalignments between the restored image and the \textit{ground-truth} candidates, the images are registered with a translation before computing the metrics. On the RealBlur \cite{rim_2020_ECCV} dataset, there is only one \textit{ground-truth} candidate per restored image. Instead of registering the restored image and the \textit{ground-truth} with a translation,  \cite{rim_2020_ECCV} computes a homography. %
In both cases, intensities mismatches are taken into account by computing a scale factor that matches the energies of the restored and \textit{ground-truth} image.   

\section{Reference-free metrics based on sharpness}

\label{sec:no_reference_metrics}
\input{supp_mat_no_reference_metrics}

\section{Improving distortion metrics by blurring our restorations}

To further confirm that the restorations produced by our deblurring pipeline are sharper than RealBlur ground-truth images, we analyze how the PSNR and the SSIM distortion metrics vary as we blur our results. 

\begin{figure}[h]
    \centering
     \begin{tabular}{*{2}{c}}
    \includegraphics[width=0.45\textwidth]{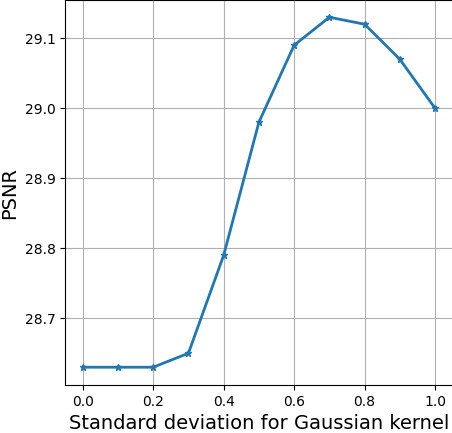} &
    \includegraphics[width=0.45\textwidth]{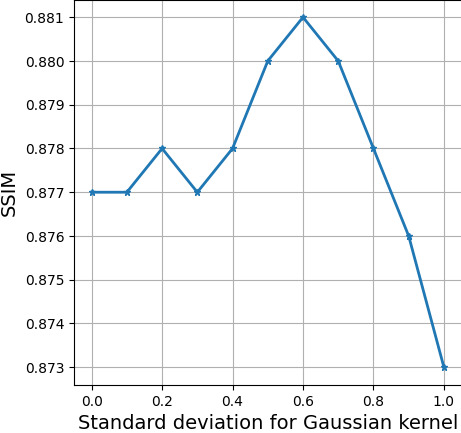}
    \end{tabular}
    \caption{When we blur our restored images on the RealBlur dataset \cite{rim_2020_ECCV} with a gaussian filter, we improve our PSNR and SSIM. The peak is reached with a Gaussian filter of standard deviation 0.7 and 0.6, respectively. This corresponds to a blurriness of approximately two pixels. The relative increment in PSNR is more significant than in SSIM.
    \label{fig:distortion_vs_blur}.}
\end{figure}

\cref{fig:distortion_vs_blur} shows that the PSNR and SSIM of our restored images do not decrease monotonically as we intentionally blur them with Gaussian filters of increasing standard deviation $\sigma$.

\cref{fig:blurring_restored} shows, for a selection of images from the RealBlur dataset, that the PSNR of the restored image increases when we intentionally blur them. This shows that, at least on RealBlur, where the target image is not totally sharp, a higher PSNR does not necessarily means that the restoration is sharper. %

\label{sec:blur_on_purpose}
\input{figure_RealBlur_blurred}

\section{Kernels on K\"{o}hler's dataset \cite{kohler2012recording}}

In \cref{fig:KernelEstimationKohler_1,fig:KernelEstimationKohler_2,fig:KernelEstimationKohler_3,fig:KernelEstimationKohler_4} we show the kernels fields estimated by our method on the Kohler's dataset. We also show the kernels estimated by other methods and the ground-truth kernels. Each figure corresponds to an original scene. For each scene, we show results on two different motion blurs. %
Note that our kernel prediction network can recover motion kernels in low-textured regions successfully, and produces motion kernel fields that do not suffer from correlation with the scene content.   
\label{sec:more_kernels_kohler}
\input{figure_kernels_Kohler_sup_mat}

\subsection{Impact of training on {\em SBDD}}

We trained the motion trajectory estimation network proposed by Zhang \emph{et al.} \cite{zhang2021exposure}, which was trained on the GoPro dataset in the original paper, with the synthetic dataset we introduce in the article. A comparison of the resulting trajectories when trained on both datasets is shown in \cref{fig:kernels_Zhang_vs_ours}, along with the results obtained by our model.  

We observe that the motion trajectories obtained by the model proposed in \cite{zhang2021exposure} are significantly better when trained on our synthetic dataset. However, the motion blur kernels obtained with our Kernel Prediction Network are more accurate. Indeed, our model does not impose constant motion speed nor any constraint on the kernels’ shape, whereas \cite{zhang2021exposure} is designed to recover motion trajectories
(quadratic curves) of constant speed.

\input{figure_kernels_Zhang_vs_ours}

\section{Deblurring results on Lai's dataset \cite{lai2016comparative}}

In \cref{fig:Lai_suppmat,fig:Lai2_suppmat,fig:LaiHS_suppmat,fig:LaiHS2_suppmat}, we compare our restorations with those of state-of-the-art methods on blurry images from Lai's datasets \cite{lai2016comparative}. 

\input{figure_Lai_suppmat}

\section{Laroche \etal ~deblurring network and its adaptation to handle saturation}

We used the deblurring network proposed by~\cite{laroche2022deep}. The network was generated by unfolding an iterative algorithm.  First, we review the algorithm, which is a deep plug-and-play algorithm based on a linearized ADMM (Alternating Direction Method of Multipliers) splitting technique~\cite{liu2019linearized}. Then we describe how the algorithm is transformed into a deblurring network that can be used to restore images degraded with spatially-varying blur. Finally, we explain how to modify the formulation to account for saturation.

In this section, we denote by $\mathbf{y}$ the observed, degraded image, and by $\mathbf{x}$ the image to be recovered. $\mathbf{H}$ is a linear operator, that represents the image degradation process. 

\paragraph{The iterative algorithm:} Model-based variational methods usually solve the deblurring problem by imposing a prior with density $p(\mathbf{x}) \propto \exp(-\lambda \phi(\mathbf{x}))$ to the latent image. Then, the maximization of the posterior density $p(\mathbf{x}|\mathbf{y}) \propto p(\mathbf{y}|\mathbf{x}) p(\mathbf{x})$ leads to the following optimization problem: 
\begin{equation}
    \mathbf{x}^* = \arg \min_{\mathbf{x}} \frac{1}{2 \sigma^2} \Vert \mathbf{H}\mathbf{x} - \mathbf{y} \Vert_2^2 + \lambda \Phi(\mathbf{x}).  
\end{equation}
This optimization problem can be solved using iterative alternate minimization schemes, like ADMM, which leads to the iterative solution
\begin{equation}
    \begin{array}{ll}
    \mathbf{x}_{k+1} &= \arg \min_{\mathbf{x}} \lambda \Phi(\mathbf{x}) + \frac{\mu}{2} \Vert \mathbf{x} - (\mathbf{v}_k - \mathbf{u}_k)\Vert_2^2 = \mathcal{P}_{\sqrt{\lambda / \mu}} (\mathbf{v}_k - \mathbf{u}_k)  \\
    \mathbf{v}_{k+1} &= \arg \min_{\mathbf{v}} \frac{1}{2 \sigma^2} \Vert \mathbf{H}\mathbf{v} - \mathbf{y}  \Vert_2^2  + \frac{\mu}{2} \Vert \mathbf{v} - (\mathbf{x}_{k+1} + \mathbf{u}_k) \Vert_2^2   \\
    \mathbf{u}_{k+1} & = \mathbf{u}_{k} + (\mathbf{x}_{k+1} - \mathbf{v}_{k+1}), 
    \end{array}
\end{equation}
for steps $k=0,1, \ldots, N$. 

The last equality in the $\mathbf{x}$-update step makes it explicit that the middle term is the proximal operator of $\Phi$. Unlike classical methods, deep plug-and-play methods replace the $\mathbf{x}$-update step by a denoising network $\mathcal{P}_{\beta}$ pre-trained to remove zero-mean Gaussian noise with variance $\beta^2$, on a set of natural images. This way, the denoiser imposes an implicit prior $\Phi$ for natural images. 

As for the $\mathbf{v}$-update, when the blur is uniform, it can be efficiently computed using the Fast Fourier Transform since the blur operator $\mathbf{H}$ is diagonal in the Fourier basis. However, this is no longer the case for the spatially-varying blur model
\begin{equation}
    \mathbf{H}\mathbf{x} = \sum_{b=1}^{B} \mathbf{m}_b \cdot \left( \mathbf{k}_b * \mathbf{x} \right).
\end{equation}
In~\cite{laroche2022deep}, the authors consider this very same motion blur model, and to solve this issue they propose to use linearized ADDM. To this end, first they substitute the splitting variable $\mathbf{v}=\mathbf{x}$ by $\mathbf{z} = \mathbf{H}\mathbf{v}$, and then the introduce the linear approximation $\Vert \mathbf{H} \mathbf{v} - \mathbf{z}_k\Vert ^2 \approx \mu\left( \mathbf{H}^T\mathbf{H}\mathbf{x}_k - \mathbf{H}^T \mathbf{z}_k \right)^T\mathbf{x} + \frac{\rho}{2} \Vert \mathbf{x} - \mathbf{x}_k \Vert^2$, leading to the following iteration:
\begin{equation}
    \begin{array}{ll}
    \mathbf{x}_{k+1} &= \arg \min_{\mathbf{x}} \lambda \Phi(\mathbf{x}) + \frac{\rho}{2} \Vert \mathbf{x} - (\mathbf{x}_k - \frac{\mu}{\rho} \mathbf{H}^T\left( \mathbf{H}\mathbf{x}_k - \mathbf{z}_k + \mathbf{u}_k \right) )\Vert_2^2 = \mathcal{P}_{\sqrt{\lambda / \rho}} \left(\mathbf{x}_k - \frac{\mu}{\rho} \mathbf{H}^T\left( \mathbf{H}\mathbf{x}_k - \mathbf{z}_k + \mathbf{u}_k \right) \right) \\
    \mathbf{z}_{k+1} &= \arg \min_{\mathbf{z}} \frac{1}{2 \sigma^2} \Vert \mathbf{z} - \mathbf{y}  \Vert_2^2  + \frac{\mu}{2} \Vert \mathbf{z} - (\mathbf{H} \mathbf{x}_{k+1} + \mathbf{u}_k) \Vert_2^2   \\
    \mathbf{u}_{k+1} & = \mathbf{u}_{k} + (\mathbf{H} \mathbf{x}_{k+1} - \mathbf{z}_{k+1}). 
    \end{array}
\end{equation}

The $\mathbf{z}$-update can be computed using the following closed-form solution:
\begin{equation}
    \mathbf{z_{k+1}} = \frac{\mathbf{y} + \sigma^2 \mu \left( \mathbf{H}\mathbf{x_{k+1} + \mathbf{u}_k} \right)}{\sigma^2 \mu + 1}.
\end{equation}
Finally, the $\mathbf{u}$-update is straightforward. 

\paragraph{The unfolded network:}  Following the linearized
ADMM formulation described above, the architecture proposed in~\cite{laroche2022deep} alternates between a prior-enforcing step $\mathcal{P}$ corresponding to the $\mathbf{x}$-update,
a data-fitting step $\mathcal{D}$ corresponding to the $\mathbf{z}$-update, and finally, an update block $\mathcal{U}$.

For the $\mathbf{x}$-update, the authors retrained a ResUnet~\cite{diakogiannis2020resunet} architecture. Denoting by $\beta_k=\sqrt{\lambda_k/\rho_k}$ and $\gamma_k=\mu_k / \rho_k$, the update becomes
\begin{equation}    \mathbf{x}_{k+1}=\mathcal{P}_{\beta_k} \left( \mathbf{x_k} - \gamma_k \mathbf{H}^T\left( \mathbf{H}\mathbf{x}_k - \mathbf{z}_k + \mathbf{u}_k \right) \ \right).
\end{equation}

The data-term module $\mathcal{D}$ computes the $\mathbf{z}$-update by defining $\alpha_k=\sigma^2 \mu_k$ and computing 
\begin{equation}
    \mathbf{z_{k+1}} = \frac{\mathbf{y} + \alpha_k \left( \mathbf{H}\mathbf{x_{k+1} + \mathbf{u}_k} \right)}{\alpha_k + 1}.
\end{equation}

The update module $\mathcal{U}$ updates the dual variable $\mathbf{u}$ and does not have trainable parameters.

\begin{figure}[h]
    \centering
    \includegraphics[width=0.8\textwidth]{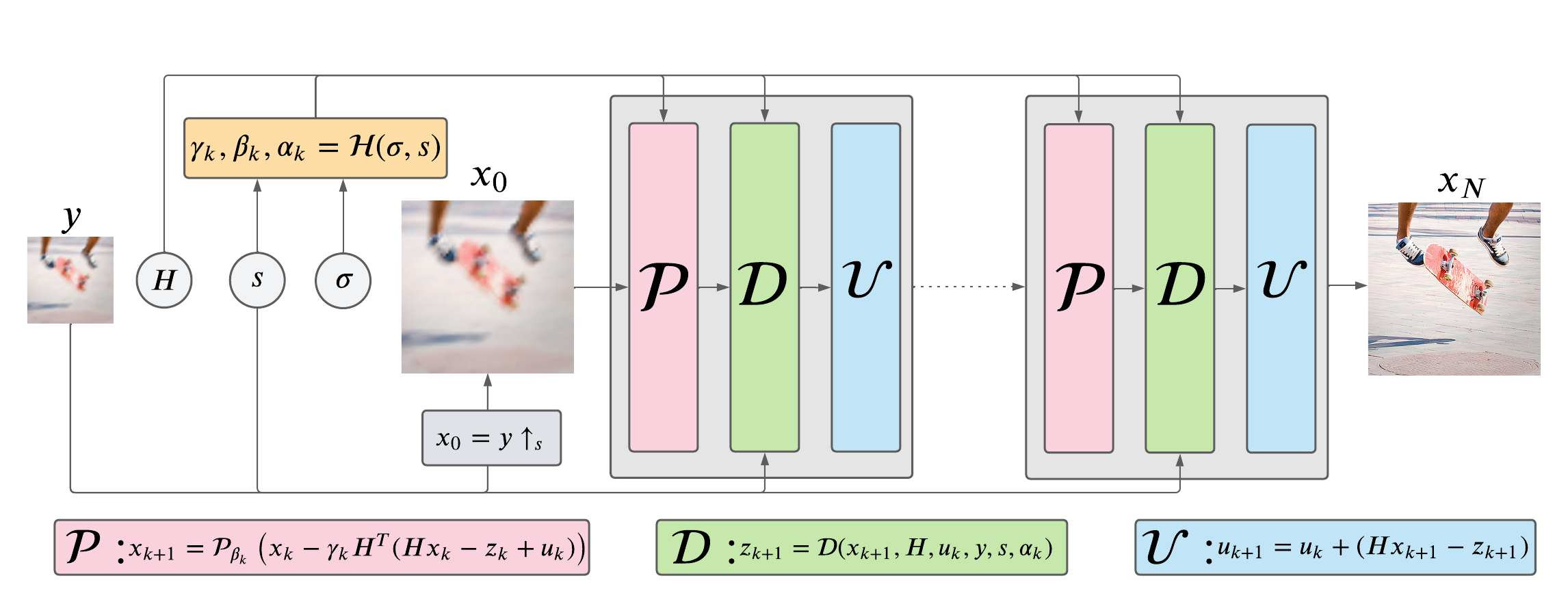}
    \caption{Deconvolution network proposed by Laroche \etal. Image reproduced from~\cite{laroche2022deep}.}
    \label{fig:deconvolution_network}
\end{figure}

These blocks are respectively stacked eight times, corresponding to the
number of iterations. The optimization process requires the hyper-parameter triplets ($\lambda$, $\mu$, $\rho$) that are predicted by the hyper-parameters block $\mathcal{H}$ at each step. The authors used three fully connected layers for the architecture of $\mathcal{H}$. The resulting architecture is
presented in \cref{fig:deconvolution_network}.    

\subsection{Adaptation to handle saturation}

Adapting the method to handle saturation is straightforward. In this case the optimization problem writes
\begin{equation}
    \mathbf{x}^* = \arg \min_{\mathbf{x}} \frac{1}{2 \sigma^2} \Vert R(\mathbf{H}\mathbf{x}) - \mathbf{y} \Vert_2^2 + \lambda \Phi(\mathbf{x}), 
\end{equation}
where $R$ is the differentiable approximation of the captor response function defined in Section III of the article. Considering the saturation function $R(\mathbf{z})$, with $\mathbf{z} = \mathbf{H}\mathbf{x}$, the data term
\begin{equation}
h_{sat}(\mathbf{z}) := \frac{1}{2\sigma^2} \Vert R(\mathbf{z}) - \mathbf{y} \Vert^2     
\end{equation}
can be linearized around $\mathbf{z}_k$ as
\begin{equation}
\hat{h}_{sat}(\mathbf{z}) =h_{sat}(\mathbf{z}_k) +  <\mathbf{z}-\mathbf{z}_k, \frac{ 1}{\sigma^2} (R(\mathbf{z}_k)-\mathbf{y})R'(\mathbf{z}_k)> + \frac{L_z}{2}\Vert \mathbf{z}-\mathbf{z}_k  \Vert^2 
\end{equation}
Therefore, the (linearized) cost to be minimized is given by
\begin{equation}
\hat{F}(\mathbf{z}) = <\mathbf{z}-\mathbf{z}_k, \frac{(R(\mathbf{z}_k)-\mathbf{y})R'(\mathbf{z}_k)}{\sigma^2}>  + \frac{L_z}{2}\Vert \mathbf{z}-\mathbf{z}_k  \Vert^2 + \frac{\beta}{2} \Vert \mathbf{z} -  (\mathbf{H}\mathbf{x}_{k+1} + \mathbf{u}_k)) \Vert^2, 
\end{equation}
where we have introduced the dual variable $\mathbf{u}$ used in ADMM. 
Again, we want $\nabla \hat{F}(\mathbf{z})  = 0$:
\begin{equation}
\begin{array}{cl}
     \nabla  \hat{F}(\mathbf{z}) & = \frac{1}{\sigma^2}(R(\mathbf{z}_k)-\mathbf{y})R'(\mathbf{z}_k) + L_z ( \mathbf{z}-\mathbf{z}^k) + \beta ( \mathbf{z}- (\mathbf{H}\mathbf{x}_{k+1} + \mathbf{u}_k) )\\
     & = (L_z + \beta) \mathbf{z} + \frac{1}{\sigma^2}(R(\mathbf{z}_k)-\mathbf{y})R'(\mathbf{z}_k) - L_z \mathbf{z}_k -\beta (\mathbf{H}\mathbf{x}_{k+1} + \mathbf{u}_k ) \\
     &= 0 
\end{array}
\end{equation}
Finally, the $\mathbf{z}$-update becomes
\begin{equation}
\begin{array}{cl}
     \mathbf{z}_{k+1} & = \frac{1}{(L_z + \beta)\sigma^2} \left( \left(\mathbf{y} - R(\mathbf{z}_k) \right) R'(\mathbf{z}_k) + \sigma^2 L_z  \mathbf{z}_k + \beta \sigma^2 (\mathbf{H}\mathbf{x}_{k+1} + \mathbf{u}_k ) \right). 
\end{array}
\label{eq:LinADMM_saturated}
\end{equation}

\newpage 

\bibliographystyle{IEEEtran}
\bibliography{refs}

%% file: section_experiments_kernels.tex
\section{Assessment of the Motion Blur Kernel Field Estimation \label{sec:kernel_estimation_experiments}}

\subsection{Qualitative Evaluation}

\cref{fig:KernelsAndMasks} shows examples of the set of kernel basis and corresponding mixing coefficients predicted for different images. Note that the predicted basis is image-dependent, especially for those kernel basis elements that are more active in the decomposition (i.e. the corresponding mixing coefficients have high values throughout the scene).

\input{figure_kernels_Kohler}

\cref{fig:KernelEstimationKohler} shows the motion field predicted by our method on a real blurry image from K\"ohler's dataset~\cite{kohler2012recording} and compares it with other existing DL-based non-uniform motion blur estimation methods: Gong \etal \cite{gong2017motion}, Sun \etal~\cite{sun2015learning} and Zhang \etal~\cite{zhang2021exposure}. Although the blur kernel fields on the K\"ohler's dataset are slightly spatially-variant, it is still an interesting dataset since, to our knowledge, it is the only standard benchmark of real blurred images for which the motion kernels are known. A visual comparison with the ground truth kernels shows that our method significantly outperforms the others. More examples can be found in the Supplementary Material.

\cref{fig:KernelEstimationExamples} depicts typical spatially-variant motion blur kernel estimates produced by our method. Despite being trained on synthetically blurred images, our approach (last column) generalizes remarkably well to real blurry images, as well as blurry images synthesized from video sequences as in GoPro~\cite{Nah_2017_CVPR} and REDs~\cite{Nah_2019_CVPR_Workshops_REDS} datasets.

\input{figure_kernels}
 Our model can characterize different types of camera and object motions. Note also that the motion blur kernels estimated by the compared methods suffer from the \textit{aperture} problem on low variance regions and tend to correlate with the scene geometry instead of revealing the underlying motion field. Moreover, our model predicts continuous free-form arbitrary motion kernels, whereas~\cite{sun2015learning}, ~\cite{gong2017motion}, and \cite{zhang2021exposure} are restricted to parametric shapes (linear o quadratic curves). %
 
\subsection{Evaluation through Motion Segmentation \label{sec:blur_detection}}
\input{BlurDetection}

\subsection{Impact of training on SBDD}

The motion blur estimation methods discussed in this work were trained on different datasets. To evaluate how these methods perform when trained on {\em SBDD}, we selected the motion trajectory estimation method proposed by Zhang~\cite{zhang2021exposure}, as it achieves the best reblur performance on the GoPro dataset. Qualitative results, presented in the Supplementary Material, show that the spatial coherence of the motion trajectories is significantly better when trained on SBDD. Still, the recovered motion kernels are not as accurate as ours, since our model does not impose constant motion speed nor any constraint on the kernels' shape, whereas \cite{zhang2021exposure} is designed to recover constant speed motion trajectories (quadratic curves). 

As a quantitative comparison between both methods, we performed the blur detection experiment reported in \cref{sec:blur_detection} with the motion estimation proposed by~\cite{zhang2021exposure}, trained on the GoPro dataset and on {\em SBDD}. The amount of blur at each pixel was computed as the motion trajectory length. The mean average precision was 0.735 and 0.754, respectively.

%% file: figure_kernels_Kohler.tex
\begin{figure}
  \centering
\setlength{\tabcolsep}{2pt}
  \begin{tabular}{cc}
    Blurry & Gong \cite{gong2017motion} \\
     \includegraphics[trim=150 170 150 300, clip,width=0.235\textwidth]{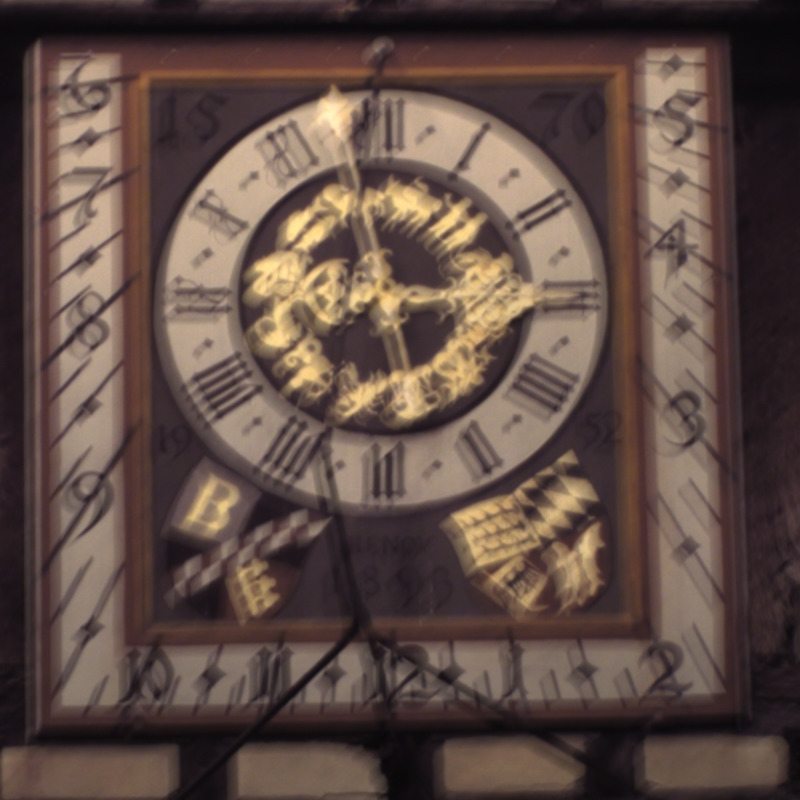}  &
    \includegraphics[trim=150 170 150 300, clip,width=0.235\textwidth]{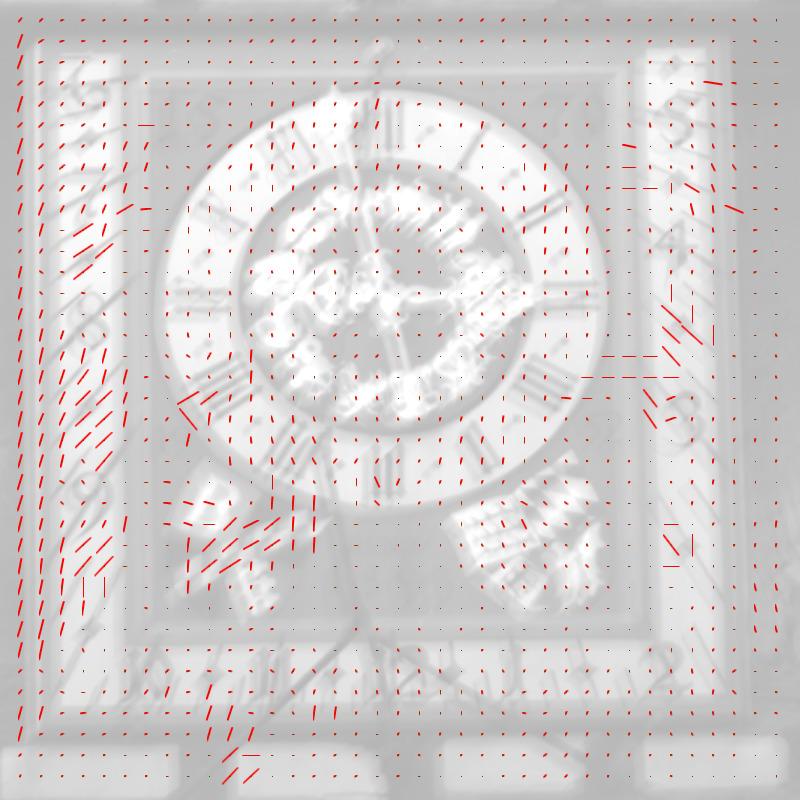}   \\
    Sun \cite{sun2015learning}   &  Zhang \cite{zhang2021exposure}\\ 
      \includegraphics[trim=150 170 150 300, clip,width=0.235\textwidth]{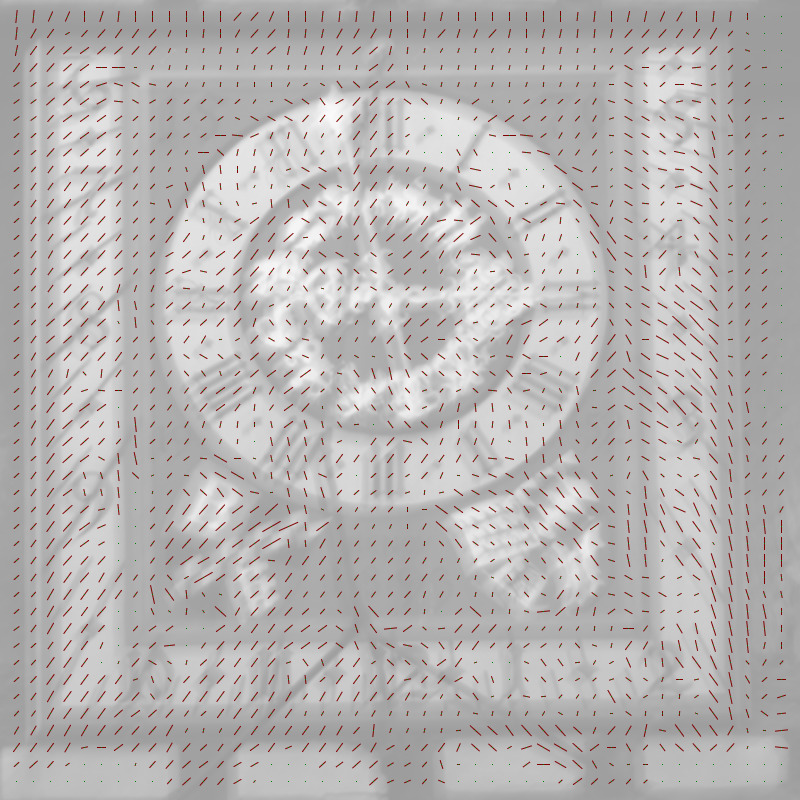} &    \includegraphics[trim=150 170 150 300, clip,width=0.235\textwidth]{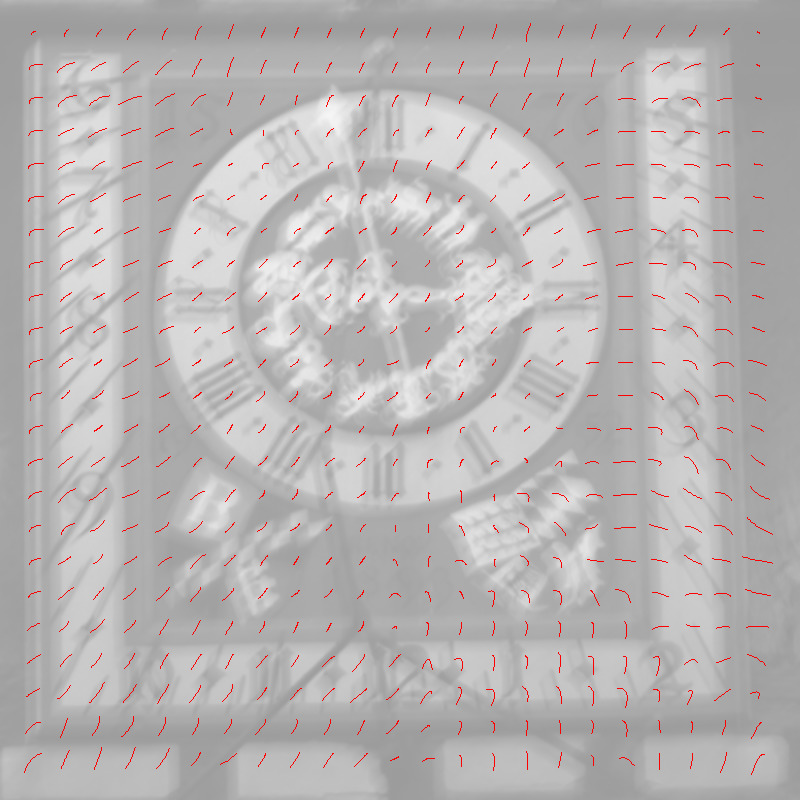}   \\ 
    Ours & GT \\
     \includegraphics[trim=150 170 150 300, clip,width=0.235\textwidth]{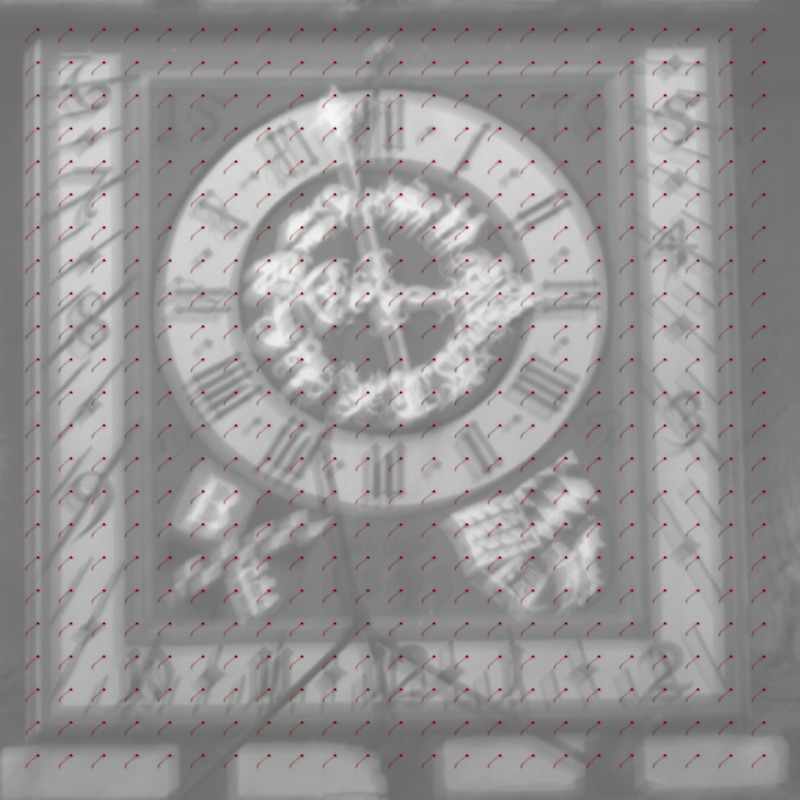}  &
     {\setlength{\fboxsep}{0pt}%
       \setlength{\fboxrule}{0.05pt}%
      \framebox{\includegraphics[trim=150 170 150 300, clip,width=0.235\textwidth]{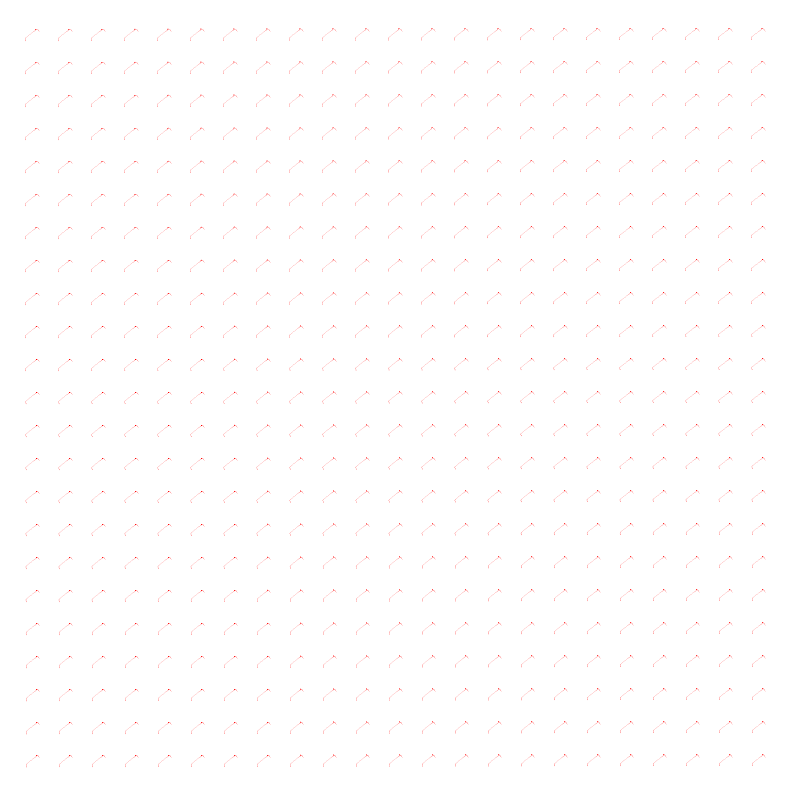} 
     }}    
  \end{tabular}
  \caption{\textbf{Visual comparison of per-pixel kernel estimation on a real image with known ground-truth (GT).} %
  Other motion estimation methods do not generalize to datasets not used during training. Motion fields are correlated with the image structure, suffer from the aperture problem, or predict deltas in blurry images. More examples can be found in the Supplementary Material. }
  \label{fig:KernelEstimationKohler}
\end{figure}

%% file: figure_kernels.tex
\begin{figure*}[ht]
  \centering
\setlength{\tabcolsep}{2pt}
  \begin{tabular}{*{4}{c}}
    Gong \cite{gong2017motion} & Sun \cite{sun2015learning} &  Zhang \cite{zhang2021exposure} & Ours  \\

    \includegraphics[trim=20 80 20 20,clip, width=0.23\textwidth]{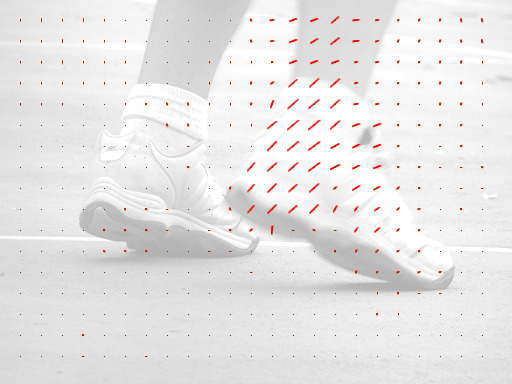}   &
      \includegraphics[trim=20 80 20 20,clip, width=0.23\textwidth]{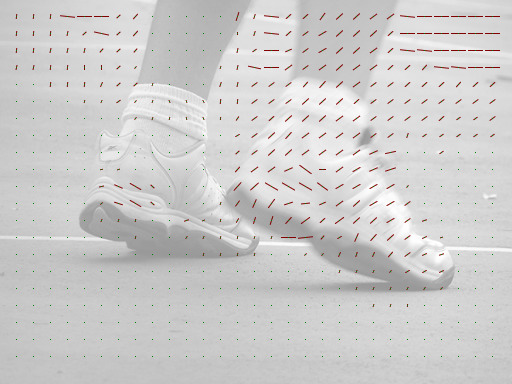}   &  
    \includegraphics[trim=20 80 20 20,clip, width=0.23\textwidth]{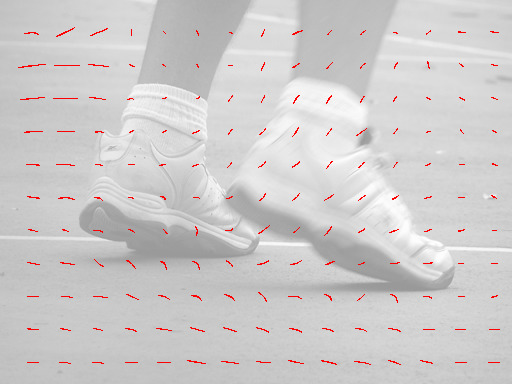}  &
   \includegraphics[trim=20 80 20 20,clip, width=0.23\textwidth]{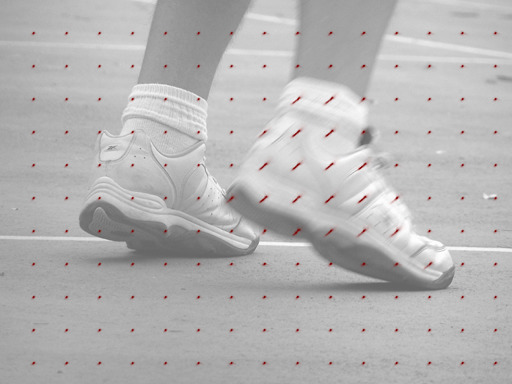}  \\

    \includegraphics[trim=50 100 700 250, clip,width=0.23\textwidth]{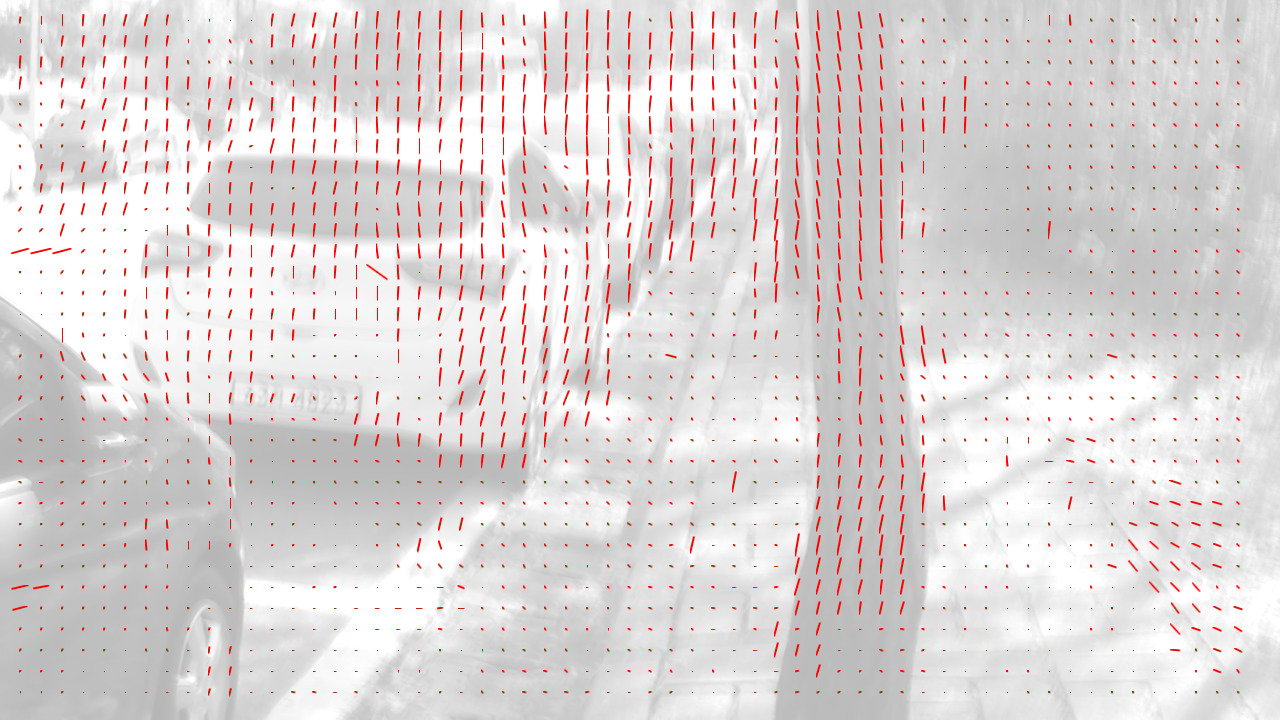}   &
      \includegraphics[trim=50 100 700 250, clip,width=0.23\textwidth]{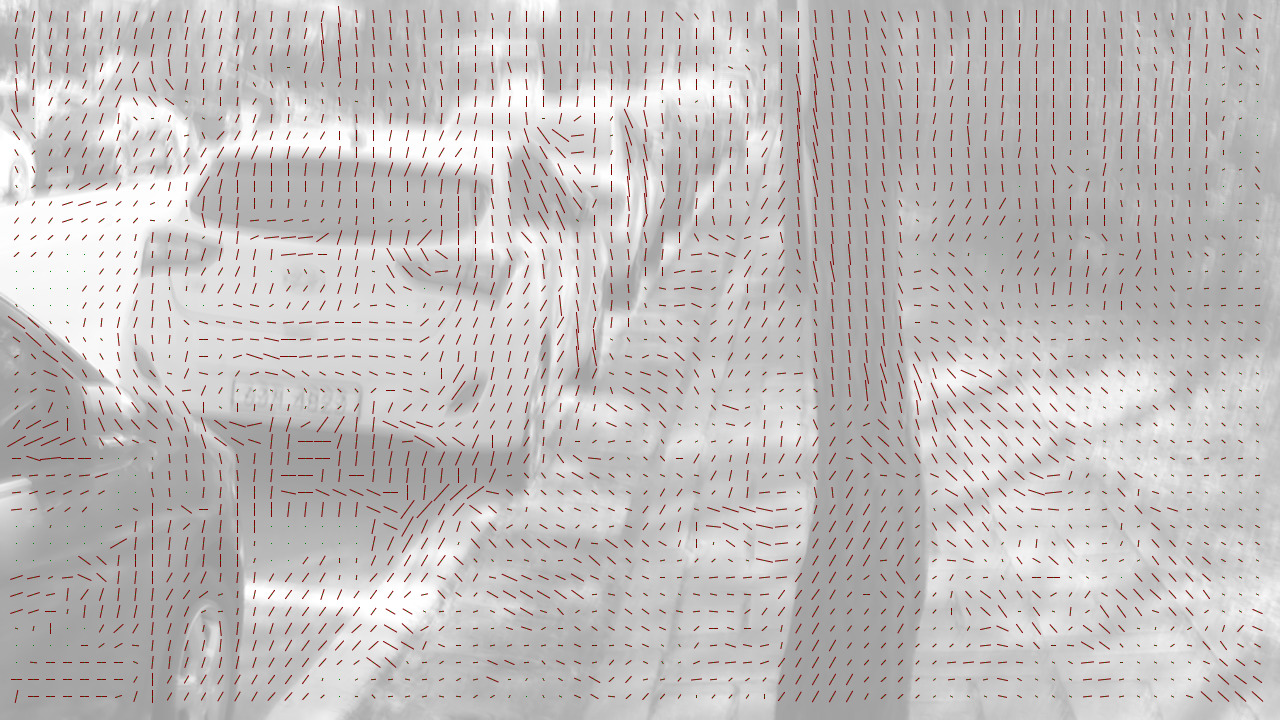} &  
     \includegraphics[trim=50 100 700 250, clip,width=0.23\textwidth]{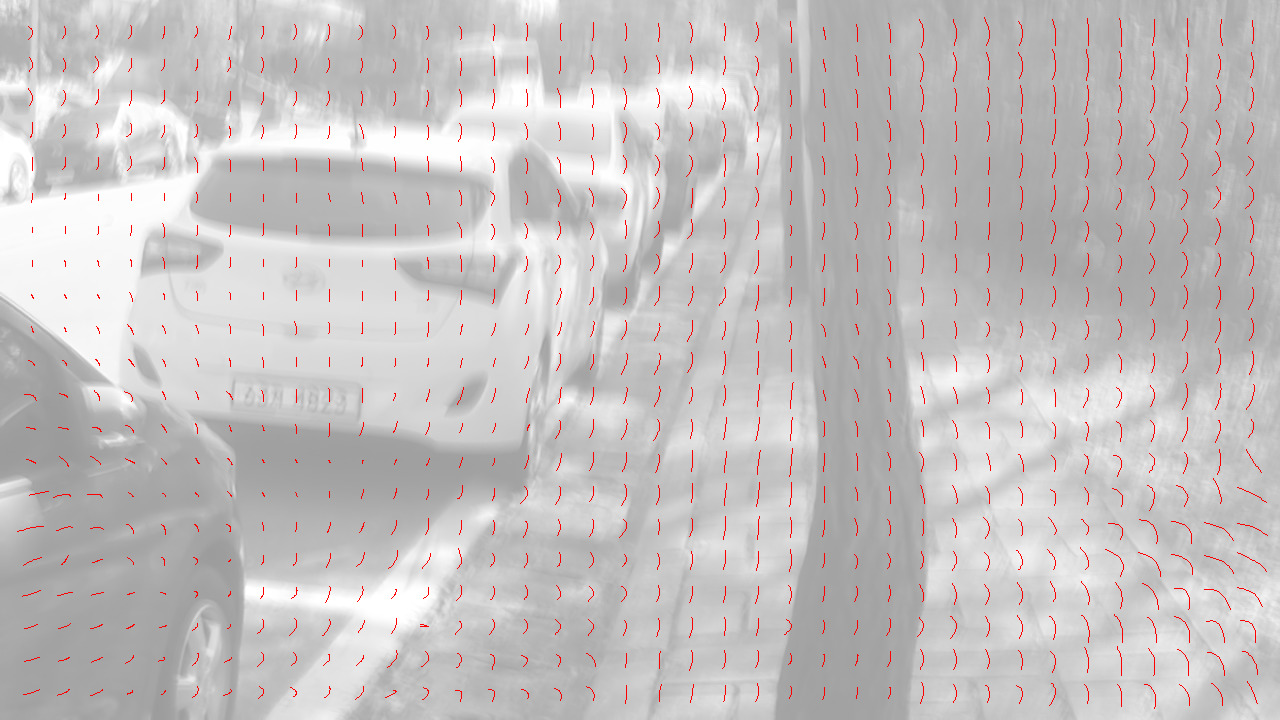} &  
        \includegraphics[trim=50 100 700 250, clip,width=0.23\textwidth]{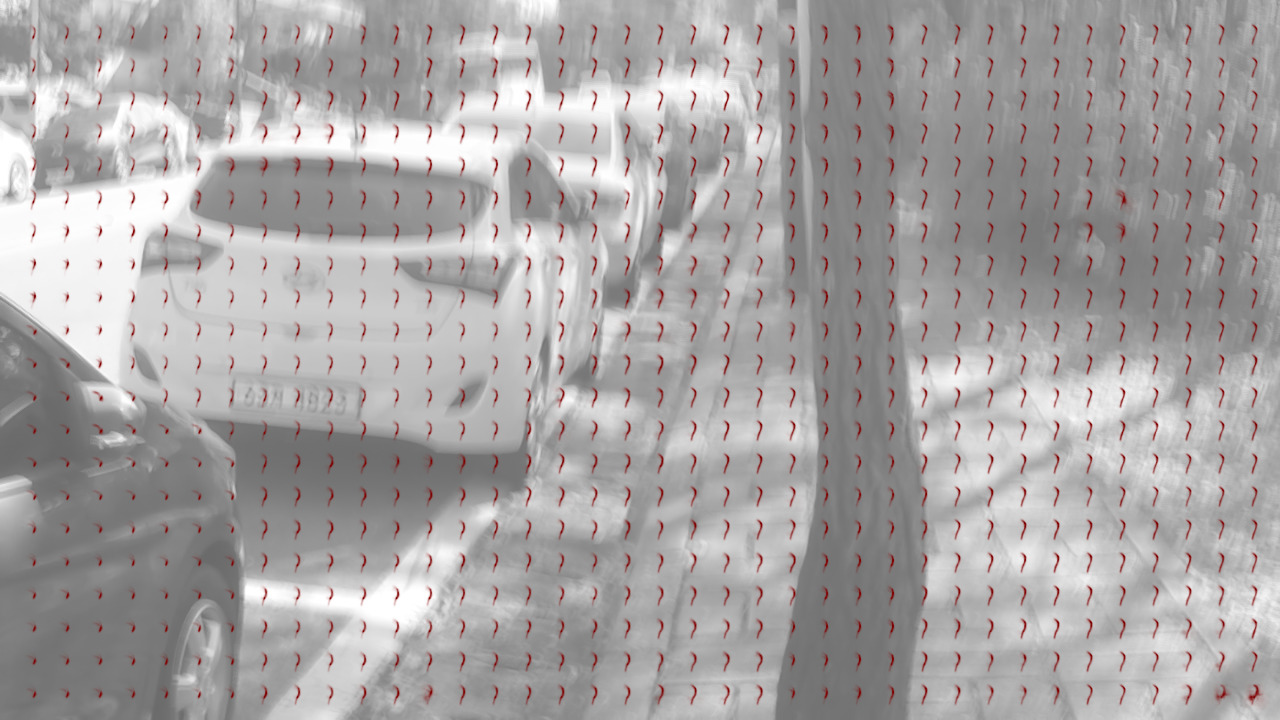}\\
    
    \includegraphics[trim=10 120 700 190, clip,width=0.23\textwidth]{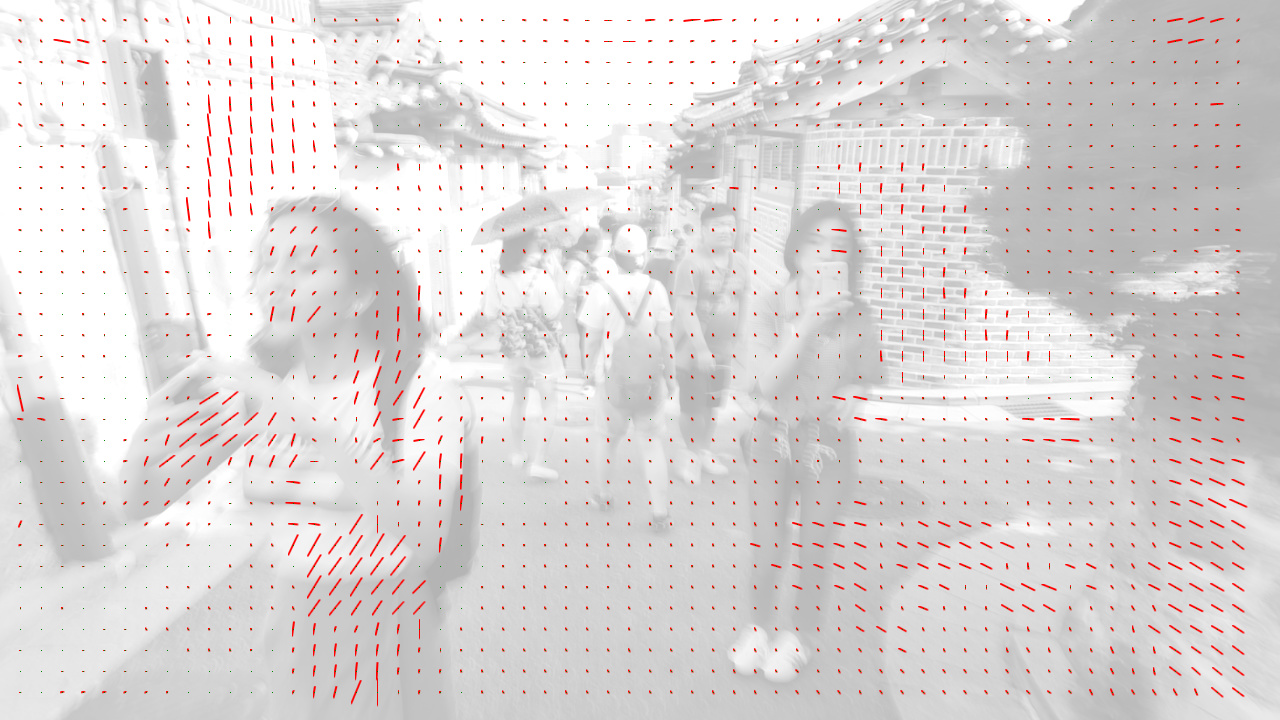}&
     \includegraphics[trim=10 120 700 190, clip,width=0.23\textwidth]{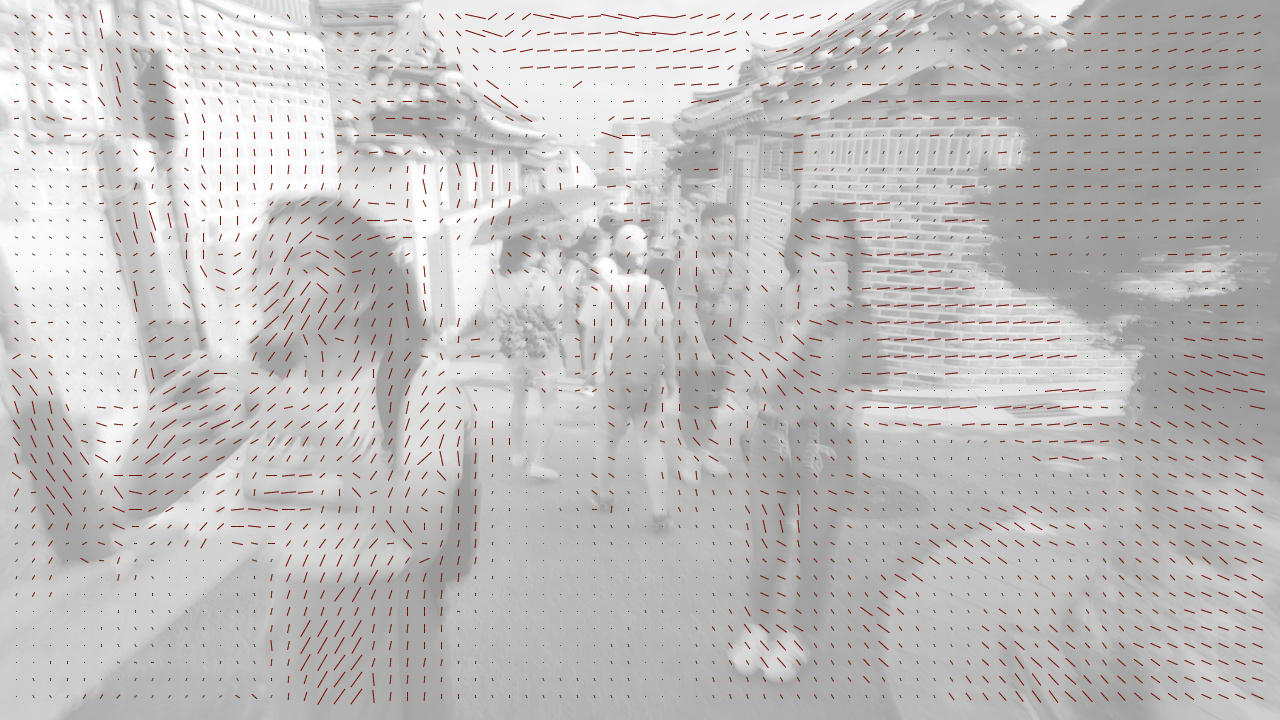}&  
      \includegraphics[trim=10 136 700 190, clip,width=0.23\textwidth]{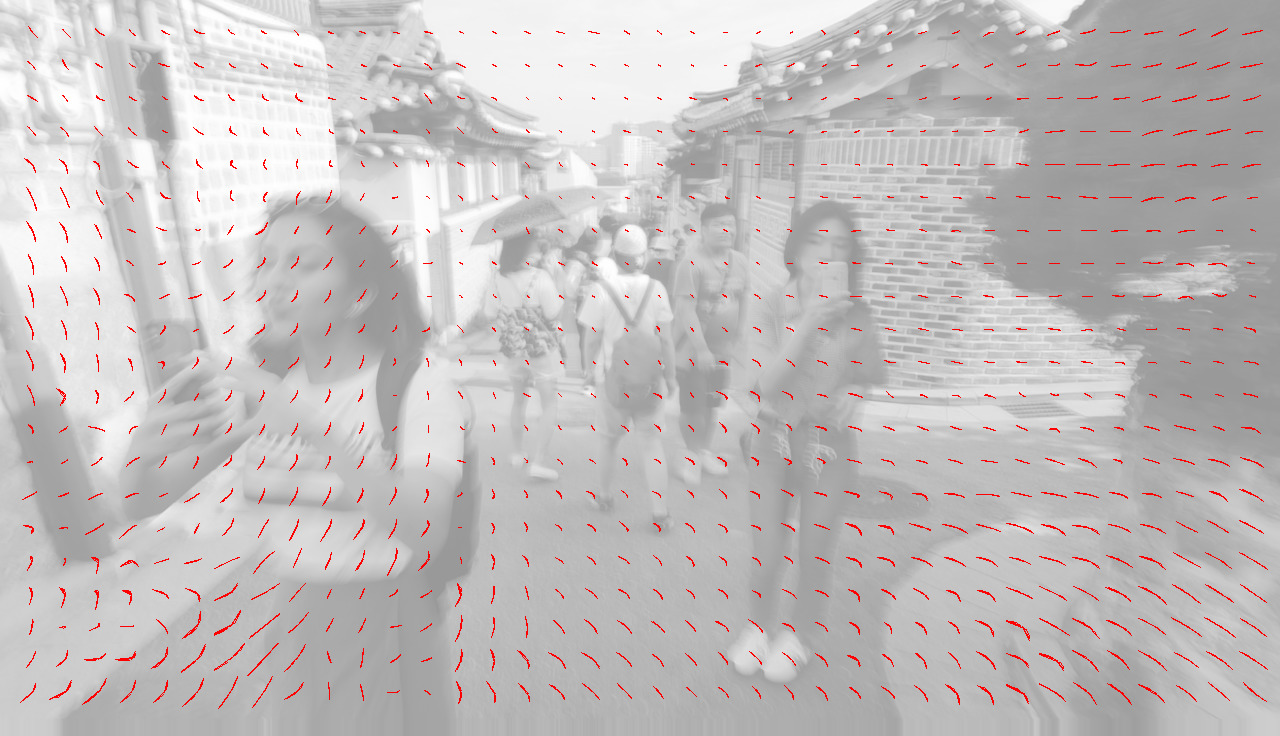}   &
     \includegraphics[trim=10 120 700 190, clip,width=0.23\textwidth]{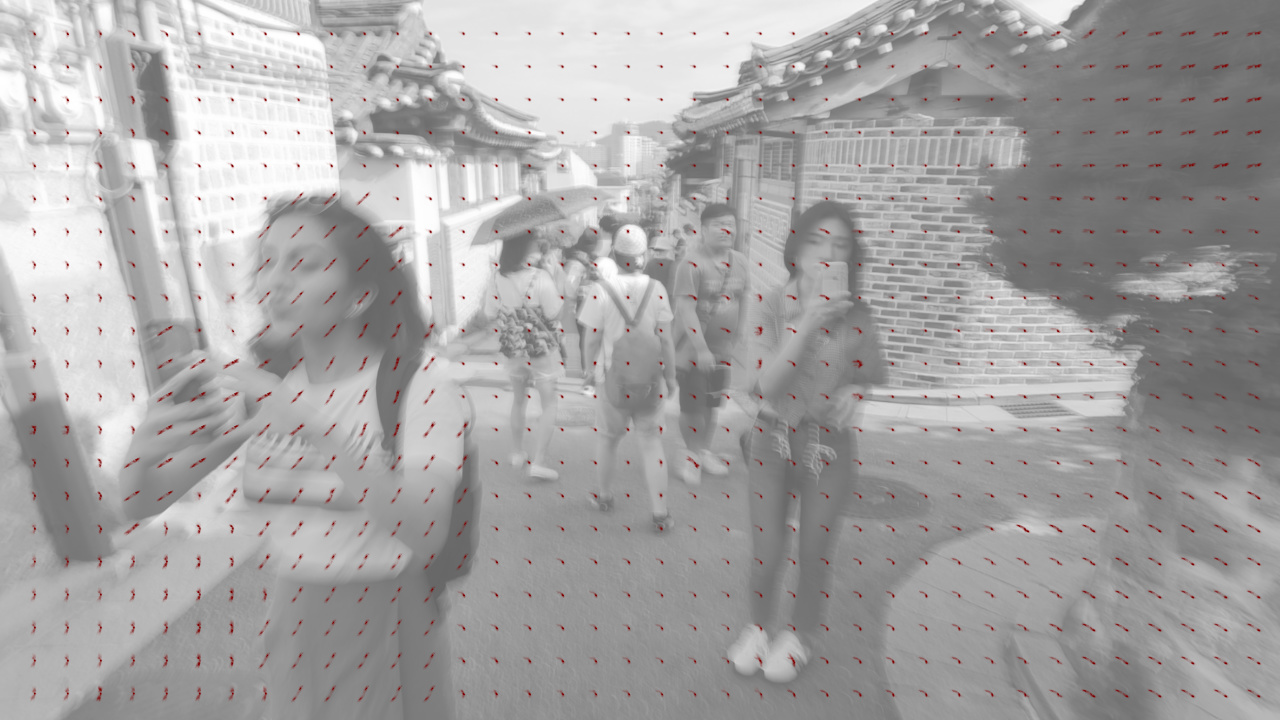} \\  

    \includegraphics[trim=10 50 740 350, clip,width=0.23\textwidth]{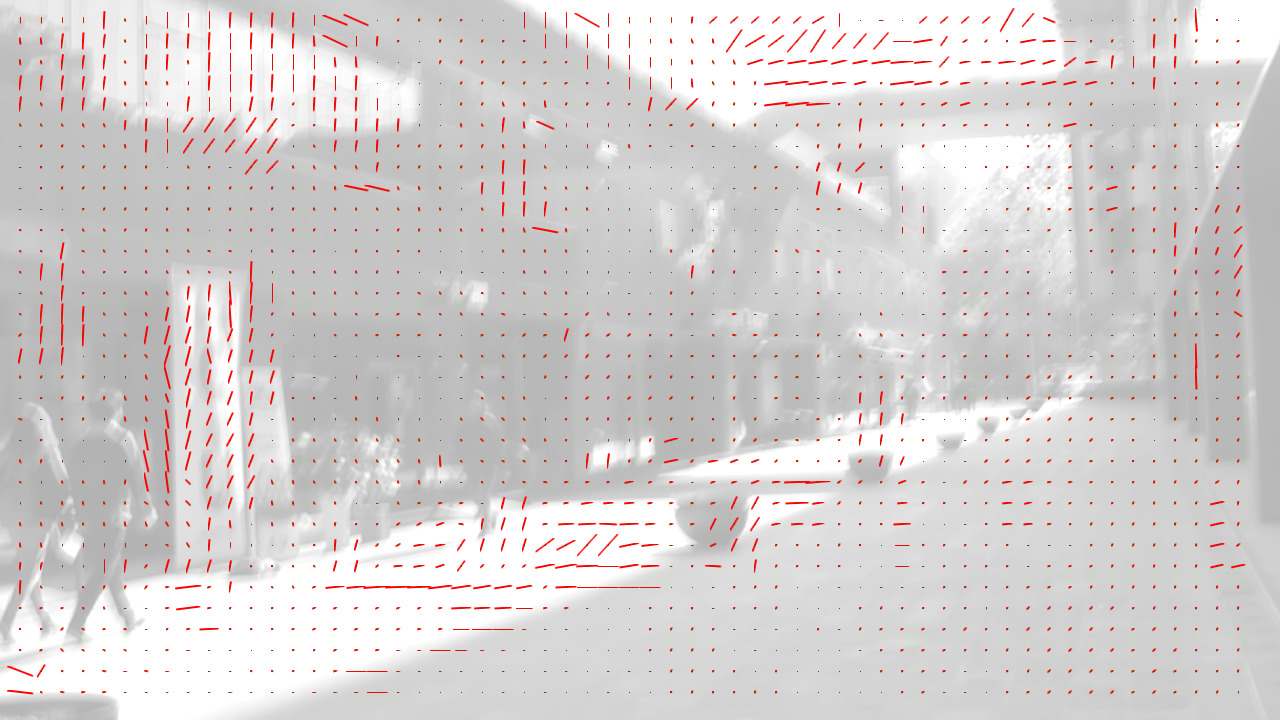}   &
      \includegraphics[trim=10 50 740 350, clip,width=0.23\textwidth]{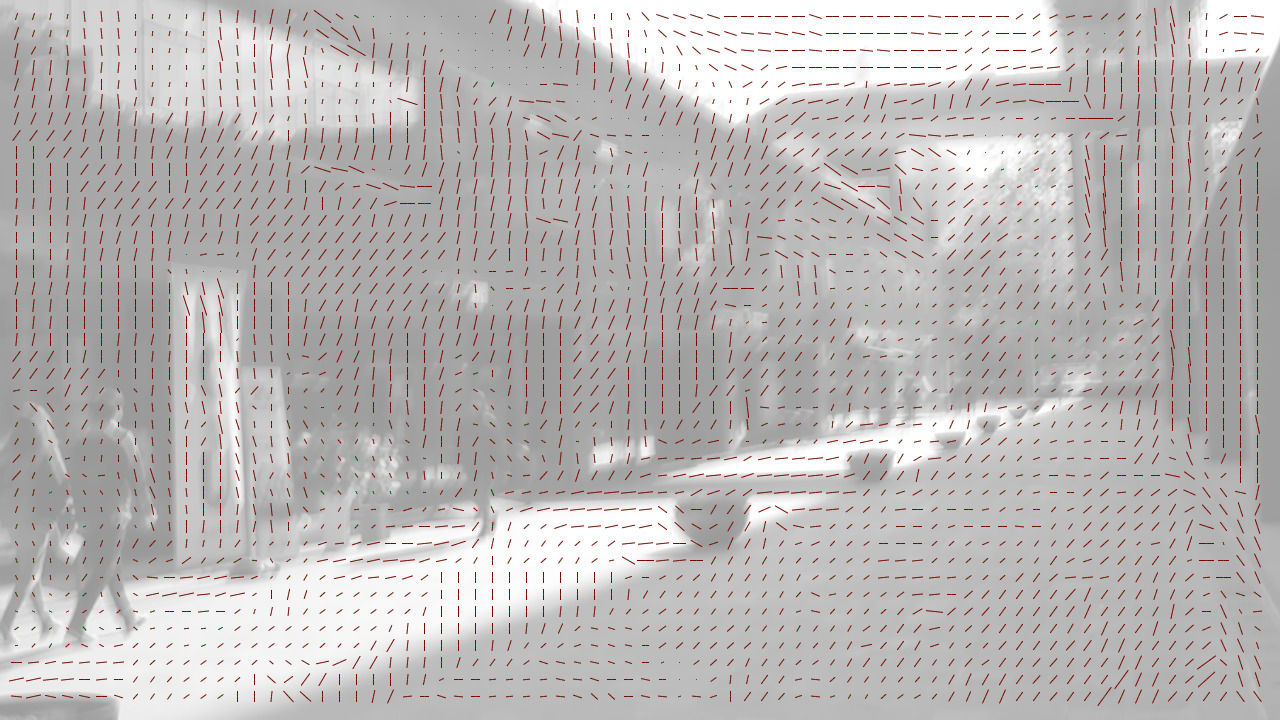} &  
     \includegraphics[trim=10 50 740 350, clip,width=0.23\textwidth]{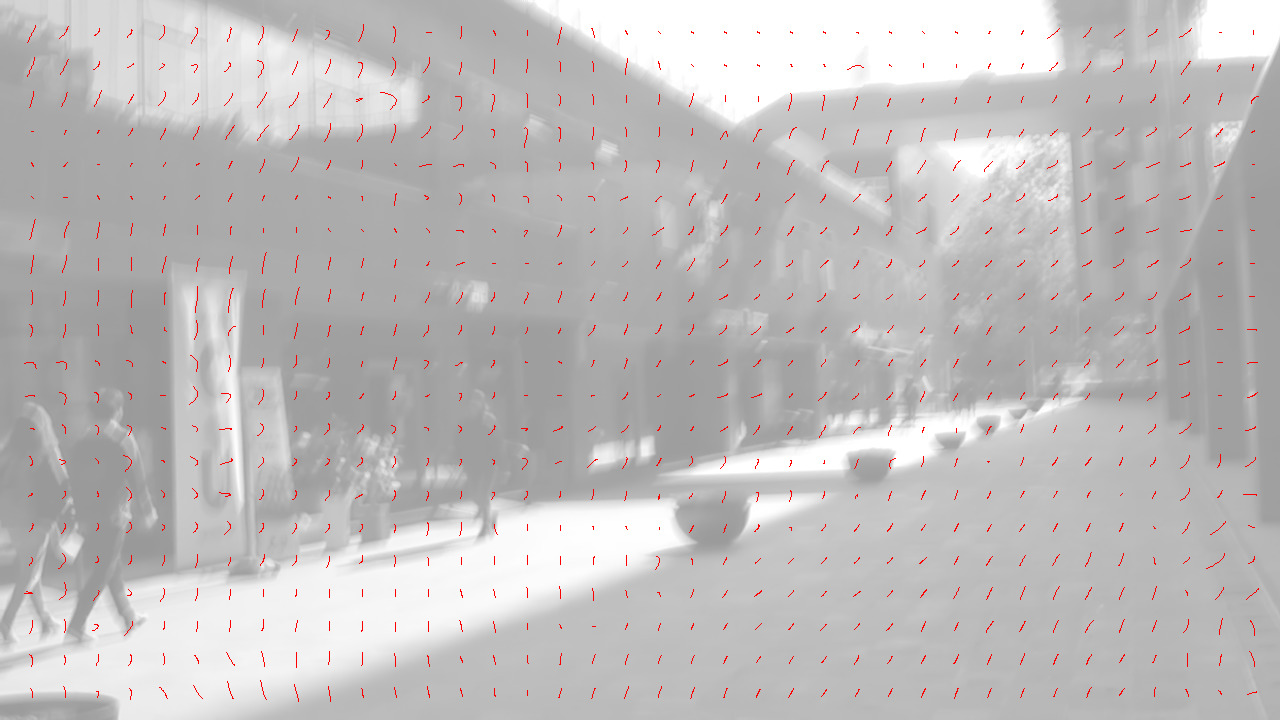} &  
        \includegraphics[trim=10 50 740 350, clip,width=0.23\textwidth]{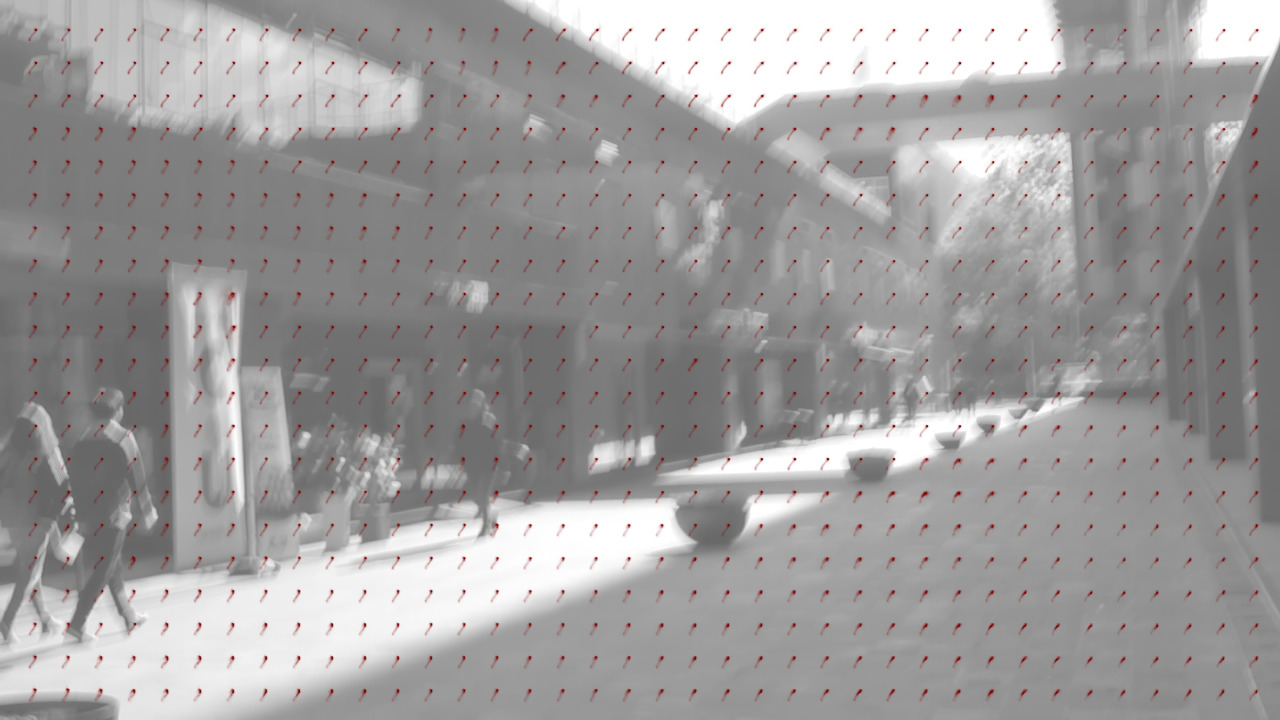}  \\

  \end{tabular}
  \caption{\textbf{Visual comparison of non-uniform motion blur kernel estimation.} %
   Our method distinguishes between different blur regions. Motion fields correlate less with the scene structure than other methods. The constraint imposed by the degradation model, promotes spatial coherence in low-texture regions.}
  \label{fig:KernelEstimationExamples}
\end{figure*}

%% file: BlurDetection.tex
\begin{figure}
    \centering
            \begin{tabular}{c@{\hspace{2px}}c@{\hspace{2px}}c@{\hspace{2px}}c}
            Input & GT &  Ours & Blur field \\
\includegraphics[width=0.11\textwidth]{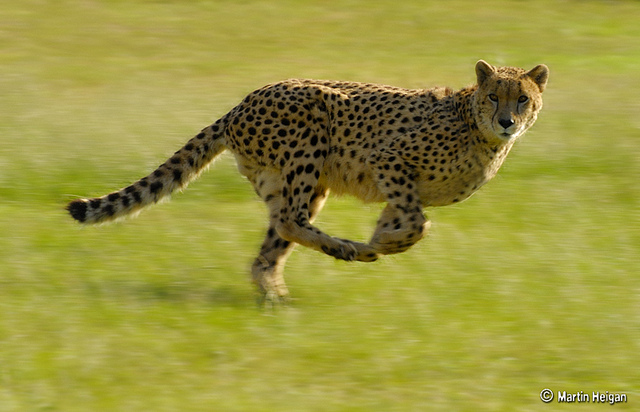}&
\includegraphics[width=0.11\textwidth]{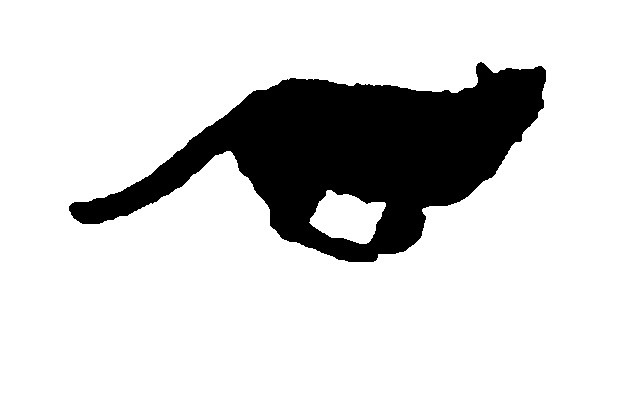}&
\includegraphics[width=0.11\textwidth]{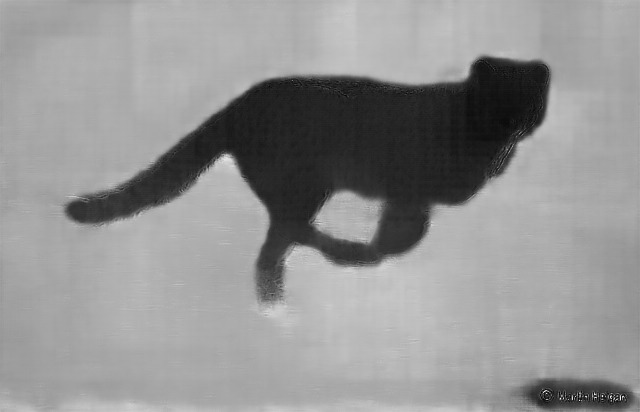} &
\includegraphics[width=0.11\textwidth]{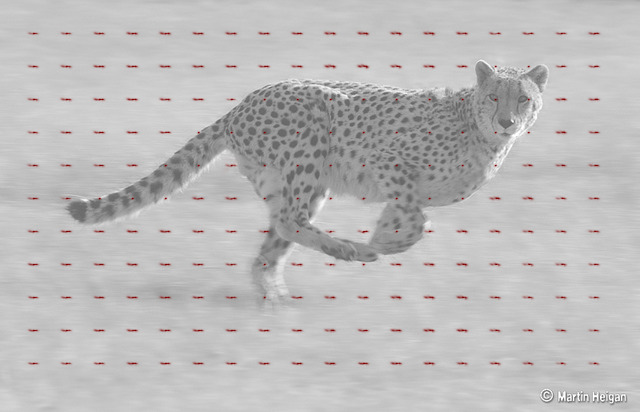}\\
\includegraphics[width=0.11\textwidth]{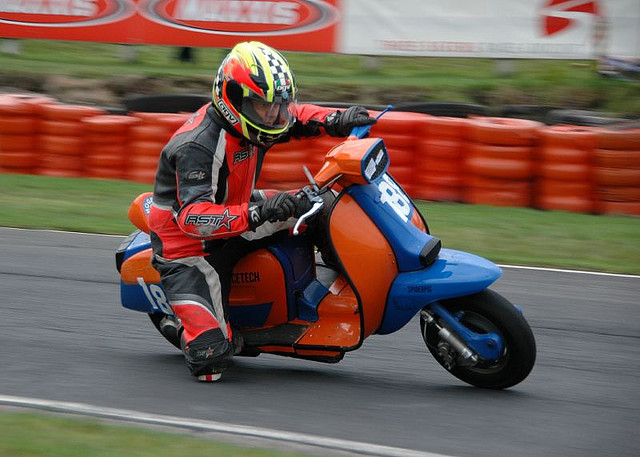}&
\includegraphics[width=0.11\textwidth]{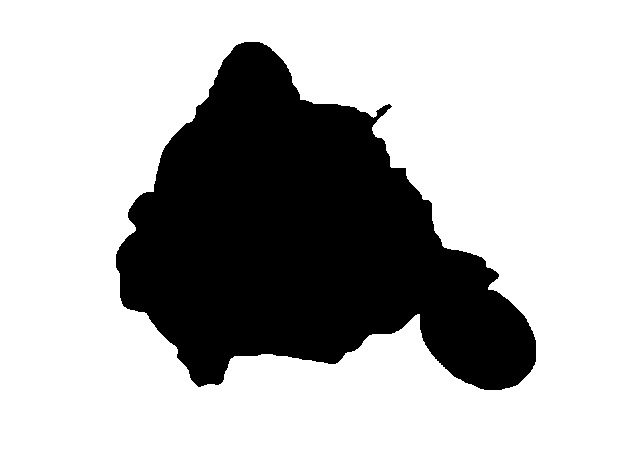} & 
\includegraphics[width=0.11\textwidth]{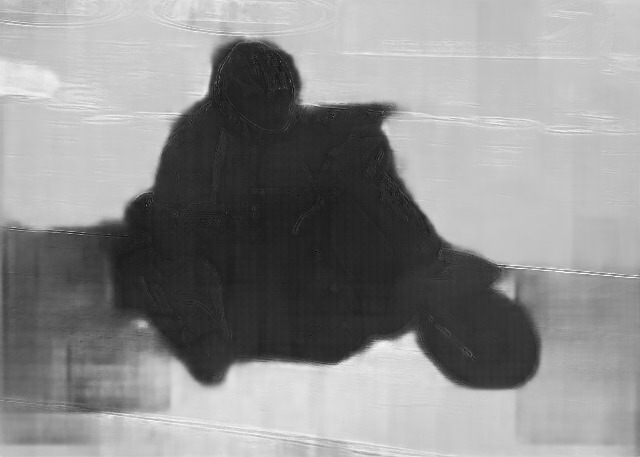} &
\includegraphics[width=0.11\textwidth]{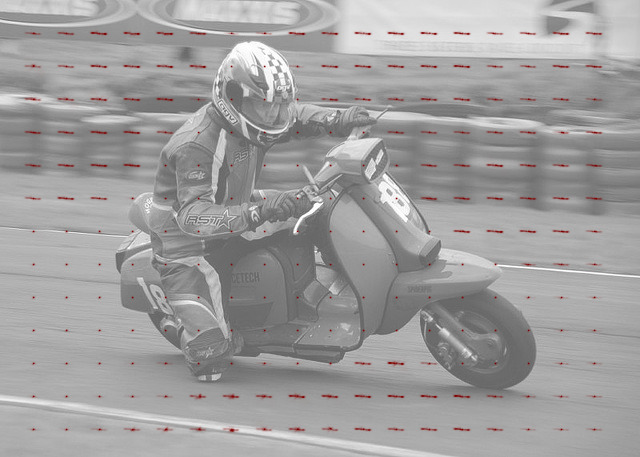}\\
\includegraphics[width=0.11\textwidth]{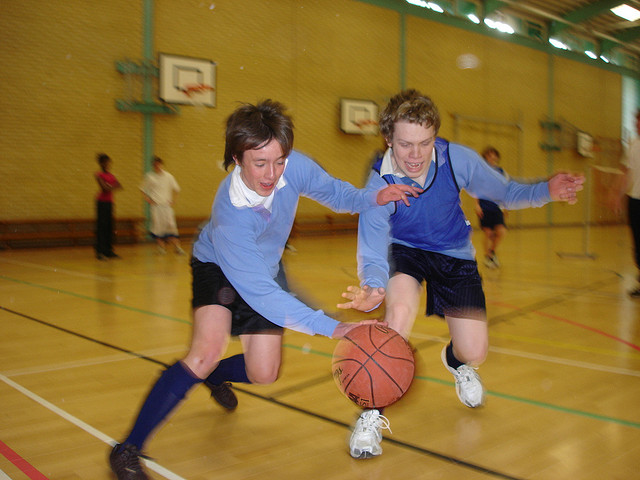}&
\includegraphics[width=0.11\textwidth]{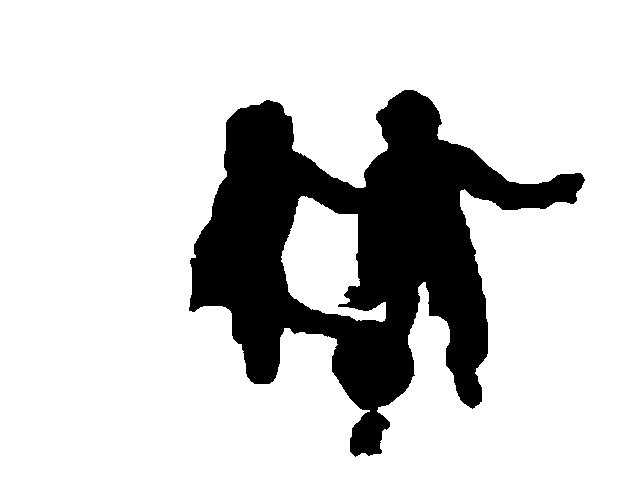} & 
\includegraphics[width=0.11\textwidth]{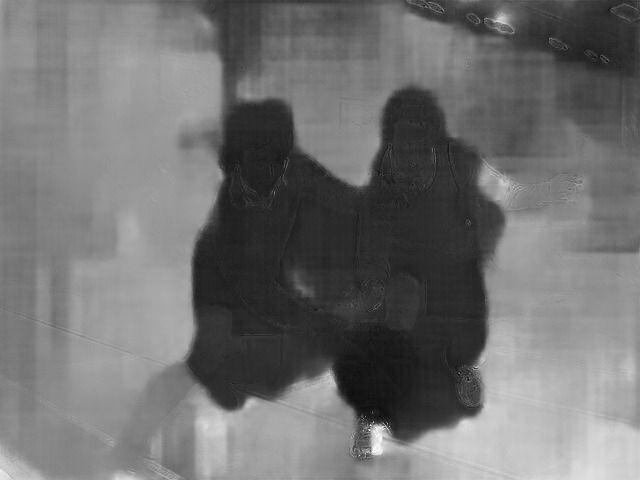}& 
\includegraphics[width=0.11\textwidth]{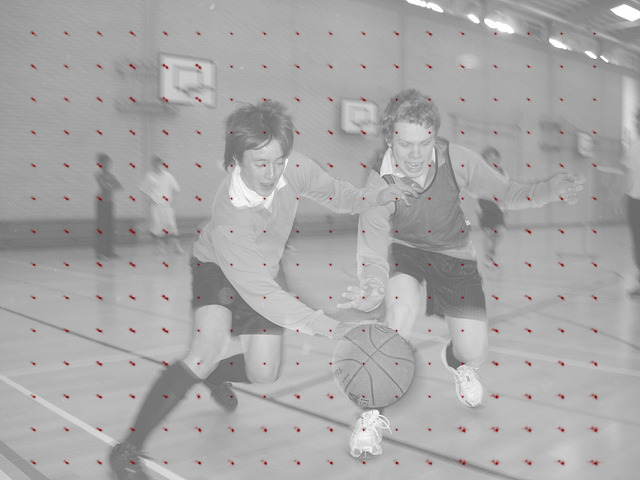}\\
\includegraphics[width=0.11\textwidth]{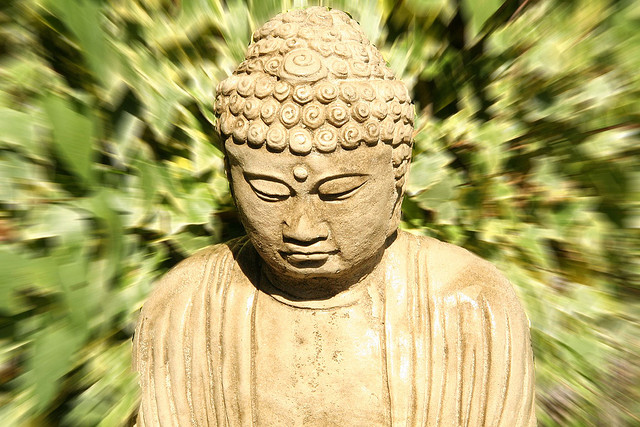}&
\includegraphics[width=0.11\textwidth]{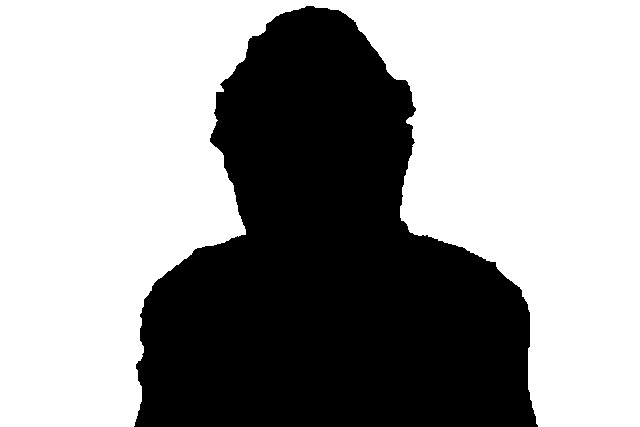} & 
\includegraphics[width=0.11\textwidth]{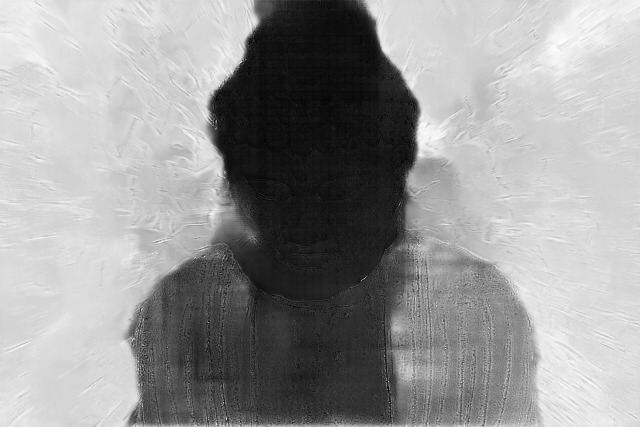}&
\includegraphics[width=0.11\textwidth]{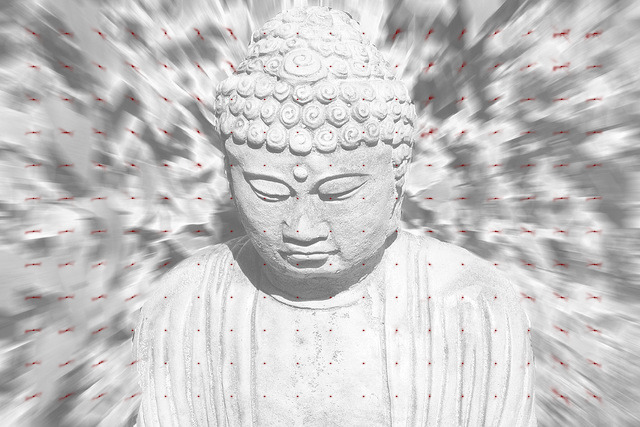}\\
\includegraphics[width=0.11\textwidth]{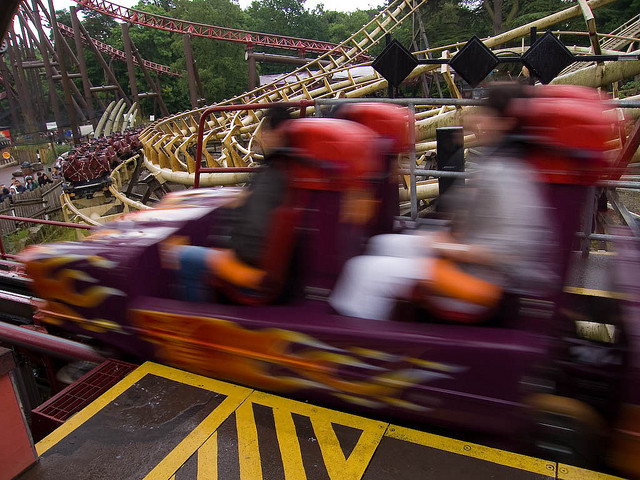}&
\includegraphics[width=0.11\textwidth]{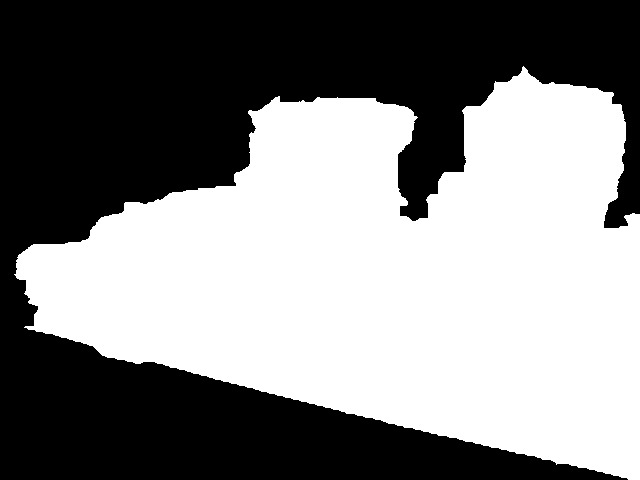} & 
\includegraphics[width=0.11\textwidth]{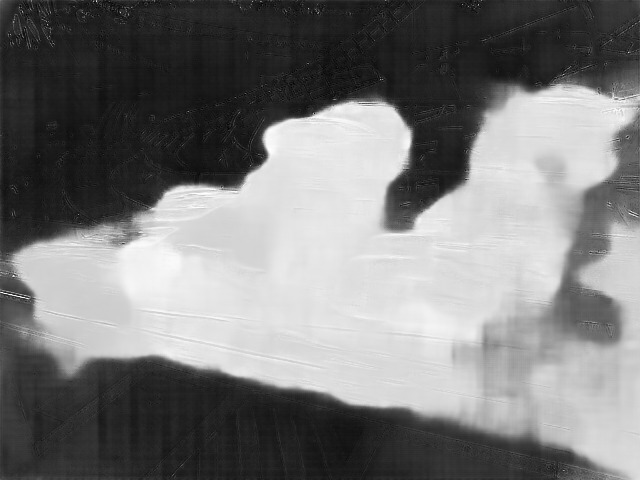}&
\includegraphics[width=0.11\textwidth]{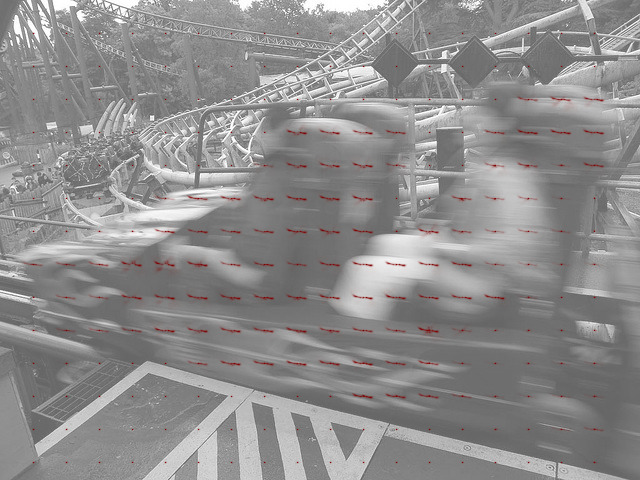}\\
\end{tabular}
    \caption{\textbf{Blur segmentation.} We use the norm of the predicted non-uniform motion blur kernels to detect regions with motion blur. Images from the CUHK blur detection dataset \cite{shi2014discriminative}.}
    \label{fig:blur_detection}
\end{figure}

Blur region detection aims at segmenting the blurred areas of a given image \cite{alvarez2019self,shi2014discriminative,golestaneh2017spatially}. Recently, the use of synthetic datasets \cite{alvarez2019self} for training blur detection networks has allowed deep learning methods to surpass the performance of methods based on local features \cite{shi2014discriminative,golestaneh2017spatially}. 
This is a straightforward application of our dense kernel field estimation method. To build a segmentation mask, we compute the per-pixel kernels and define the sharpness of the pixel as the $L^2$-norm of the kernel. Low $L^2$-norm indicates a more spread kernel. We evaluate this strategy on the standard CUHK blur detection dataset \cite{shi2014discriminative}, under the motion blur category. Figure~\ref{fig:blur_detection} shows that our approach can effectively segment regions of the image with motion blur. Quantitatively, following \cite{alvarez2019self}, we measure the mean average precision across the evaluation dataset, shown in Table~\ref{tab:blur_detection}. Despite not being trained for this task, our method is competitive with existing state-of-the-art methods specifically designed for this task.

\begin{table}[h]
  \setlength\tabcolsep{1.5pt} %
  \centering
    \caption{\textbf{Comparison on the CUHK blur detection dataset \cite{shi2014discriminative}, motion blur category.} Following \cite{alvarez2019self}, we report mean average precision across images of the  evaluation split. Values in parenthesis are taken from \cite{alvarez2019self}.  \label{tab:blur_detection}}
  \begin{tabular}{cccccc}
    \toprule 
    CUHK \cite{shi2014discriminative}  &  HiFST \cite{golestaneh2017spatially} & Ma \etal \cite{ma2018deep} & Self-sup. \cite{alvarez2019self}  & KPN (ours)\\
    \midrule 
     0.6944 & 0.7484  & (0.784) & (0.838) & \textbf{0.840} \\
    \bottomrule 
  \end{tabular}
\end{table}

%% file: section_experiments_deblurring.tex
\section{Motion Deblurring of Real Images \label{sec:deblurring_experiments}}

In this section, we evaluate the capability of the proposed deblurring procedure to generalize to real motion-blurred photographs. We compare our results with state-of-the-art DL-based and classical variational methods. In all cases, we used the models provided by the authors.  Following the main trend in the literature, methods that do not require knowledge of the motion blur kernels for training, are trained on the GoPro dataset \cite{Nah_2017_CVPR}. 
Ours was trained on the synthetic dataset presented in Section~\ref{subsec:DatasetGen}. We compare the deblurring results on the standard datasets: Köhler~\cite{kohler2012recording}, Lai~\cite{lai2016comparative}, and RealBlur~\cite{rim_2020_ECCV}. This cross-dataset evaluation allows to validate the generalization of the deblurring performance and avoids evaluating dataset-specific mappings.

\subsection{Quantitative results}

Quantifying restoration quality is a difficult task because the target image is generally unknown. K\"{o}hler's~\cite{kohler2012recording} and RealBlur \cite{rim_2020_ECCV} datasets are, to our knowledge, the only two datasets of real images with available ground truth. Since a target image is available in both cases, it is common practice to report reference-based metrics such as PSNR and SSIM. Although extremely useful, these reference-based distortion metrics are sensitive to unavoidable geometric and chromatic misalignment between the blurry/sharp pairs. This is particularly relevant for the Real Blur dataset, since there is only one target image for each blurry image, and many pictures are taken outdoors under uncontrolled lighting conditions. In K\"{o}hler's dataset, misalignment is mitigated using 192 sharp candidates for each blurry image taken in an indoor controlled environment. In the supplementary material, we present a detailed explanation of how blurry/sharp pairs are generated for each dataset, and we explain why such distortion metrics may not correctly reflect the restoration quality. 

To cope with the limitations of PSNR and SSIM, we propose to complement the quantitative evaluation with  {\em perceptual metrics}. %
To this end, we consider the LPIPS~\cite{zhang2018perceptual} as a reference-based metric, and three reference-free 
metrics:

\begin{itemize}
    \item {\em Blur Strength (BS)~\cite{crete2007blur}.} This metric, implemented in scikit-image as \verb|measure.blur_effect|, builds on the observation that when an image is blurred on purpose, the sharper the original image, the more it changes. The key idea of the blur estimation principle is to blur the initial image and analyze the neighboring pixels’ variation. The metric score varies between 0 (no blur) and 1 (maximum blur). 
    \item {\em Cumulative Probability Blur Distribution (CPBD)~\cite{narvekar2011no}.} This metric is motivated by a perceptual study that determined, for a given step edge,  under which combinations of contrast and width the edge is perceived as blurry. Given an image, CPBD is the percentage of edges that we do not perceive as blurry. Hence, a higher CPBD score corresponds to a sharper image.
    \item {\em Sharpness Index (SI)~\cite{blanchet2012explicit}.} This measure is based on the observation that, the sharper an image, the more it gets affected when perturbing the phase of its Fourier Transform with random noise. Intuitively, when an image is blurred, its high-frequency components are attenuated, and therefore the oscillations generated by the phase distortion are lower.
\end{itemize}
Because of their nature, these sharpness metrics provide complementary information that is very useful to assess the results. For instance, the blur strength captures well sharp edges but does not penalize high-frequency artifacts. On the other side, the sharpness index also favors sharp edges but penalizes textures or high-frequency artifacts.

When reporting quantitative results on tables, we highlight the \colorbox{Cyan1}{best} and \colorbox{Yellow1}{second-best} values for each metric.

\subsubsection{RealBlur Dataset~\cite{rim_2020_ECCV}}

\input{deblurringRealBlurAndLai}

\input{perceptual_metrics_results}

\subsubsection{K\"ohler's Dataset~\cite{kohler2012recording}}

\input{deblurringKohler}

\subsection{Impact of joint-training}

The kernels estimated by our Kernel Prediction Network can be directly plugged into the restoration network. Still, restored images will present several artifacts due to a mismatch between both separately trained networks. Jointly fine-tuning the KPN and the restoration network allows us to remove artifacts and produce high-quality results. This is illustrated in \cref{fig:ft_comparison}: the images in the  third column (restored before joint training) present a considerable number of artifacts, that are no longer visible in the images in the fifth column (restored after joint training). \cref{fig:ft_comparison} also displays the motion blur kernels predicted before and after joint training; the changes are subtle, and slightly more noticeable in those regions presenting artifacts in the non-fine-tuned restorations (third column).

\input{figure_kernels_before_and_after_ft.tex}

\subsection{Impact of training on \emph{SBDD}:} The experiments presented so far compare the results of {\em J-MKPD} with other approaches by running the models provided by the authors. Following the current trend, most of these methods are trained on the GoPro dataset. Since our method is blind for testing but requires ground truth kernels for training, we cannot effectively train on the GoPro dataset. To analyze the impact of training on {\em SBDD} on generalization, we selected one of the best-performing end-to-end deblurring networks (MIMO-UNet+~\cite{cho2021rethinking}), originally trained on GoPro, and retrained it on {\em SBDD}. Deblurring results on K\"{o}hler and RealBlur datasets are presented in \cref{tab:MIMO_our_dataset_Kohler_op1} and \cref{tab:MIMO_our_dataset_RealBlur_op1}. The first observation is that MIMO-UNet+ achieves higher generalization performance when training on {\em SBDD}, instead of on the GoPro dataset.  As for the comparison of MIMO-UNet+ and our model when both are trained on {\em SBDD}, MIMO-UNet+ performs slightly better in terms of PSNR and SSIM, while our method performs better in terms of LPIPS and in all the reference-free metrics. 
Qualitative results on both datasets are shown in \cref{fig:MIMO_vs_JMKPD_Kohler_better} and \cref{fig:MIMO_vs_JMKPD_RealBlur_better}. In general, when the motion structure of the scene can be well approximated by our spatially-varying motion blur model, our method outperforms MIMO-UNet+. On the RealBlur dataset, we found that the constraints imposed by our model are helpful to avoid artifacts on repetitive patterns, and also to achieve more precise restoration under mild blur conditions. 

\begin{table}[h]
    \setlength\tabcolsep{3 pt}
    \centering
    \caption{Results on Kohler dataset}
    \label{tab:MIMO_our_dataset_Kohler_op1}
    \begin{tabular}{lccccccc}
        \toprule 
        Method & Tr. Set &  PSRN & SSIM & LPIPS$\downarrow$ & BS$\downarrow$ & CPBD & SI \\
        \midrule
        MIMO-UNet+    & GoPro & 25.05 & 0.746 & 0.369 & 0.632  & 0.101 & 214 \\
        MIMO-UNet+    & \emph{SBDD} &  \textbf{28.72} & 0.830 & 0.272 & 0.566 & 0.147 & 741 \\
        J-MKPD    & \emph{SBDD} &28.65 & \textbf{0.832} & \textbf{0.250} & \textbf{0.559} & \textbf{0.178} & \textbf{939} \\
        \bottomrule 
    \end{tabular}
\end{table}
\begin{table}[h]
    \setlength\tabcolsep{3 pt}
    \centering
    \caption{Results on RealBlur dataset}
    \label{tab:MIMO_our_dataset_RealBlur_op1}    
    \begin{tabular}{lccccccc}
        \toprule
        Method & Tr. Set &  PSRN & SSIM & LPIPS$\downarrow$ & BS$\downarrow$ & CPBD & SI \\
        \midrule
        MIMO-UNet+    & GoPro & 27.64 & 0.836 & 0.199 & 0.459 & 0.237 & 1484 \\
        MIMO-UNet+    & \emph{SBDD} & \textbf{28.94} & \textbf{0.887} & 0.151 & 0.384 & 0.400 & 3290 \\
        JMKPD (180k) & \emph{SBDD} & 28.72 & 0.878 & \textbf{0.147} & \textbf{0.378} & \textbf{0.433} & \textbf{3323} \\
        \bottomrule 
    \end{tabular}
\end{table}

\input{figure_MIMO_vs_JMKPD_Kohler}

\input{figure_MIMO_vs_JMKPD_RealBlur}

%% file: deblurringRealBlurAndLai.tex
We evaluate our method following the  benchmark presented in \cite{rim_2020_ECCV}. Results for the distortion metrics and for the reference-free perceptual metrics are presented in \cref{tab:RealBlurDataset} and \cref{tab:perceptual_RealBlur}, respectively.  
Among the with-reference metrics, our method outperforms previous methods in terms of SSIM and performs second in terms of PSNR and LPIPS. Visual inspection of several restored images shown in \cref{fig:RealBlurComparison} reveals that, even when the outcomes of our method exhibit lower PNSR, they are clearly better and sharper in qualitative terms. We argue that these inconsistencies between PSNR and qualitative subjective evaluation are mainly caused by blur in the (supposedly sharp) target images. The reader can readily verify this by comparing the rightmost column (ground truth sharp images) with the first to rightmost column (our restorations) in \cref{fig:RealBlurComparison}. 

\input{figure_RealBlur}

\cref{tab:perceptual_RealBlur} also confirms that when evaluating the restorations' sharpness over the whole restorations from the RealBlur dataset, J-MKPD ranks first. 

Finally, to further evidence the impact of the blurred target images, the last row in \cref{tab:RealBlurDataset} shows that our method increases its performance on reference-based metrics when blurring the restored images with a Gaussian kernel of $\sigma = 0.5$. \cref{tab:perceptual_RealBlur} shows that all the perceptual reference-free metrics decrease their performance when restored images are blurred, as expected.

\begin{table}[h]
    \caption{Cross-dataset evaluation on the RealBlurDataset \cite{rim_2020_ECCV}.    \label{tab:RealBlurDataset}} %

     \centering
    \begin{tabular}{lccc}
    \toprule
    Method & PSNR & SSIM & LPIPS $\downarrow$ \\
    \midrule
    Nah\cite{Nah_2017_CVPR}  & 28.06 & 0.855 & 0.186 \\
    DMPHN \cite{Zhang_2019_CVPR}  & 27.78 & 0.841  & 0.197 \\
    DeblurGAN-v2 \cite{kupyn2019deblurgan}  & 28.70 & 0.866 & \colorbox{Cyan1}{0.139} \\
    MIMO-UNet \cite{cho2021rethinking} & 27.76  &  0.836  & 0.189 \\
    MIMO-UNet+ \cite{cho2021rethinking} & 27.64  &  0.836  & 0.199 \\
    NAFNet \cite{NAFNet} & 28.32  & 0.857 & 0.164 \\
    SRN \cite{tao2018scale} & 28.56 & 0.867 & 0.151 \\
    MPRNet \cite{Zamir2021MPRNet}  & 28.70 & 0.873  & 0.153\\
    Analysis-Synthesis \cite{kaufman2020deblurring}  & \colorbox{Yellow1}{28.78} & 0.864 & 0.197\\
    \hline
{J-MKPD} & 28.72 &  \colorbox{Yellow1}{0.878} & 0.147 \\

\hdashline
{J-MKPD (blurred)} & \colorbox{Cyan1}{29.08} &  \colorbox{Cyan1}{0.882}  & \colorbox{Yellow1}{0.146}  \\
    \bottomrule
    \end{tabular}
\end{table}

\begin{table}[ht]
    \caption{Reference-free perceptual metrics results on the RealBlur dataset~\cite{rim_2020_ECCV}.}
    \centering
    \begin{tabular}{lccc}
        \toprule
        Method &  BS(11)$\downarrow$ & CPBD $\uparrow$ & SI$\uparrow$   \\
        \midrule
        None (input image)     & 0.504 & 0.226 &  1039 \\
        DeblurGAN-v2 \cite{kupyn2019deblurgan}    & \colorbox{Yellow1}{0.397} & 0.329 & 1571 \\   
        MIMO-UNet \cite{cho2021rethinking}   & 0.453 &0.259 & 1497 \\
        MIMO-UNet+ \cite{cho2021rethinking}         & 0.459 & 0.237 & 1484 \\
        NAFNet \cite{NAFNet}        & 0.428 & 0.291 & 1592 \\
      SRN \cite{tao2018scale} & 0.413& 0.275 &  1850    \\
      MPRNet \cite{Zamir2021MPRNet}    & 0.421 & 0.288 & 1748  \\   
    Analysis-Synthesis \cite{kaufman2020deblurring}  & 0.465 & 0.219 & 2084  \\
    \hline
 {J-MKPD}            & \colorbox{Cyan1}{0.378} & \colorbox{Cyan1}{0.433} & \colorbox{Cyan1}{3323}  \\
 \hdashline
 {J-MKPD (blurred)} & 0.399 & \colorbox{Yellow1}{0.338} & \colorbox{Yellow1}{2195}  \\
     \midrule
    Ground truth    & 0.329 & 0.477 & 1926  \\
     \bottomrule
    \end{tabular}
    \label{tab:perceptual_RealBlur}
\end{table}

\input{figure_Lai}

\subsubsection{Lai's Dataset~\cite{lai2016comparative}} This is another standard benchmark that contains real blurry images in challenging situations, such as variable depth scenes and highly saturated images. The dataset has no corresponding ground truth, so it only enables visual comparison and evaluation with reference-free metrics. A comparison of different methods on images from Lai's dataset is presented in \cref{fig:LaiDeepLearning} and the Supplementary Material. The blur strength~\cite{crete2007blur} and the sharpness index~\cite{blanchet2012explicit} for the restored images at full and half resolution (Lai(HS)) are shown in \cref{tab:blur_strength} and \cref{tab:sharpness_index}, respectively\footnote{Since the motion blur in several images from Lai exceeds our method's maximum manageable kernel size, we halve the resolution to evaluate how our method performs when this design condition is met for more images.}.

%% file: figure_RealBlur.tex
\begin{figure*}[ht]
  \centering
  \scriptsize
\setlength{\tabcolsep}{2pt}
  \begin{tabular}{*{6}{c}}
            Blurry & Ana.-Synth.
          \cite{kaufman2020deblurring} & SRN \cite{tao2018scale} & MPRNet \cite{Zamir2021MPRNet} & J-MKPD & Ground truth \\
    \includegraphics[trim=250 370 200 200, clip, width=0.15\textwidth]{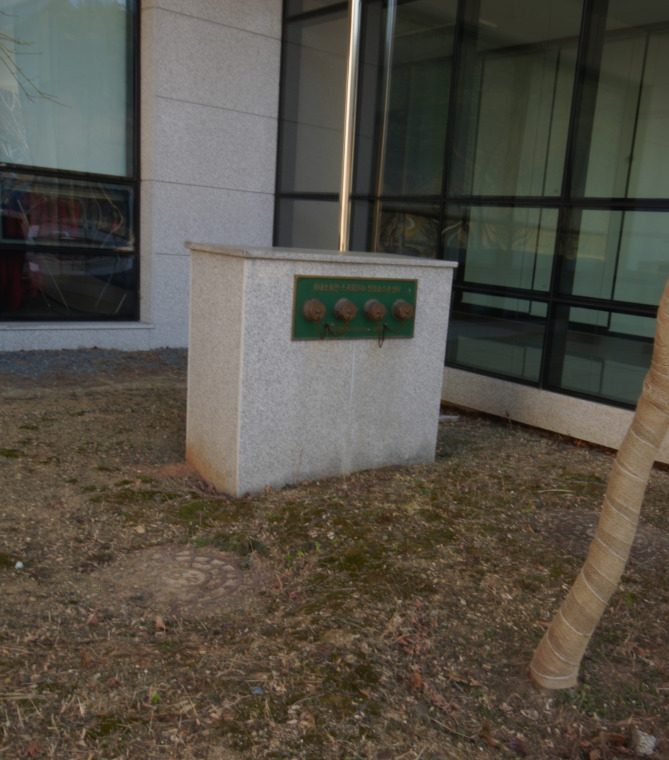}   &
      \includegraphics[trim=250 370 200 200, clip,width=0.15\textwidth]{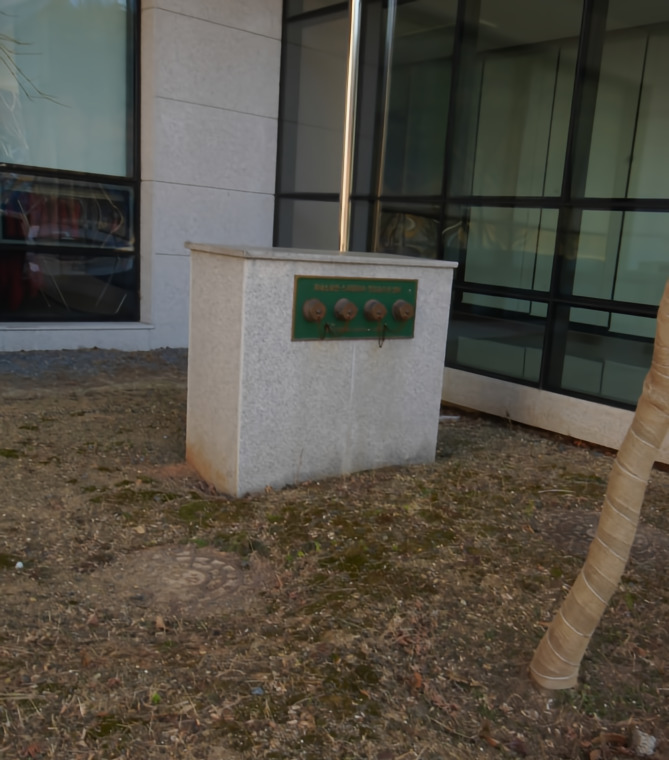}   &
   \includegraphics[trim=250 370 200 200, clip,width=0.15\textwidth]{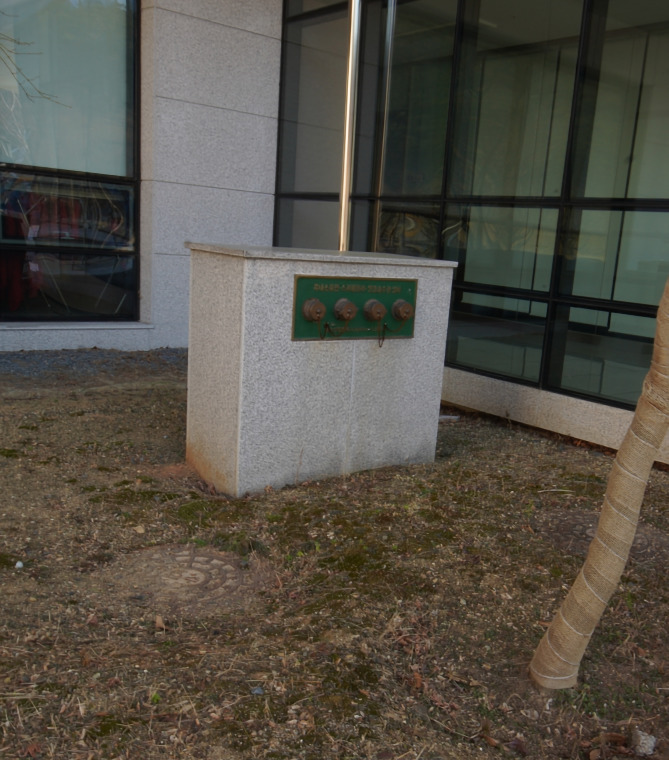}  &
    \includegraphics[trim=250 370 200 200, clip,width=0.15\textwidth]{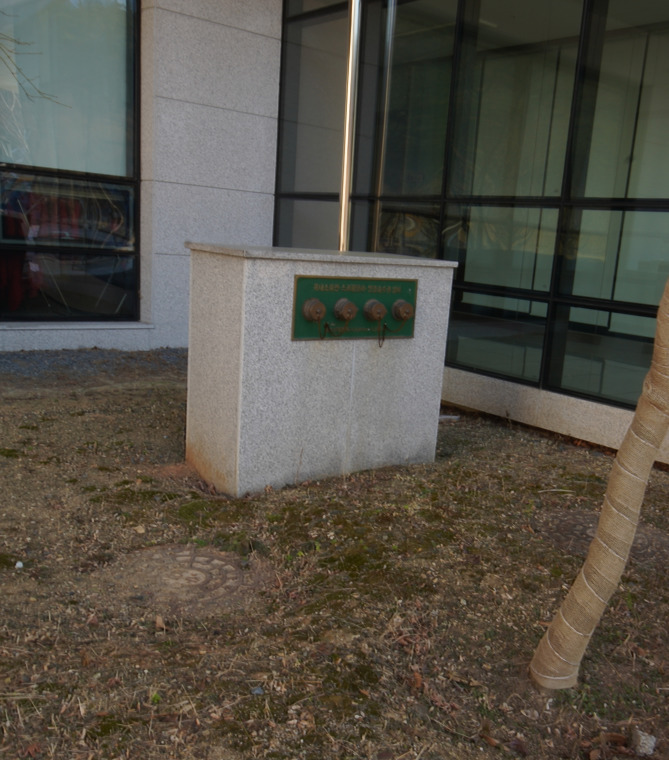}   &
      \includegraphics[trim=250 370 200 200, clip,width=0.15\textwidth]{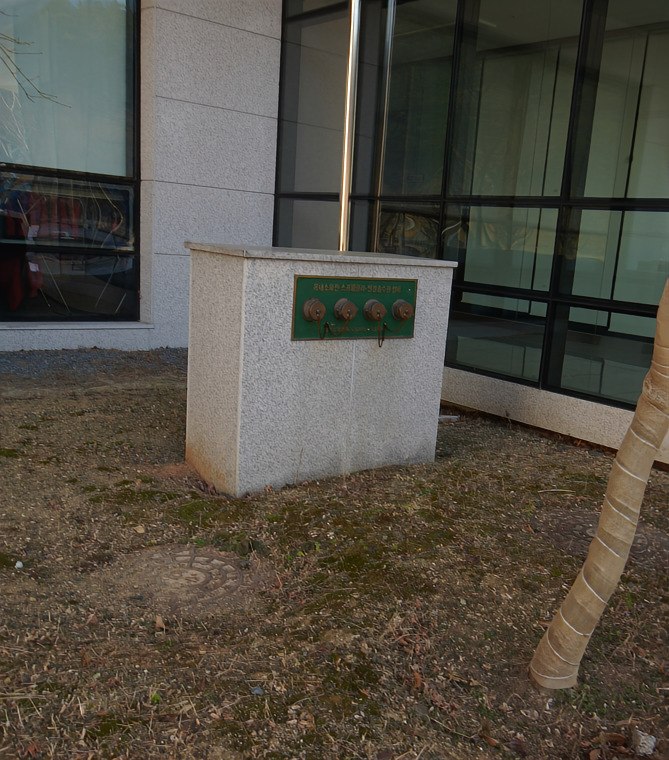} &  
        \includegraphics[trim=250 370 200 200, clip,width=0.15\textwidth]{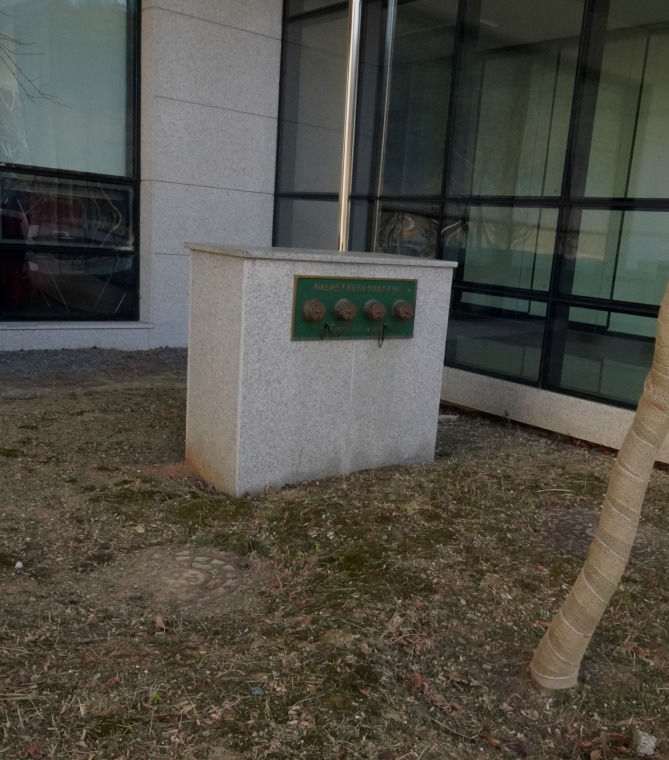}\\
    30.35 dB  &   30.94 dB & 31.08 dB & 31.26 dB & 28.81 dB & \\
    \includegraphics[trim=20 220 350 320, clip, width=0.15\textwidth]{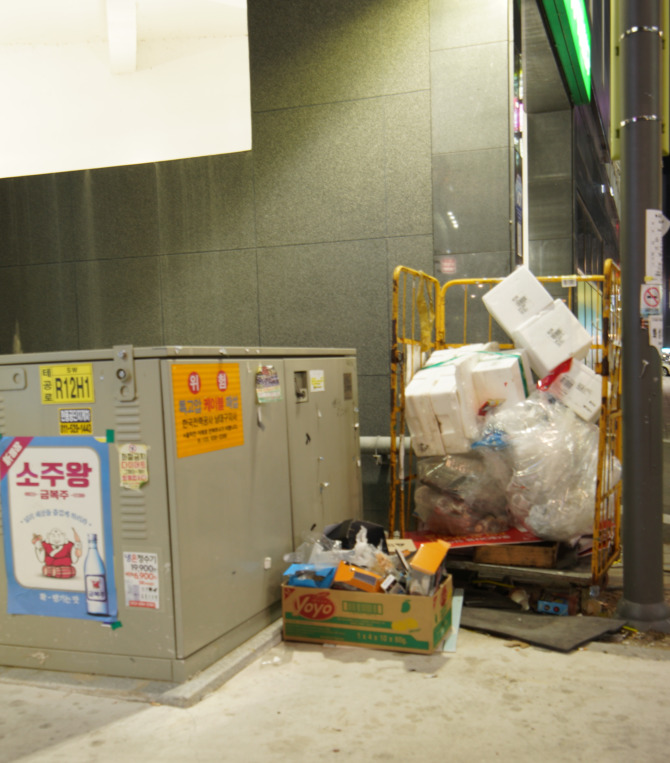}   &
      \includegraphics[trim=20 220 350 320, clip,width=0.15\textwidth]{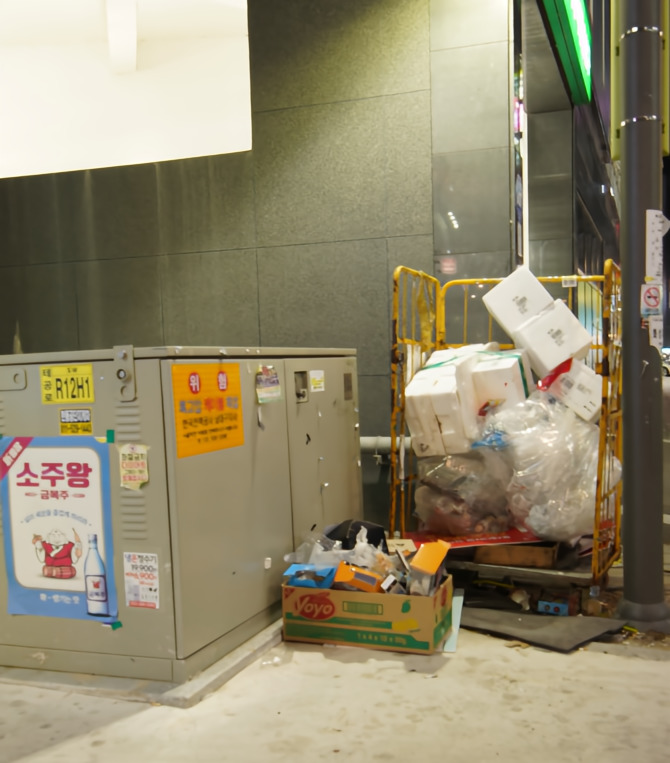}   &
   \includegraphics[trim=20 220 350 320, clip,width=0.15\textwidth]{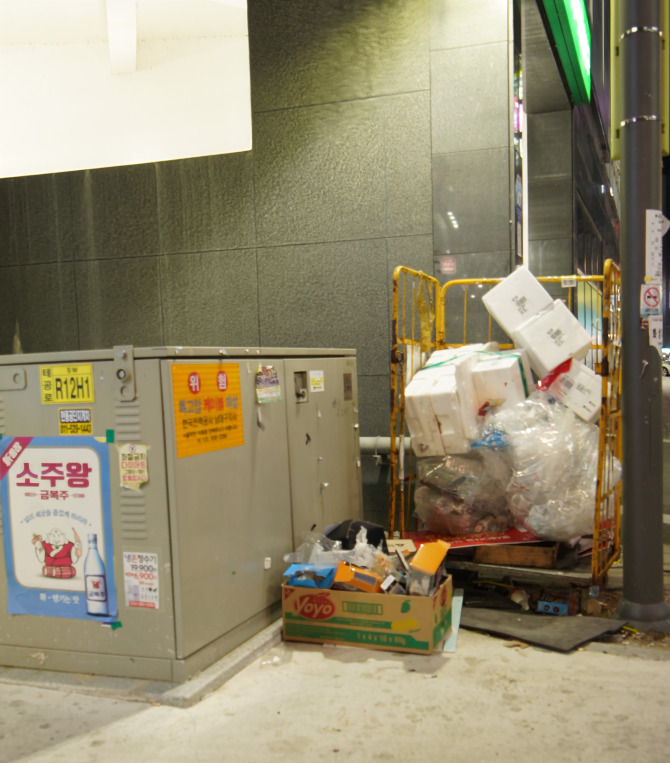}  &
    \includegraphics[trim=20 220 350 320, clip,width=0.15\textwidth]{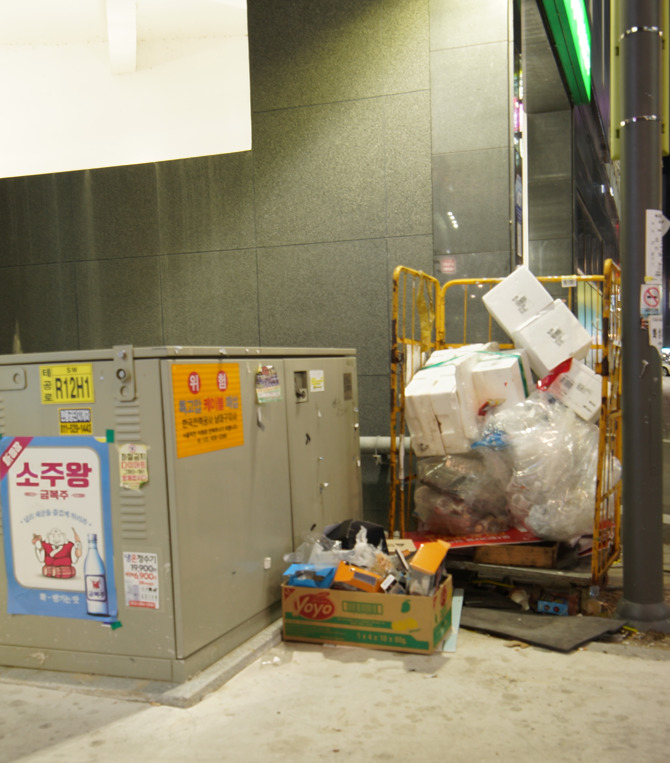}   &
      \includegraphics[trim=20 220 350 320, clip,width=0.15\textwidth]{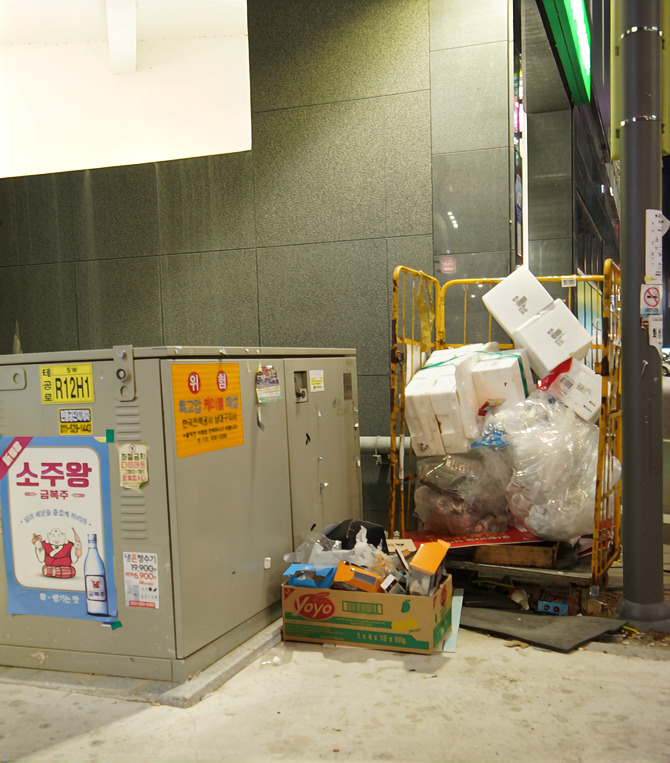} &  
        \includegraphics[trim=20 220 350 320, clip,width=0.15\textwidth]{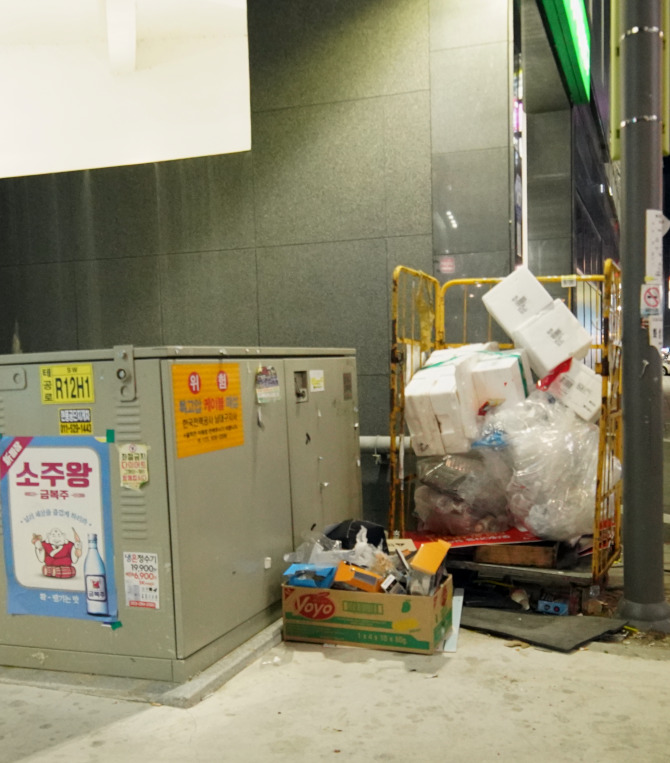}\\
29.64 dB &  30.11 dB & 29.9 dB & 30.14 dB & 27.26 dB & \\
    \includegraphics[trim=0 170 350 320, clip, width=0.15\textwidth]{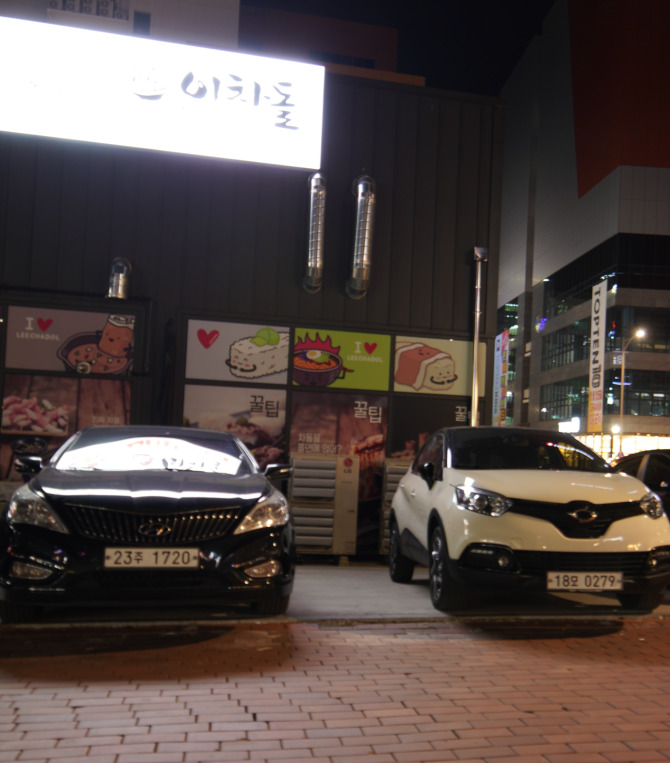}   &
      \includegraphics[trim=0 170 350 320, clip,width=0.15\textwidth]{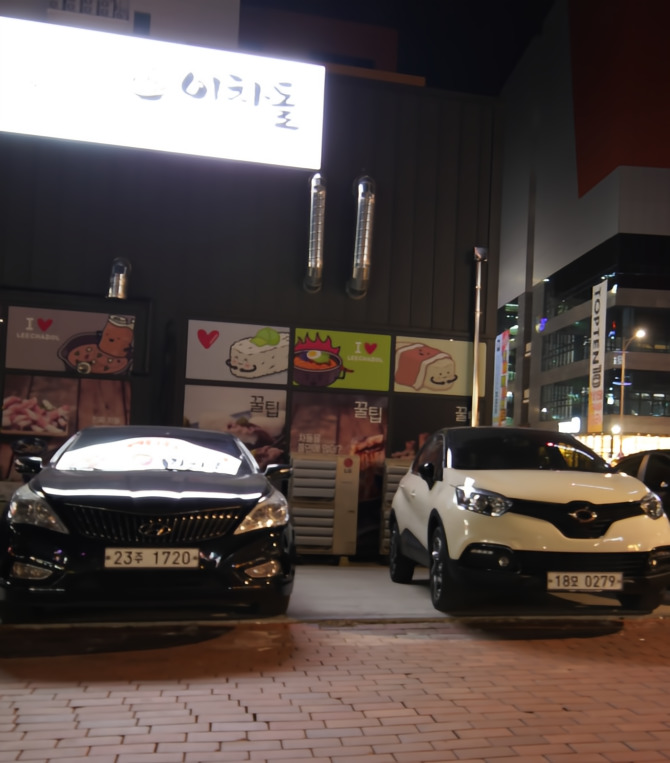}   &
   \includegraphics[trim=0 170 350 320, clip,width=0.15\textwidth]{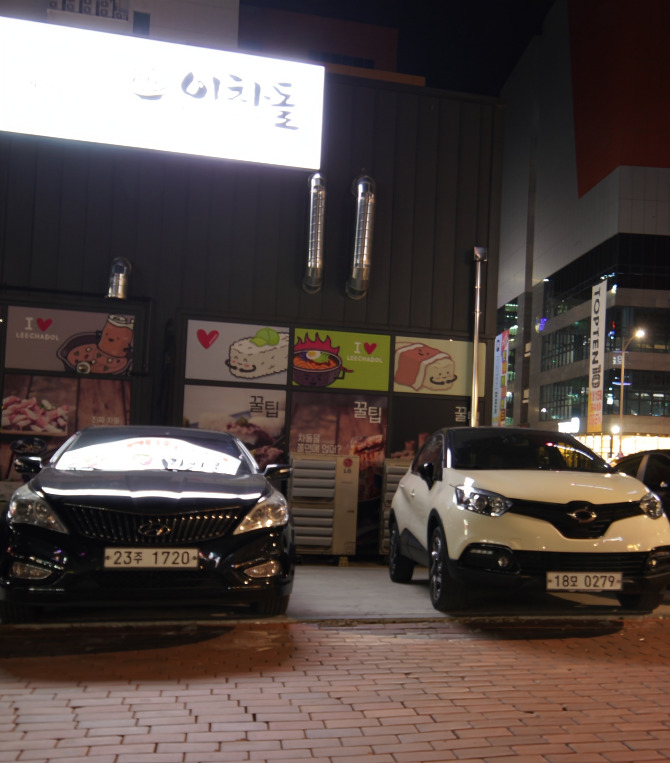}  &
    \includegraphics[trim=0 170 350 320, clip,width=0.15\textwidth]{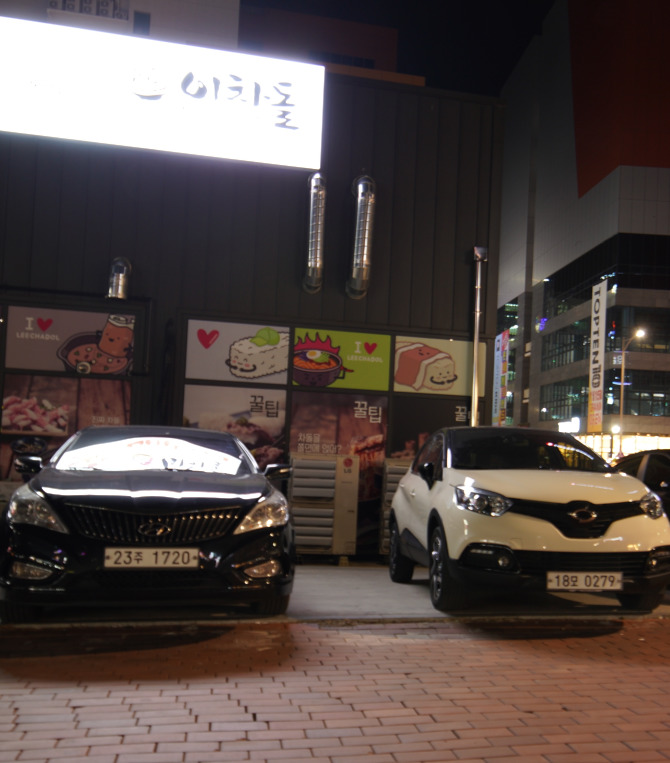}   &
      \includegraphics[trim=0 170 350 320, clip,width=0.15\textwidth]{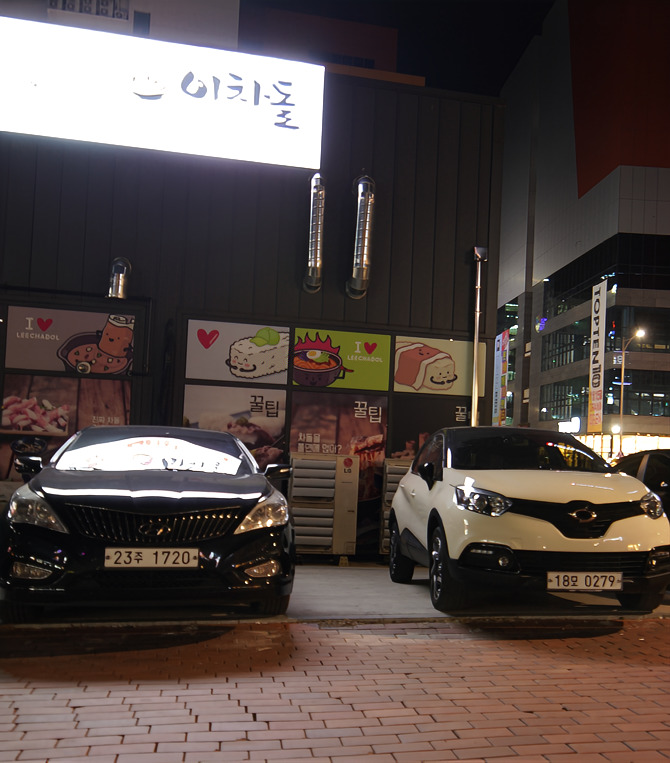} &  
        \includegraphics[trim=0 170 350 320, clip,width=0.15\textwidth]{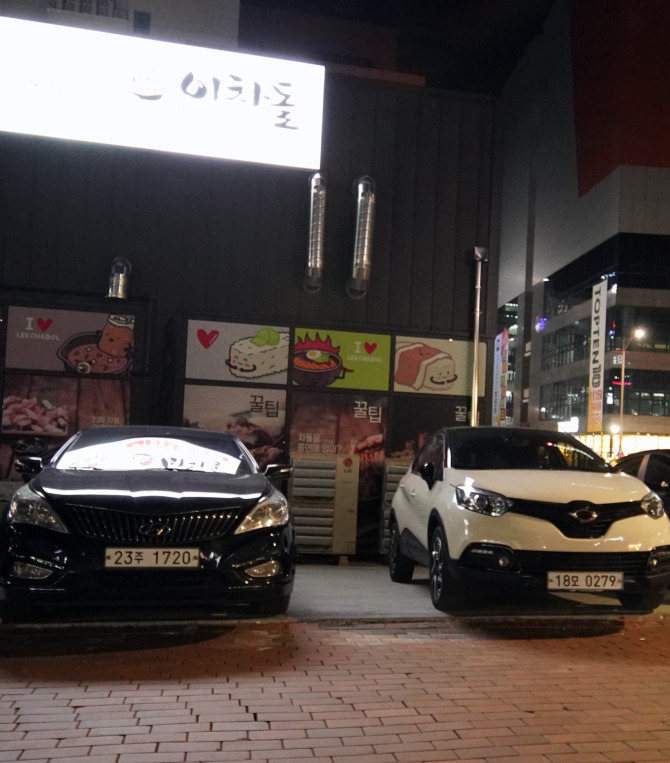}\\
30.25 dB & 30.84 dB & 30.76 dB & 30.83 dB & 28.61 dB & \\
    \includegraphics[trim=200 350 100 120, clip, width=0.15\textwidth]{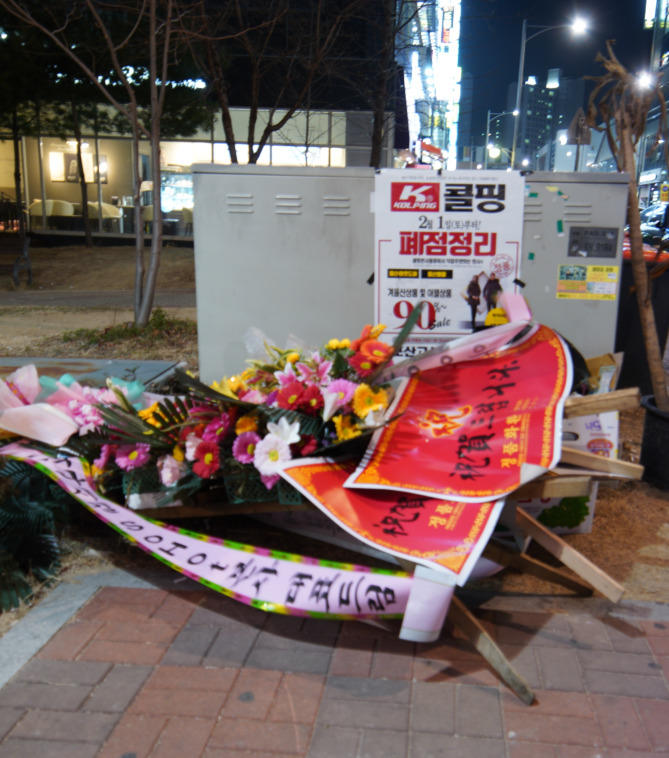}   &
      \includegraphics[trim=200 350 100 120, clip,width=0.15\textwidth]{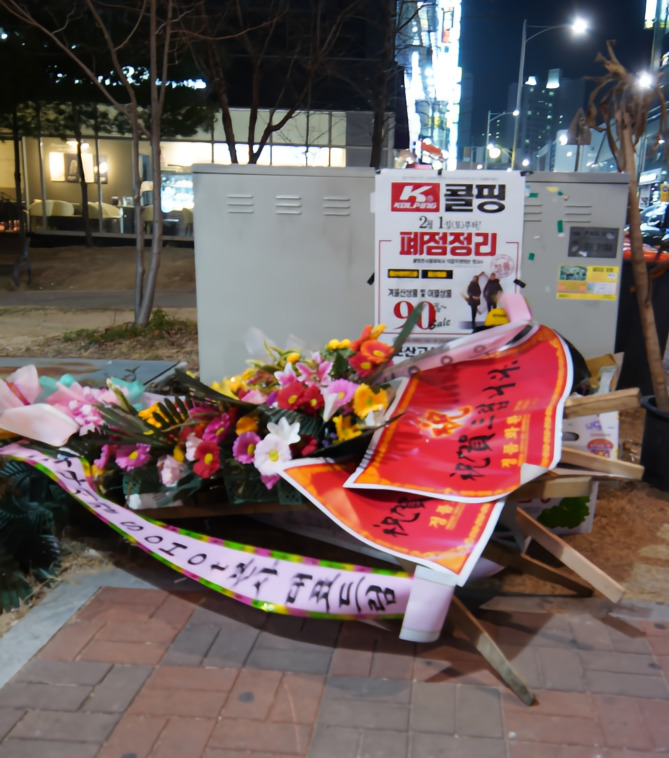}   &
   \includegraphics[trim=200 350 100 120, clip,width=0.15\textwidth]{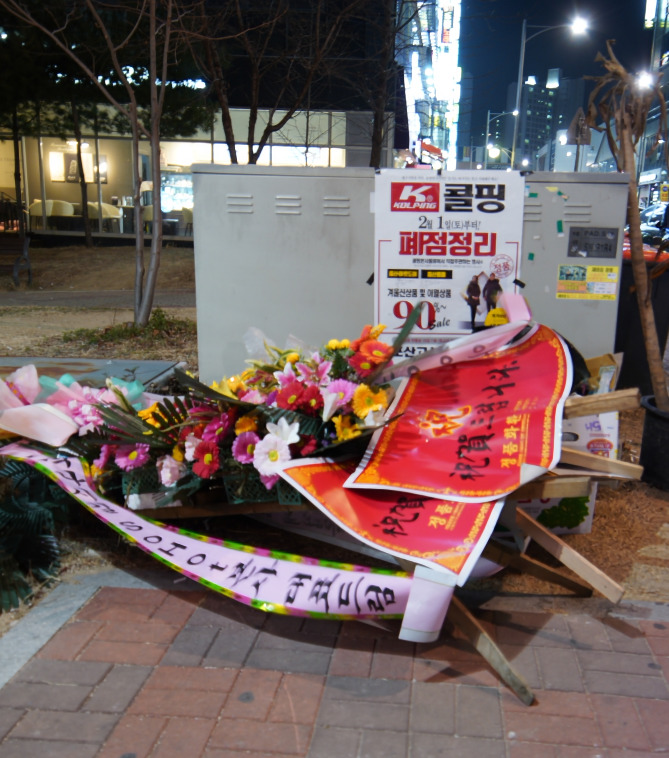}  &
    \includegraphics[trim=200 350 100 120, clip,width=0.15\textwidth]{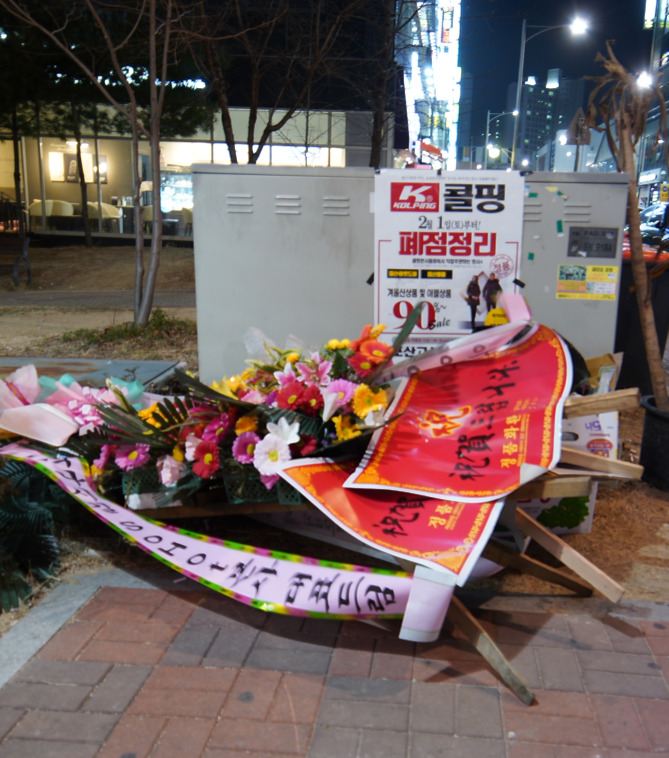}   &
      \includegraphics[trim=200 350 100 120, clip,width=0.15\textwidth]{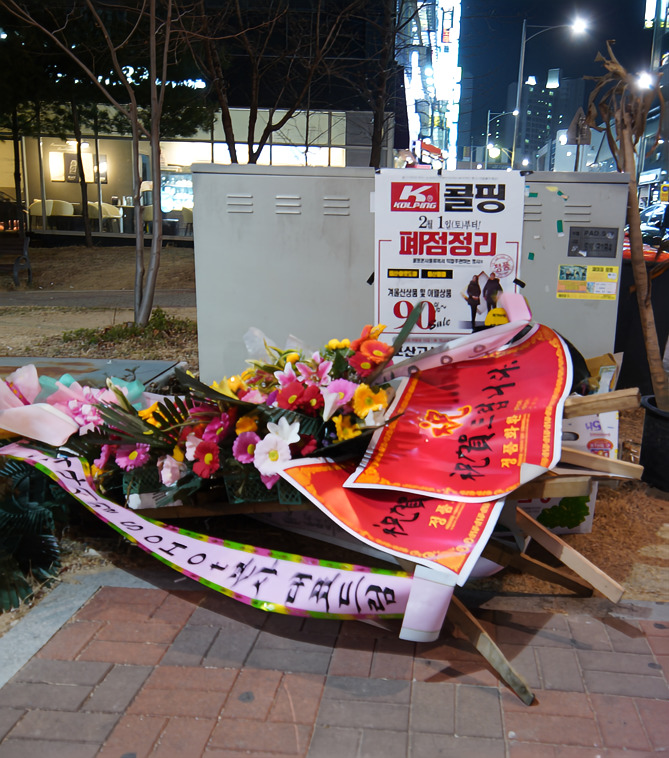} &  
        \includegraphics[trim=200 350 100 120, clip,width=0.15\textwidth]{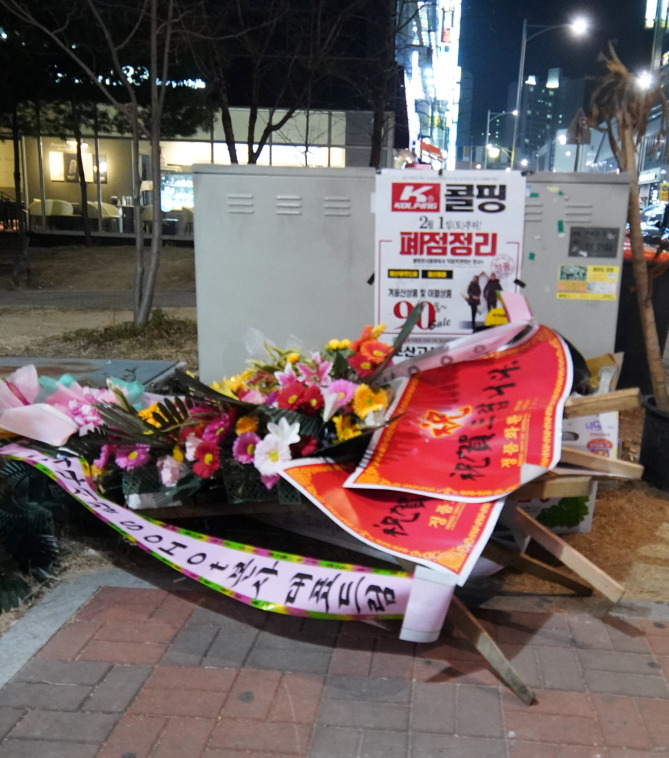}\\
28.47 dB & 28.64 dB & 28.26 dB & 28.57 dB & 26.01 dB & \\
    \includegraphics[trim=200 200 50 240, clip, width=0.15\textwidth]{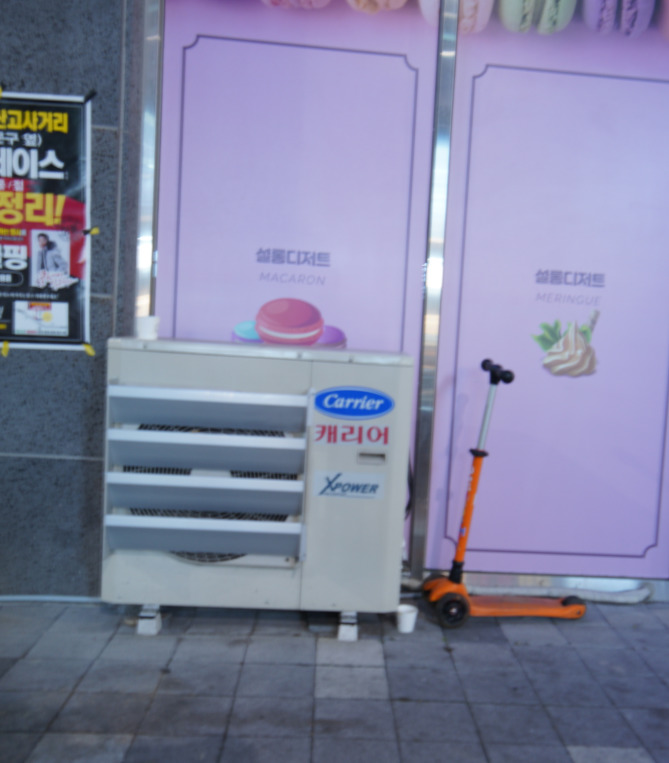}   &
      \includegraphics[trim=200 200 50 240, clip,width=0.15\textwidth]{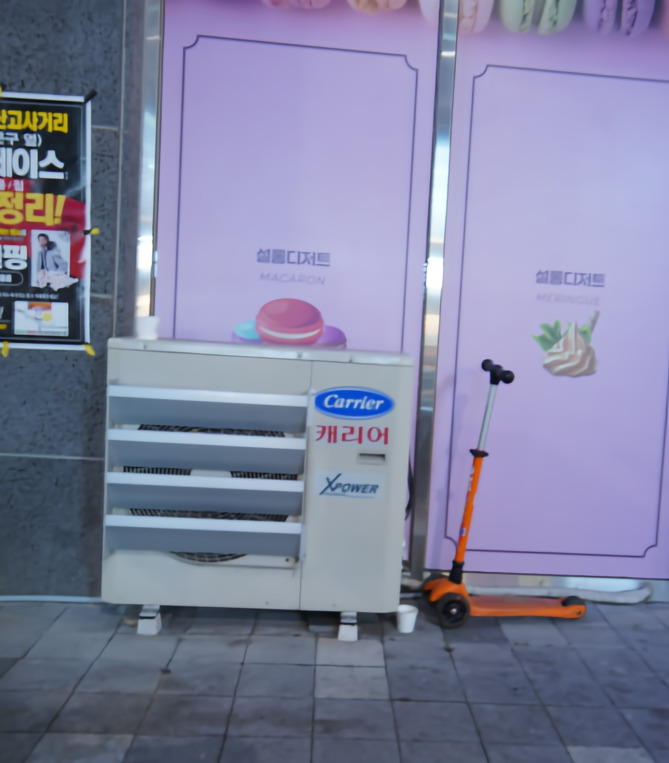}   &
   \includegraphics[trim=200 200 50 240, clip,width=0.15\textwidth]{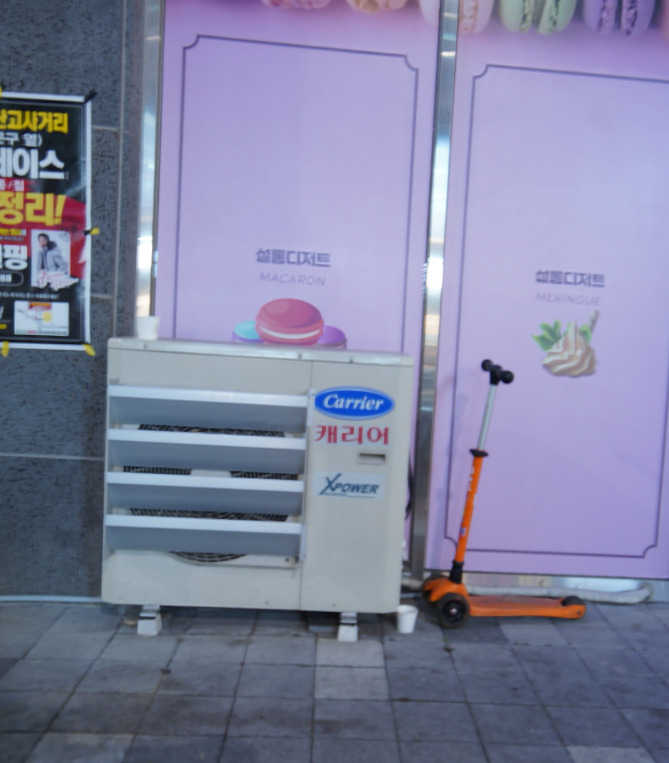}  &
    \includegraphics[trim=200 200 50 240, clip,width=0.15\textwidth]{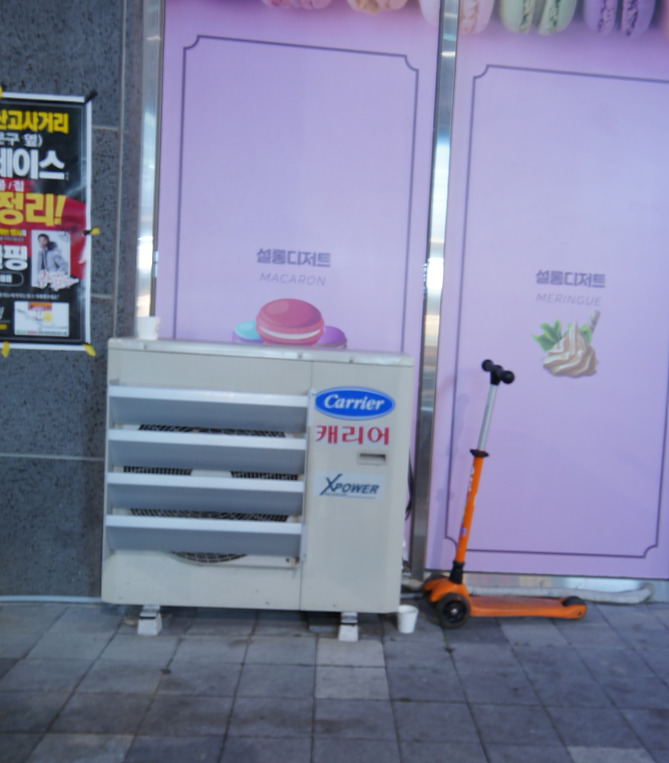}   &
      \includegraphics[trim=200 200 50 240, clip,width=0.15\textwidth]{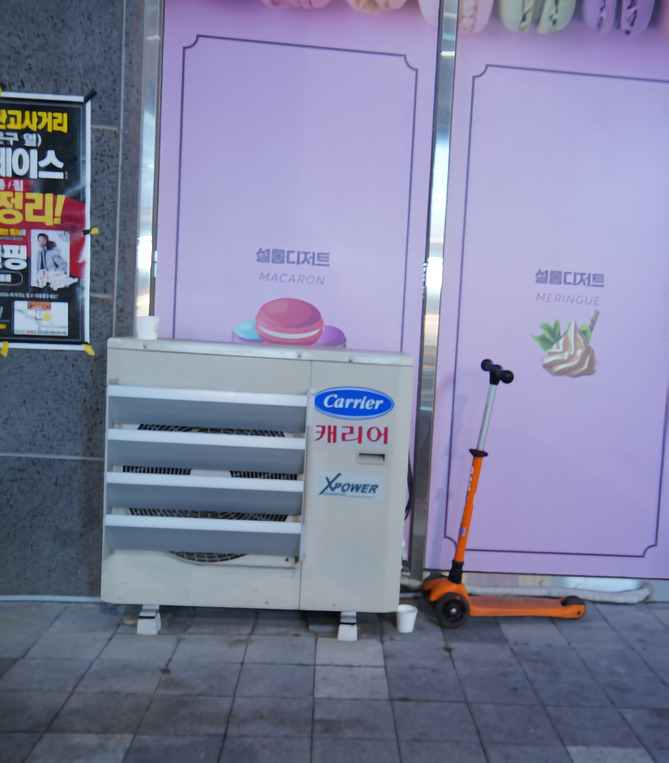} &  
        \includegraphics[trim=200 200 50 240, clip,width=0.15\textwidth]{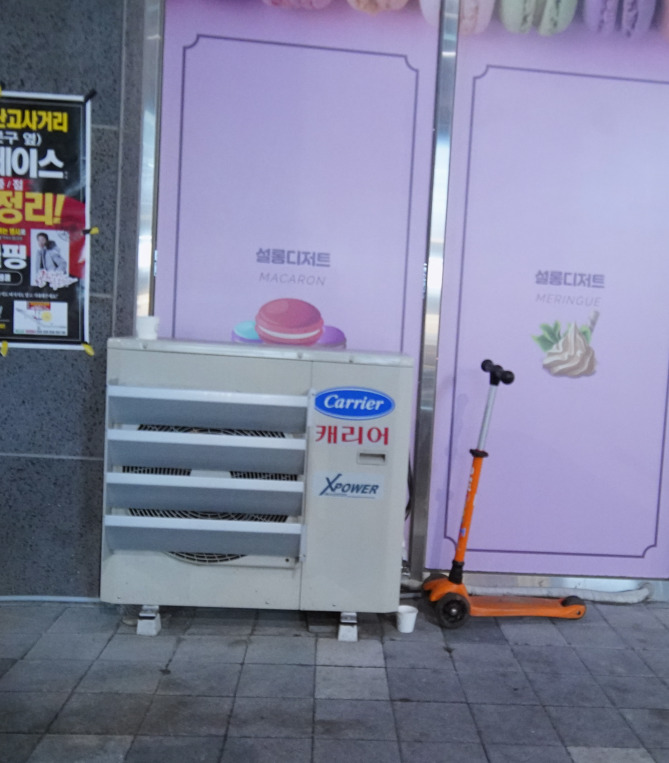}\\
      30.08 dB  &  31.17 dB & 31.03 dB & 31.43 dB & 29.97 dB & \\
    \includegraphics[trim=200 200 50 250, clip, width=0.15\textwidth]{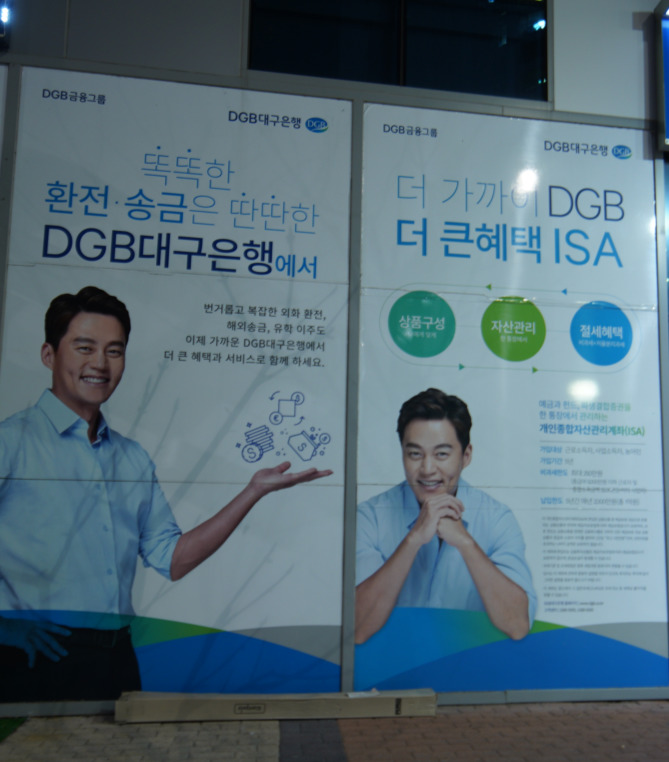}   &
      \includegraphics[trim=200 200 50 250, clip,width=0.15\textwidth]{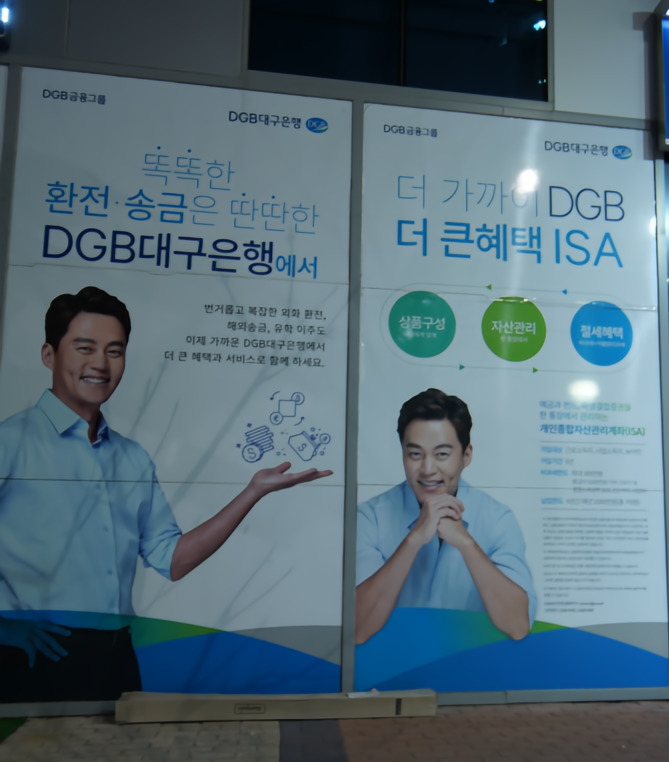}   &
   \includegraphics[trim=200 200 50 250, clip,width=0.15\textwidth]{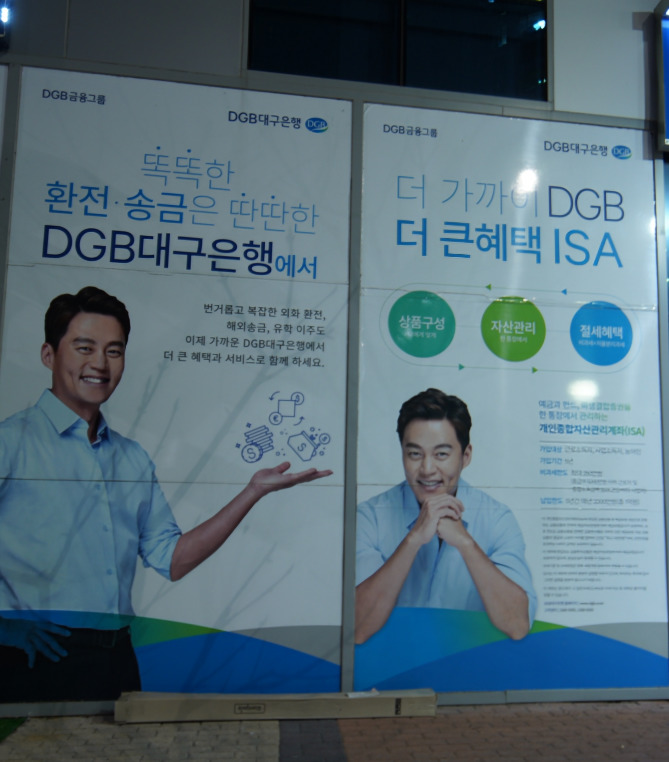}  &
    \includegraphics[trim=200 200 50 250, clip,width=0.15\textwidth]{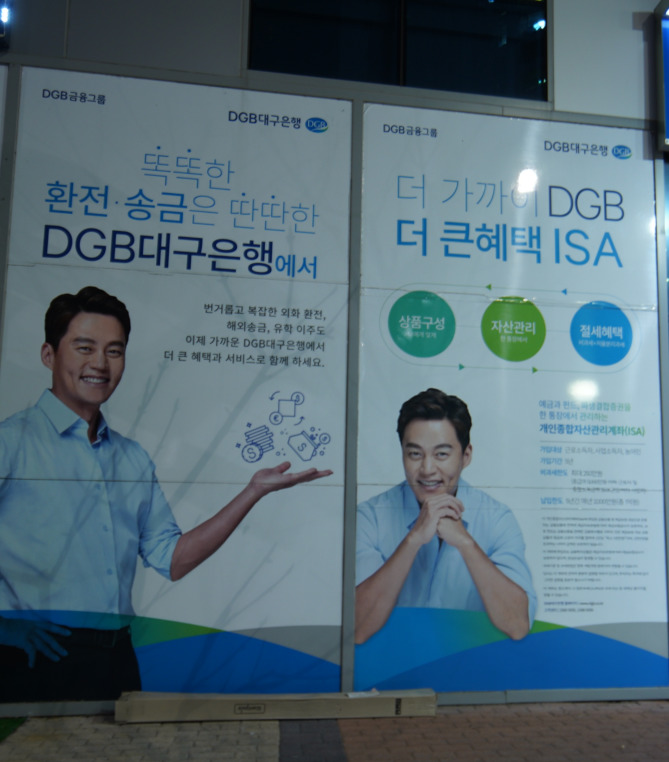}   &
      \includegraphics[trim=200 200 50 250, clip,width=0.15\textwidth]{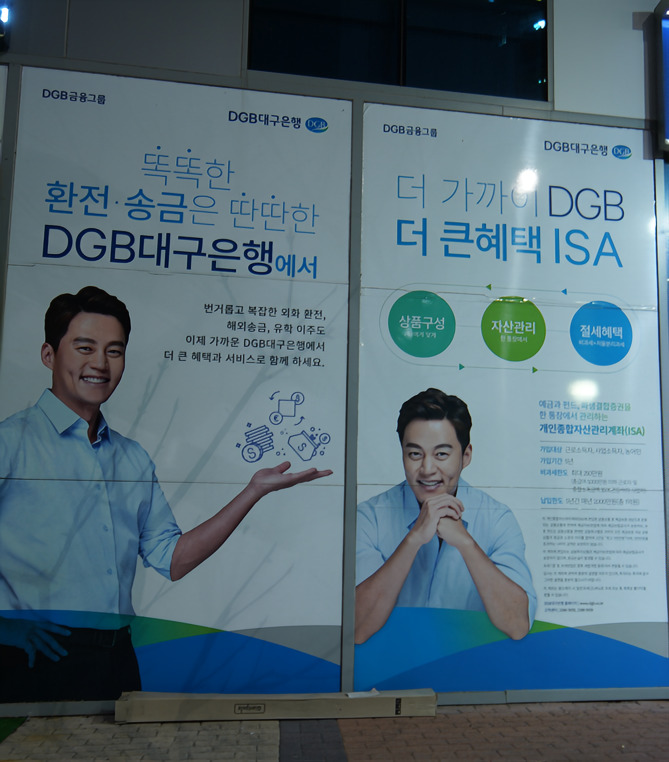} &  
        \includegraphics[trim=200 200 50 250, clip,width=0.15\textwidth]{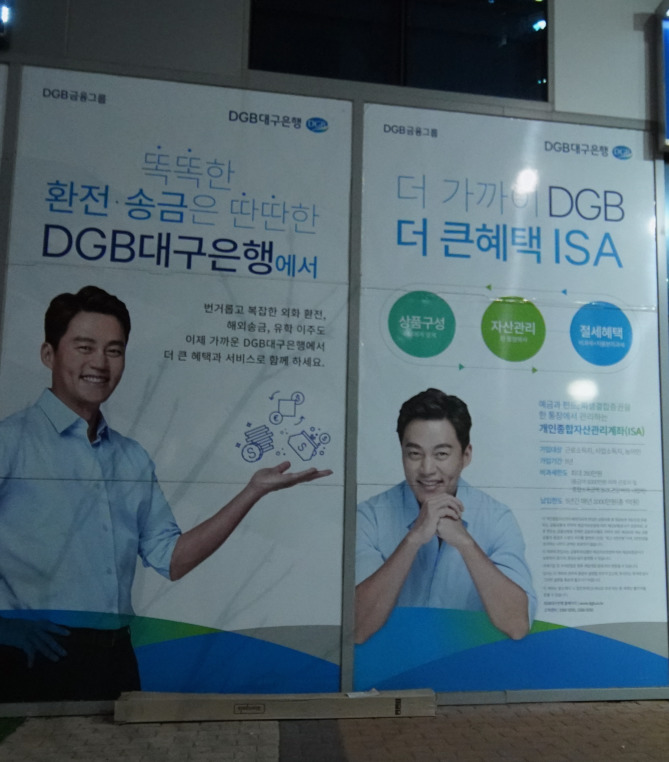}\\
29.20 dB &  32.22 dB & 32.17 dB & 32.42 dB & 29.53 dB & \\
    \includegraphics[trim=250 150 50 320, clip, width=0.15\textwidth]{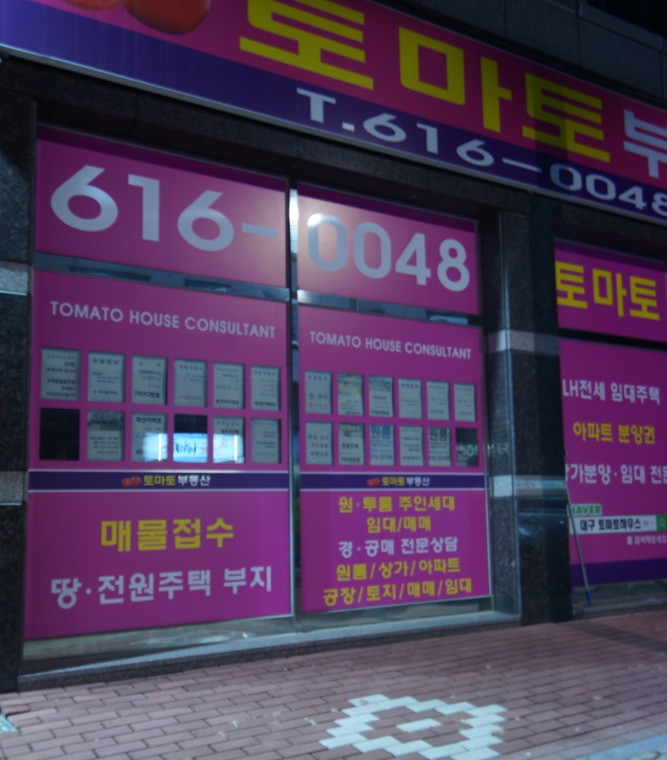}   &
      \includegraphics[trim=250 150 50 320, clip,width=0.15\textwidth]{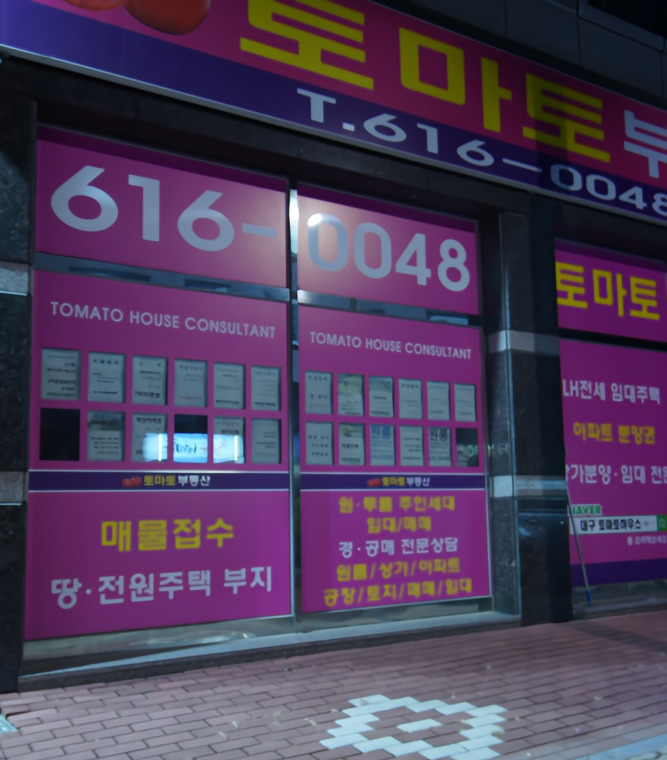}   &
   \includegraphics[trim=250 150 50 320, clip,width=0.15\textwidth]{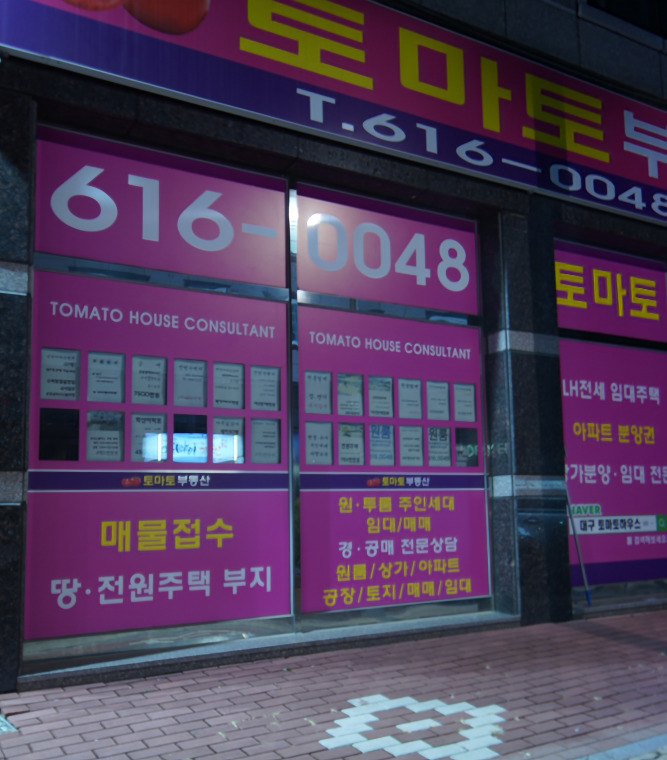}  &
    \includegraphics[trim=250 150 50 320, clip,width=0.15\textwidth]{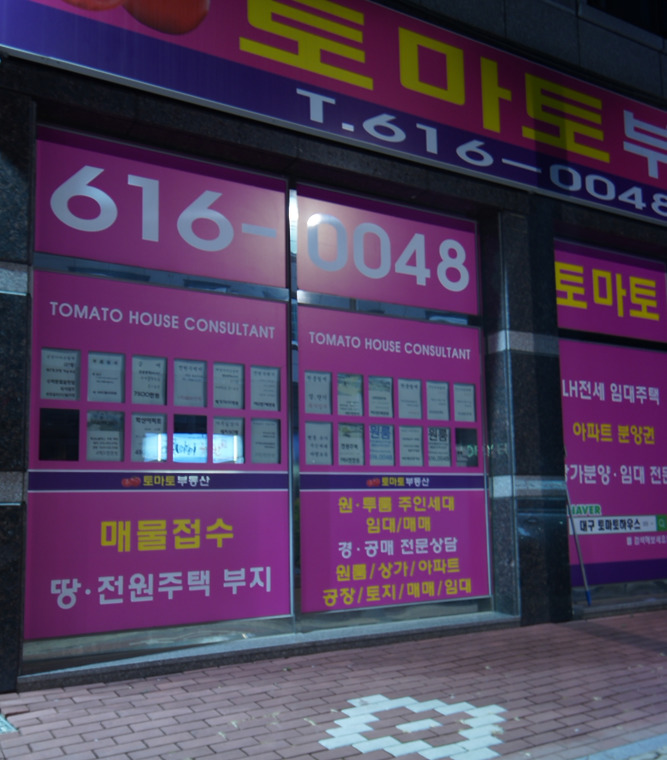}   &
      \includegraphics[trim=250 150 50 320, clip,width=0.15\textwidth]{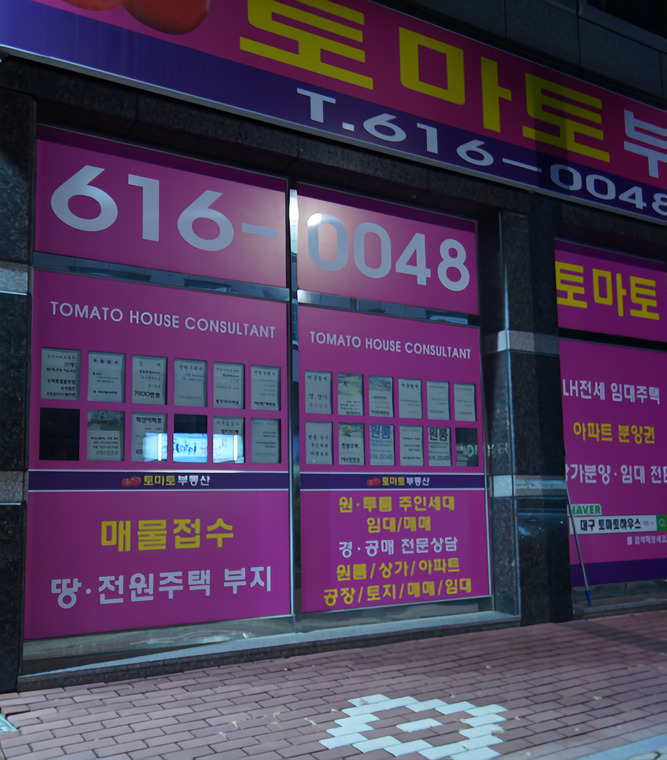} &  
        \includegraphics[trim=250 150 50 320, clip,width=0.15\textwidth]{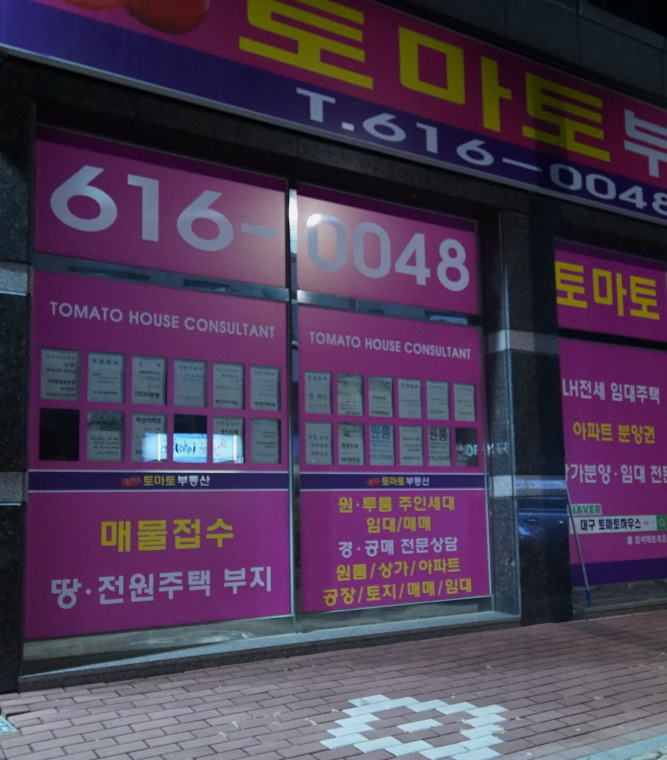}\\
29.24 dB &  30.61 dB & 30.52 dB & 30.64 dB & 28.91 dB & \\
  \end{tabular}
  \caption{\textbf{Comparison of deblurring on the RealBlur dataset~\cite{rim_2020_ECCV}.} Our method produces perceptually better images on photos with mild motion blur than the others.  High-frequency content, such as letters and textures are better appreciated in our deblurring result than in the ground-truth image provided by the dataset. As the ground truth is slightly blurred, the PSNR does not correlate well with the quality of the restorations. Best viewed in electronic format. }
  \label{fig:RealBlurComparison}
\end{figure*}

%% file: figure_Lai.tex
\begin{figure*}[ht]

    \setlength\tabcolsep{1pt} %
    \centering
    \scriptsize
\begin{tabular}{cccccc}

Blurry & Analysis-Synthesis \cite{kaufman2020deblurring} & SRN \cite{tao2018scale}  & DGAN-v2 \cite{kupyn2019deblurgan} & MPRNet \cite{Zamir2021MPRNet} & J-MKPD \\

\includegraphics[width=0.16\textwidth]{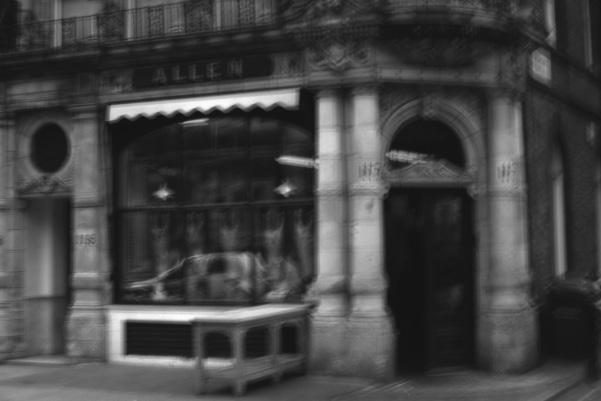}   &
\includegraphics[width=0.16\textwidth]{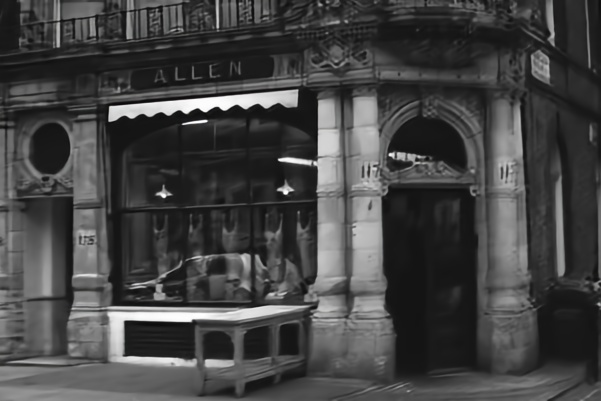} & 
\includegraphics[width=0.16\textwidth]{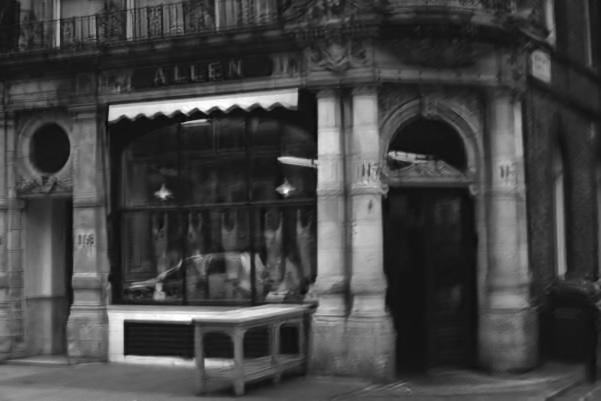}  &
\includegraphics[width=0.16\textwidth]{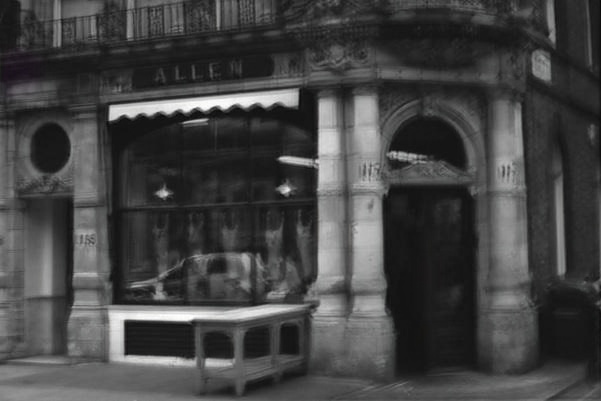}  &
\includegraphics[width=0.16\textwidth]{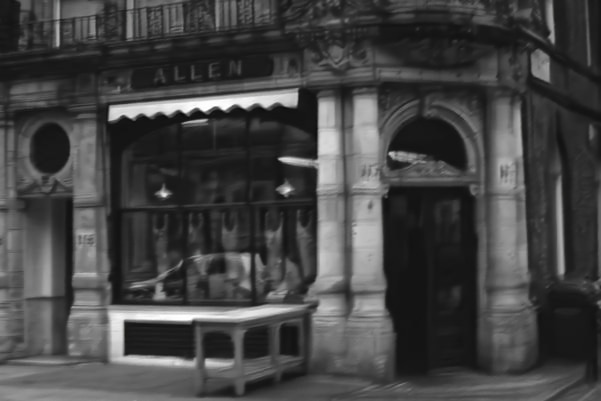}  &
\includegraphics[width=0.16\textwidth]{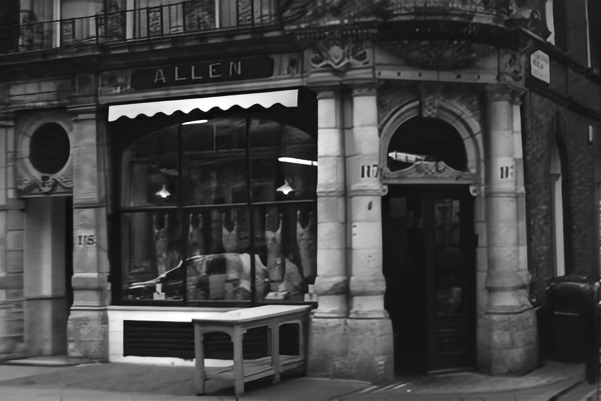}\\ 
\includegraphics[trim=50 140 450 220, clip,width=0.075\textwidth]{imgs/Lai/Blurry/butchershop.jpg}  
\includegraphics[trim=325 210 175 150, clip,width=0.075\textwidth]{imgs/Lai/Blurry/butchershop.jpg}  &
\includegraphics[trim=50 140 450 220, clip,width=0.075\textwidth]{imgs/Lai/AnaSyn/butchershop_deblurred.jpg} 
\includegraphics[trim=325 210 175 150, clip,width=0.075\textwidth]{imgs/Lai/AnaSyn/butchershop_deblurred.jpg}  &
\includegraphics[trim=50 140 450 220, clip,width=0.075\textwidth]{imgs/Lai/SRN/butchershop.jpg}  
\includegraphics[trim=325 210 175 150, clip,width=0.075\textwidth]{imgs/Lai/SRN/butchershop.jpg}  &
\includegraphics[trim=50 140 450 220, clip,width=0.075\textwidth]{imgs/Lai/DeblurGAN_V2/butchershop.jpg}
\includegraphics[trim=325 210 175 150, clip,width=0.075\textwidth]{imgs/Lai/DeblurGAN_V2/butchershop.jpg} &
\includegraphics[trim=50 140 450 220, clip,width=0.075\textwidth]{imgs/Lai/MPRNet/butchershop.jpg}
\includegraphics[trim=325 210 175 150, clip,width=0.075\textwidth]{imgs/Lai/MPRNet/butchershop.jpg} &
\includegraphics[trim=50 140 450 220, clip,width=0.075\textwidth]{imgs/Lai/COCO900_restL2_aug_all_loss_gf1_80k/butchershop_NIMBUSR.jpg}
\includegraphics[trim=325 210 175 150, clip,width=0.075\textwidth]{imgs/Lai/COCO900_restL2_aug_all_loss_gf1_80k/butchershop_NIMBUSR.jpg}\\

\includegraphics[width=0.160\textwidth]{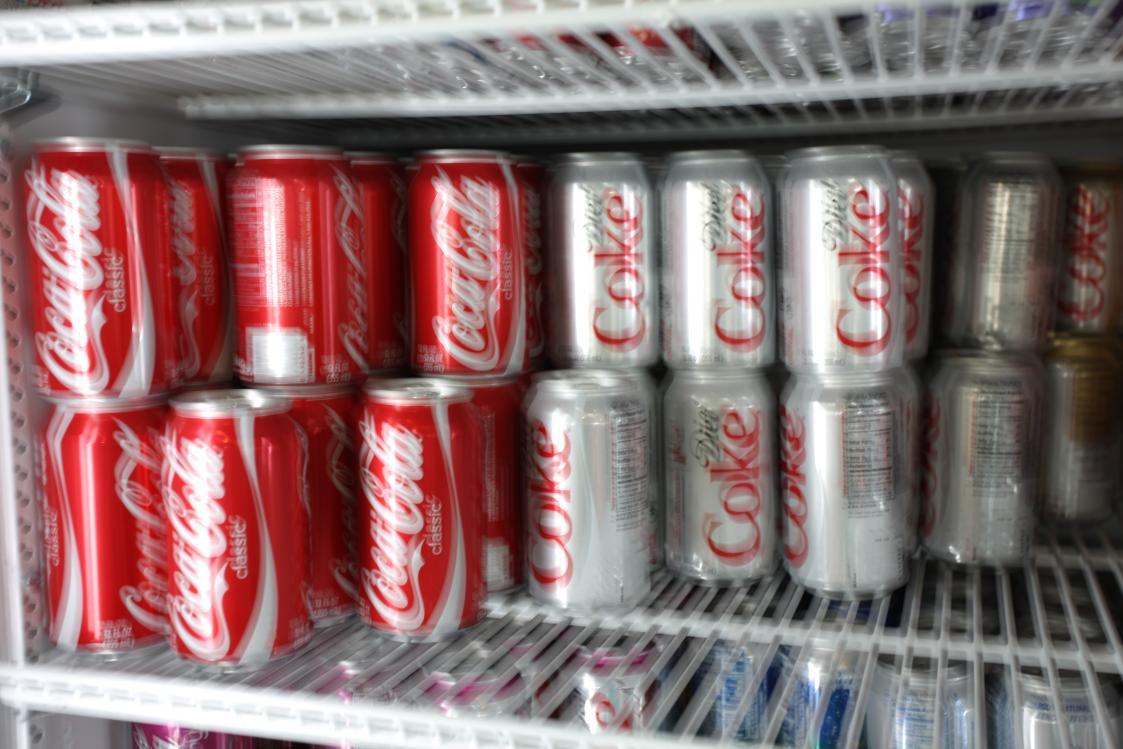}   &
\includegraphics[width=0.160\textwidth]{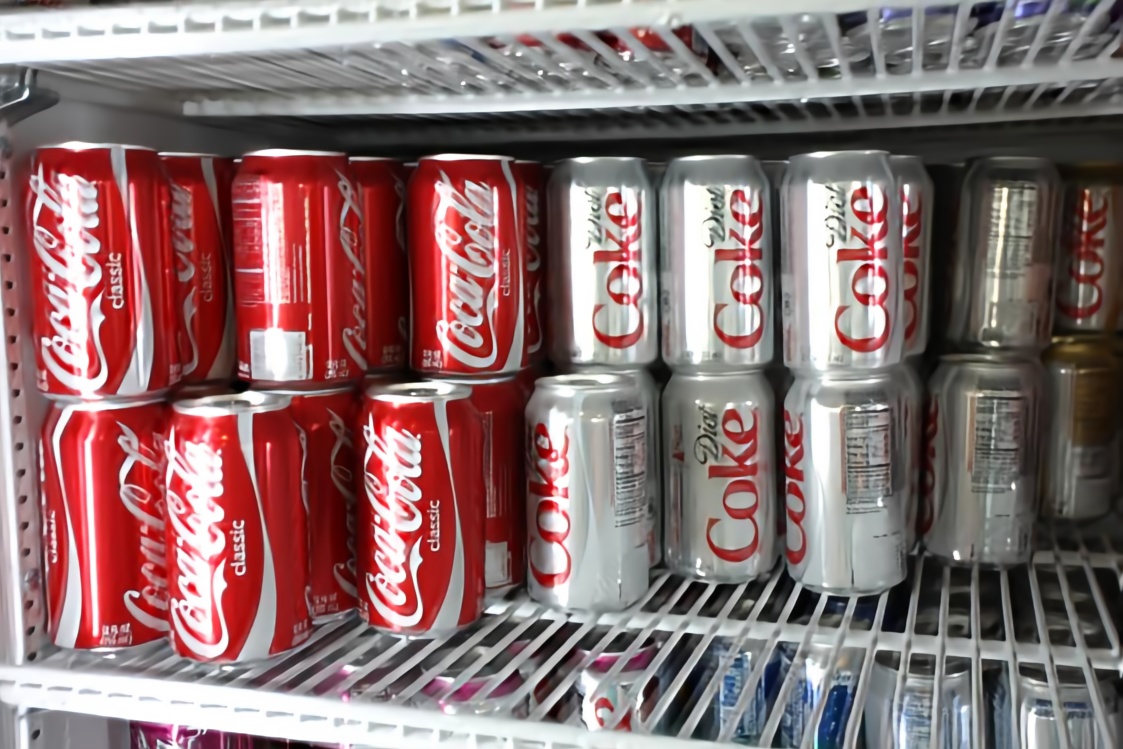} & 
\includegraphics[width=0.160\textwidth]{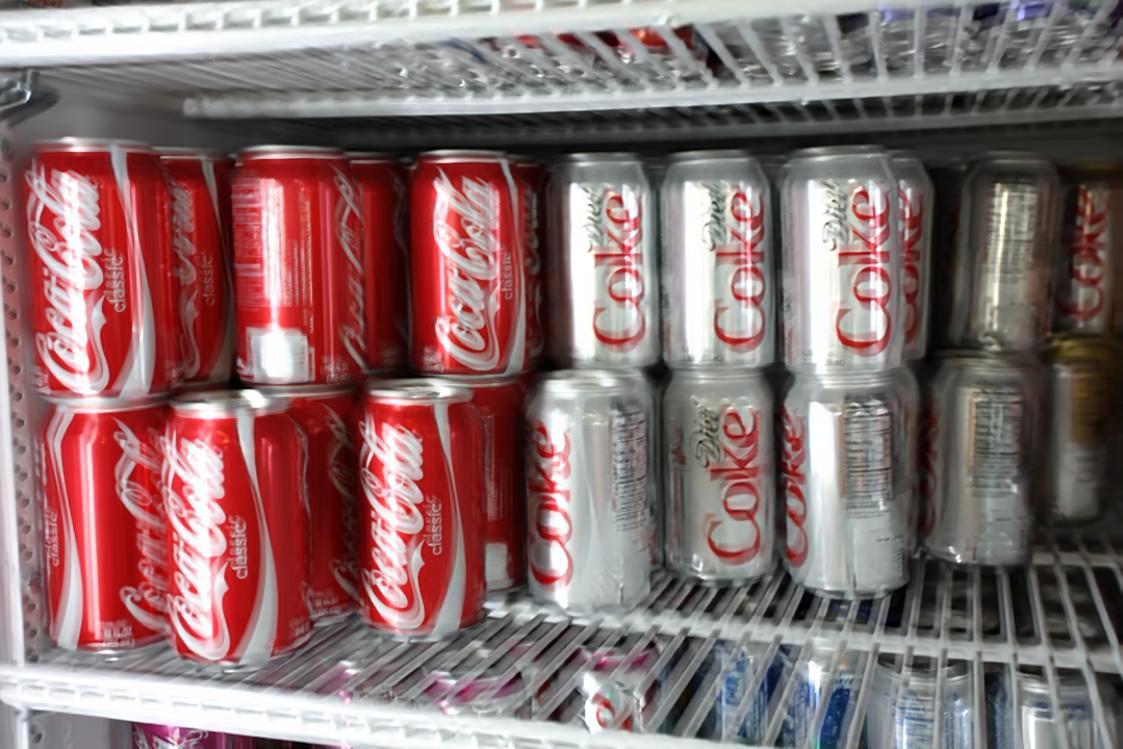}  &
\includegraphics[width=0.160\textwidth]{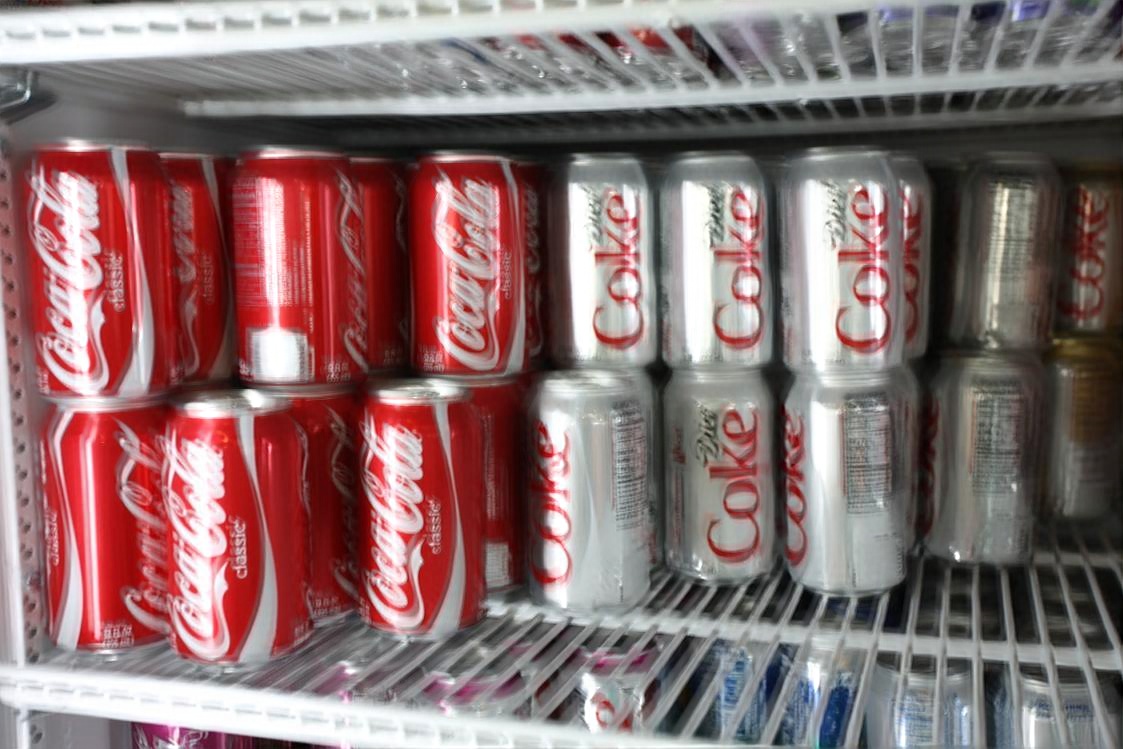}  &
\includegraphics[width=0.160\textwidth]{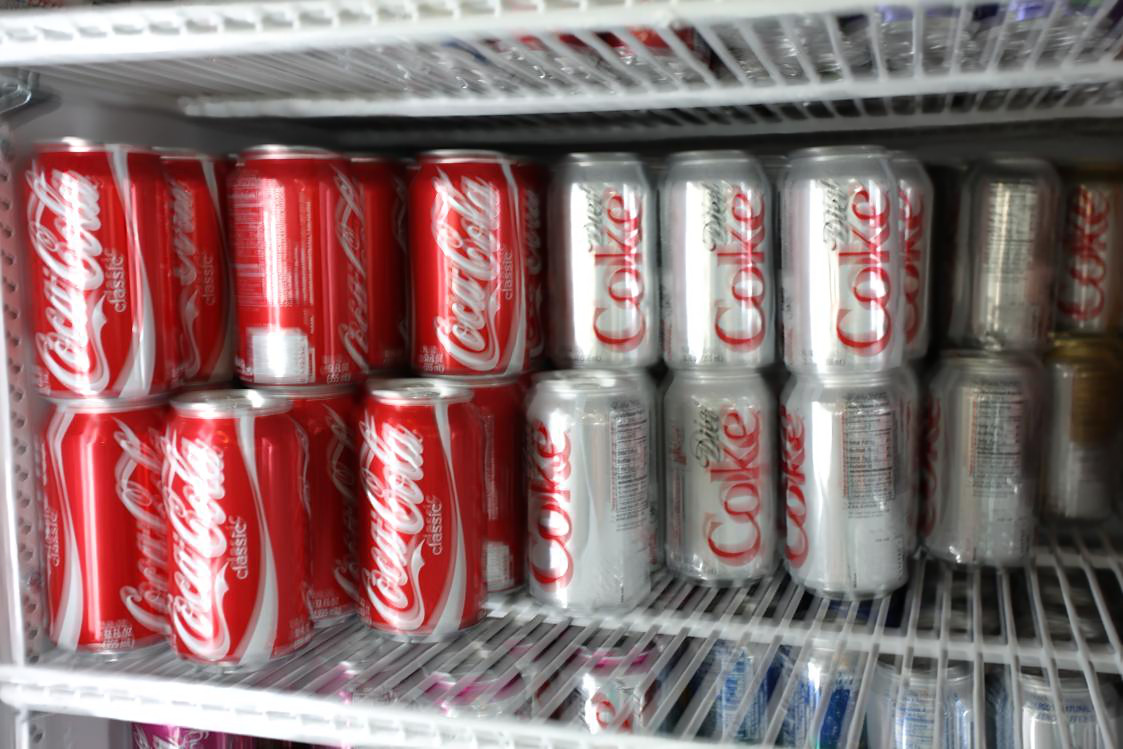}  &
\includegraphics[width=0.160\textwidth]{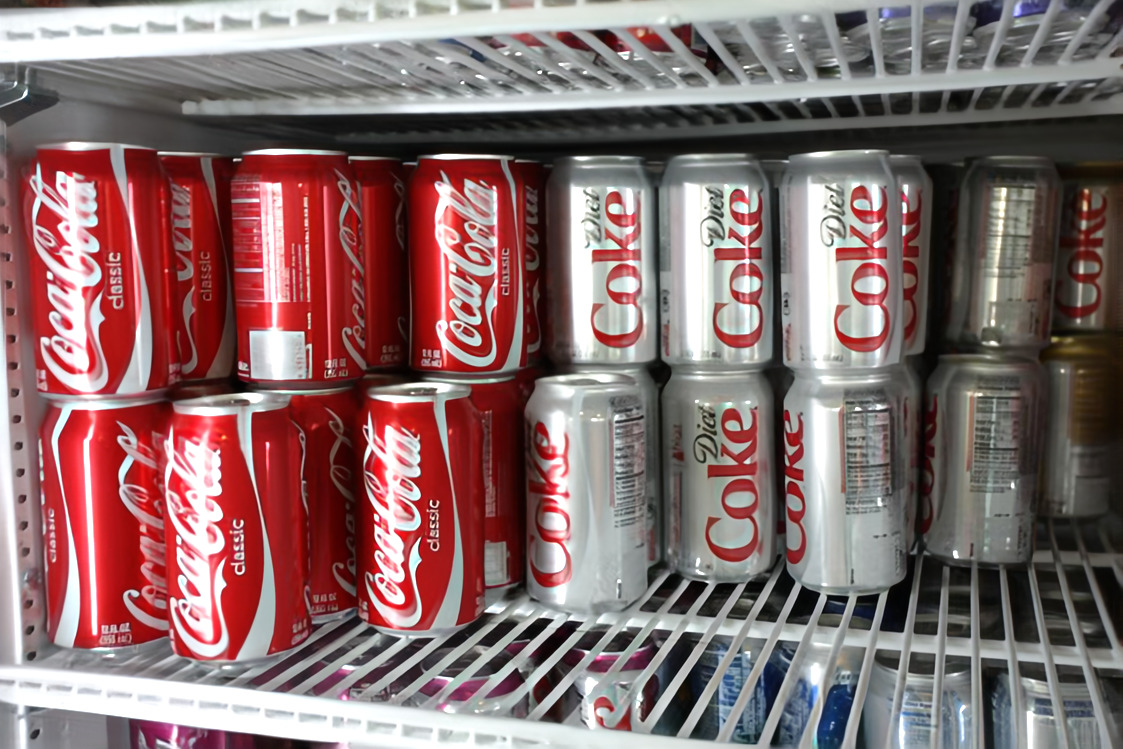}  \\
\includegraphics[trim=300 75 600 440, clip,width=0.075\textwidth]{imgs/Lai/Blurry/coke.jpg}  
\includegraphics[trim=600 200 300 315, clip,width=0.075\textwidth]{imgs/Lai/Blurry/coke.jpg}  &
\includegraphics[trim=300 75 600 440, clip,width=0.075\textwidth]{imgs/Lai/AnaSyn/coke_deblurred.jpg}  
\includegraphics[trim=600 200 300 315, clip,width=0.075\textwidth]{imgs/Lai/AnaSyn/coke_deblurred.jpg}  &
\includegraphics[trim=300 75 600 440, clip,width=0.075\textwidth]{imgs/Lai/SRN/coke.jpg}  
\includegraphics[trim=600 200 300 315, clip,width=0.075\textwidth]{imgs/Lai/SRN/coke.jpg}  &
\includegraphics[trim=300 75 600 440, clip,width=0.075\textwidth]{imgs/Lai/DeblurGAN_V2/coke.jpg}
\includegraphics[trim=600 200 300 315, clip,width=0.075\textwidth]{imgs/Lai/DeblurGAN_V2/coke.jpg} &
\includegraphics[trim=300 75 600 440, clip,width=0.075\textwidth]{imgs/Lai/MPRNet/coke.jpg}
\includegraphics[trim=600 200 300 315, clip,width=0.075\textwidth]{imgs/Lai/MPRNet/coke.jpg} &
\includegraphics[trim=300 75 600 440, clip,width=0.075\textwidth]{imgs/Lai/COCO900_restL2_aug_all_loss_gf1_80k/coke_NIMBUSR.jpg}
\includegraphics[trim=600 200 300 315, clip,width=0.075\textwidth]{imgs/Lai/COCO900_restL2_aug_all_loss_gf1_80k/coke_NIMBUSR.jpg}\\
\includegraphics[trim=0 200 0 0, clip, width=0.160\textwidth]{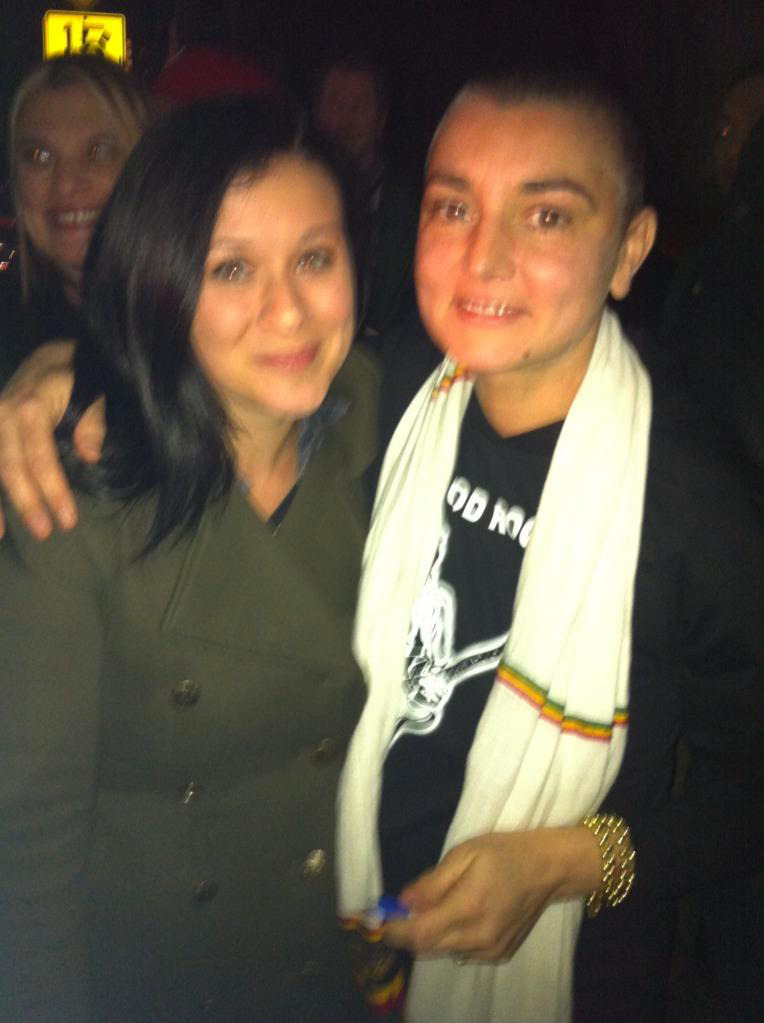}   &
\includegraphics[trim=0 200 0 0, clip, width=0.160\textwidth]{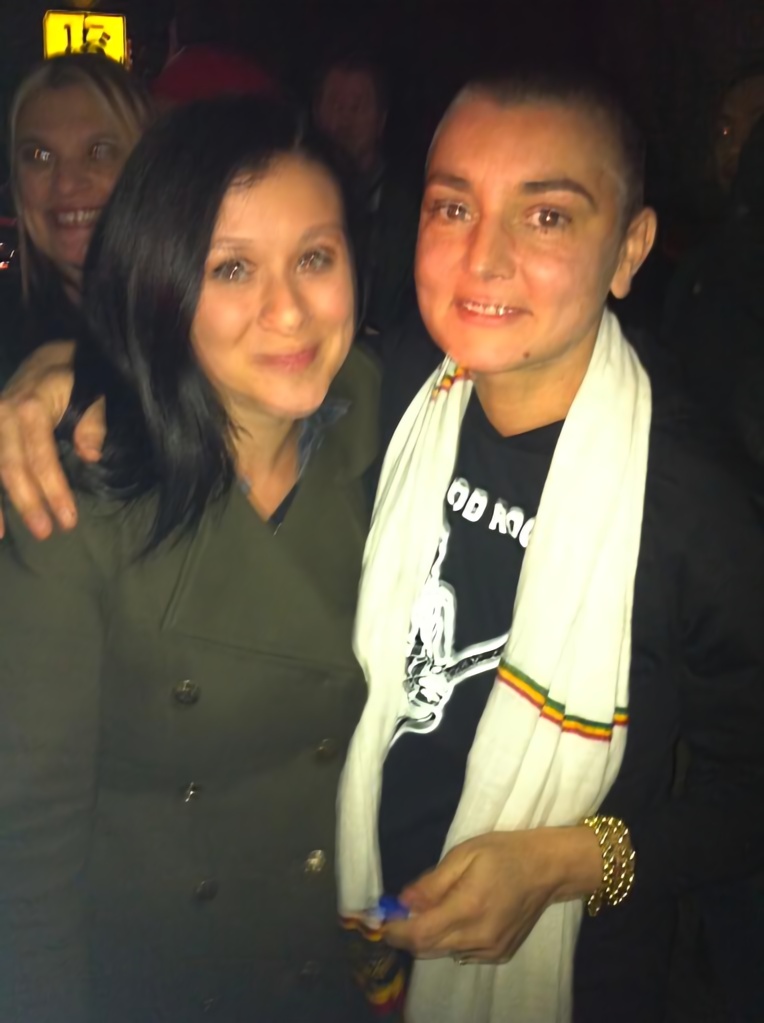} & 
\includegraphics[trim=0 200 0 0, clip, width=0.160\textwidth]{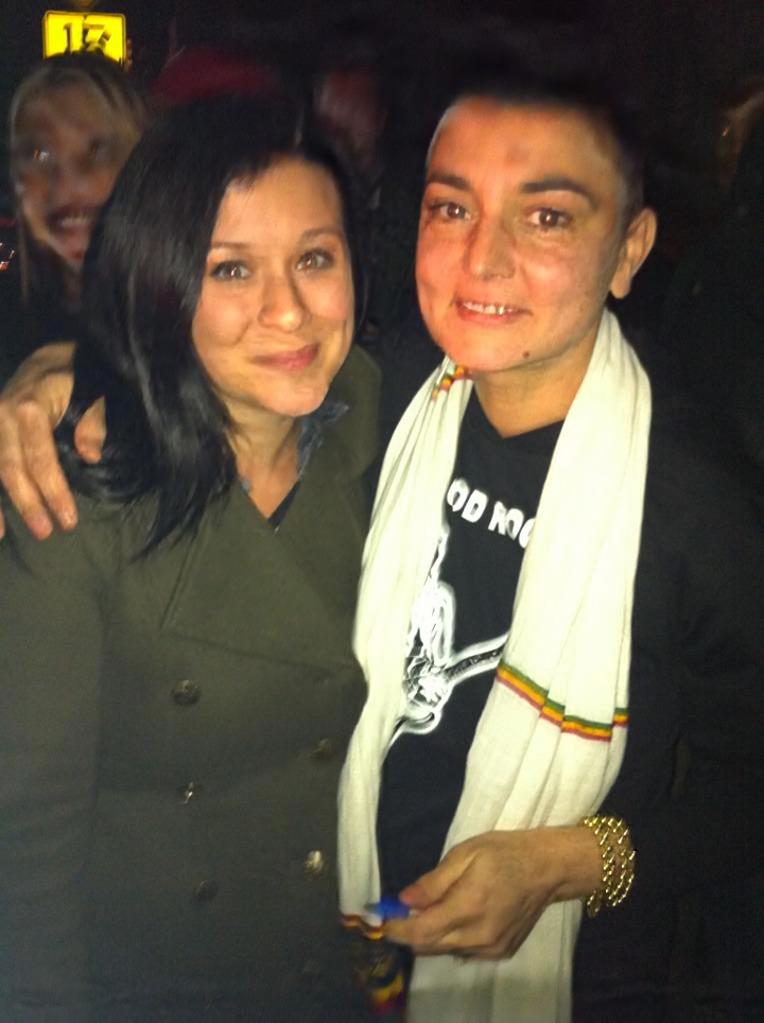}  &
\includegraphics[trim=0 200 0 0, clip, width=0.160\textwidth]{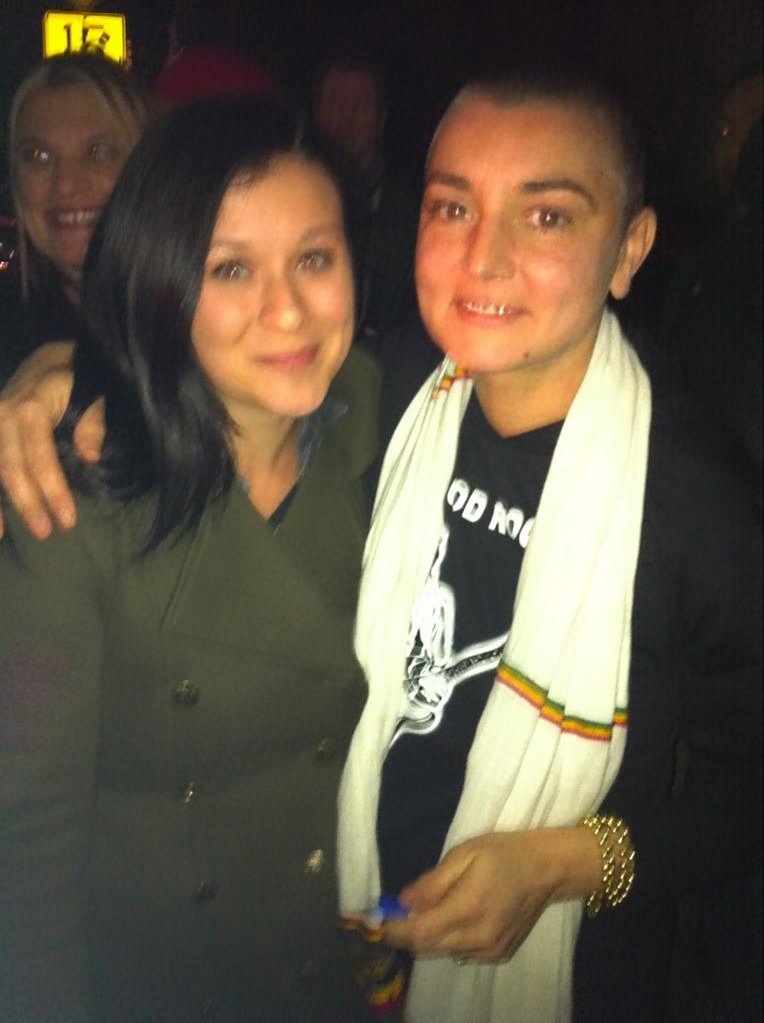}  &
\includegraphics[trim=0 200 0 0, clip, width=0.160\textwidth]{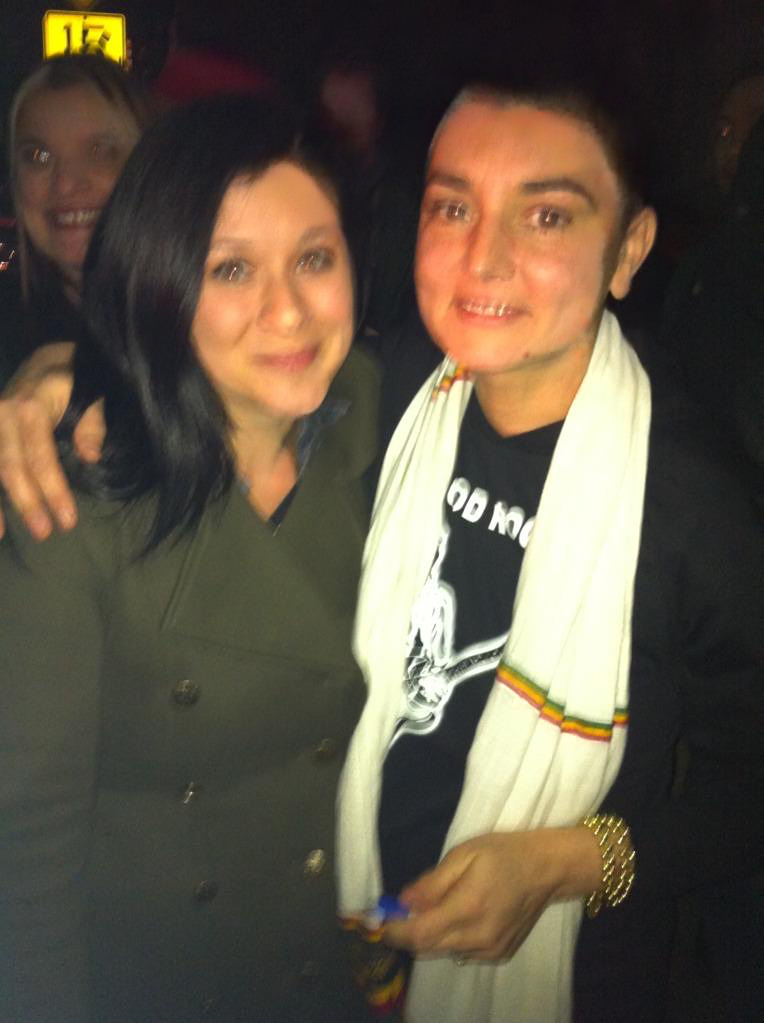}  &
\includegraphics[trim=0 200 0 0, clip, width=0.160\textwidth]{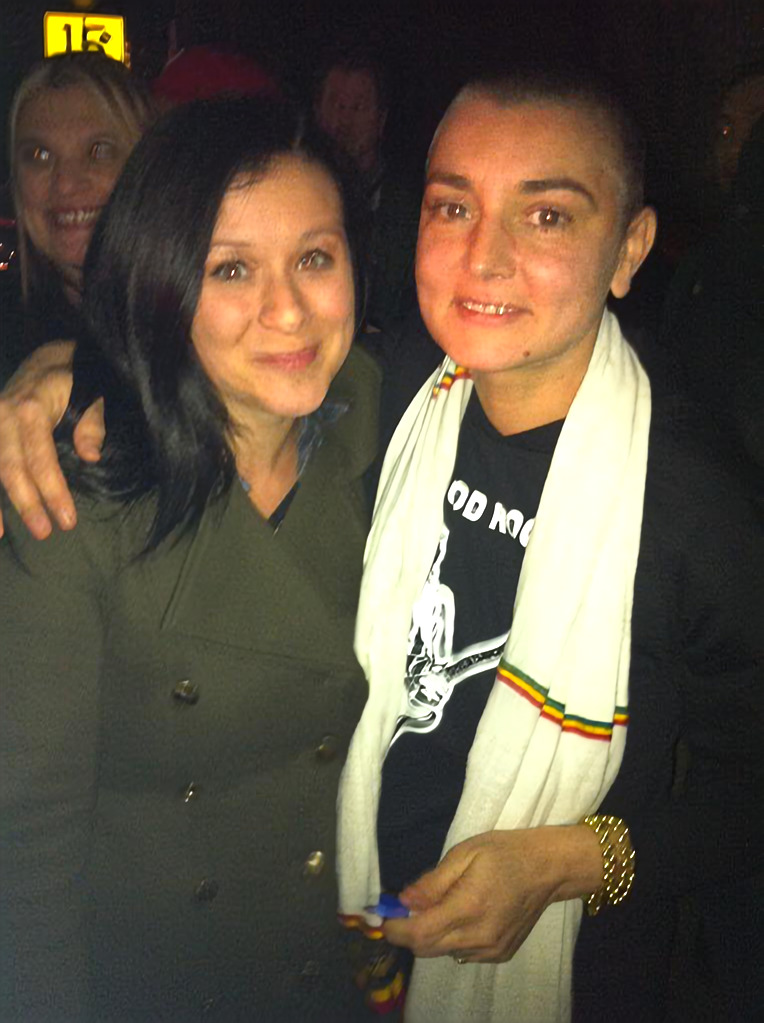}  \\

\end{tabular} 
    \caption{\textbf{Deblurring examples on real blurry images from Lai's  dataset  \cite{lai2016comparative}.} Our model-based approach is competitive with state-of-the-art data-driven methods. Best viewed in electronic format.}
    \label{fig:LaiDeepLearning}
\end{figure*}

%% file: perceptual_metrics_results.tex
\begin{table}[h]
    \caption{Blur strength ($\downarrow$) of restored images on real datasets}
    \centering
    \setlength\tabcolsep{3 pt}
    \begin{tabular}{lccccc}
     \toprule 
        Method & K\"{o}hler  & K\"{o}hler-S & Lai  & Lai(HS) & RealBlur \\
        \midrule 
        None (input image)& 0.649 & 0.635 &0.489 & 0.453 & 0.504   \\
        SRN \cite{tao2018scale} & 0.606 & 0.550 & 0.450 & 0.388 & 0.413  \\
        DGAN-v2 \cite{kupyn2019deblurgan}    & \colorbox{Cyan1}{0.525}   & \colorbox{Yellow1}{0.491} &\colorbox{Yellow1}{0.416}  & \colorbox{Yellow1}{0.380} & \colorbox{Yellow1}{0.397} \\
        MPRNet \cite{Zamir2021MPRNet}   & 0.597 & 0.568& 0.478 & 0.412 & 0.421 \\ 
        MIMO-UNet \cite{cho2021rethinking} & 0.615  & 0.578 &0.484   & 0.443 & 0.453 \\
        MIMO-UNet+ \cite{cho2021rethinking} & 0.632   & 0.590& 0.486 & 0.446 & 0.459 \\
        NAFNet \cite{NAFNet}        & 0.583 & 0.537 & 0.467 & 0.393 & 0.428 \\
        Ana.-Synth. \cite{kaufman2020deblurring}        &  0.562 & 0.511  & 0.473   & 0.425 & 0.465 \\
        \hline 
{J-MKPD}  &   \colorbox{Yellow1}{0.559}&  \colorbox{Cyan1}{0.433}  &  \colorbox{Cyan1}{0.410}  & \colorbox{Cyan1}{0.370} & \colorbox{Cyan1}{0.378} \\
        \midrule 
        Ground truth  &  -  & -   & - & - & 0.329 \\
        \bottomrule 
    \end{tabular}
    \label{tab:blur_strength}
\end{table}

\begin{table}[h]
    \caption{Sharpness index of restored images on real datasets}
    \centering
    \setlength\tabcolsep{3 pt}
    \begin{tabular}{lccccc}
     \toprule
        Method & K\"{o}hler  & K\"{o}hler-S & Lai  & Lai(HS) & RealBlur \\
        \midrule
        None (input image)& 112 & 157 & 1978 & 717 & 1039  \\
        SRN \cite{tao2018scale} &314 & 438 & 2092 & \colorbox{Yellow1}{1113} & 1850  \\
        DGAN-v2 \cite{kupyn2019deblurgan}    & 432   & 624 &\colorbox{Yellow1}{2700}  & 915 & 1571 \\
        MPRNet \cite{Zamir2021MPRNet}   & 302 & 421 & 2569 & 988 & 1748 \\ 
        MIMO-UNet \cite{cho2021rethinking} & 222  & 312 & 2292   & 963 & 1497\\
        MIMO-UNet+ \cite{cho2021rethinking} & 213  & 294 & 2267 & 928 & 1483 \\
        NAFNet \cite{NAFNet}        & 295 & 406 & 1863 & 808 & 1592\\
        Ana.-Synth. \cite{kaufman2020deblurring}        &  \colorbox{Cyan1}{1247} & \colorbox{Cyan1}{1739}  & 3145   & 982 & \colorbox{Yellow1}{2084} \\
        \midrule  
{J-MKPD}  &   \colorbox{Yellow1}{939}&  \colorbox{Yellow1}{1345}  &  \colorbox{Cyan1}{4672}  & \colorbox{Cyan1}{1656} & \colorbox{Cyan1}{3323} \\
        \midrule 
        Ground truth  &  -  & -   & - & - & 1926 \\
        \bottomrule 
    \end{tabular}
    \label{tab:sharpness_index}
\end{table}

%% file: deblurringKohler.tex
Deblurring results on this dataset, shown in \cref{tab:kohler_results}, illustrate the benefit of having an image degradation model. Since the blur in this dataset is only slightly non-uniform, methods that assume uniform blur and can deal with large kernels, such as Xu and Jia~\cite{xu2010two},  Cho and Lee~\cite{cho2009fast}, and Whyte \etal~\cite{whyte2014deblurring}, not included in the table,  outperform most DL end-to-end methods by a large margin. %
 The main drawbacks of these variational algorithms are the severe artifacts that arise when the assumed conditions are not met (e.g. non-uniform blur), and the orders of magnitude slower inference time than DL-based methods. 

Interestingly, despite being a DL-based method, Analysis-Synthesis~\cite{kaufman2020deblurring} performs remarkably well. Indeed, this method is well adapted to K{\"o}hler's images, as the analysis network is trained to predict spatially-uniform blur models, which is close to the kernel field generated in this dataset. 

Our method, contrary to the other kernel estimation-based methods, does not assume a uniform or global parametric blur model. However, the structure imposed by the degradation model is enough to outperform end-to-end methods by a significant margin. Its performance is close to that of Analysis-Synthesis~\cite{kaufman2020deblurring}, outperforming it in terms of LPIPS. 

For completeness, the rightmost part of \cref{tab:kohler_results} reports the results obtained when discarding images with blur kernels larger than the maximum kernel size estimated by our method. 

The blur strength and sharpness index of the restored images on the whole K\"ohler's dataset, and on the small kernels subset (K{\"o}hler-S), are also reported in\cref{tab:blur_strength} and \cref{tab:sharpness_index}. Note that globally, over all the datasets considered in this study, J-MKPD ranks first on both metrics. It is also interesting to note that DeblurGAN-v2 ranks second in terms of blur strength, but its performance in terms of sharpness index decreases; this is due to the fact that DeblurGAN-v2 produces sharp images, but it is known to hallucinate textures and high-frequency artifacts. On the other side, Analysis-Synthesis lags behind DeblurGAN-v2 in terms of blur strength, but outperforms it in terms of sharpness index; this results from the fact that Analysis-Synthesis tends to produce cartoon-like restorations. 

\begin{table*}[h]
    \caption{\textbf{Comparison on K\"{o}hler's dataset~\cite{kohler2012recording}}. We report the metrics average results for the whole dataset and for a subset that excludes the images blurred with kernel sizes bigger than the maximum kernel size predicted by our KPN. We report the values for the two registration methods previously reported: pure translation \cite{kohler2012recording} and homography transformation \cite{rim_2020_ECCV}.  %
      \label{tab:kohler_results}}  

  \setlength\tabcolsep{1pt} %
  \centering
  \begin{tabular}{l|l|c c |c c c|c c|c c c}
    \toprule 
          \multicolumn{2}{c}{} &  \multicolumn{5}{|c|}{All images} &  \multicolumn{5}{|c}{Except \#8, \#9, \#10, \#11} \\
          \multicolumn{2}{c}{} & \multicolumn{2}{|c|}{Trans-reg} & \multicolumn{3}{c}{Homo-reg} & \multicolumn{2}{|c|}{Trans-reg} & \multicolumn{3}{c}{Homo-reg} \\
          \hline
           & \hspace{2cm} Model & PSNR & MSSIM  & PSNR & SSIM & LPIPS $\downarrow$ & PSNR & MSSIM &  PSNR & SSIM & LPIPS $\downarrow$ \\
    \hline   
    DeblurGAN \cite{kupyn2018deblurgan} &  \multirow{10}{*}{Fully data-driven, end-to-end}  & 25.81  & 0.801 & 25.71 &  0.748 & 0.303 & 28.29 & 0.899 & 28.04 & 0.802 & 0.195 \\
    DeepDeblur \cite{Nah_2017_CVPR}  &  & 25.71 & 0.811 & 25.66 & 0.763 & 0.341 & 27.97 & 0.901 & 27.76 & 0.813 & 0.236     \\
    SRN \cite{tao2018scale} & & 26.85 & 0.839  &  26.91 & 0.789 & 0.327 & 29.50 & 0.932 & 29.45 & 0.848 & 0.221       \\
    DMPHN 1-2-4 \cite{Zhang_2019_CVPR} &  & 25.39  & 0.790 &  24.35& 0.697 & 0.698 & 27.62 & 0.882 & 26.07 & 0.733  & 0.369 \\
    DeblurGANv2 Inc. \cite{kupyn2019deblurgan} &  & 26.97 & 0.830 & 26.95 & 0.788 &0.295 & 29.84 & 0.939 & 29.71 & 0.851 & 0.165 \\
    DeblurGANv2 Mob. \cite{kupyn2019deblurgan}& &25.67&0.801  &  25.51 & 0.742 & 0.307 &28.08&0.896& 27.74 & 0.793 & 0.194 \\
    MIMO-UNet \cite{cho2021rethinking} & & 25.34 &   0.79 & 25.21  & 0.746 & 0.361 & 27.53 & 0.882 & 27.20 & 0.792 & 0.263 \\
    MIMO-UNet+ \cite{cho2021rethinking} &  & 25.22 & 0.790  & 25.05  & 0.746 & 0.369 & 26.83 & 0.874 & 27.49 & 0.790  & 0.277     \\
    MPRNet \cite{Zamir2021MPRNet}&  &  26.32 & 0.829& 26.16 & 0.779 & 0.328 & 28.63& 0.906 & 28.28 & 0.827 & 0.226 \\
    NAFNet \cite{NAFNet} &&26.07& 0.816 & 25.97 & 0.778 & 0.309 &28.32&0.894& 28.06 & 0.809  & 0.216 \\         
    \hline
        Sun \etal \cite{sun2015learning}   & Spatially-varying linear kernels    &24.96 & 0.789 & 24.85 & 0.738 & 0.383 & 27.08 & 0.878  & 26.79 & 0.78 & 0.294 \\
    Gong \etal \cite{gong2017motion}        & Spatially-varying linear kernels  &       24.92 & 0.784 & 24.82 & 0.734 & 0.394 & 27.03 & 0.872 & 26.74 & 0.775 & 0.306 \\
    Zhang \etal \cite{zhang2021exposure}        & Spatially-varying quadratic kernels  &     25.57 & 0.798 & 25.42 & 0.754 & 0.358 & 27.72 & 0.883 & 27.41 & 0.802 & 0.263 \\
    \hline
    Analysis-Synthesis\cite{kaufman2020deblurring} & Spatially uniform kernel-guided convolutions & \colorbox{Cyan1}{29.37} & \colorbox{Cyan1}{0.892} & \colorbox{Cyan1}{29.86} &\colorbox{Cyan1}{0.851}& \colorbox{Yellow1}{0.266} & \colorbox{Cyan1}{32.55} & \colorbox{Cyan1}{0.972} & \colorbox{Cyan1}{33.05} & \colorbox{Cyan1}{0.915} & \colorbox{Yellow1}{0.159} \\ 
    \hline   
J-MKPD  &     Joint kernel prediction and restoration  &  \colorbox{Yellow1}{28.30} &   \colorbox{Yellow1}{0.860}   &   \colorbox{Yellow1}{28.65} &  \colorbox{Yellow1}{0.832} &  \colorbox{Cyan1}{0.250} & \colorbox{Yellow1}{31.96}  & \colorbox{Yellow1}{0.968} & \colorbox{Yellow1}{32.36}& \colorbox{Yellow1}{0.911} &\colorbox{Cyan1}{0.133} \\  
    \bottomrule 
  \end{tabular}
\end{table*}

%% file: figure_kernels_before_and_after_ft.tex
\begin{figure*}[t!]
    \centering
    \scriptsize
    \setlength{\tabcolsep}{2pt}
    \begin{tabular}{ccccc}
    Blurry & Kernels without JT & Restored without JT & Kernels after JT & Restored after JT  \\
\includegraphics[trim=20 70 170 350, clip,width=0.19\textwidth]{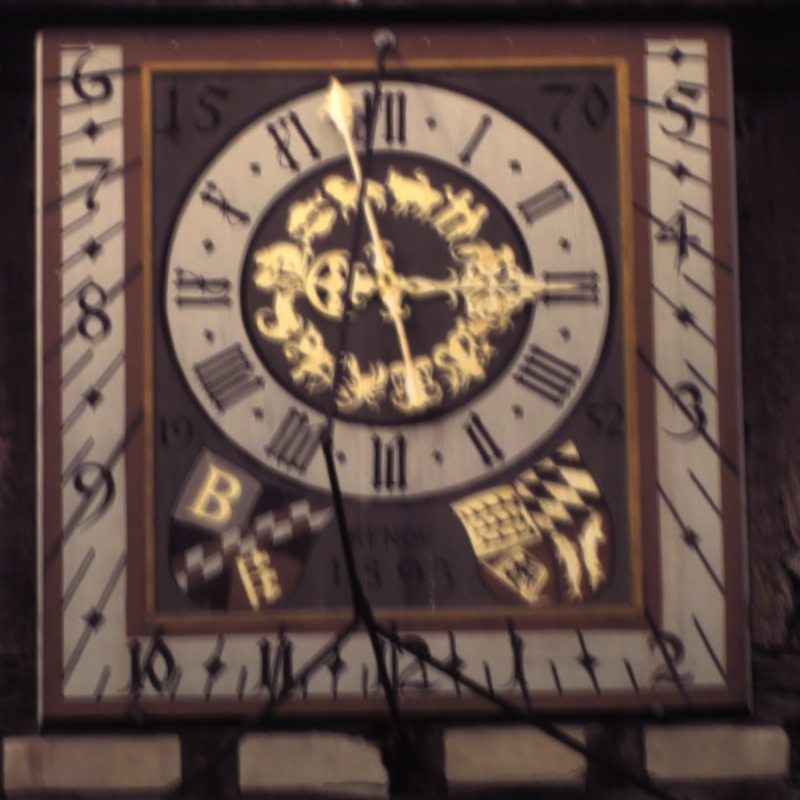} &
\includegraphics[trim=20 70 170 350, clip,width=0.19\textwidth]{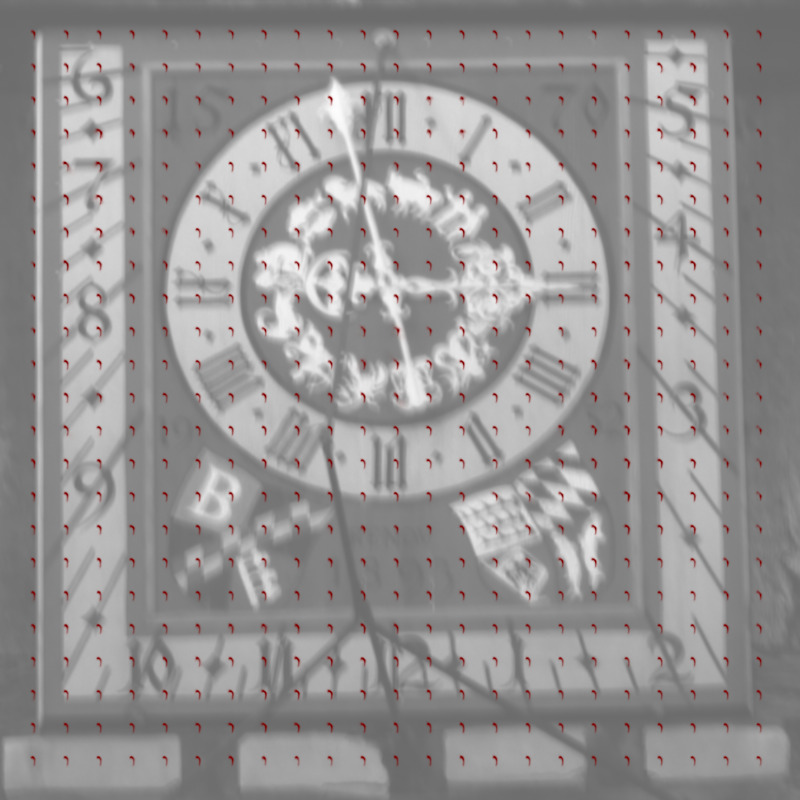} &   
\includegraphics[trim=20 70 170 350, clip,width=0.19\textwidth]{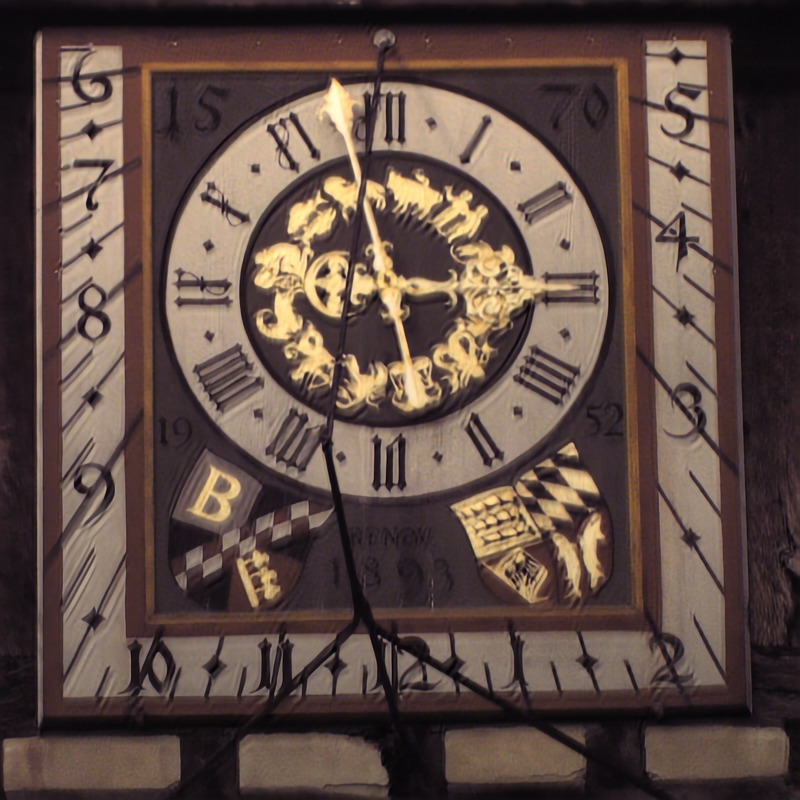} &
\includegraphics[trim=20 70 170 350, clip,width=0.19\textwidth]{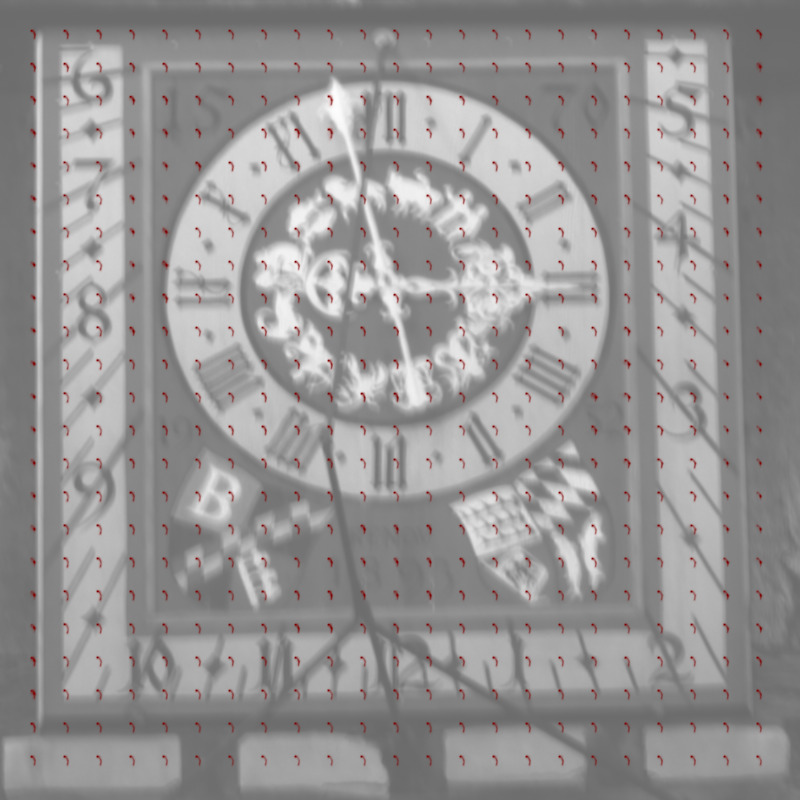} &
\includegraphics[trim=20 70 170 350, clip,width=0.19\textwidth]{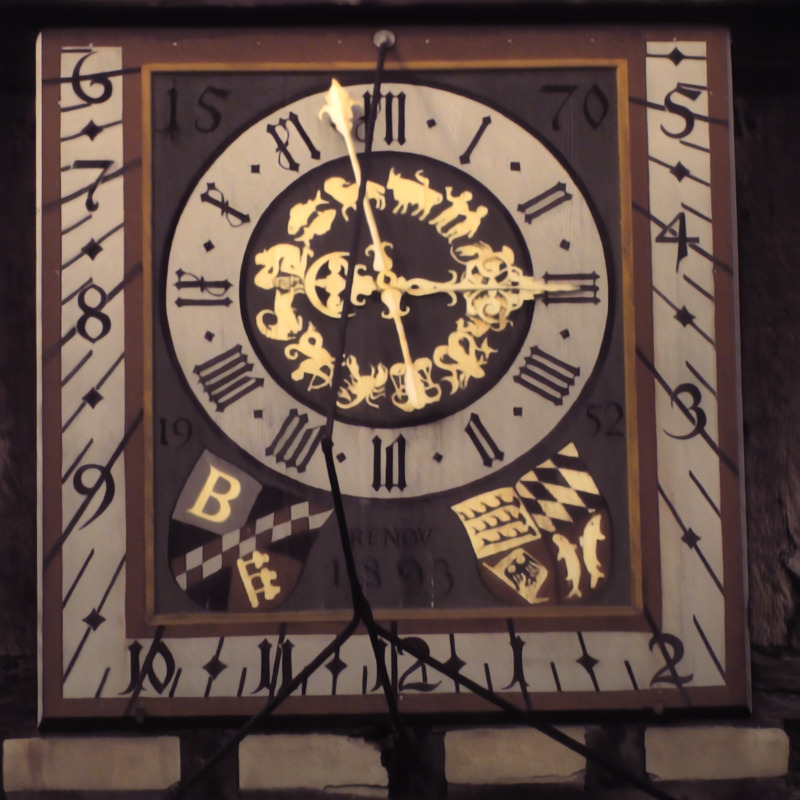}\\ 
\includegraphics[trim=0 120 350 410, clip,width=0.19\textwidth]{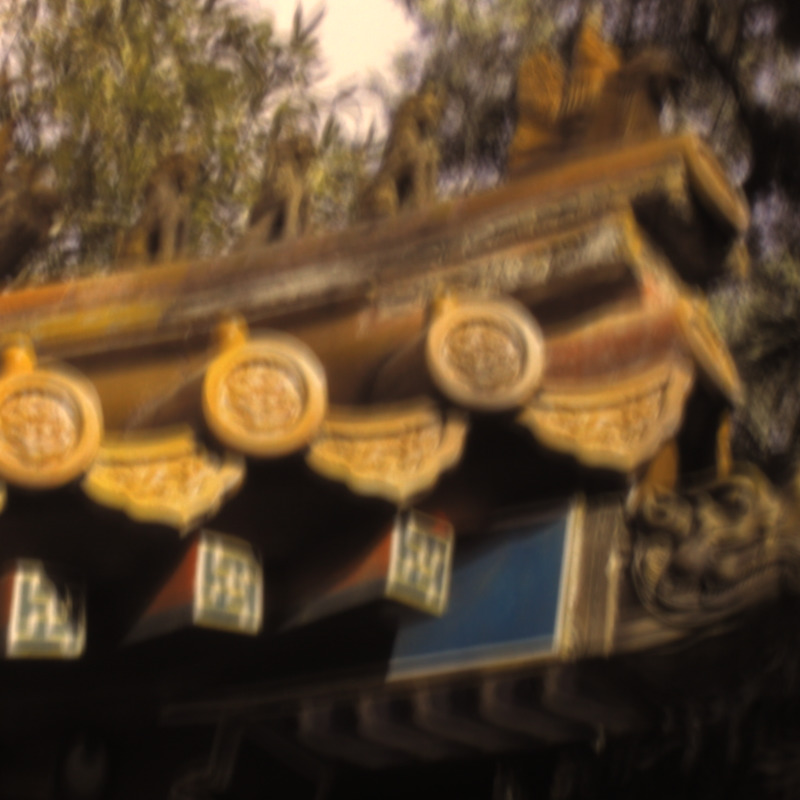} &
\includegraphics[trim=0 120 350 410, clip,width=0.19\textwidth]{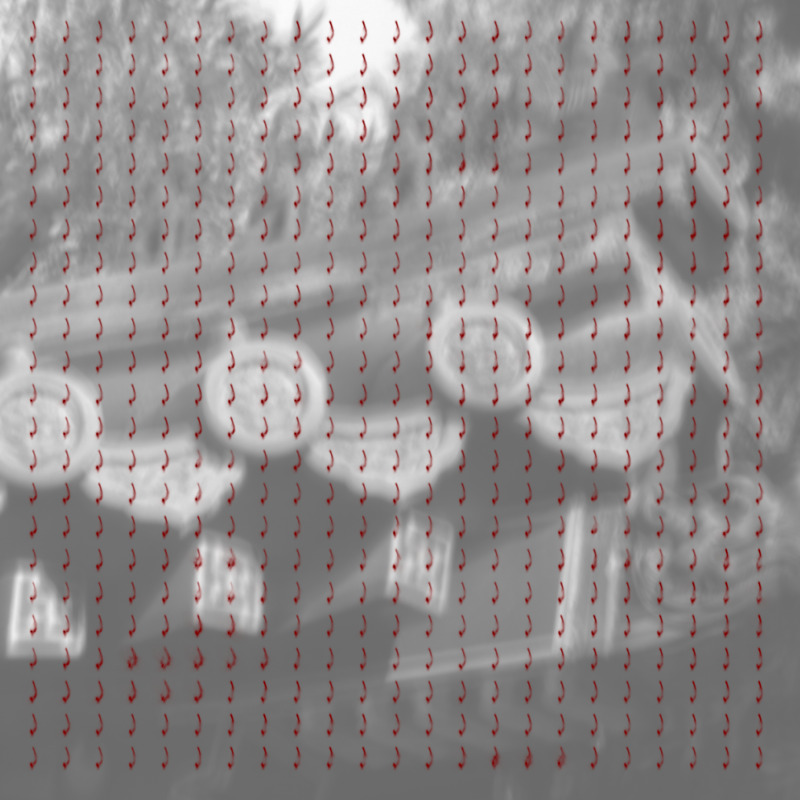} &   
\includegraphics[trim=0 120 350 410, clip,width=0.19\textwidth]{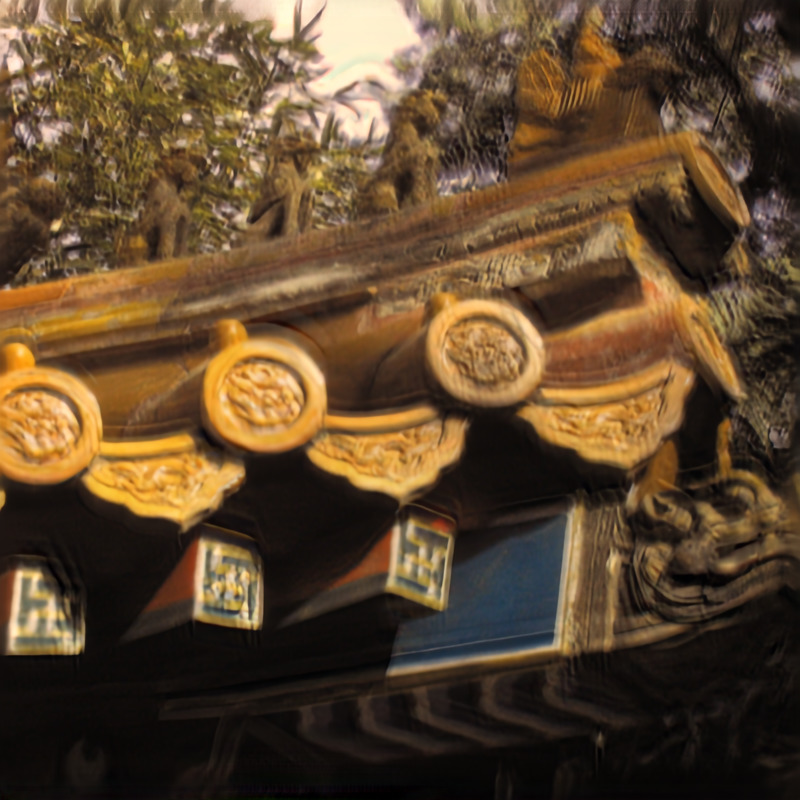} &
\includegraphics[trim=0 120 350 410, clip,width=0.19\textwidth]{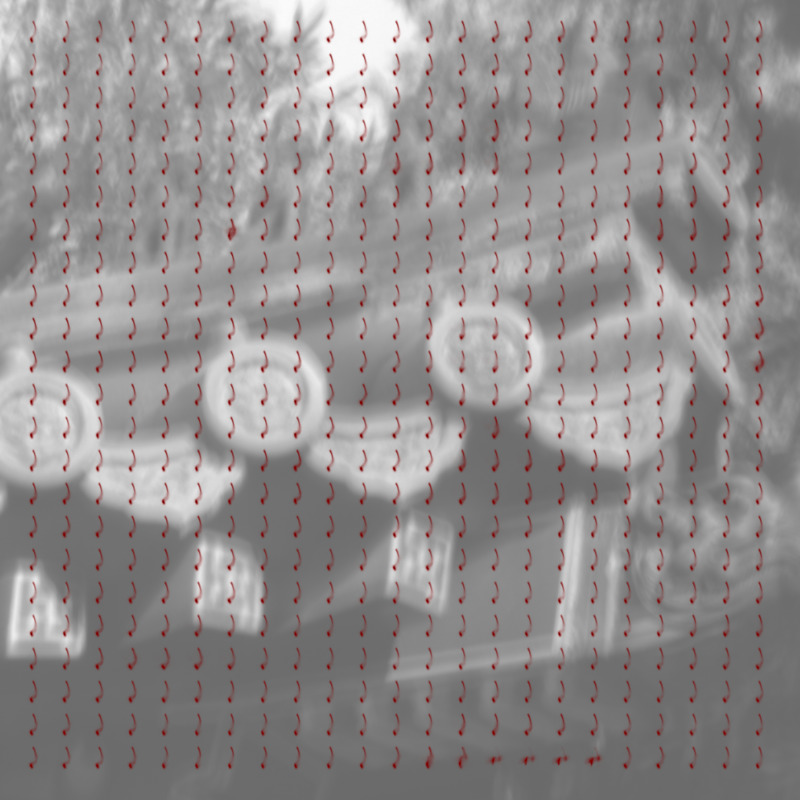} &
\includegraphics[trim=0 120 350 410, clip,width=0.19\textwidth]{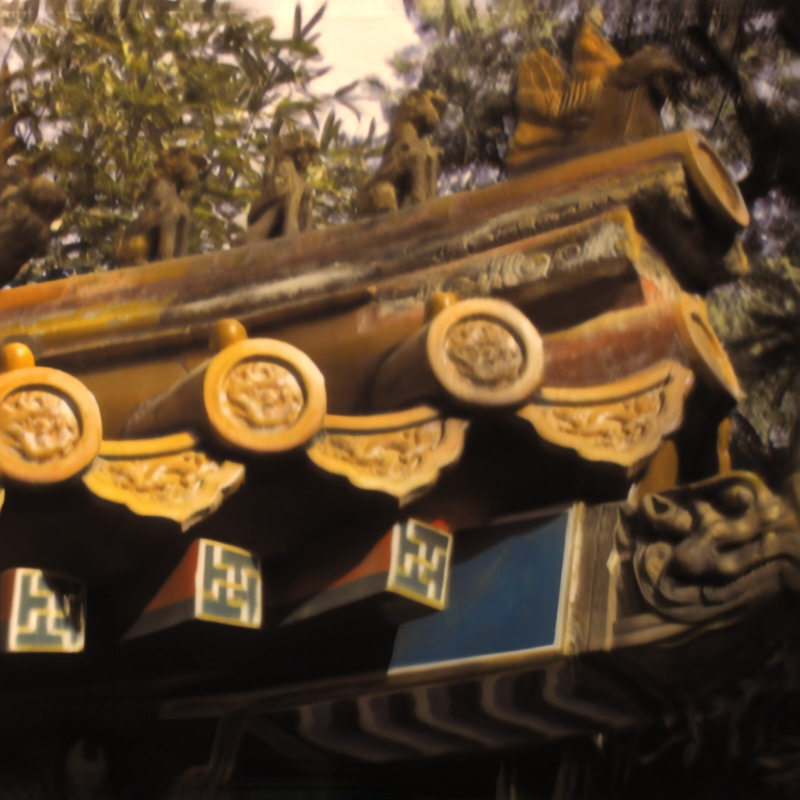}\\ 
\includegraphics[trim=140 200 150 200, clip,width=0.19\textwidth]{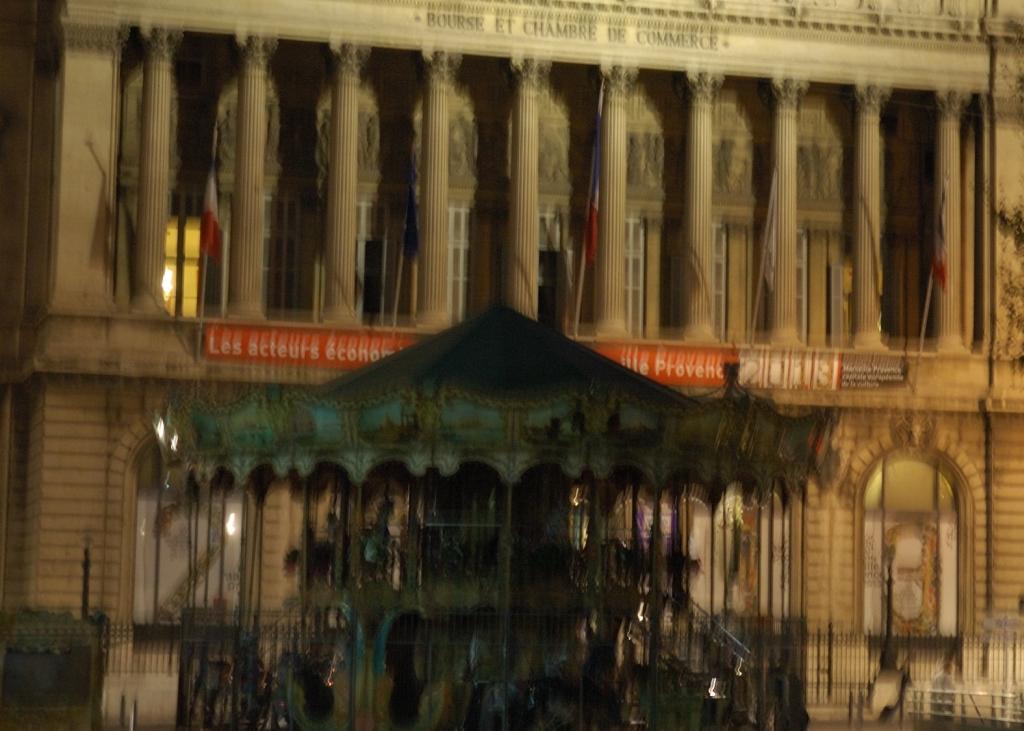} &
\includegraphics[trim=140 200 150 200, clip,width=0.19\textwidth]{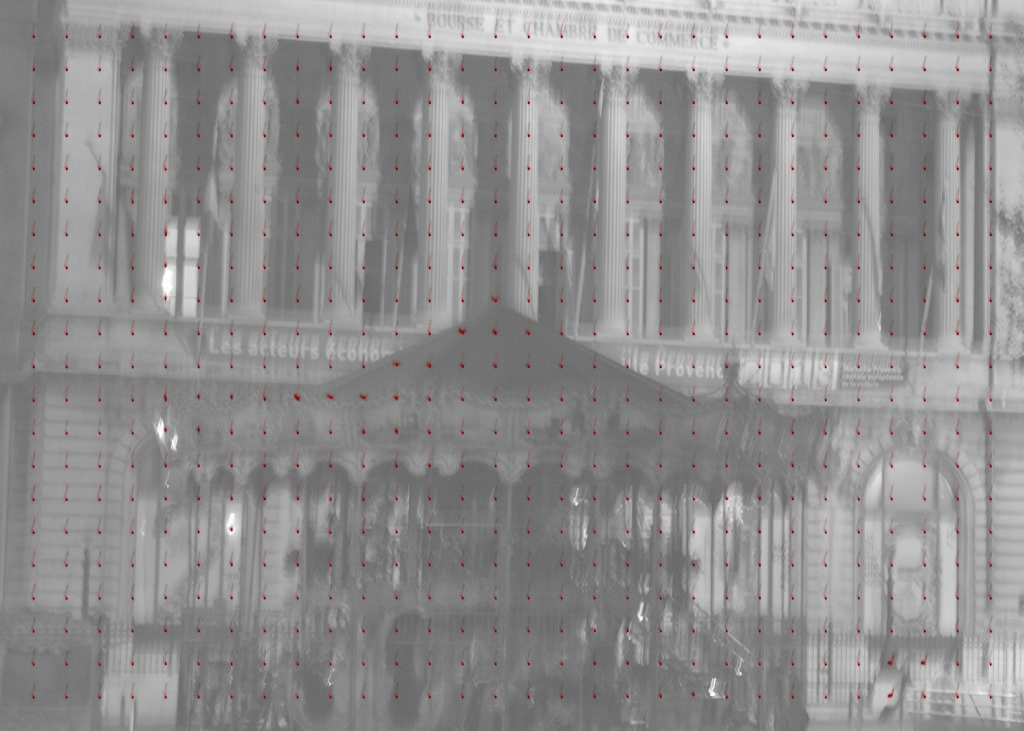} &   
\includegraphics[trim=140 200 150 200, clip,width=0.19\textwidth]{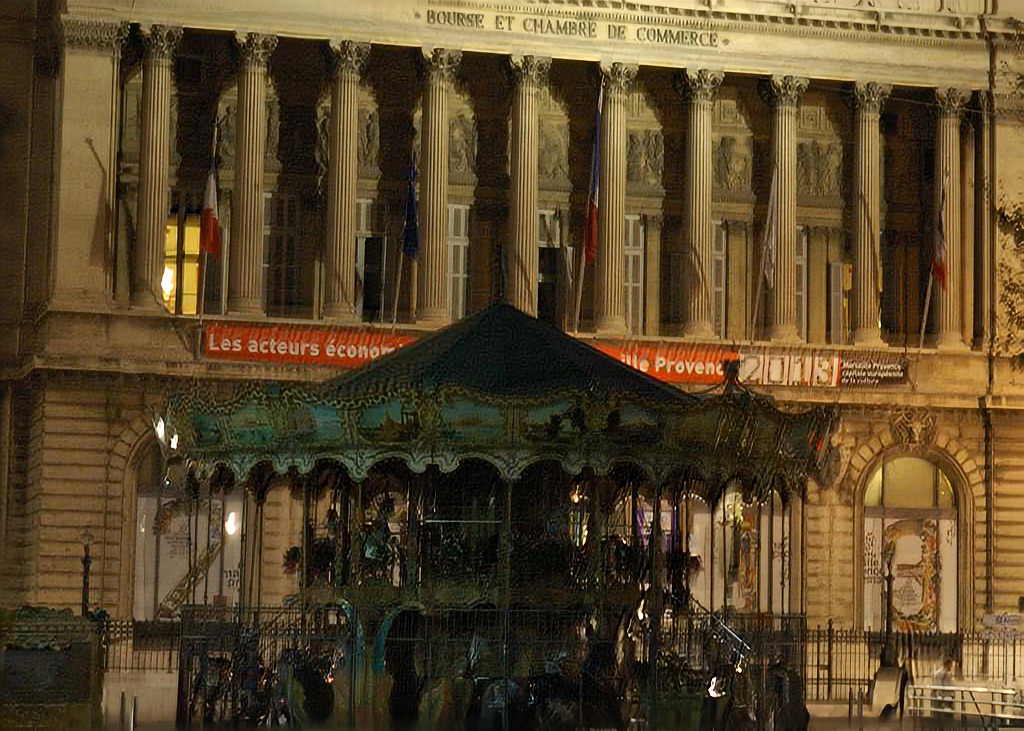} &
\includegraphics[trim=140 200 150 200, clip,width=0.19\textwidth]{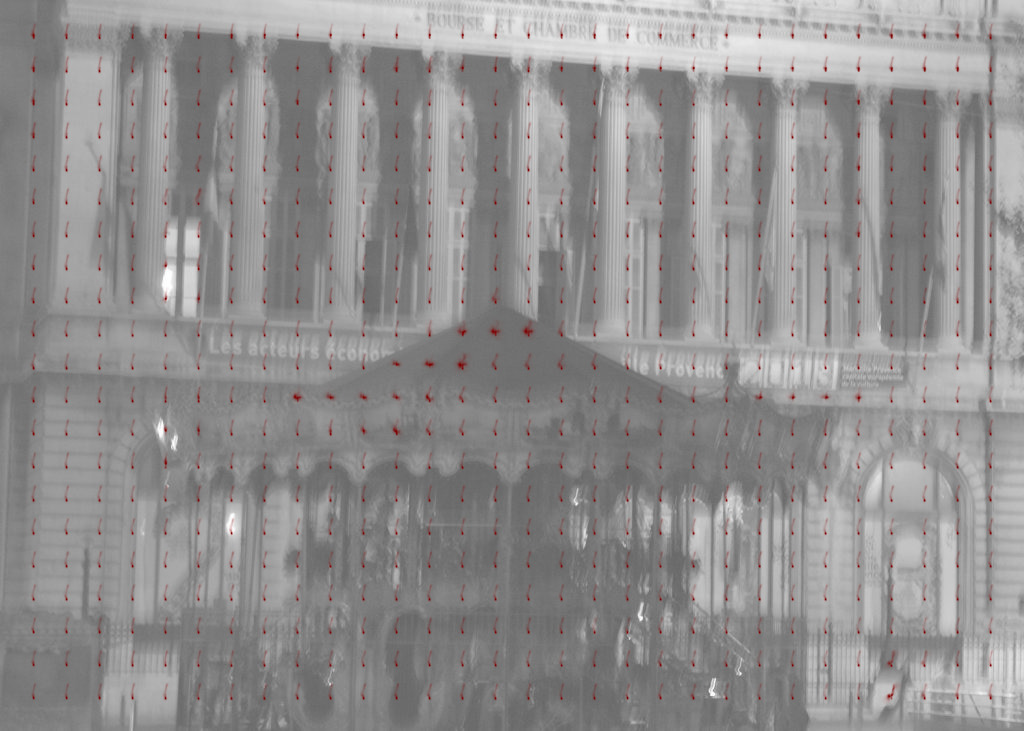} &
\includegraphics[trim=140 200 150 200, clip,width=0.19\textwidth]{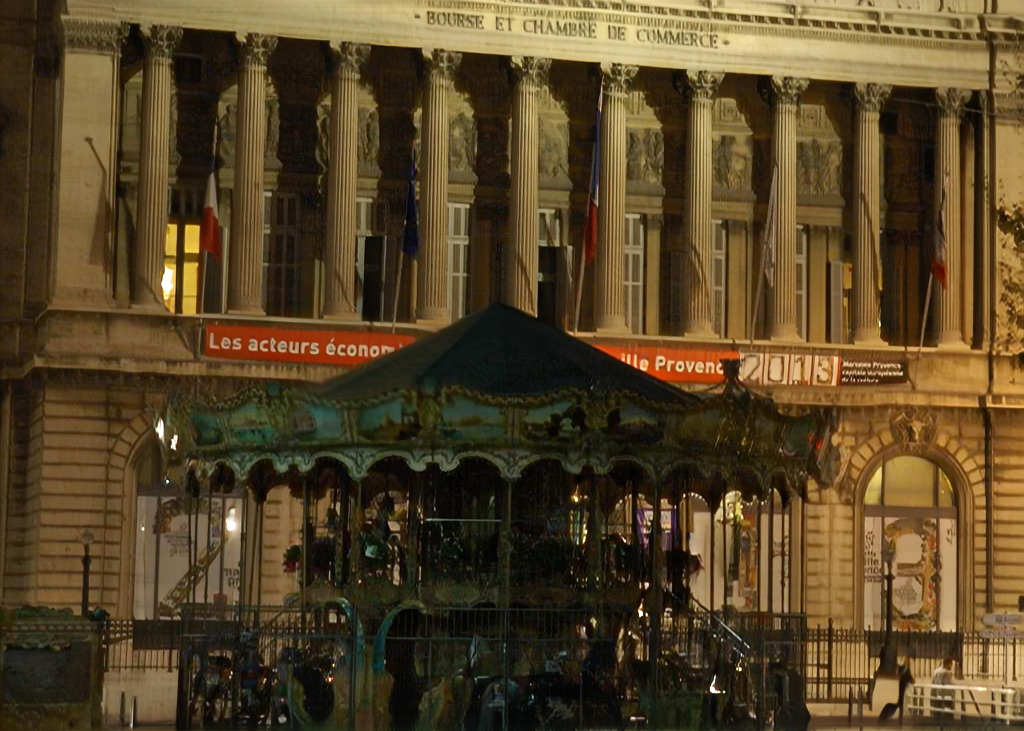} \\ 
\includegraphics[trim=40 95 100 50, clip,width=0.19\textwidth]{imgs/Lai/Blurry/butchershop.jpg} &
\includegraphics[trim=40 95 100 50, clip,width=0.19\textwidth]{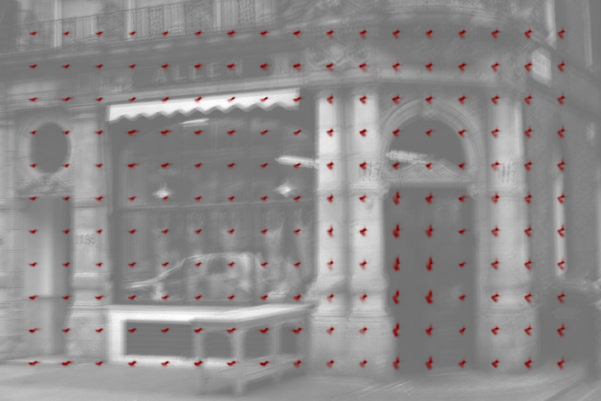} &   
\includegraphics[trim=40 95 100 50, clip,width=0.19\textwidth]{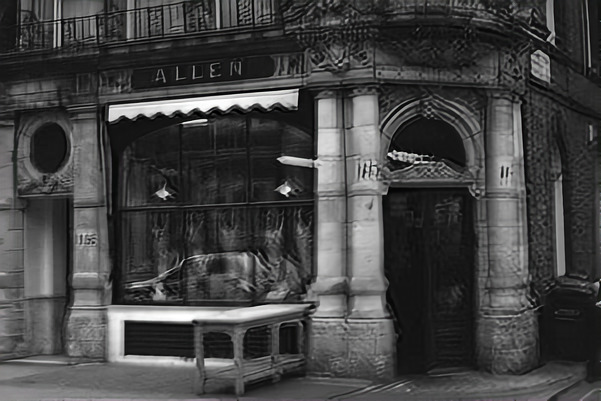} &
\includegraphics[trim=40 95 100 50, clip,width=0.19\textwidth]{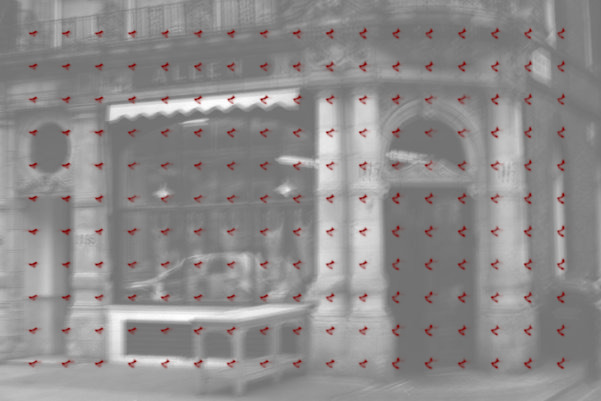} &
\includegraphics[trim=40 95 100 50, clip,width=0.19\textwidth]{imgs/Lai/COCO900_restL2_aug_all_loss_gf1_80k/butchershop_NIMBUSR.jpg}  \\
\includegraphics[trim=300 280 80 110, clip,width=0.19\textwidth]{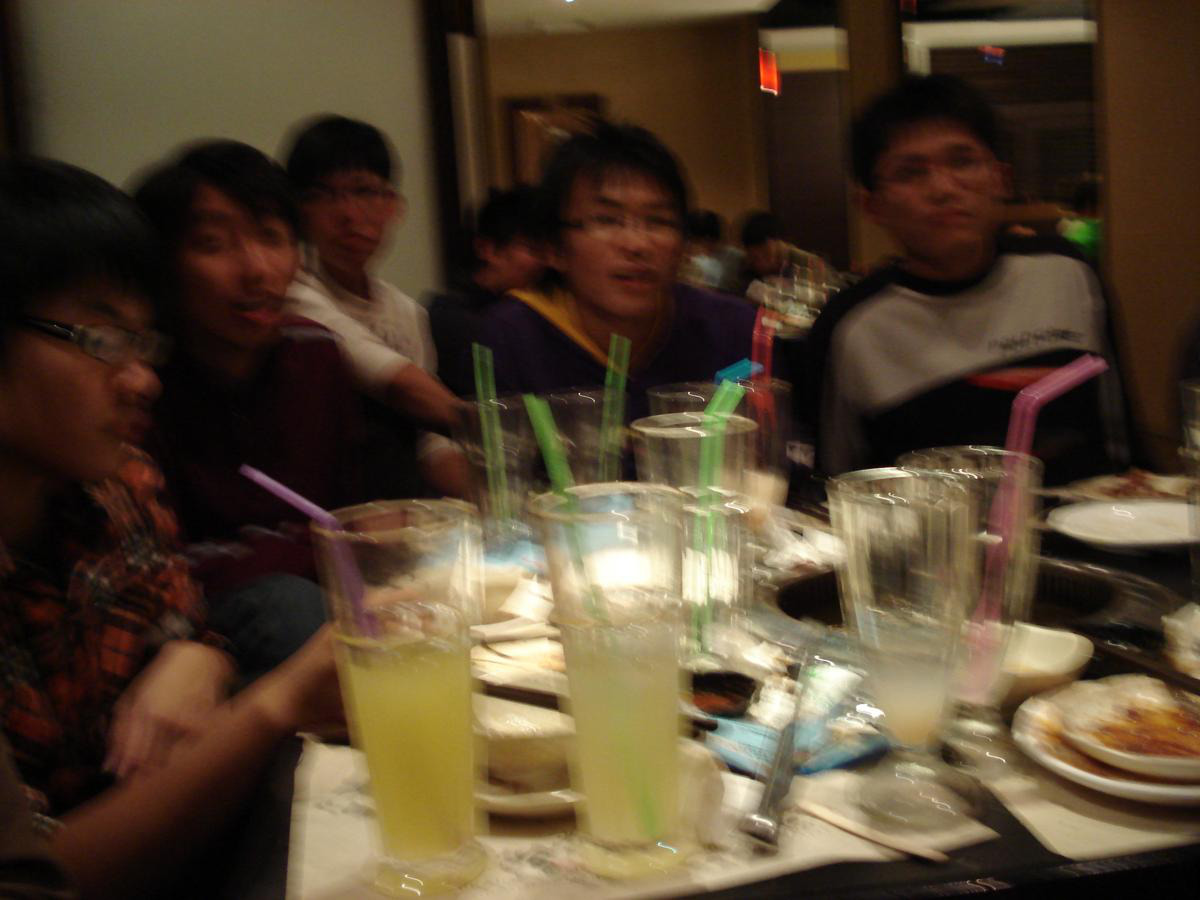} &
\includegraphics[trim=300 280 80 110, clip,width=0.19\textwidth]{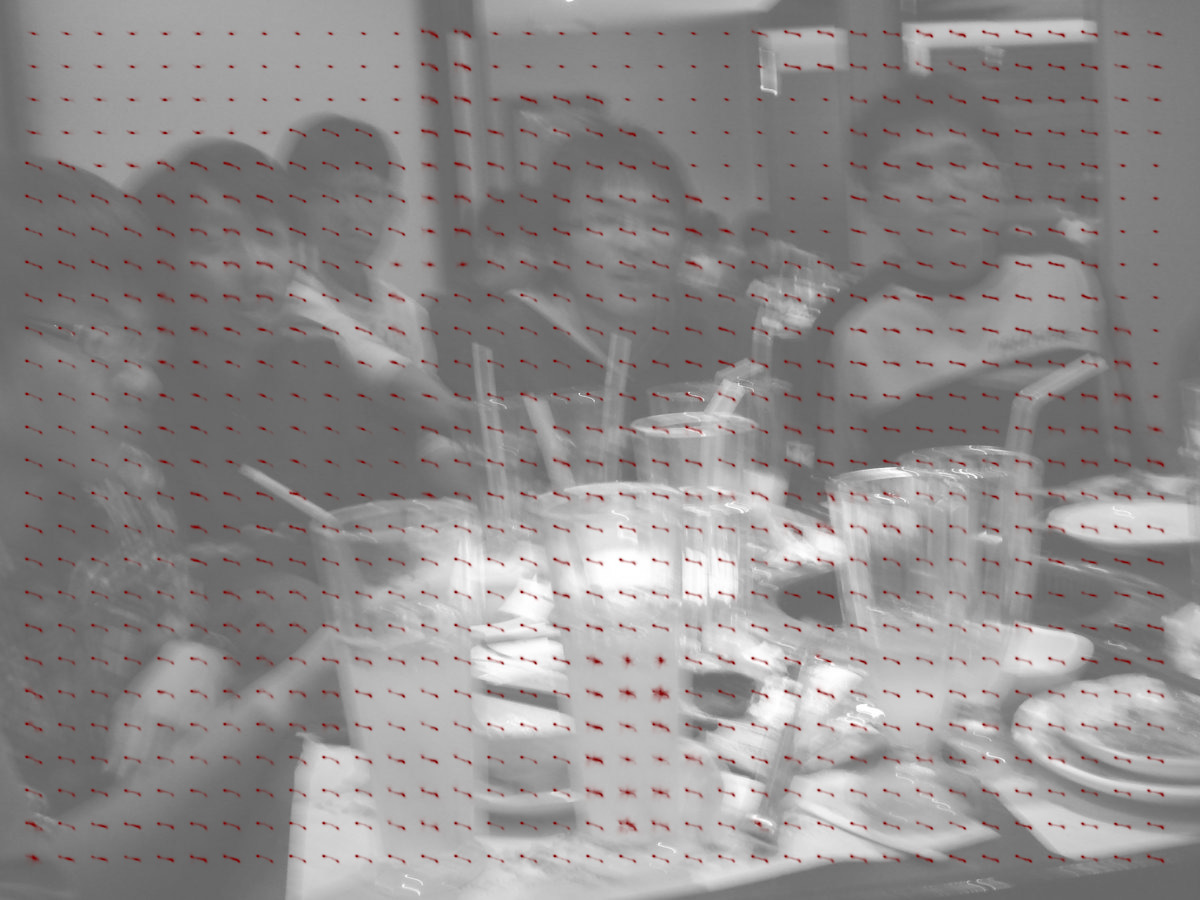} &   
\includegraphics[trim=300 280 80 110, clip,width=0.19\textwidth]{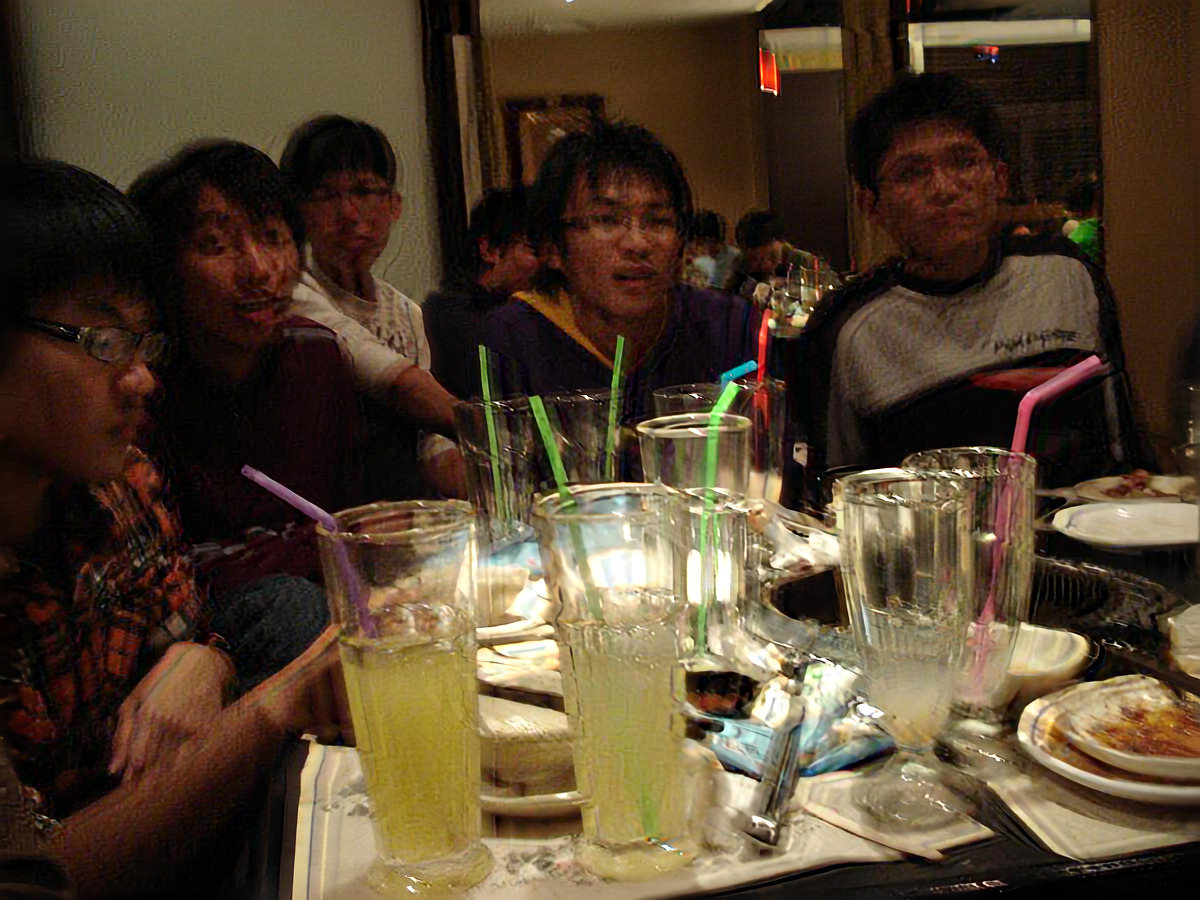} &
\includegraphics[trim=300 280 80 110, clip,width=0.19\textwidth]{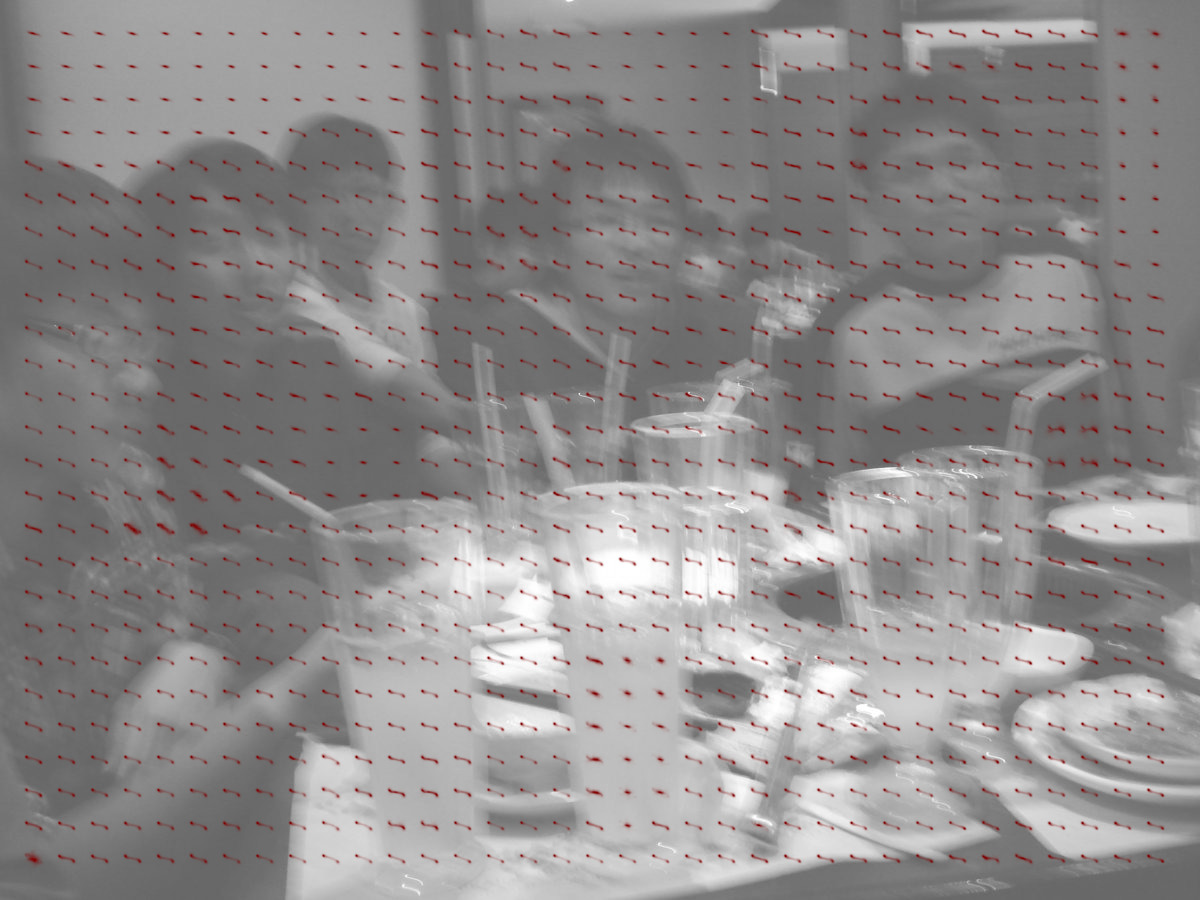} &
\includegraphics[trim=300 280 80 110, clip,width=0.19\textwidth]{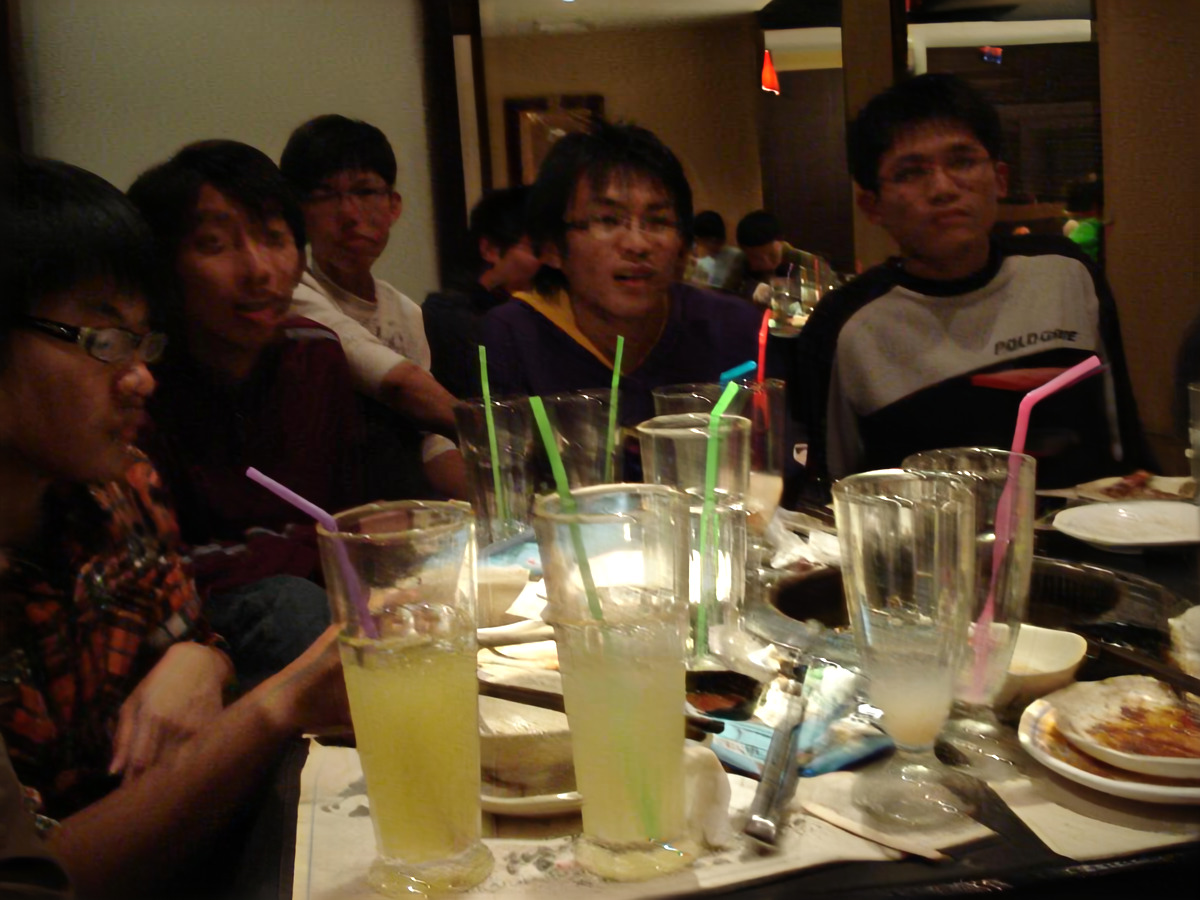} \\
    \end{tabular}
    \caption{\textbf{Comparison of blur kernels fields and restorations before and after joint training.} Without joint training, restorations are affected by artifacts, such as ringings and high-frequency patterns. Those are more prominent in regions where kernels are not properly estimated. Both the kernels and restorations improve after joint training, especially on low-variance regions and saturated pixels. Best viewed in electronic format.  %
    }
    \label{fig:ft_comparison}
\end{figure*}

%% file: figure_MIMO_vs_JMKPD_Kohler.tex
\begin{figure*}[h]
  \centering
  \scriptsize
\setlength{\tabcolsep}{2pt}
  \begin{tabular}{*{4}{c}}
            Blurry & MIMO-UNet+ (SBDD) & J-MKPD & Kernels \\
    \includegraphics[trim=250 370 200 100, clip, width=0.23\textwidth]{imgs/Kohler/Blurry/Blurry2_4.jpg}   &
      \includegraphics[trim=250 370 200 100, clip,width=0.23\textwidth]{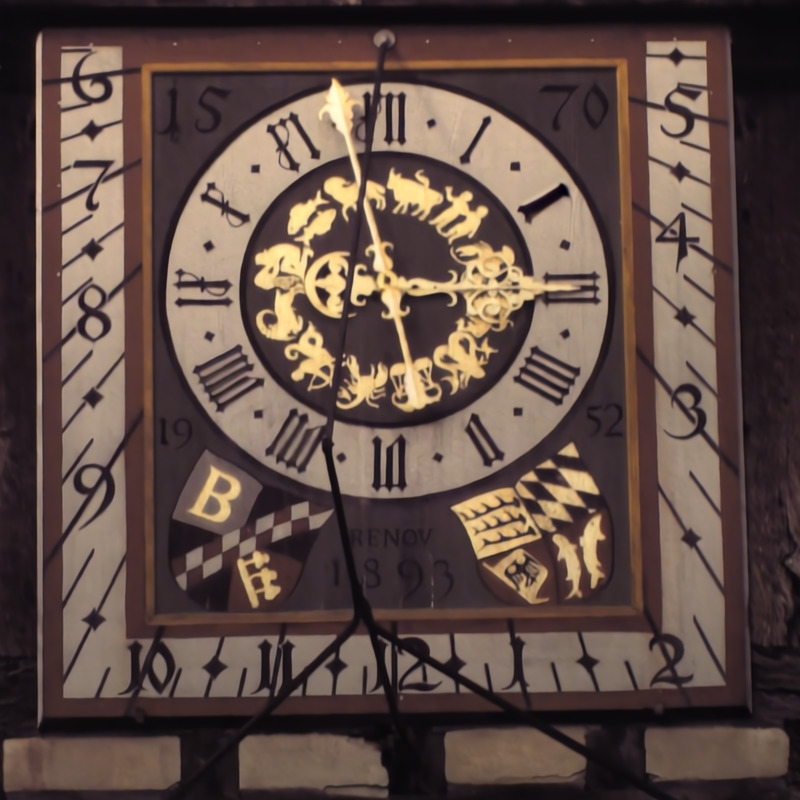}   &
   \includegraphics[trim=250 370 200 100, clip,width=0.23\textwidth]{imgs/Kohler/COCO900_restL2_aug_all_loss_gf1_80k/restored/Blurry2_4.jpg}   &
   \includegraphics[trim=250 370 200 100, clip,width=0.23\textwidth]{imgs/Kohler/COCO900_restL2_aug_all_loss_gf1_80k/kernels/Blurry2_4_kernels.jpg} \\
   &   30.84 dB & 31.79 dB &  \\
    \includegraphics[trim=250 120 200 350, clip, width=0.23\textwidth]{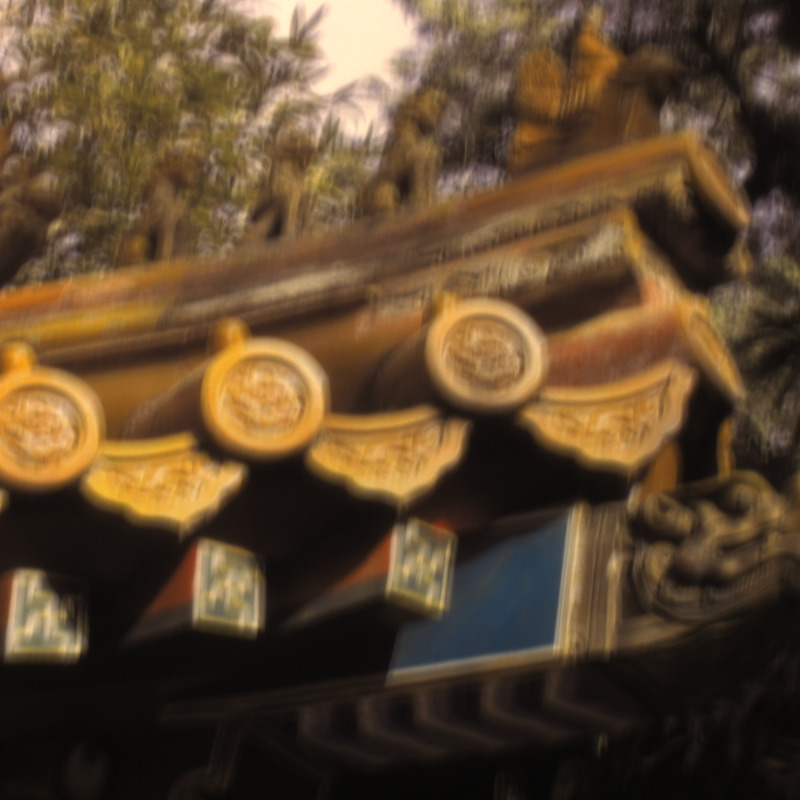}   &
      \includegraphics[trim=250 120 200 350, clip,width=0.23\textwidth]{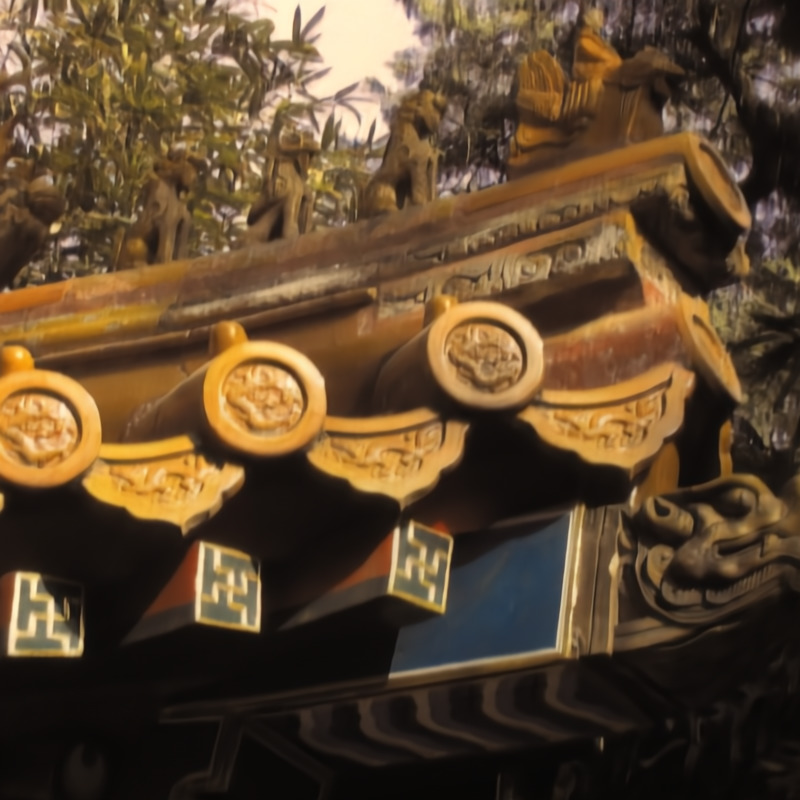}   &
   \includegraphics[trim=250 120 200 350, clip,width=0.23\textwidth]{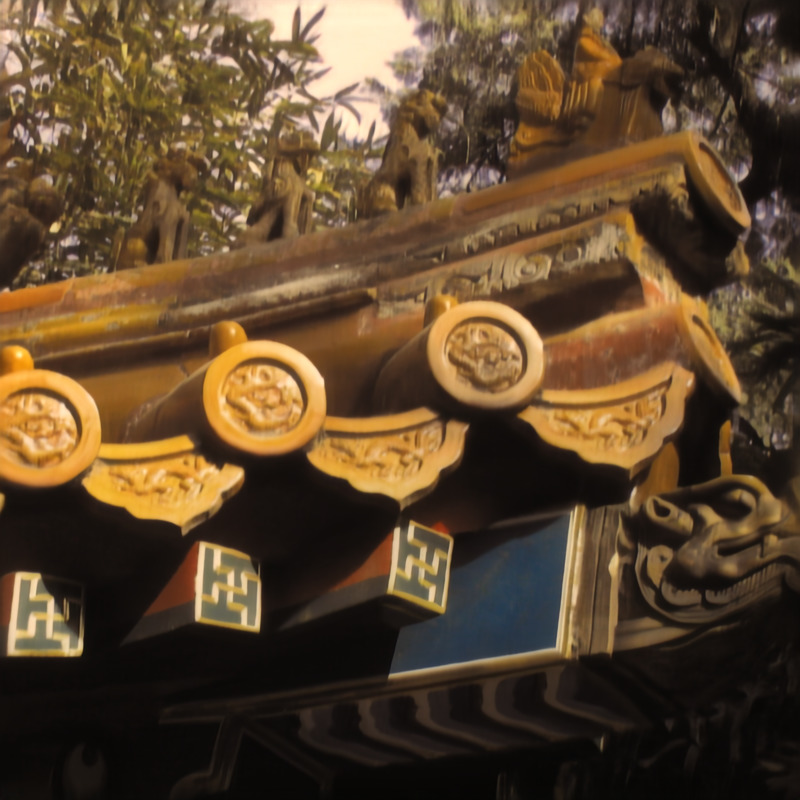}   &
   \includegraphics[trim=250 120 200 350, clip,width=0.23\textwidth]{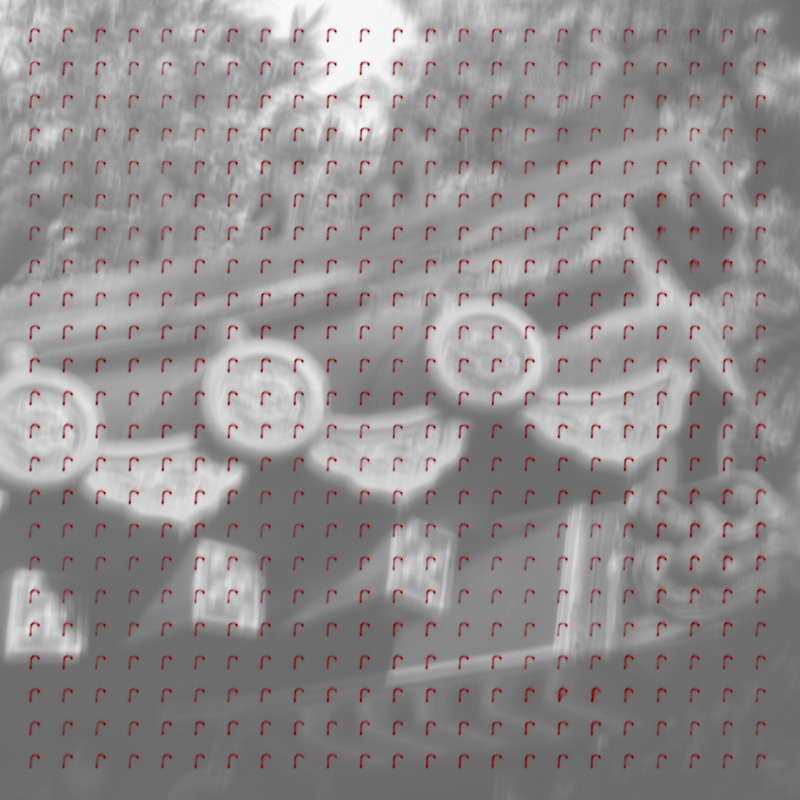} \\
   &   29.21 dB & 29.99 dB &  \\
  \end{tabular}
  \caption{\textbf{Comparison of deblurring on K\"ohler's dataset.} \textbf{First row:} The global constraint imposed by the model favors smooth kernel fields that in turn are helpful to avoid artifacts. MIMO-UNet+ restoration around number II looks sharp but implies an unrealistic motion field. \textbf{Second row:} Well-estimated kernels in combination with a model-based restoration tend to produce sharper natural results as reflected by all the reference-free metrics.}
  \label{fig:MIMO_vs_JMKPD_Kohler_better}
\end{figure*}

%% file: figure_MIMO_vs_JMKPD_RealBlur.tex
\begin{figure*}[h]
  \centering
  \scriptsize
\setlength{\tabcolsep}{2pt}
  \begin{tabular}{cc|*{3}{c}}
            Blurry & MIMO-UNet+ & MIMO-UNet+ (\emph{SBDD}) & J-MKPD & Kernels  \\
    \includegraphics[trim=330 330 100 200, clip, width=0.19\textwidth]{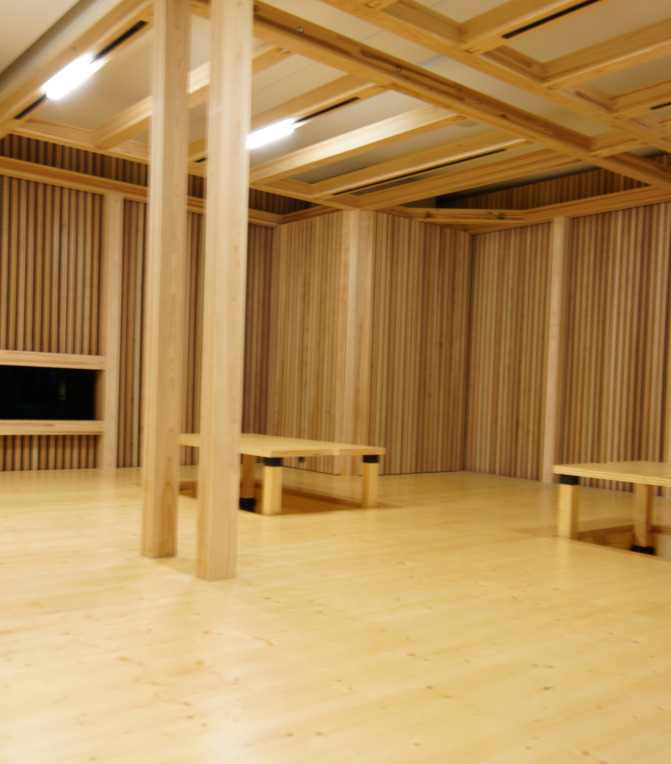}   &
        \includegraphics[trim=330 330 100 200, clip,width=0.19\textwidth]{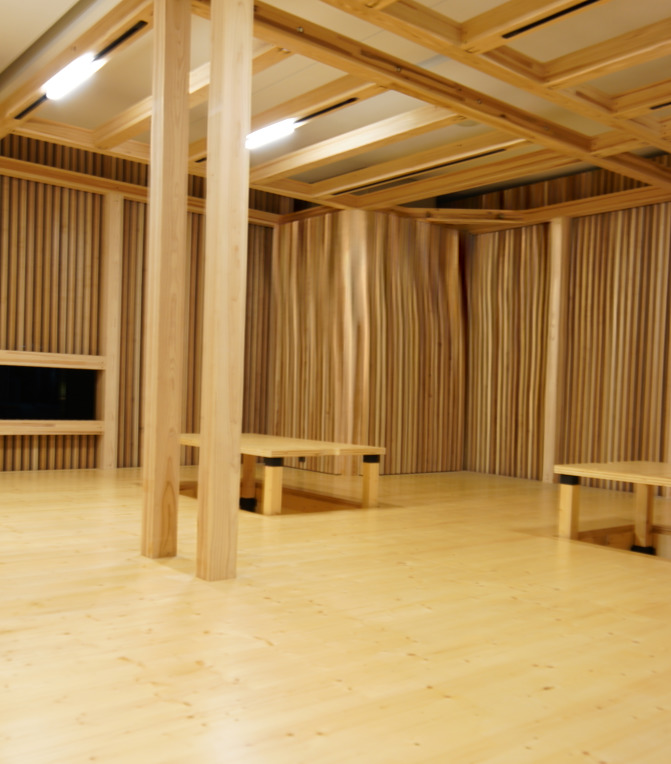}       &
      \includegraphics[trim=330 330 100 200, clip,width=0.19\textwidth]{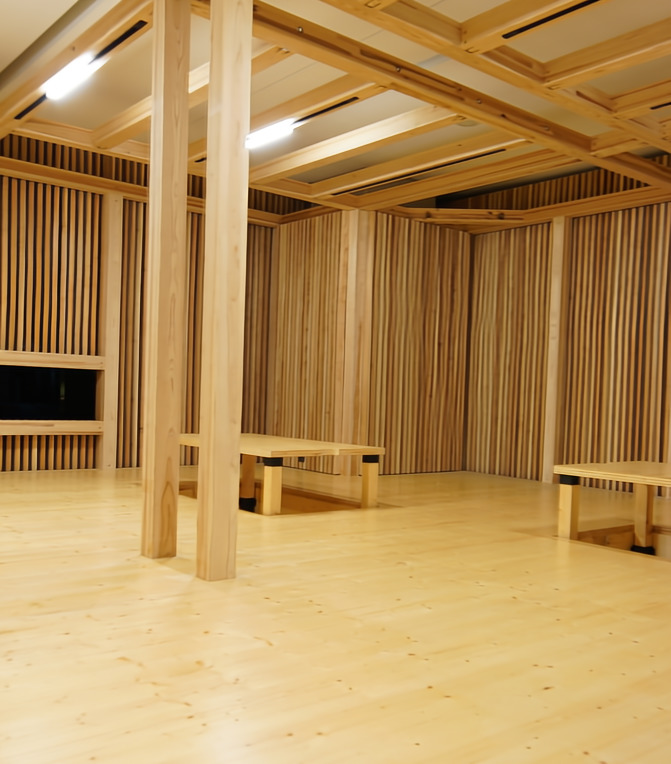}   &
   \includegraphics[trim=330 330 100 200, clip,width=0.19\textwidth]{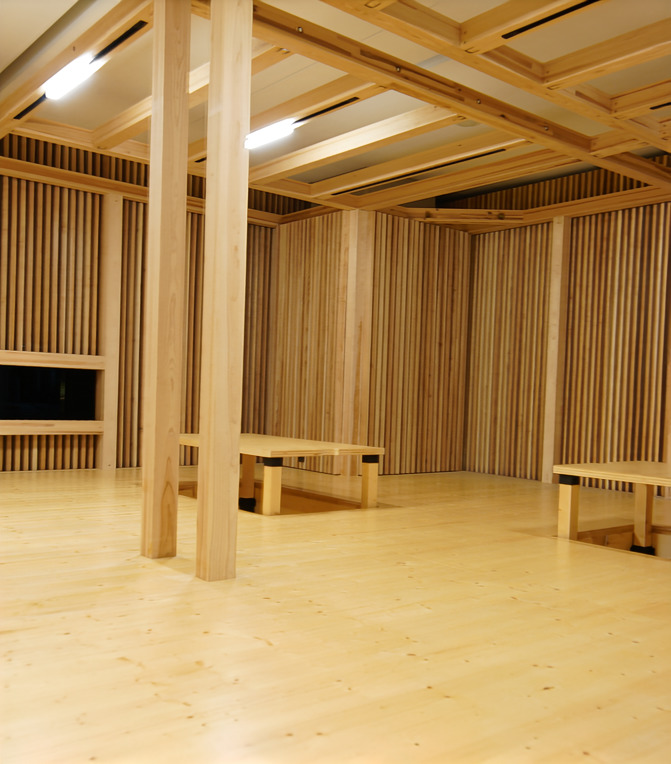} &
      \includegraphics[trim=330 330 100 200, clip,width=0.19\textwidth]{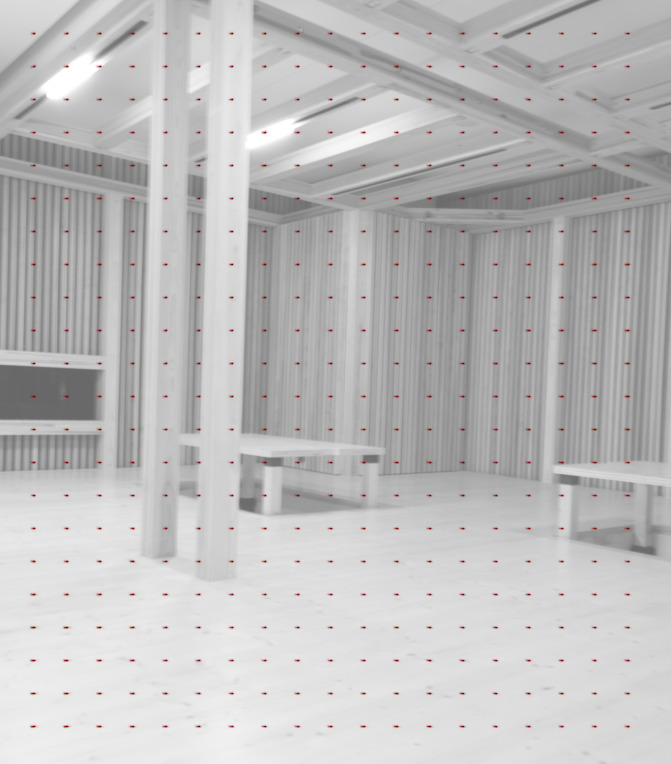} 
   \\
   & 25.21 dB &  24.11 dB & 27.67 dB &  \\
      \includegraphics[trim=220 390 160 120, clip, width=0.19\textwidth]{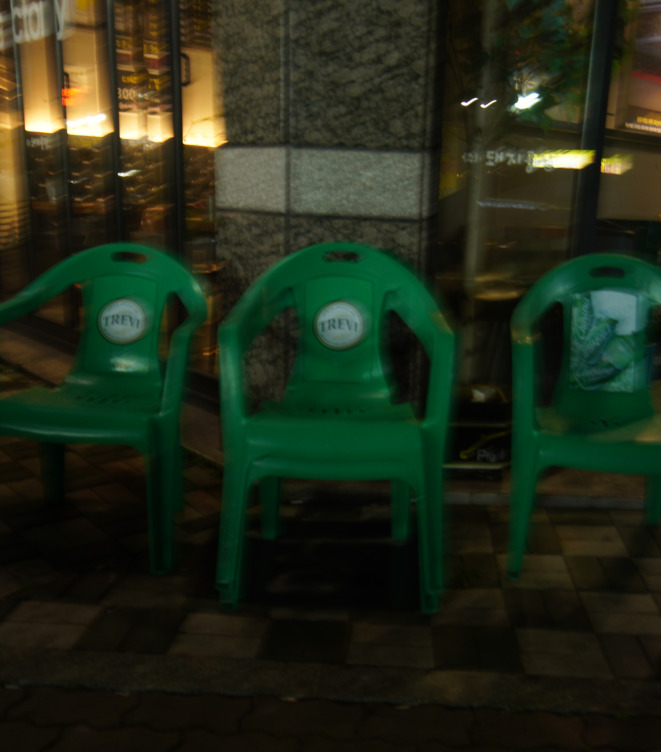}   &
       \includegraphics[trim=220 390 160 120, clip,width=0.19\textwidth]{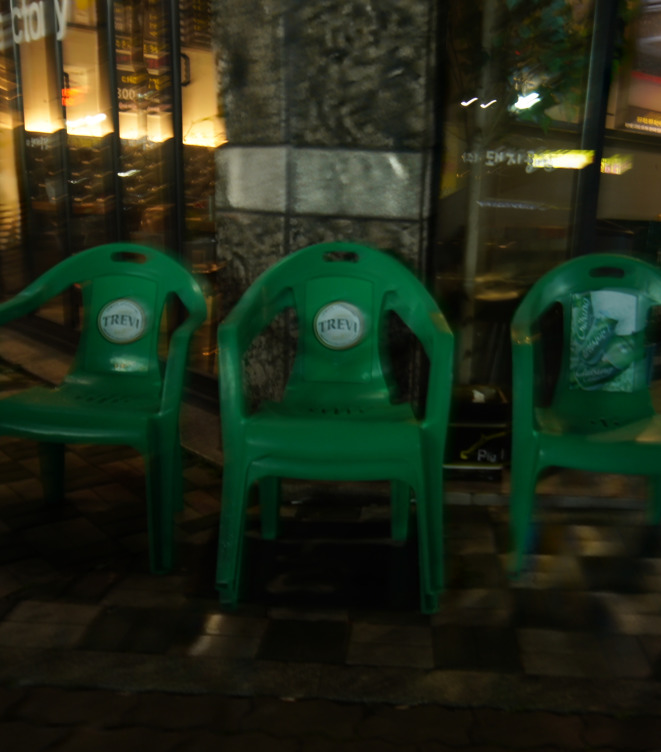}        &
      \includegraphics[trim=220 390 160 120, clip,width=0.19\textwidth]{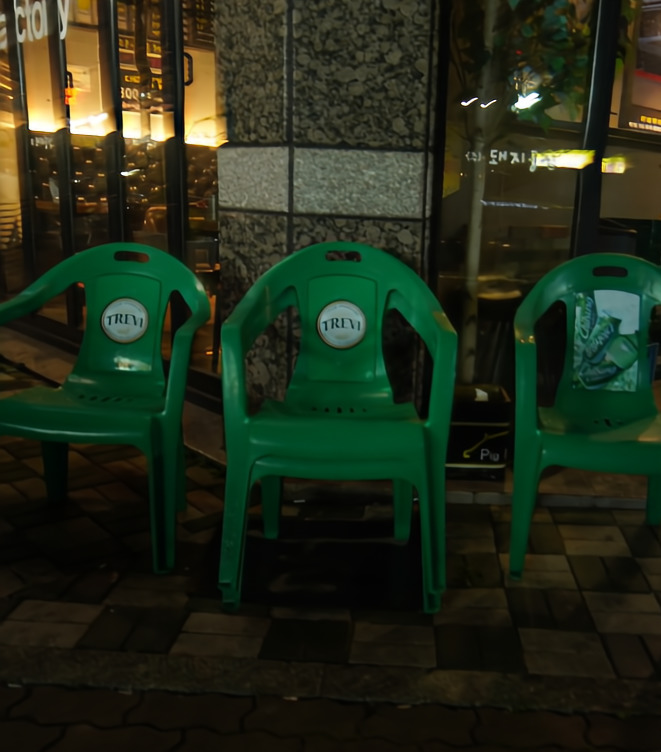}   &
   \includegraphics[trim=220 390 160 120, clip,width=0.19\textwidth]{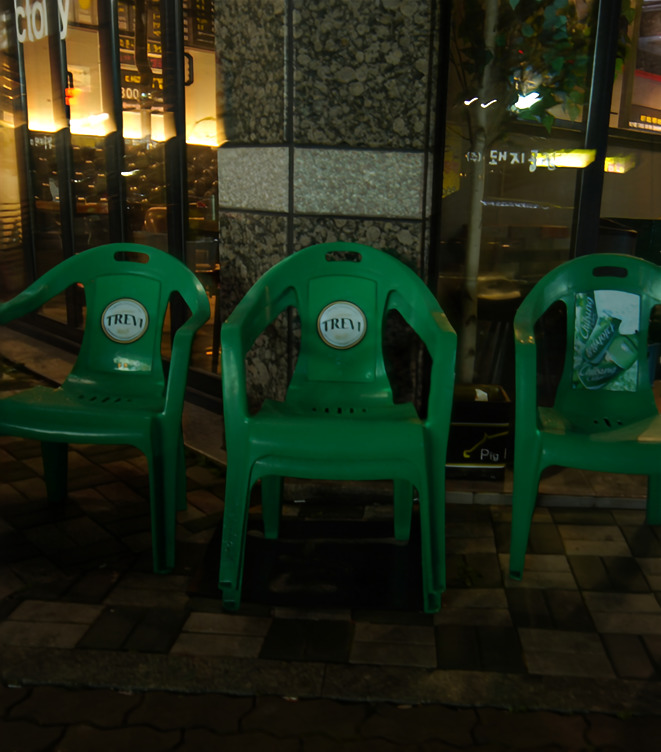} &
      \includegraphics[trim=220 390 160 120, clip,width=0.19\textwidth]{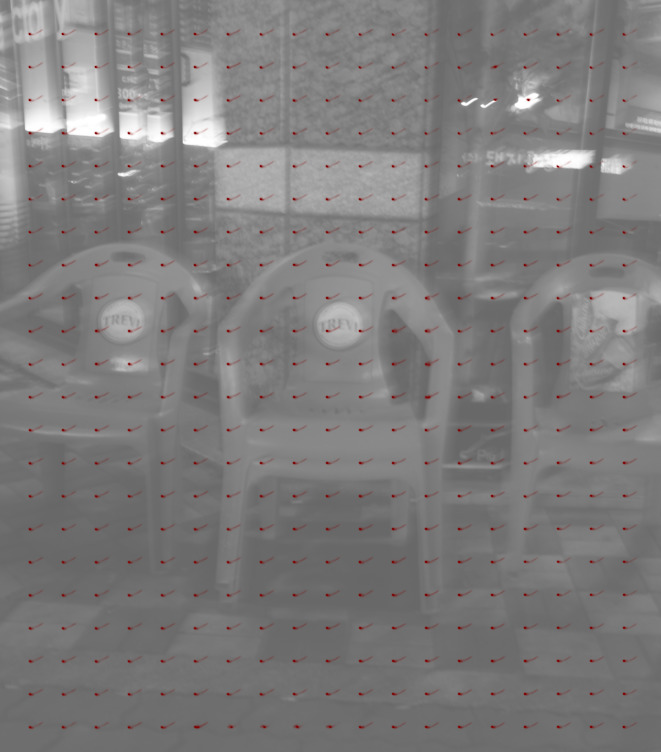} \\
   & 26.25 dB  &   28.56 dB & 28.38 dB &  \\
      \includegraphics[trim=0 290 260 200, clip, width=0.19\textwidth]{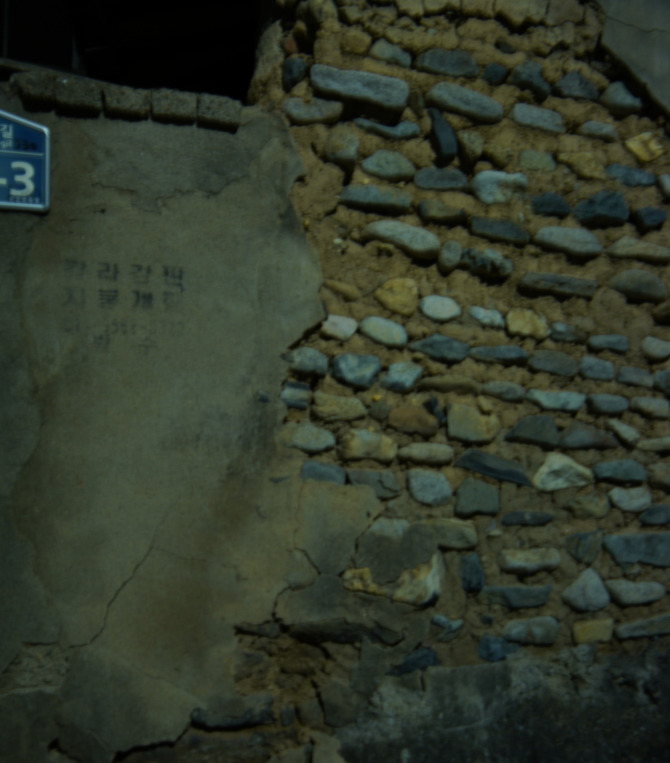}   &
       \includegraphics[trim=0 290 260 200, clip,width=0.19\textwidth]{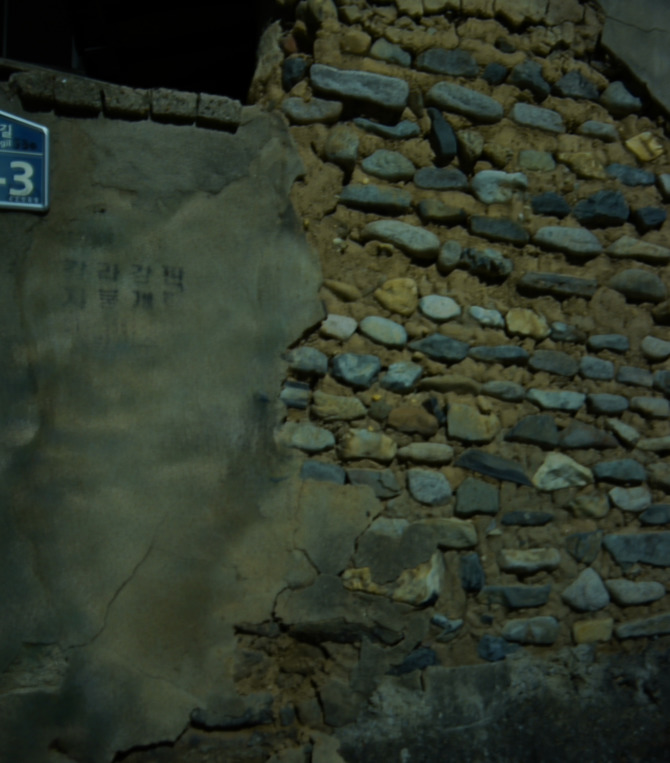}        &
      \includegraphics[trim=0 290 260 200, clip,width=0.19\textwidth]{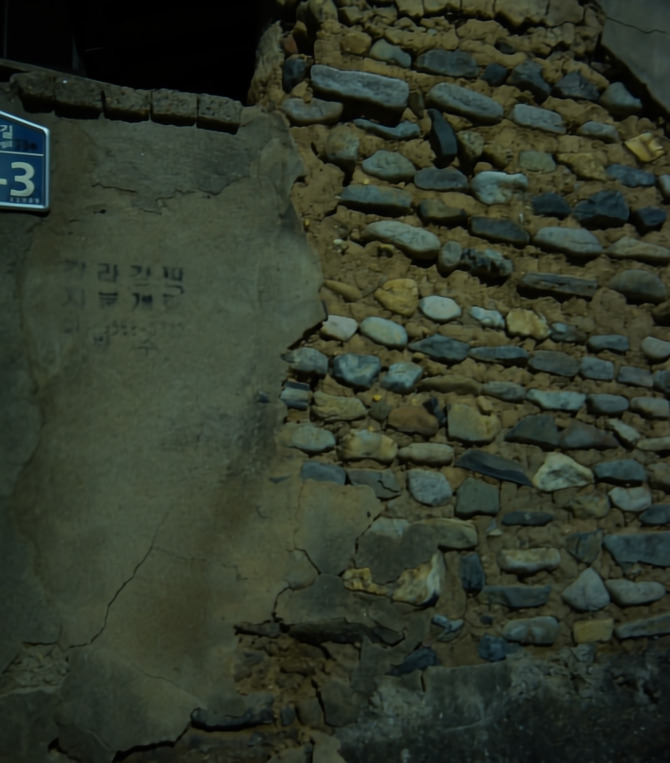}   &
   \includegraphics[trim=0 290 260 200, clip,width=0.19\textwidth]{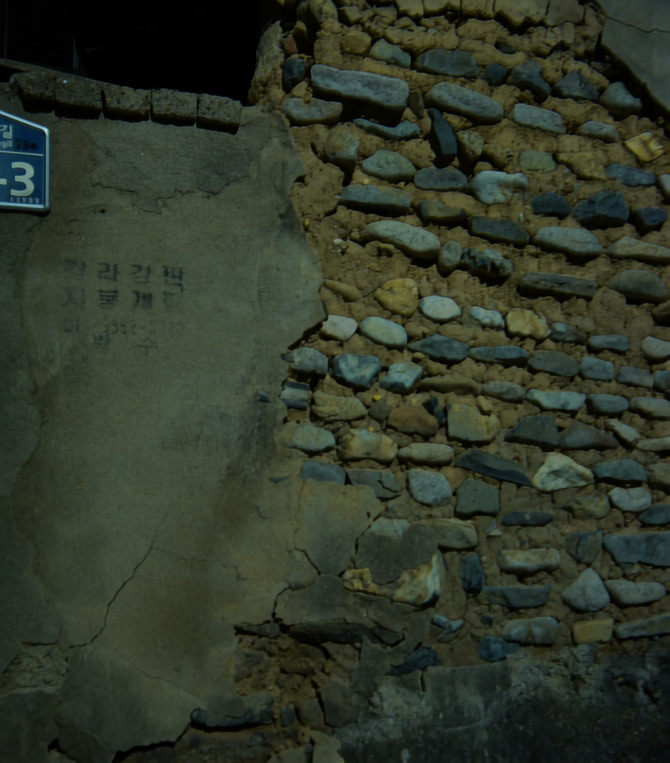} &
      \includegraphics[trim=0 290 260 200, clip,width=0.19\textwidth]{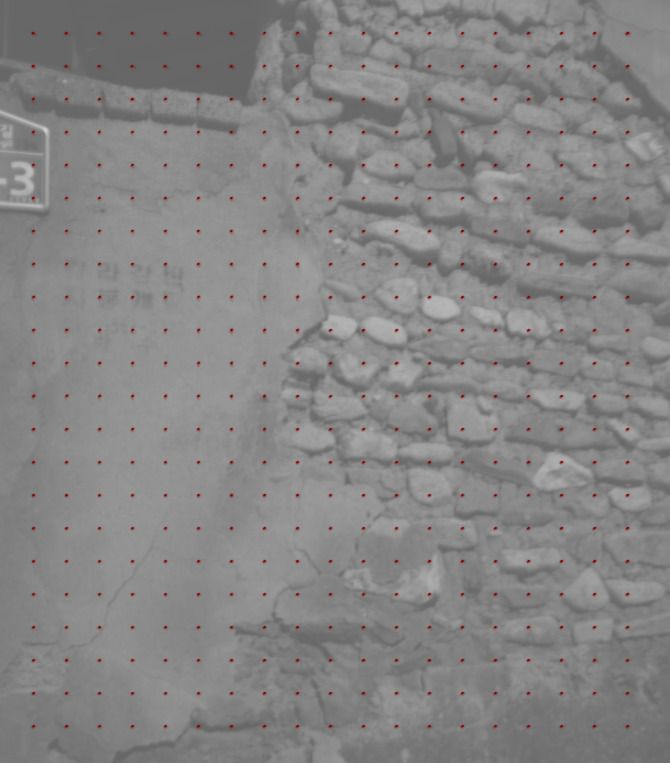} 
   \\
      & 33.51 dB & 33.94 dB & 33.32 dB &  \\
    \includegraphics[trim=250 370 200 200, clip, width=0.19\textwidth]{imgs/RealBlur/Blurry/scene002_blur_10.jpg}   &
    \includegraphics[trim=250 370 200 200, clip,width=0.19\textwidth]{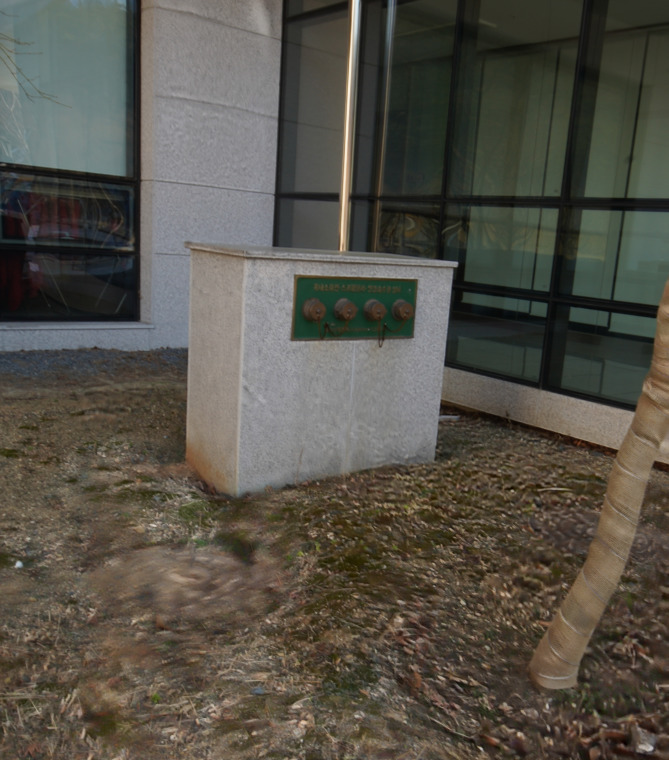}   &
      \includegraphics[trim=250 370 200 200, clip,width=0.19\textwidth]{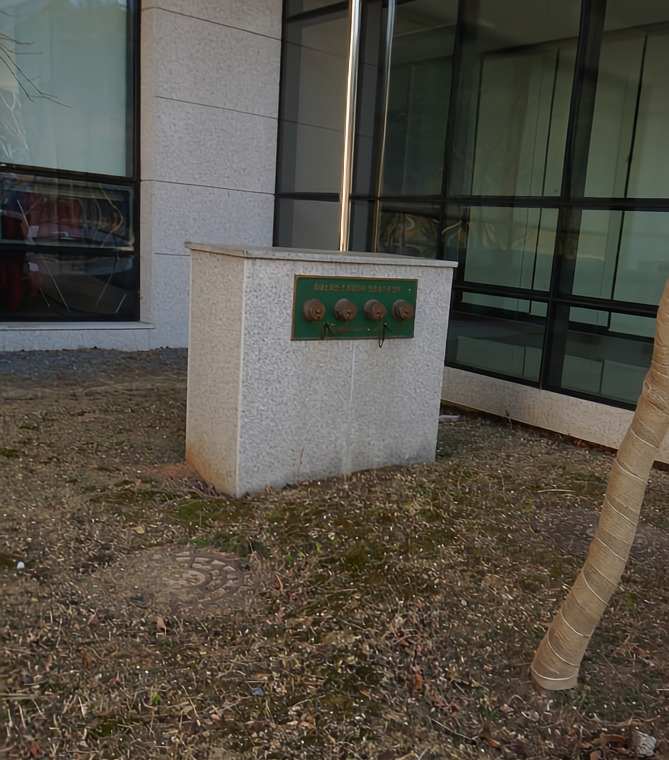}   &
   \includegraphics[trim=250 370 200 200, clip,width=0.19\textwidth]{imgs/RealBlur/COCO900_restL2_aug_all_loss_gf1_80k/restored/scene002_blur_10.jpg}   &
   \includegraphics[trim=250 370 200 200, clip,width=0.19\textwidth]{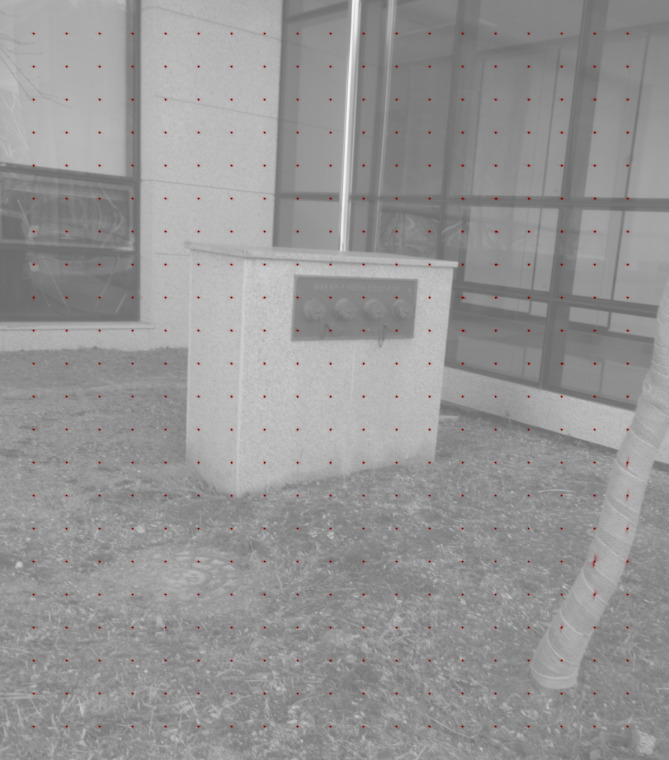} \\
      &  26.61 dB &  27.76 dB & 29.50 dB &  \\
    \includegraphics[trim=170 320 300 310, clip, width=0.19\textwidth]{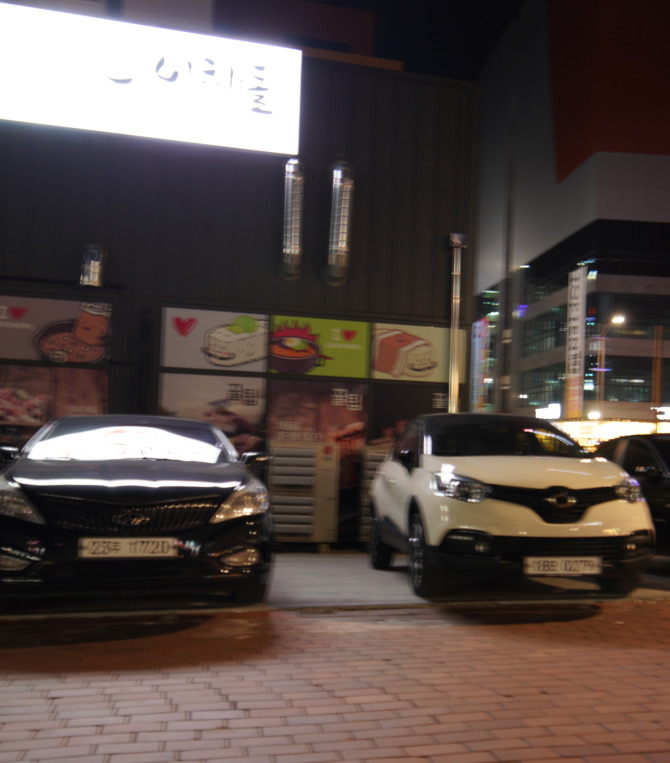}   &
       \includegraphics[trim=170 320 300 310, clip,width=0.19\textwidth]{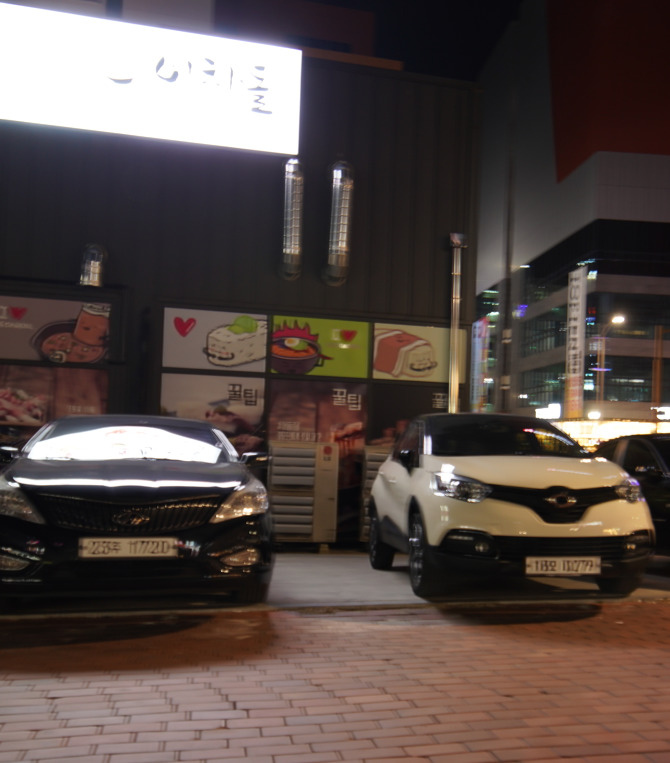}   &                                                         \includegraphics[trim=170 320 300 310, clip,width=0.19\textwidth]{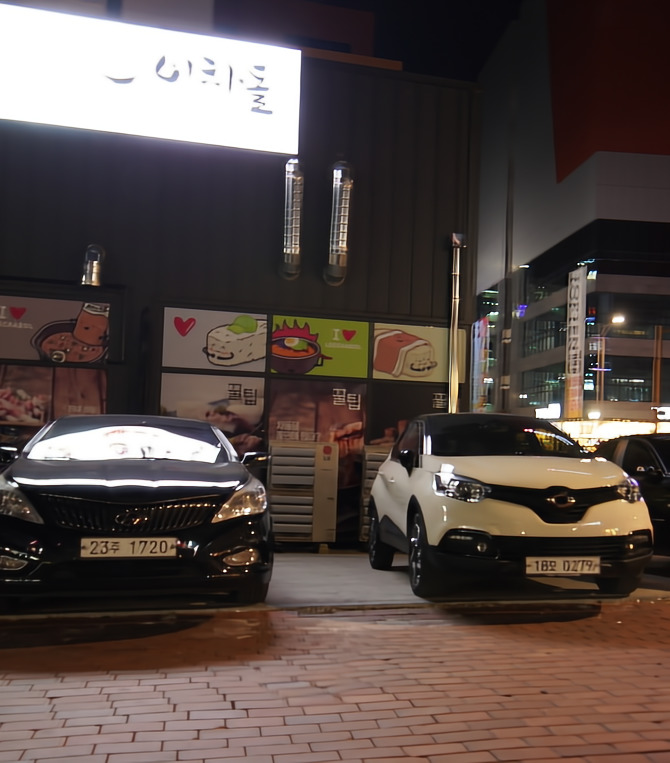}   &
   \includegraphics[trim=170 320 300 310, clip,width=0.19\textwidth]{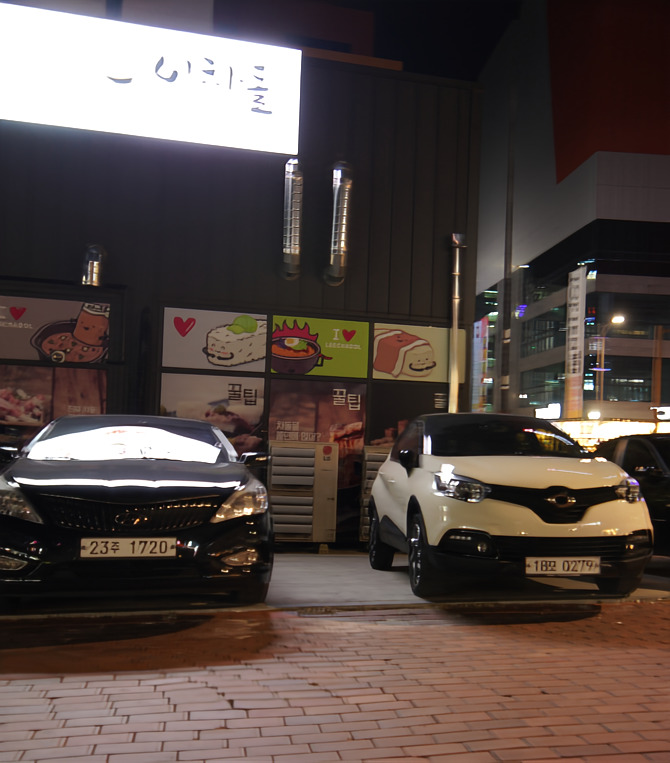} &
      \includegraphics[trim=170 320 300 310, clip,width=0.19\textwidth]{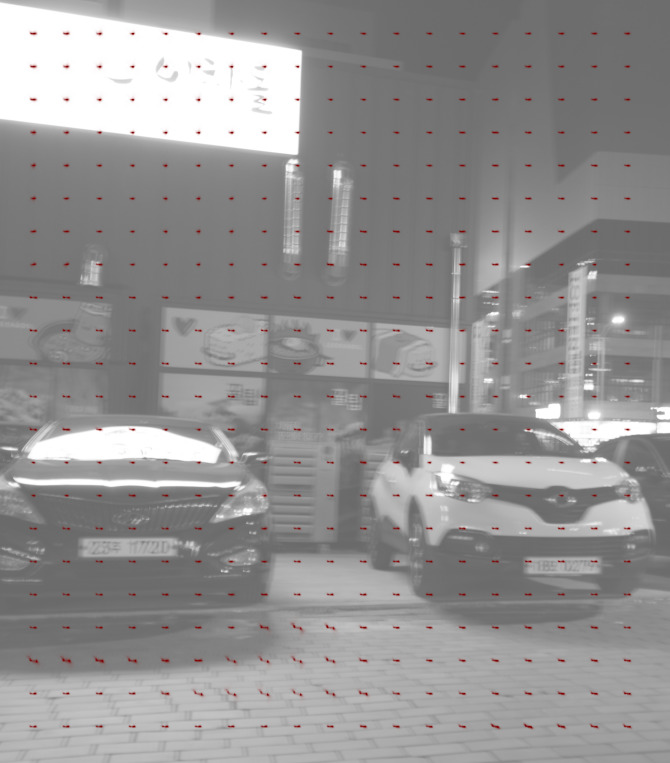} 
   \\
   &  26.58 dB &   26.24 dB & 26.84 dB &  \\
    \includegraphics[trim=170 500 210 140, clip, width=0.19\textwidth]{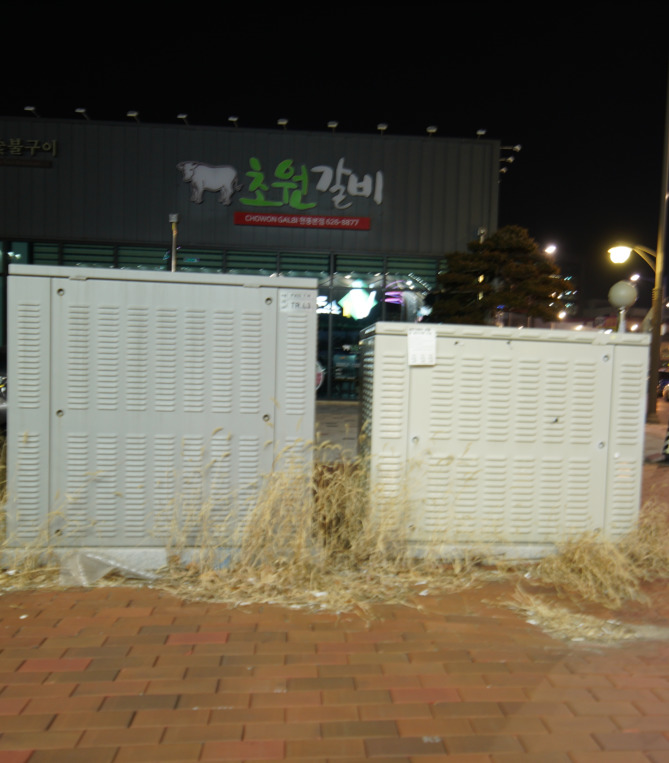}   &
       \includegraphics[trim=170 500 210 140, clip,width=0.19\textwidth]{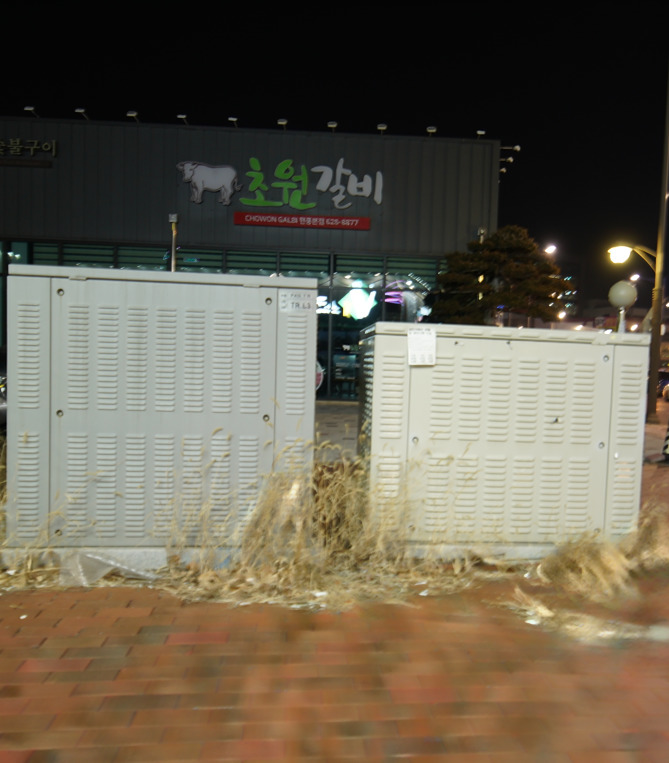}                                                                                                                         &
      \includegraphics[trim=170 500 210 140, clip,width=0.19\textwidth]{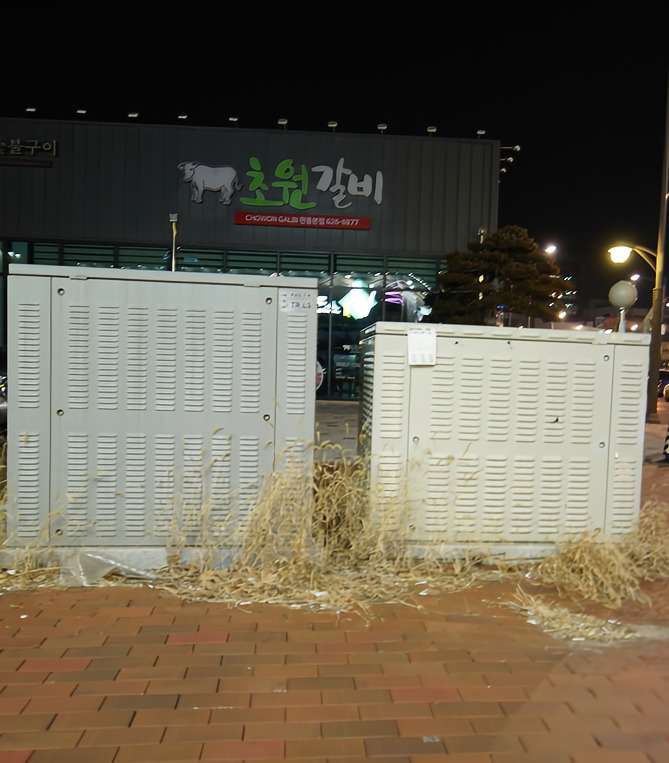}   &
   \includegraphics[trim=170 500 210 140, clip,width=0.19\textwidth]{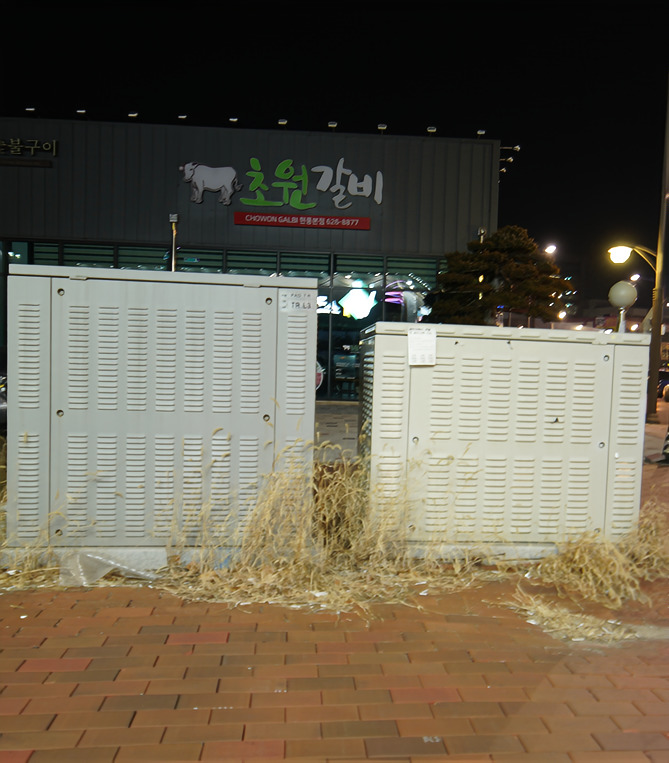} &
      \includegraphics[trim=170 500 210 140, clip,width=0.19\textwidth]{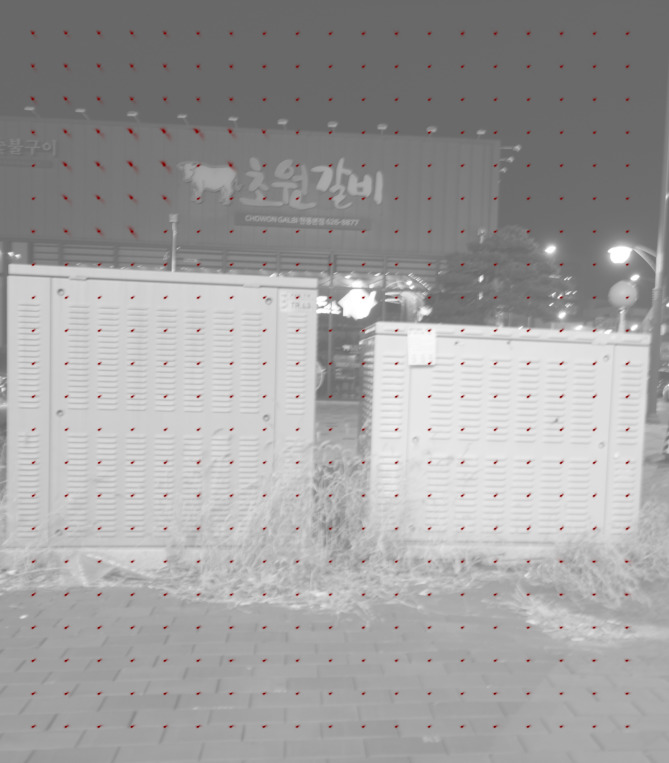} 
   \\
   &  &   29.66 dB & 30.02 dB &  \\
    \includegraphics[trim=310 170 100 420, clip, width=0.19\textwidth]{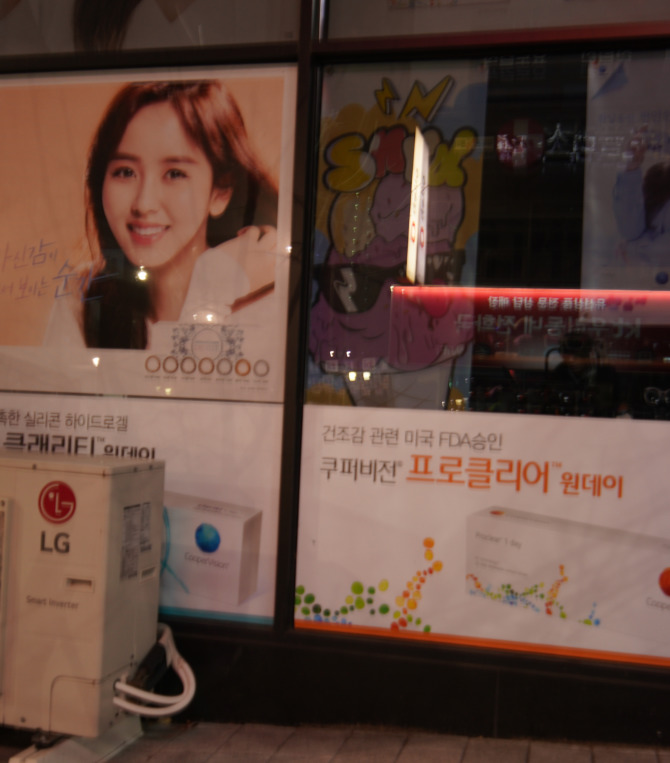}   &
       \includegraphics[trim=310 170 100 420, clip,width=0.19\textwidth]{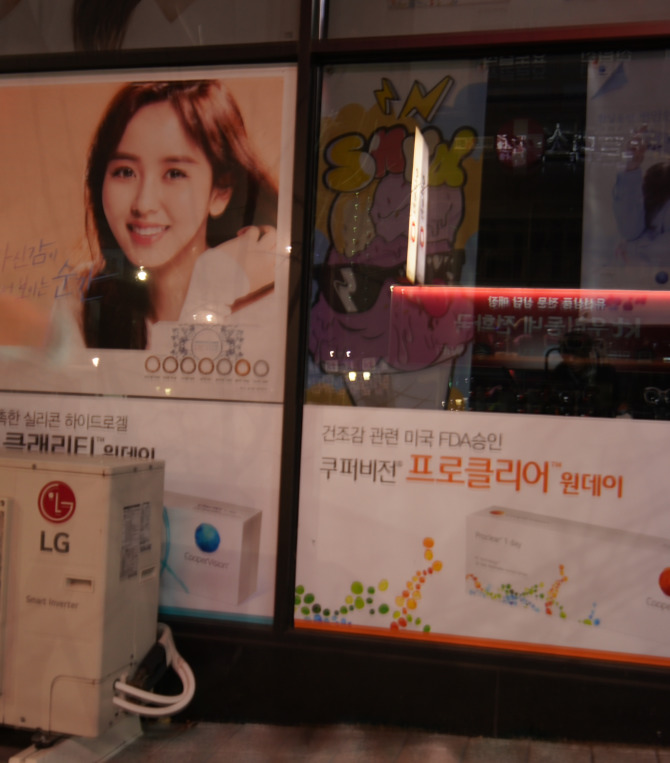}                                                                                                                         &
      \includegraphics[trim=310 170 100 420, clip,width=0.19\textwidth]{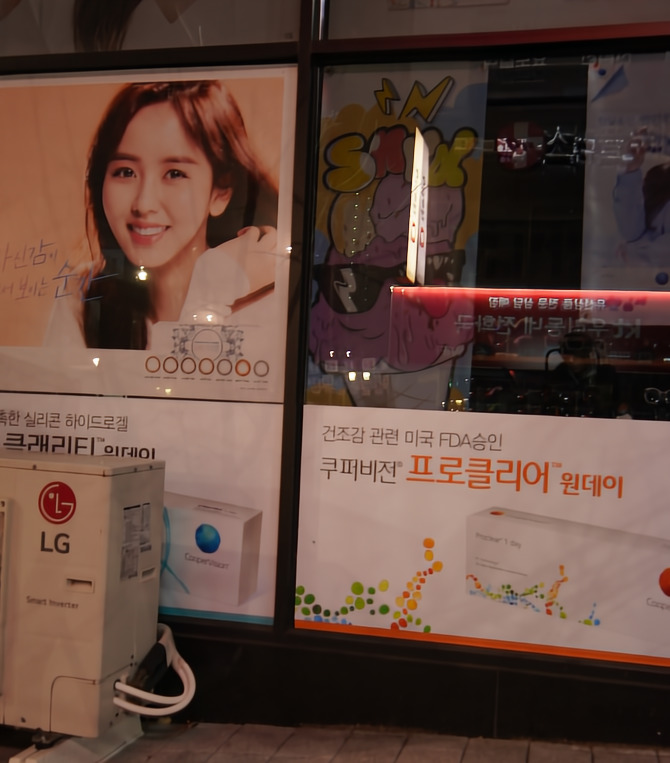}   &
   \includegraphics[trim=310 170 100 420, clip,width=0.19\textwidth]{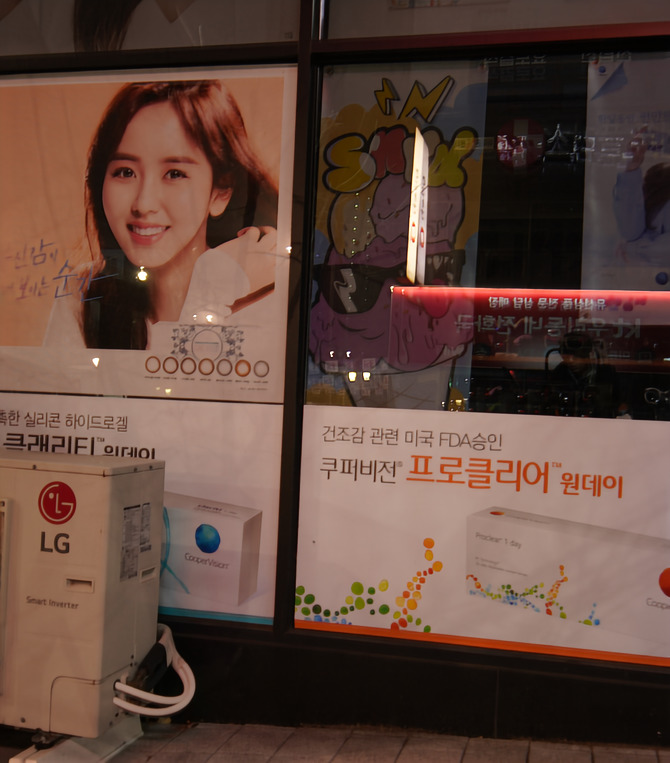} &
      \includegraphics[trim=310 170 100 420, clip,width=0.19\textwidth]{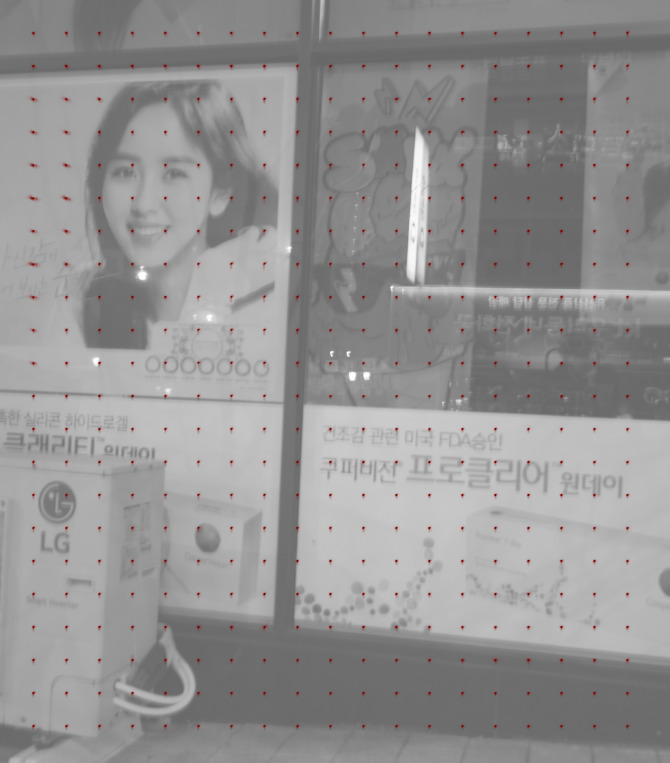} 
   \\
   &  30.66 dB &   30.40 dB & 30.56 dB &  \\
   \end{tabular}
  \caption{\textbf{Comparison of deblurring on the RealBlur dataset~\cite{rim_2020_ECCV}.} The original MIMO-UNet+ model, trained on GoPro, often does not generalize well. When trained on SBDD, its generalization performance increases, but the method is still prone to produce artifacts on repetitive patterns. J-MKPD enforces a spatially-varying motion kernel field that prevents this kind of artifact. J-MKPD also produces sharper and more precise image restorations under mild blur conditions. In case artifacts may arise due to the difficulty in the estimation, in J-MKPD they are easily detected by observing inconsistencies in the kernel fields. As a side comment, note how the PSNR fails to reflect the quality of the restorations. Best viewed in electronic format. \label{fig:MIMO_vs_JMKPD_RealBlur_better}}
\end{figure*}

%% file: section_extensions.tex
\section{Extensions to the method \label{sec:extensions}}

The proposed approach is flexible enough to be adapted to other image restoration problems, thanks to the possibility to modify, at the same time, the synthetic dataset generation process and the degradation model. Two such problems are:

\paragraph{Motion deblurring in the presence of light streaks} Although our model can deal well with saturated pixels, we notice that the KPN is prone to estimate \textit{delta kernels} in the presence of highly saturated punctual lighting sources.   

As a preliminary exploration, we trained our model introducing a data augmentation step that randomly adds light streaks in randomly selected sharp images. {We refer to this augmented dataset as \emph{SBDD-LS}. As shown in \cref{fig:sat_images}, this minor modification improves the estimation of kernels in saturated regions and the restorations. Without this augmentation step, kernels predicted in saturated regions are often deltas. On the RealBlur dataset, which includes a significant number of saturated images, the PSNR gets increased by 0.1 dB. This improvement seems incremental, but this is mainly due to the fact that light streaks usually cover small areas. The visual impact in photographs captured under low-light conditions is significant, as shown in \cref{fig:sat_images} \footnote{The \cref{fig:sat_images} are at half resolution because the size of the kernels at full resolution is larger than we can predict with our network.}. 
\input{figure_saturated_images}

 \paragraph{Joint motion deblurring and super-resolution} The proposed deblurring pipeline can be easily modified to deblur and super-resolve images simultaneously. Adding a down-sampling by a factor of $\alpha$ in the degradation model (\ref{eq:model}) yields
    \begin{equation}
    \mathbf{v}_{i,j} = \langle \mathbf{u}_{i,j}, \mathbf{k}_{i,j} \rangle \downarrow_{\alpha} + n_{i,j}.
    \label{eq:model_sr}
\end{equation}
This way, the high-resolution versions of the motion blur kernels and the latent sharp images can be recovered from the low-resolution, blurry observations. To illustrate the capacity of the method to jointly deblur and super-resolve images, we trained a model using the previous degradation model. We refer to this model as J-MKPSR (Joint Motion Kernels Prediction and Super Resolution). The training set was generated following the same procedure described in \cref{subsec:DatasetGen}, but using \cref{eq:model_sr} instead of \cref{eq:model_sat} to generate the low-resolution images.  We refer to this dataset as \emph{SBDD-SR}.

This simple strategy produces excellent results, c.f. \cref{fig:superresolution}. 
\input{figure_super_resolution}

%% file: figure_saturated_images.tex
\begin{figure*}
  \centering
  \scriptsize
\setlength{\tabcolsep}{2pt}
  \begin{tabular}{*{5}{c}}
     Blurry & Kernels (\emph{SBDD}) & Restored (\emph{SBDD}) & Kernels (S\emph{SBDD-LS}) & Restored (\emph{SBDD-LS}) \\
     \includegraphics[trim= 0 0 0 10, clip, width=0.19\textwidth]{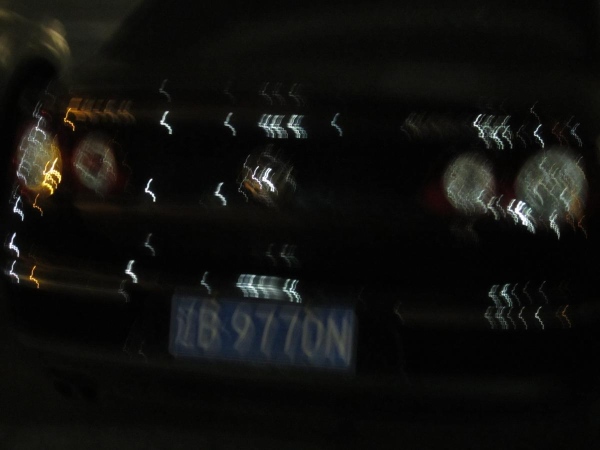}   &
    \includegraphics[trim= 0 0 0 10, clip,width=0.19\textwidth]{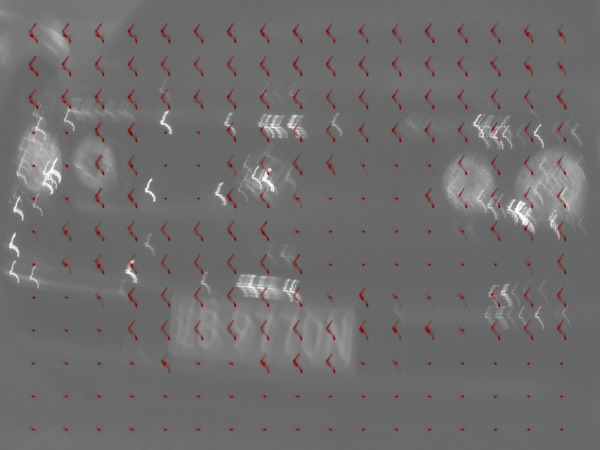}   &
    \includegraphics[trim= 0 0 0 10, clip,width=0.19\textwidth]{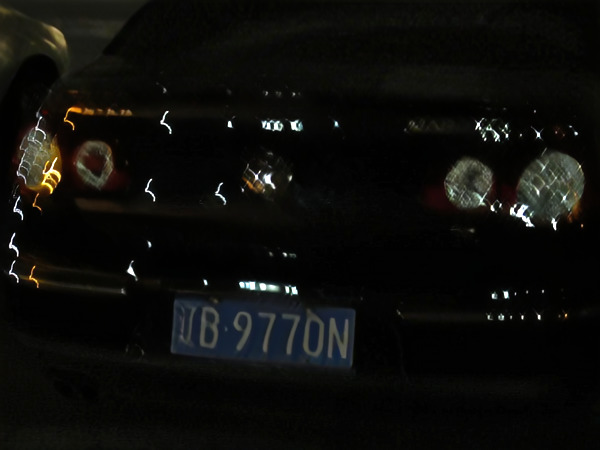}  &
    \includegraphics[trim= 0 0 0 10, clip,width=0.19\textwidth]{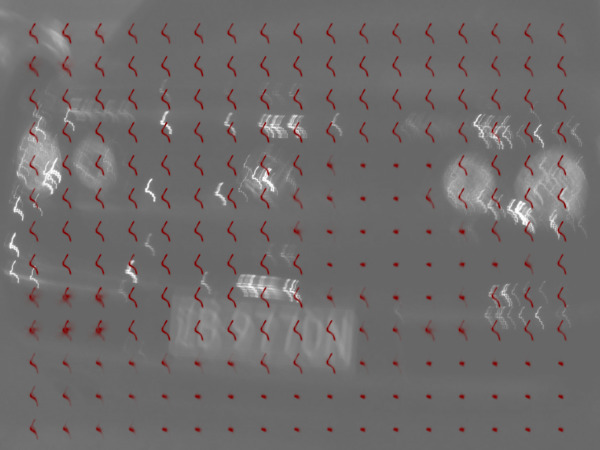}   &
   \includegraphics[trim= 0 0 0 10, clip,width=0.19\textwidth]{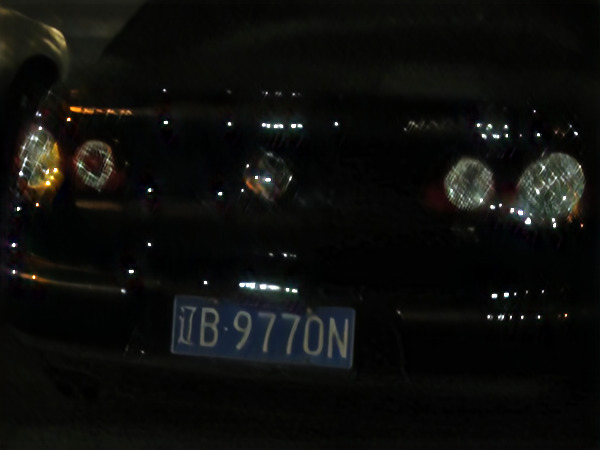}  \\
       \includegraphics[trim= 0 0 0 10, clip,width=0.19\textwidth]{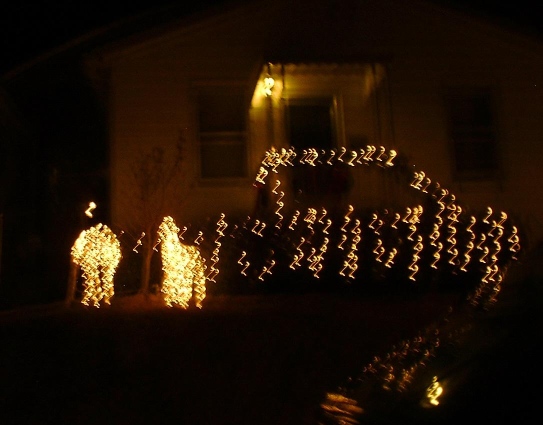}   &
      \includegraphics[trim= 0 0 0 10, clip,width=0.19\textwidth]{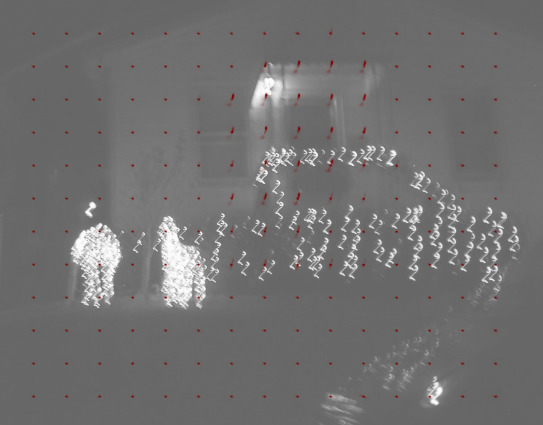}   &
   \includegraphics[trim= 0 0 0 10, clip,width=0.19\textwidth]{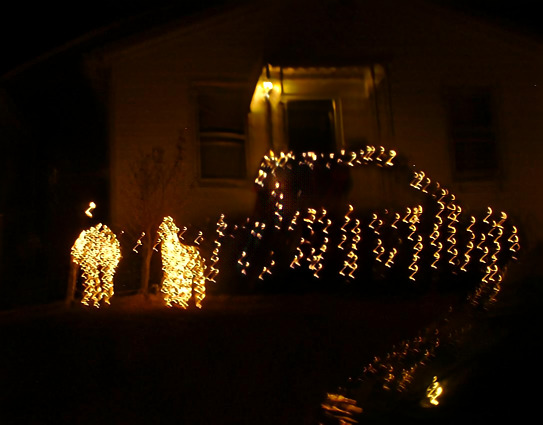}  &
      \includegraphics[trim= 0 0 0 10, clip,width=0.19\textwidth]{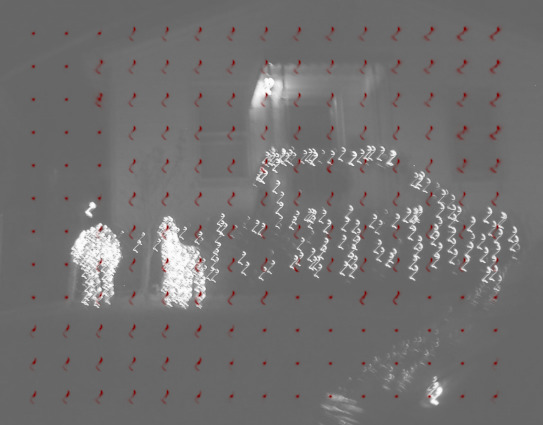}   &
   \includegraphics[trim= 0 0 0 10, clip,width=0.19\textwidth]{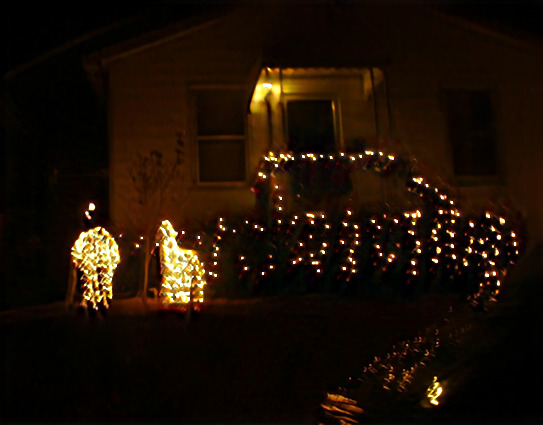}  \\
       \includegraphics[trim= 0 0 0 10, clip,width=0.19\textwidth]{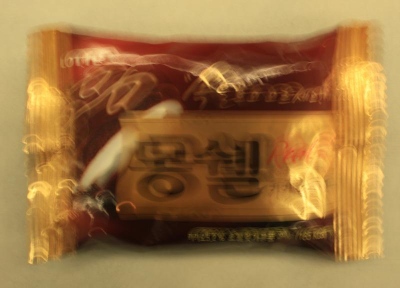}   &
      \includegraphics[trim= 0 0 0 10, clip,width=0.19\textwidth]{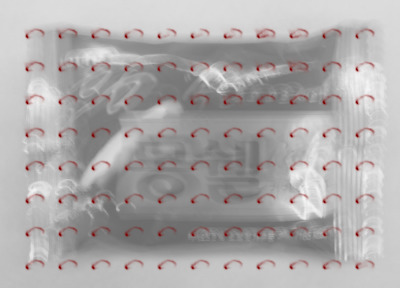}   &
   \includegraphics[trim= 0 0 0 10, clip,width=0.19\textwidth]{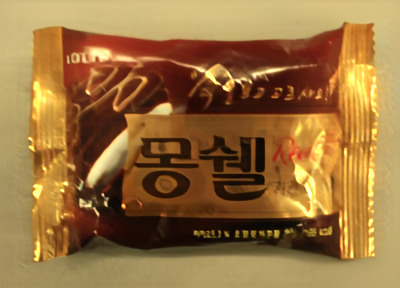}  &
      \includegraphics[trim= 0 0 0 10, clip,width=0.19\textwidth]{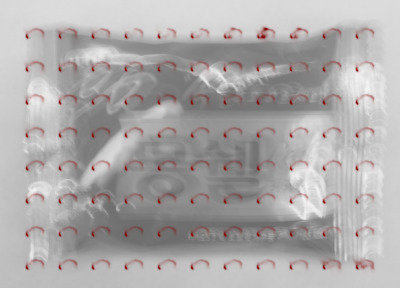}   &
   \includegraphics[trim= 0 0 0 10, clip,width=0.19\textwidth]{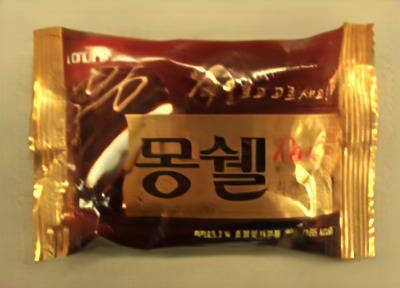}  \\
    \end{tabular}
    \caption{\textbf{Examples of restoration of real saturated images from Lai's dataset~\cite{lai2016comparative}}. Restorations obtained by J-MKPD when trained on two different synthetic datasets: the original \emph{SBDD} described in \cref{subsec:DatasetGen}, and an augmented version of SBDD which emphasizes the presence of light streaks (\emph{SBDD-LS}). 
    Although our model accounts for saturated pixels, when trained on \emph{SBDD} the network often predicts delta kernels on flat saturated regions. Results improve considerably by using the same degradation model and artificially adding light streaks on the training pairs. Best viewed in electronic format. \label{fig:sat_images}}
    \end{figure*}

%% file: figure_super_resolution.tex
\begin{figure*}
  \centering
  \scriptsize
\setlength{\tabcolsep}{2pt}

  \begin{tabular}{*{5}{c}}
     Blurry & J-MKPD Kernels & J-MKPSR Kernels  & J-MKPD restoration upsampled & J-MKPSR restoration \\
    \includegraphics[trim= 0 10 0 0, clip,width=0.19\textwidth]{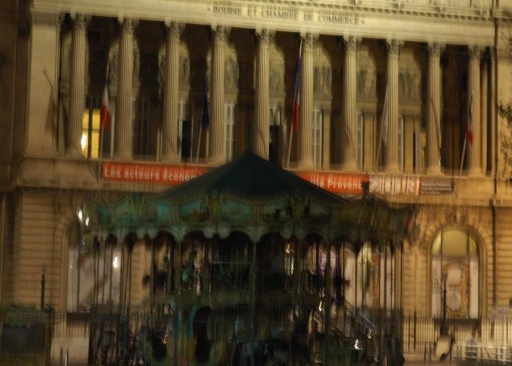}   &
      \includegraphics[trim= 0 10 0 0, clip, width=0.19\textwidth]{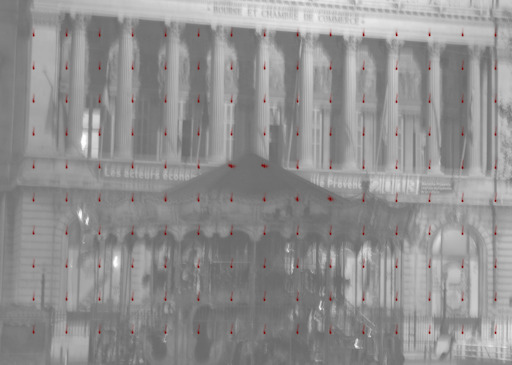}   &
   \includegraphics[trim= 0 10 0 0, clip, width=0.19\textwidth]{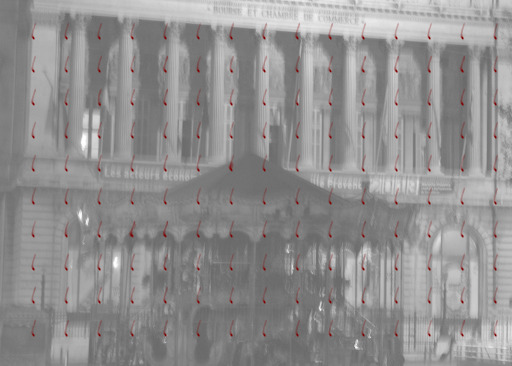}  &
      \includegraphics[trim= 0 10 0 0, clip, width=0.19\textwidth]{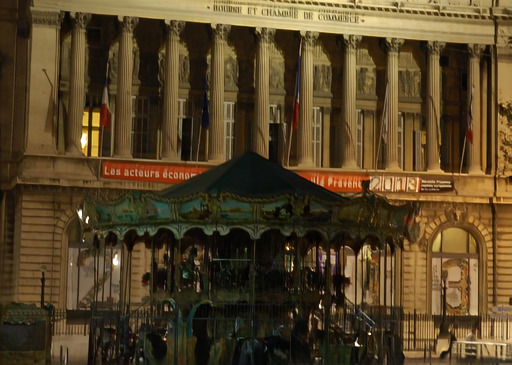}   &
   \includegraphics[trim= 0 20 0 0, clip, width=0.19\textwidth]{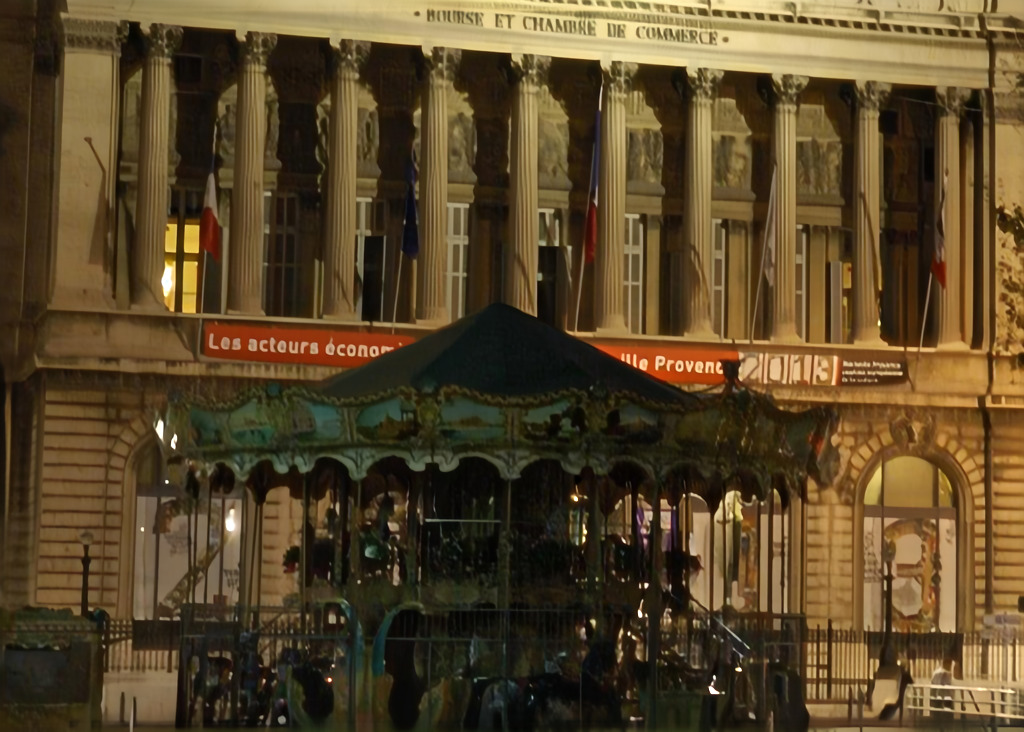}  \\
       \includegraphics[trim= 0 20 0 15, clip, width=0.19\textwidth]{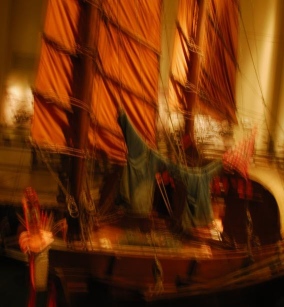}   &
      \includegraphics[trim= 0 25 0 20, clip, width=0.19\textwidth]{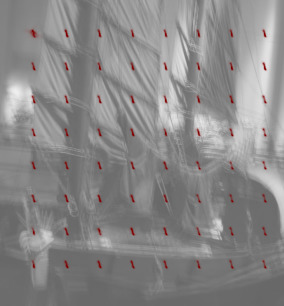}   &
   \includegraphics[trim= 0 25 0 20, clip, width=0.19\textwidth]{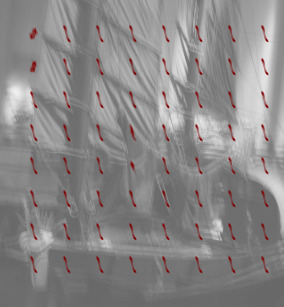}  &
      \includegraphics[trim= 0 25 0 20, clip, width=0.19\textwidth]{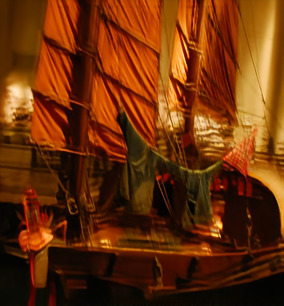}   &
   \includegraphics[trim= 0 50 0 40, clip, width=0.19\textwidth]{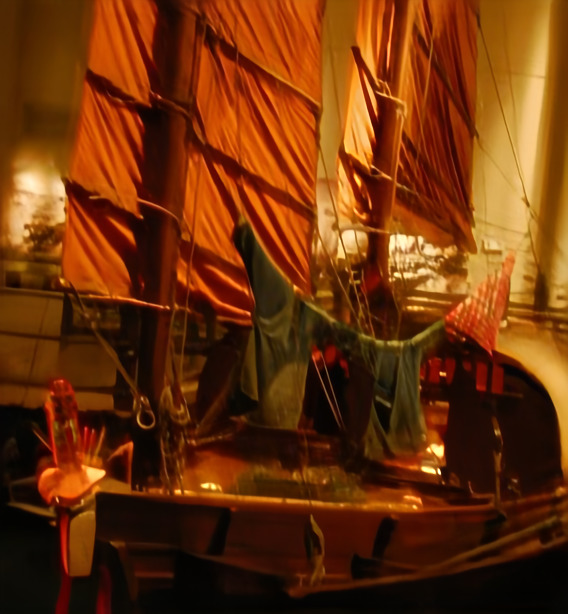}  \\
    \includegraphics[width=0.19\textwidth]{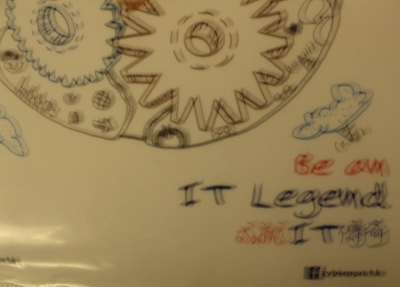}   &
    \includegraphics[width=0.19\textwidth]{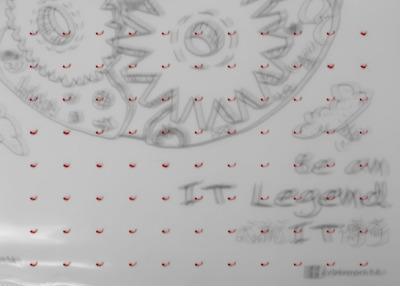}   &
   \includegraphics[width=0.19\textwidth]{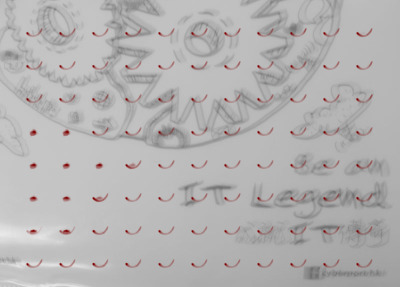}  &
      \includegraphics[width=0.19\textwidth]{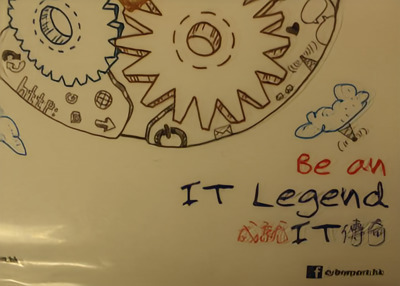}   &
   \includegraphics[width=0.19\textwidth]{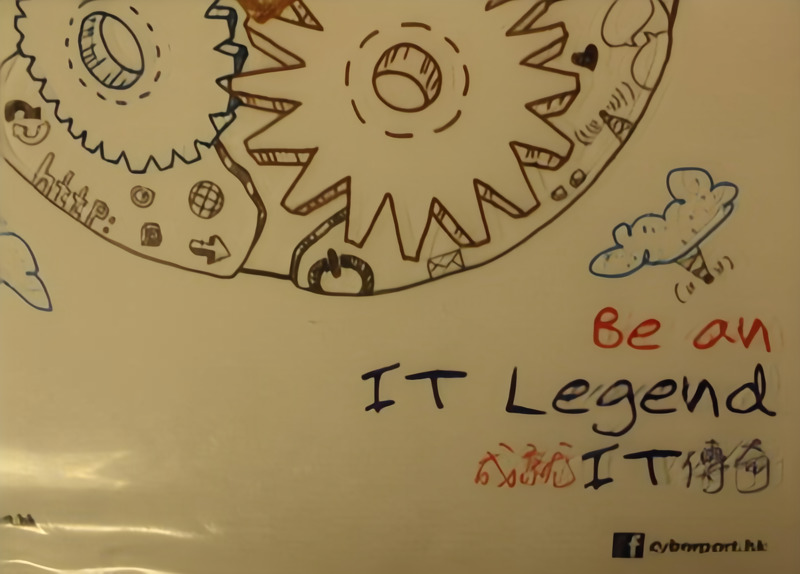}  \\
    \includegraphics[trim= 0 5 0 5, clip,width=0.19\textwidth]{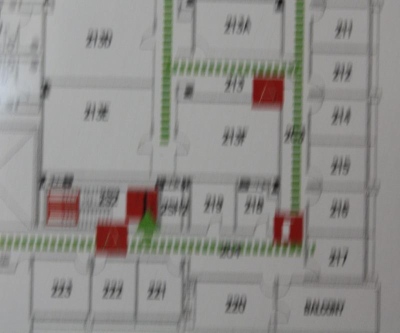}   &
    \includegraphics[trim= 0 5 0 5, clip,width=0.19\textwidth]{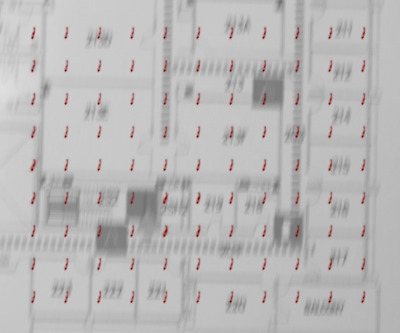}   &
   \includegraphics[trim= 0 5 0 5, clip,width=0.19\textwidth]{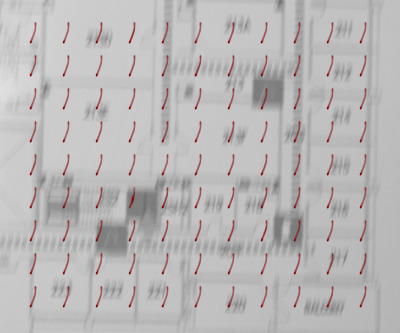}  &
      \includegraphics[trim= 0 5 0 5, clip,width=0.19\textwidth]{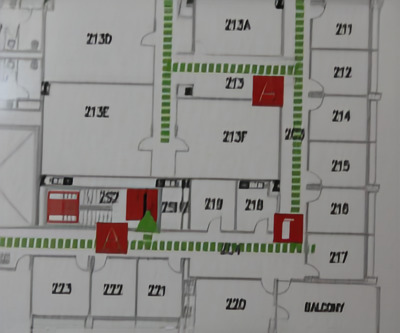}   &
   \includegraphics[trim= 0 5 0 5, clip,width=0.19\textwidth]{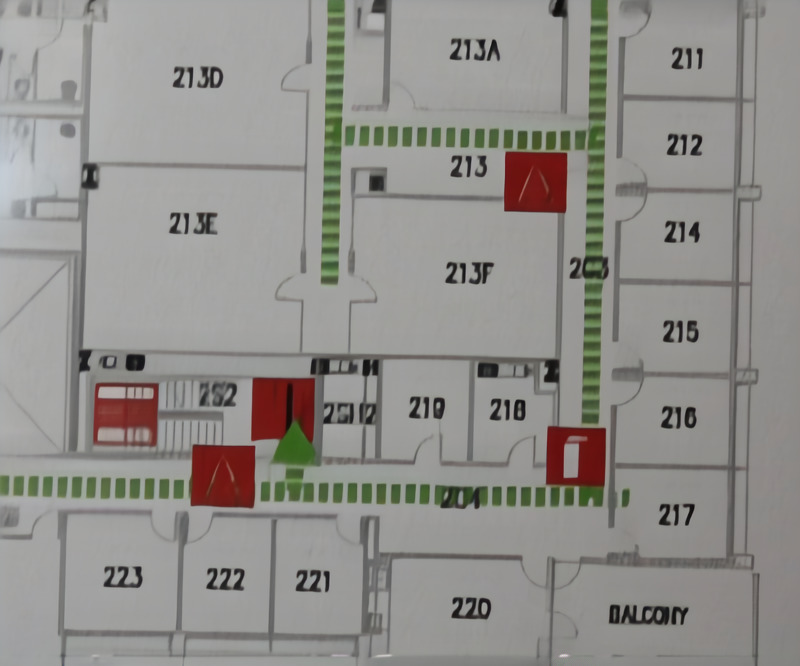}  \\
    \end{tabular}
    \caption{\textbf{Examples of combined deblurring and $\times 2$-super-resolution on images from Lai's  dataset~\cite{lai2016comparative} .} %
    J-MKPD performs motion kernel field estimation and deblurring at the resolution of the input blurry image.  By including a downsampling operator in the degradation model of J-MKPD, J-MKPSR produces high-resolution kernels and a deblurred high-resolved image.  When the estimation of HR kernels is accurate,  restoring the images using the high-resolution kernels produces better results than restoring with the low-resolution kernels and upsampling. Best viewed in electronic format.  \label{fig:superresolution}}
    \end{figure*}

%% file: biography.tex
\vspace{-33pt}

\begin{IEEEbiographynophoto}{Guillermo Carbajal} received the electrical engineering degree (2010) and MSc. (2014) from Universidad de la República, Uruguay. From 2013 to 2020, he worked in the industry as a developer and researcher on computer vision and image processing applications.  Since 2008, he has been with the Division of Electrical Engineering, Universidad de la República, where he is currently an Assistant and PhD student.
\end{IEEEbiographynophoto}
\vspace{-33pt}
\begin{IEEEbiographynophoto}
{Patricia Vitoria} received the Audiovisual Engineering degree from Universitat Pompeu Fabra (UPF), Spain (2013), a MSc.  in Informatics in Technical University of Munich, Germany  (2017) and a Ph.D. in Computer Vision at UPF (2021). Since 2020, she is a researcher at Huawei Research Center in Zurich, Switzerland. Her research interest are in the field of image processing and computer vision, more specifically camera artifacts, image and video restoration using both model-based and machine learning approaches.
\end{IEEEbiographynophoto}
\vspace{-33pt}
\begin{IEEEbiographynophoto}{José Lezama} is an assistant professor at Universidad de la República in Uruguay. He holds an MSc and PhD in applied mathematics from École Normale Supérieure Paris-Saclay, and was a postdoctoral researcher at Duke University.
\end{IEEEbiographynophoto}

\vspace{-33pt}
\begin{IEEEbiographynophoto}{Pablo Musé} received the electrical engineering degree from Universidad de la República, Uruguay (1999), an MSc. in mathematics and statistical learning (2001) and a Ph.D. in applied mathematics (2004), both from École Normale Supérieure Paris-Saclay, France. %
Since 2008, he has been with the Division of Electrical Engineering, Universidad de la República, where he is currently a Full Professor of signal processing. His research interests include machine learning, image restoration and analysis, computational photography and remote sensing.
\end{IEEEbiographynophoto}

%% file: evaluation_datasets.tex
Generating real blurry-sharp pairs for training or testing deep learning models is a difficult task that requires hard work. For this reason, most deep-learning-based models are trained on synthetic datasets. Below we present two efforts to generate high-quality real blurry-sharp pairs and explain how they address the geometric and chromatic challenges. 

\subsection{K\"ohler's dataset \cite{kohler2012recording}}

\begin{figure}[h]
    \centering
    \includegraphics[width=0.7\textwidth]{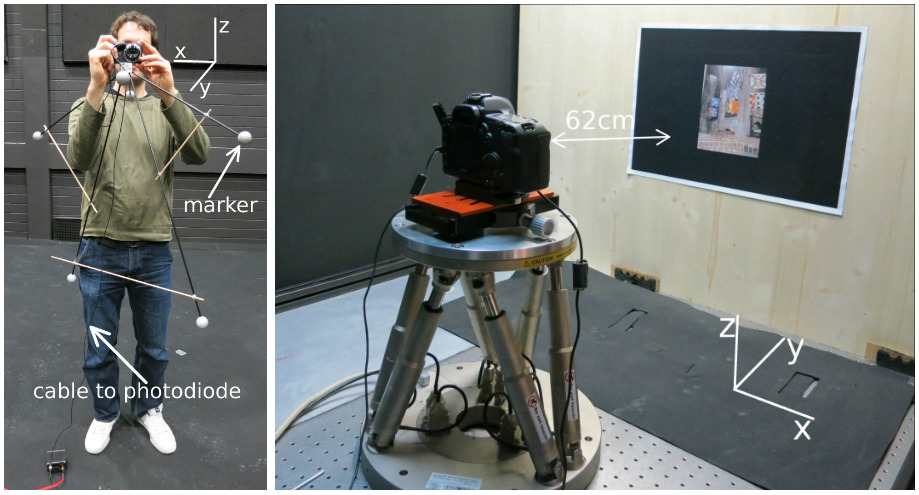}
    \caption{Image acquisition system utilised in \cite{kohler2012recording}. Camera shake trajectories are recorded with a Vicon tracker system (left). The camera attached to a high-precision robot played back camera shakes. Figure reproduced from \cite{kohler2012recording}.   }
    \label{fig:img_acq_kohler}
\end{figure}

This dataset consists of 48 blurred images captured with a robotic platform that mimics the motion of a human arm. Images were acquired by a SLR camera ((Canon Eos 5D Mark II) t) with ISO 100, aperture f/9.0, and exposure time 1sec, with Canon's raw format SRAW2. A Canon EF 50mm f/1.4 lens was used.  The robotic platform is shown on \cref{fig:img_acq_kohler}(right). The true sharp image was
mounted as a poster at a distance of 62cm from the camera. 

\paragraph{Trajectories recording:} Trajectories followed by the robot are human camera shakes previously recorded with a Vicon tracking system  at a frame rate of 500 Hz as shown in \cref{fig:img_acq_kohler} (left). The camera used by the subjects was a Samsung WB600, and the exposure time was 1/3 sec. 

\paragraph{Ground-truth images:} For each of the 48 blurry images, 167 ground-truth images were generated by playing back the camera trajectory step by step and taking an image per step.

\paragraph{Non-uniform blur} The blur kernels numbered 1,3,4,5,8,9,11 tend to be approximately uniform, while blur kernels 2,6,7,10,12 appear non-uniform. Kernels 8,9,10, and 11 turned out rather large, but the authors decided to leave those blur kernels even though those images are usually discarded. 

\subsection{Real Blur dataset generation \cite{rim_2020_ECCV}}

\begin{figure}[h]
    \centering
    \includegraphics[width=0.4\textwidth]{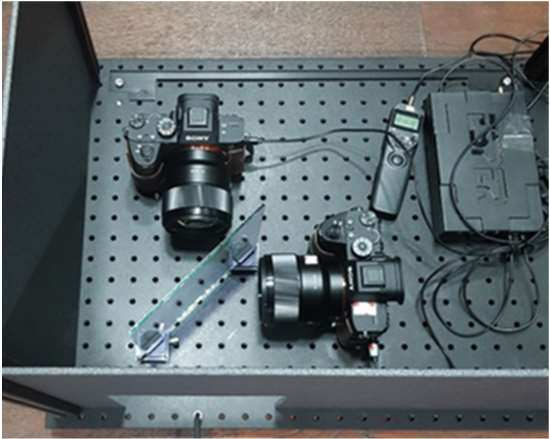}
    \caption{Image acquisition system utilised in \cite{rim_2020_ECCV}.  Figure reproduced from \cite{rim_2020_ECCV}. }
    \label{fig:img_acq_realblur}
\end{figure}

For this dataset, blurry-sharp image pairs are captured simultaneously using the dual camera
system shown in \ref{fig:img_acq_realblur}. It consists of two cameras and a beam splitter placed to allow the cameras to capture the same scene. The cameras and beam splitter are inside an optical enclosure to avoid out-of-scene ray lights interferences. One camera captures the reference image using a 1/80 seconds shutter speed, with ISO and aperture size adjusted to avoid blur caused by camera shakes. The other camera captures a blurry image with 1/2 seconds shutter speed and 40 times lower ISO to have similar brightness to the reference image. 

Although the cameras and their lenses are of the same model, and the capture is synchronized using a multi-camera trigger, careful post-processing is required to generate the final blurry-sharp pair.  

\paragraph{Crop:} Images are acquired at a resolution of 7952$\times$5304 and then cropped to 2721$\times$ 3094 to discard invalid areas along the boundaries capturing outside the beam splitter or inside the optical enclosure. 

\paragraph{Downsampling \& Denoising:} Since sharp images are captured using high ISO values, they are noisy. Also, the alignment of the cameras is not perfect; therefore, a small amount of parallax exists between sharp and blurred images. The images are downsampled by 1/4  for each axis to  680 $\times$ 733.  To further reduce noise on the sharp image, a BM3D denoiser is applied.
\paragraph{Geometric alignment:} Since the alignment of the camera is not perfect and the
cameras' positions may slightly change over time due to camera shake, a geometric alignment process is applied. First, a homography is estimated to align blurry-sharp pairs roughly. As estimating a homography from a blurry image is complex, the homography is calculated using a reference image captured with the same camera instead of the blurry image.  The multi-camera trigger synchronizes the shutters to open simultaneously, but they close at different moments due to their different shutter speeds. Thus, after the shutter of one camera is closed, the other camera still captures incoming lights while moving, causing misalignment between blurred and sharp images. In the second step, the remaining misalignment between each pair of blurred and sharp images is corrected using a phase correlation-based approach. Simply applying the previously mentioned alignment steps results in objects in the sharp image aligned to the corners of their corresponding blurry objects in the blurred image, not to the centers. A third step aligns images to match the center of an object in a sharp image with the center of its corresponding object in a blurred image. To this end, a blur kernel of the
blurred image is estimated using its corresponding sharp image, assuming that the scene is
static and blur is nearly spatially-invariant. Then, the estimated blur kernel centroid is computed and used to align the sharp image by shifting it according to the centroid. According to the authors, all the steps are essential to align the images correctly. 
\paragraph{Photometric alignment:} Although cameras and lenses are of the same models, their images may have slight intensity differences. For geometrically aligned sharp and blurred images $\mathbf{u}$ and $\mathbf{v}$, $\mathbf{u}$ is photometrically aligned to $\mathbf{v}$ by applying a linear transform $\alpha \mathbf{u} + \beta \sim \mathbf{v}$. Since the coefficients $\alpha$ and $\beta$ are difficult to estimate from $\mathbf{u}$ and $\mathbf{v}$ due to the blur in $\mathbf{v}$, they are estimated from the reference pairs corresponding to $\mathbf{u}$ and $\mathbf{v}$. 

%% file: supp_mat_no_reference_metrics.tex
Besides the usual reference metrics, we reported three reference-free perceptual metrics: the Blur Strength~\cite{crete2007blur}, the Cumulative Probability of Blur Distribution (CPBD)~\cite{narvekar2011no}, and the Sharpness Index(SI)~\cite{blanchet2012explicit}. In contrast to other commonly reported metrics, these metrics are reference-free perceptual metrics explicitly introduced to quantify the degree of blurriness or sharpness in an image. 

\paragraph{Blur Strength:} This metric is built on the observation that we have difficulties perceiving differences between a blurred image and the same re-blurred image. If we blur a sharp picture, gray levels of neighboring pixels will change with a significant variation. On the contrary, if we blur an already blurred image, gray levels of adjacent pixels will still change, but only to a lesser extent. The key idea of the blur estimation principle is to blur the initial image and analyze the neighboring pixels' variation.  The metric score varies between 0 (no blur) and 1 (maximum blur). The quality of the metric was validated with a subjective test with 15 nonexpert viewers. Each of them graded pictures according to the amount of blur. Each viewer had to grade 264 images. We used the implementation available on \textit{scikit-image}\footnote{\texttt{skimage.measure.blur\_effect}}.

\paragraph{Cumulative Probability of Blur Detection (CPBD):} This metric is motivated by a study of the sensitivity of human blur perception at different contrast values. The metric utilizes a probabilistic model to estimate the probability of detecting blur at each edge in the image, and then the information is pooled by
computing the cumulative probability of blur detection (CPBD). 

This work builds upon the notion of Just Noticeable Blur (JNB) width. According to \cite{ferzli2009no}, an edge is perceived as blurry with a probability of 0.63 when its width is

\begin{equation}
w_{JNB} = \left\lbrace \begin{array}{cc}
     5 &  C \leq 50 \\
     3 &  C > 50
\end{array}  \right. ,
\end{equation}

where $C$ is the local contrast in the neighborhood of the edge. The probability of perceiving blur diminishes when the edge width is below $w_{JNB}$ and augments when the edge width increases.   

The CPBD metric corresponds to the percentage of edges at which the probability of blur detection is below the
Just Noticeable Blur detection probability ($P_{JNB}=0.63$). Hence, a higher CPBD score corresponds to a sharper image. 

\begin{equation}
    CPBD = \sum_{P_{blur}=0}^{P_{blur}=P_{JNB}} P(P_{blur}).
\end{equation}
According to our experiments, this metric is appropriate for comparing the impact of different methods applied to the same image. However, as the metric depends on the distribution of the edges on the image, the score may not be suitable for comparing the sharpness of images corresponding to different scenes.    
 
\paragraph{Sharpness Index:} It is well-known that the geometry of an image is mainly encoded in the phase of its Fourier Transform \cite{oppenheim1981importance}. By distorting the phase information of the image, it is natural that some oscillations will appear in the regions where the original image was flat.  The key observation is that a sharp image is much more sensitive to phase perturbations than a blurry image. Therefore, it is reasonable to assume that a sharp image will be more affected by the destruction of the phase information than a blurry image. This observation inspired the definition of three sharpness metrics based on the phase coherence principle: the Global Phase Coherence~\cite{blanchet2008measuring}, the Sharpness Index~\cite{blanchet2012explicit}, and the index S~\cite{leclaire2013blind}. 

It becomes natural to expect the image's Total Variation (TV) to increase when the phase is distorted. Intuitively, when an image $\mathbf{u}$ is blurred, its high-frequency components are attenuated. Therefore, the TV increase entailed by the phase perturbation is expected to be less dramatic than in the sharp case.  

In this article, we used the Sharpness Index~\cite{blanchet2012explicit}, which is defined as
\begin{equation}
    SI(\mathbf{u}) = - \log_{10} \Phi \left( \frac{\mu - TV(\mathbf{u})}{\sigma} \right),
\end{equation}
where $\Phi(t)$ is the Gaussian tail distribution
\begin{equation}
    \Phi(t)=\frac{1}{\sqrt{2\pi}} \int_t^\infty \exp^{(-s^2/2)} ds,
\end{equation}
and
\begin{equation}
    \mu = \mathtt{E}\left[TV(\mathbf{u} * \mathbf{W}) \right], \quad \sigma^2 = \mathtt{Var}\left[TV(\mathbf{u} * \mathbf{W})\right].
\end{equation}
Here $\mathbf{u} * \mathbf{W}$ denotes the periodic convolution of the $M$ by $N$ image $\mathbf{u}$ with a Gaussian white noise image $\mathbf{W}$ with mean 0 and variance $(MN)^{-1}$. The behavior of convolving $\mathbf{u}$ with Gaussian noise is similar to that of perturbing the phase by random noise. However, it is computationally more friendly since it is possible to obtain an explicit formula for the first two moments of $TV(\mathbf{u} * \mathbf{W})$ (see~\cite{leclaire2015no}). Also, it has been observed experimentally that it is reasonable to assume that $TV(\mathbf{u} * \mathbf{W})$ is Gaussian. 

\subsection{Perceptual metrics on the Real Blur dataset}

\begin{figure}[h]
    \centering
     \begin{tabular}{*{2}{c}}
    \includegraphics[width=0.45\textwidth]{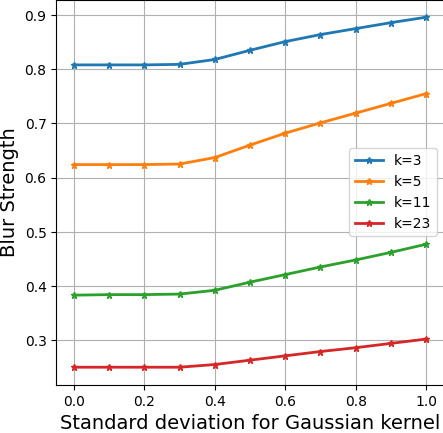} &
    \includegraphics[width=0.45\textwidth]{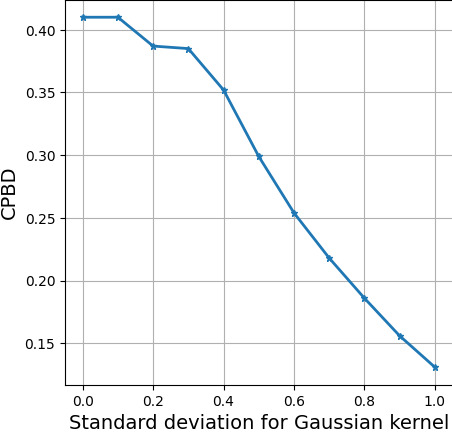}
    \end{tabular}
    \caption{When we blur our restored images from the RealBlur dataset \cite{rim_2020_ECCV} with a Gaussian filter, the blur strength increases and diminishes the CPBD. The behavior of blur perceptive metrics is as expected. 
    \label{fig:perceptual_vs_blur}}
\end{figure}

In \cref{fig:perceptual_vs_blur} left, we show that the blur strength increases when we blur the results produced by our method on the Real Blur datasets. As expected, the blur strength increases monotonically with the gaussian kernel size. We show results for different sizes ($k$-values) of the re-blurring filter. In the article, we used \texttt{scikit-image} default values $k=11$ since this value correlates best with human perception. 

In \cref{fig:perceptual_vs_blur} right, we show that the $CPBD$ score monotonically decreases when we blur the results produced by our method on the Real Blur datasets. This is the expected behavior. 

\begin{figure*}[h]
  \centering
  \scriptsize
\setlength{\tabcolsep}{2pt}

  \begin{tabular}{*{6}{c}}
            Blurry & Ana.-Synth.
          \cite{kaufman2020deblurring} & SRN \cite{tao2018scale} & MPRNet \cite{Zamir2021MPRNet} & Ours & Ground truth \\
 \hline
    \includegraphics[trim=250 350 200 200, clip, width=0.15\textwidth]{imgs/RealBlur/Blurry/scene002_blur_10.jpg}   &
      \includegraphics[trim=250 350 200 200, clip,width=0.15\textwidth]{imgs/RealBlur/AnaSyn/scene002_blur_10.jpg}   &
   \includegraphics[trim=250 350 200 200, clip,width=0.15\textwidth]{imgs/RealBlur/SRN/scene002_blur_10.jpg}  &
    \includegraphics[trim=250 350 200 200, clip,width=0.15\textwidth]{imgs/RealBlur/MPRNet/scene002_blur_10.jpg}   &
      \includegraphics[trim=250 350 200 200, clip,width=0.15\textwidth]{imgs/RealBlur/COCO900_restL2_aug_all_loss_gf1_80k/restored/scene002_blur_10.jpg} &  
        \includegraphics[trim=250 350 200 200, clip,width=0.15\textwidth]{imgs/RealBlur/GT/scene002_gt_10.jpg}\\
    BS=0.289, CPBD=0.526  &   BS=0.295, CPBD=0.453 & BS=0.261, CPBD=0.615 & BS=0.258, CPBD=0.636 & BS=0.215, CPBD=0.786 & BS= 0.258, CPBD=0.659 \\ 
    \hline
    \includegraphics[trim=220 300 0 150, clip, width=0.15\textwidth]{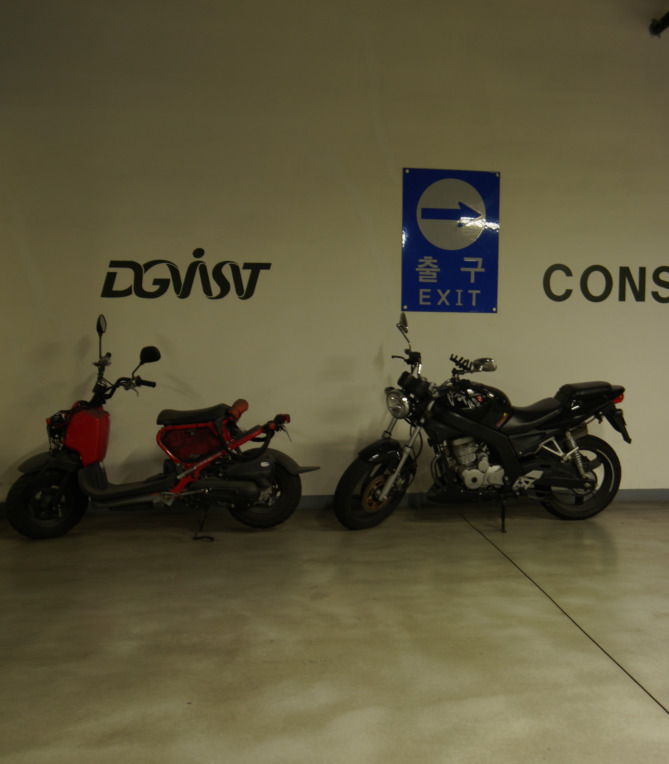}   &
      \includegraphics[trim=220 300 0 150, clip,width=0.15\textwidth]{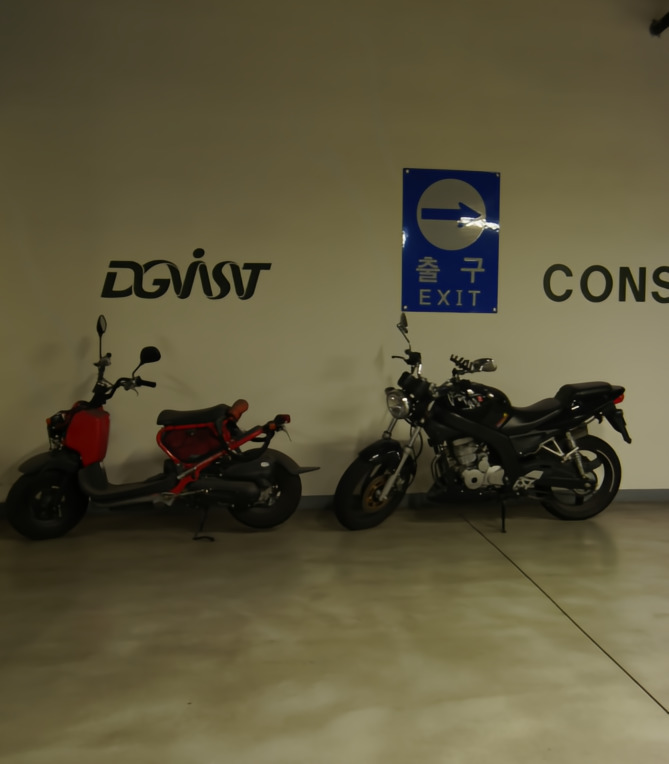}   &
   \includegraphics[trim=220 300 0 150, clip,width=0.15\textwidth]{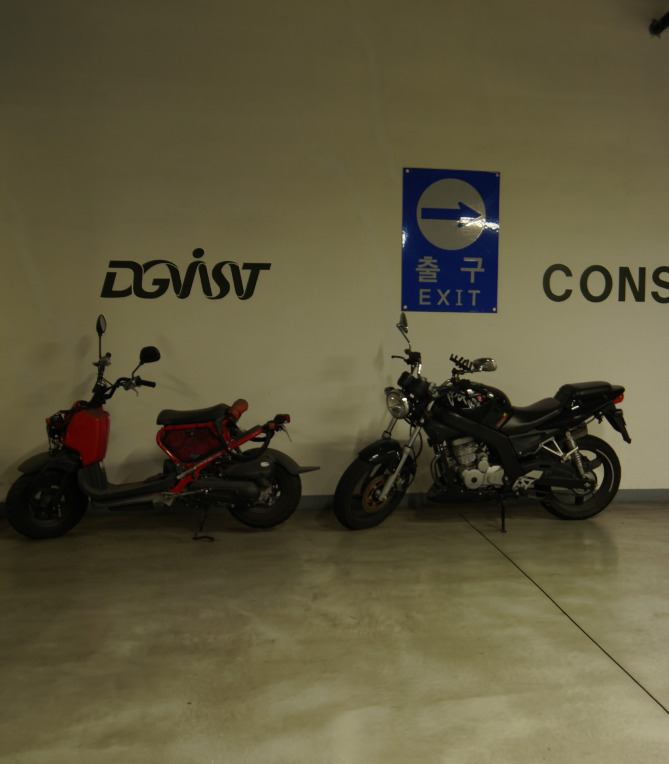}  &
    \includegraphics[trim=220 300 0 150, clip,width=0.15\textwidth]{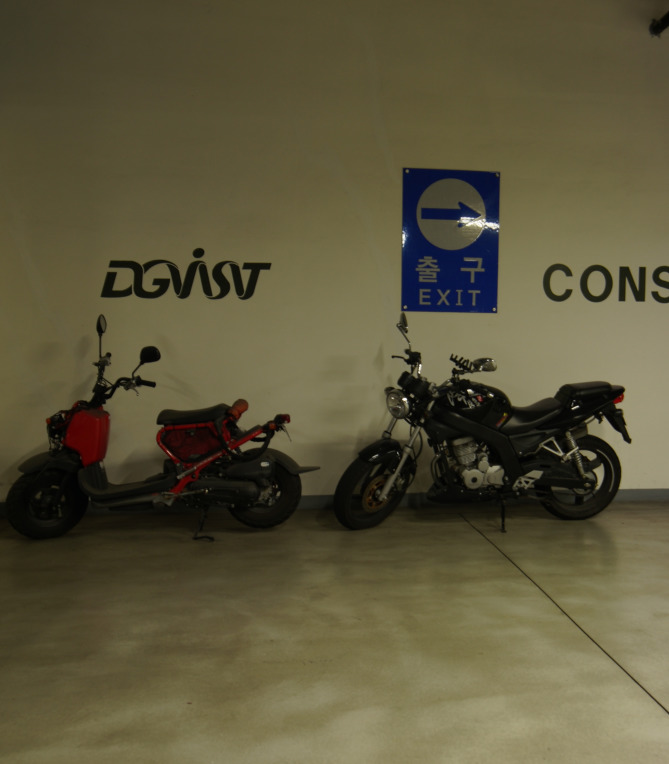}   &
      \includegraphics[trim=220 300 0 150, clip,width=0.15\textwidth]{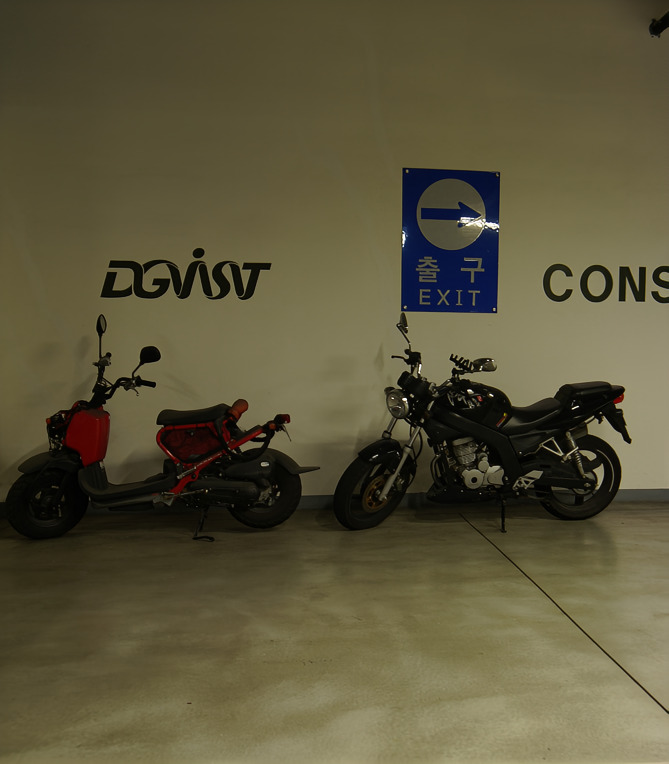} &
    \includegraphics[trim=220 300 0 150, clip,width=0.15\textwidth]{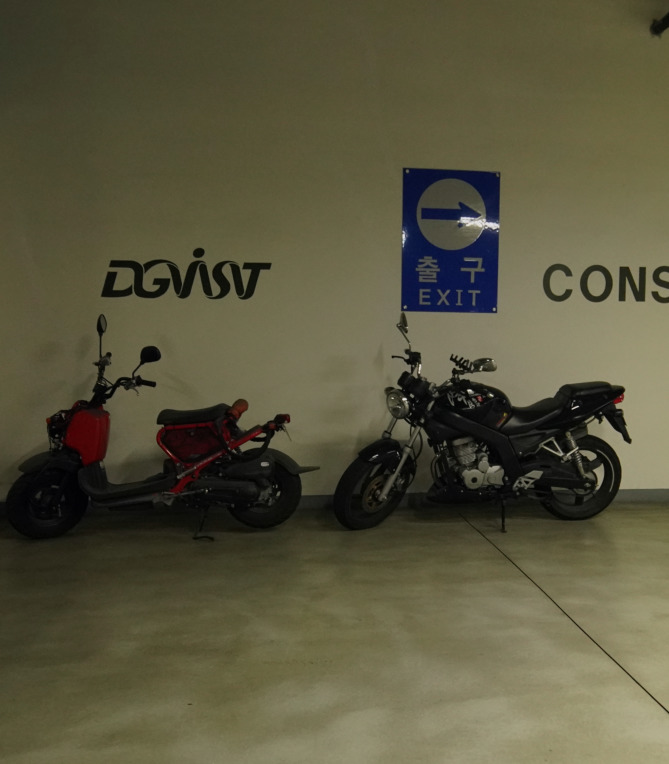}\\
    BS=0.378, CPBD=0.305  &   BS=0.399, CPBD=0.272 & BS=0.372, CPBD=0.383 & BS=0.374, CPBD=0.335 & BS=0.316, CPBD=0.734 & BS=0.338, CPBD=0.503 \\ 
    \includegraphics[trim=20 200 350 320, clip, width=0.15\textwidth]{imgs/RealBlur/Blurry/scene044_blur_06.jpg}   &
      \includegraphics[trim=20 200 350 320, clip,width=0.15\textwidth]{imgs/RealBlur/AnaSyn/scene044_blur_06.jpg}   &
   \includegraphics[trim=20 200 350 320, clip,width=0.15\textwidth]{imgs/RealBlur/SRN/scene044_blur_06.jpg}  &
    \includegraphics[trim=20 200 350 320, clip,width=0.15\textwidth]{imgs/RealBlur/MPRNet/scene044_blur_06.jpg}   &
      \includegraphics[trim=20 200 350 320, clip,width=0.15\textwidth]{imgs/RealBlur/COCO900_restL2_aug_all_loss_gf1_80k/restored/scene044_blur_06.jpg} &  
        \includegraphics[trim=20 200 350 320, clip,width=0.15\textwidth]{imgs/RealBlur/GT/scene044_gt_06.jpg}\\
    BS=0.339, CPBD=0.403  &   BS=0.333, CPBD=0.415 & BS=0.310, CPBD=0.502 & BS=0.319, CPBD=0.488 & BS=0.258, CPBD=0.729 & BS=0.316, CPBD=0.491 \\ 
    \includegraphics[trim=0 150 350 320, clip, width=0.15\textwidth]{imgs/RealBlur/Blurry/scene050_blur_17.jpg}   &
      \includegraphics[trim=0 150 350 320, clip,width=0.15\textwidth]{imgs/RealBlur/AnaSyn/scene050_blur_17.jpg}   &
   \includegraphics[trim=0 150 350 320, clip,width=0.15\textwidth]{imgs/RealBlur/SRN/scene050_blur_17.jpg}  &
    \includegraphics[trim=0 150 350 320, clip,width=0.15\textwidth]{imgs/RealBlur/MPRNet/scene050_blur_17.jpg}   &
      \includegraphics[trim=0 150 350 320, clip,width=0.15\textwidth]{imgs/RealBlur/COCO900_restL2_aug_all_loss_gf1_80k/restored/scene050_blur_17.jpg} &  
        \includegraphics[trim=0 150 350 320, clip,width=0.15\textwidth]{imgs/RealBlur/GT/scene050_gt_17.jpg}\\
    BS=0.358, CPBD=0.362  &   BS=0.358, CPBD=0.368 & BS=0.327, CPBD=0.436 & BS=0.350, CPBD=0.385 & BS=0.274, CPBD=0.727 & BS=0.322, CPBD=0.496 \\ 
    \includegraphics[trim=200 350 100 120, clip, width=0.15\textwidth]{imgs/RealBlur/Blurry/scene103_blur_04.jpg}   &
      \includegraphics[trim=200 350 100 120, clip,width=0.15\textwidth]{imgs/RealBlur/AnaSyn/scene103_blur_04.jpg}   &
   \includegraphics[trim=200 350 100 120, clip,width=0.15\textwidth]{imgs/RealBlur/SRN/scene103_blur_04.jpg}  &
    \includegraphics[trim=200 350 100 120, clip,width=0.15\textwidth]{imgs/RealBlur/MPRNet/scene103_blur_04.jpg}   &
      \includegraphics[trim=200 350 100 120, clip,width=0.15\textwidth]{imgs/RealBlur/COCO900_restL2_aug_all_loss_gf1_80k/restored/scene103_blur_04.jpg} &  
        \includegraphics[trim=200 350 100 120, clip,width=0.15\textwidth]{imgs/RealBlur/GT/scene103_gt_04.jpg}\\
    BS=0.362, CPBD=0.285  &   BS=0.360, CPBD=0.249 & BS=0.337, CPBD=0.368 & BS=0.339, CPBD=0.366 & BS=0.288, CPBD=0.612 & BS=0.341, CPBD=0.361 \\ 
    \includegraphics[trim=200 200 50 200, clip, width=0.15\textwidth]{imgs/RealBlur/Blurry/scene107_blur_07.jpg}   &
      \includegraphics[trim=200 200 50 200, clip,width=0.15\textwidth]{imgs/RealBlur/AnaSyn/scene107_blur_07.jpg}   &
   \includegraphics[trim=200 200 50 200, clip,width=0.15\textwidth]{imgs/RealBlur/SRN/scene107_blur_07.jpg}  &
    \includegraphics[trim=200 200 50 200, clip,width=0.15\textwidth]{imgs/RealBlur/MPRNet/scene107_blur_07.jpg}   &
      \includegraphics[trim=200 200 50 200, clip,width=0.15\textwidth]{imgs/RealBlur/COCO900_restL2_aug_all_loss_gf1_80k/restored/scene107_blur_07.jpg} &  
        \includegraphics[trim=200 200 50 200, clip,width=0.15\textwidth]{imgs/RealBlur/GT/scene107_gt_07.jpg}\\
    BS=0.409, CPBD=0.162  &   BS=0.352, CPBD=0.257 & BS=0.360, CPBD=0.195 & BS=0.371, CPBD=0.278 & BS=0.323, CPBD=0.470 & BS=0.341, CPBD=0.361 \\ 
    \includegraphics[trim=200 200 50 250, clip, width=0.15\textwidth]{imgs/RealBlur/Blurry/scene130_blur_07.jpg}   &
      \includegraphics[trim=200 200 50 250, clip,width=0.15\textwidth]{imgs/RealBlur/AnaSyn/scene130_blur_07.jpg}   &
   \includegraphics[trim=200 200 50 250, clip,width=0.15\textwidth]{imgs/RealBlur/SRN/scene130_blur_07.jpg}  &
    \includegraphics[trim=200 200 50 250, clip,width=0.15\textwidth]{imgs/RealBlur/MPRNet/scene130_blur_07.jpg}   &
      \includegraphics[trim=200 200 50 250, clip,width=0.15\textwidth]{imgs/RealBlur/COCO900_restL2_aug_all_loss_gf1_80k/restored/scene130_blur_07.jpg} &  
        \includegraphics[trim=200 200 50 250, clip,width=0.15\textwidth]{imgs/RealBlur/GT/scene130_gt_07.jpg}\\
    BS=0.335, CPBD=0.310  &   BS=0.336, CPBD=0.291 & BS=0.323, CPBD=0.332 & BS0.326=, CPBD=0.332 & BS=0.264, CPBD=0.707 & BS=0.310, CPBD=0.413 \\ 
    \includegraphics[trim=250 150 50 300, clip, width=0.15\textwidth]{imgs/RealBlur/Blurry/scene208_blur_03.jpg}   &
      \includegraphics[trim=250 150 50 300, clip,width=0.15\textwidth]{imgs/RealBlur/AnaSyn/scene208_blur_03.jpg}   &
   \includegraphics[trim=250 150 50 300, clip,width=0.15\textwidth]{imgs/RealBlur/SRN/scene208_blur_03.jpg}  &
    \includegraphics[trim=250 150 50 300, clip,width=0.15\textwidth]{imgs/RealBlur/MPRNet/scene208_blur_03.jpg}   &
      \includegraphics[trim=250 150 50 300, clip,width=0.15\textwidth]{imgs/RealBlur/COCO900_restL2_aug_all_loss_gf1_80k/restored/scene208_blur_03.jpg} &  
        \includegraphics[trim=250 150 50 300, clip,width=0.15\textwidth]{imgs/RealBlur/GT/scene208_gt_03.jpg}\\
    BS=0.386, CPBD=0.290  &   BS=0.401, CPBD=0.291 & BS=0.360, CPBD=0.366 & BS=0.358, CPBD=0.416 & BS=0.310, CPBD=0.692 & BS=0.341, CPBD=0.422 \\ 
  \end{tabular}
  \caption{\textbf{Perceputal metrics on the RealBlur dataset \cite{rim_2020_ECCV}}. Blur strength and CPBD metrics computed on restored images for different methods correlate well with our perception of blur. The lower the blur strength and the higher the CPBD, the sharper the image. Best viewed in electronic format. }
  \label{fig:RealBlurComparison_perceptual}
\end{figure*}

In \cref{fig:RealBlurComparison_perceptual} we show, on a selection of images from the RealBlur dataset, that both perceptual metrics correlate well with our perception of the blur present on the images restored with different methods. Note that RealBlur ground-truth images are actually not completely sharp. Our restorations look sharper, as confirmed by the perceptual metrics. 

%% file: figure_RealBlur_blurred.tex
\begin{figure*}[ht]
  \centering
\setlength{\tabcolsep}{2pt}

  \begin{tabular}{*{5}{c}}
            Blurry & J-MKPD (ours) & J-MKPD blurred & GT  & \\
            \hline
    \includegraphics[trim=250 350 200 200, clip, width=0.15\textwidth]{imgs/RealBlur/Blurry/scene002_blur_10.jpg}   &
      \includegraphics[trim=250 350 200 200, clip,width=0.15\textwidth]{imgs/RealBlur/COCO900_restL2_aug_all_loss_gf1_80k/restored/scene002_blur_10.jpg} &  
      \includegraphics[trim=250 350 200 200, clip,width=0.15\textwidth]{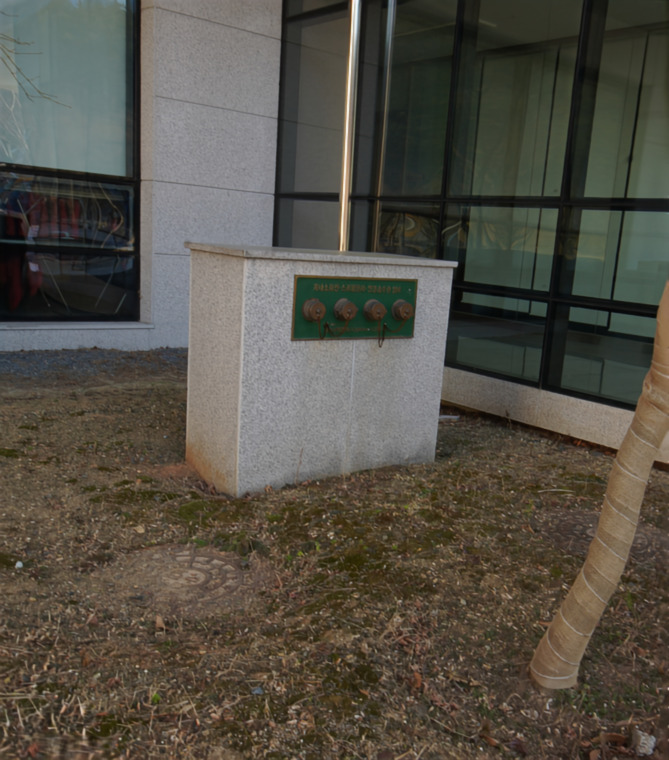} &  
        \includegraphics[trim=250 350 200 200, clip,width=0.15\textwidth]{imgs/RealBlur/GT/scene002_gt_10.jpg} &         \includegraphics[width=0.25\textwidth]{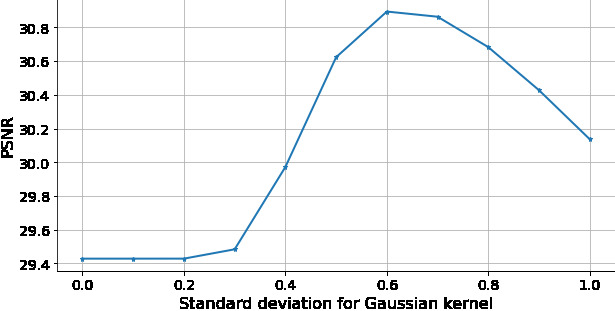}  \\
             30.35 dB  &  29.43 dB & 30.89 dB  &   &   \\
    \includegraphics[trim=220 300 0 150, clip, width=0.15\textwidth]{imgs/RealBlur/Blurry/scene011_blur_20.jpg}   &
      \includegraphics[trim=220 300 0 150, clip,width=0.15\textwidth]{imgs/RealBlur/COCO900_restL2_aug_all_loss_gf1_80k/restored/scene011_blur_20.jpg} &
      \includegraphics[trim=220 300 0 150, clip,width=0.15\textwidth]{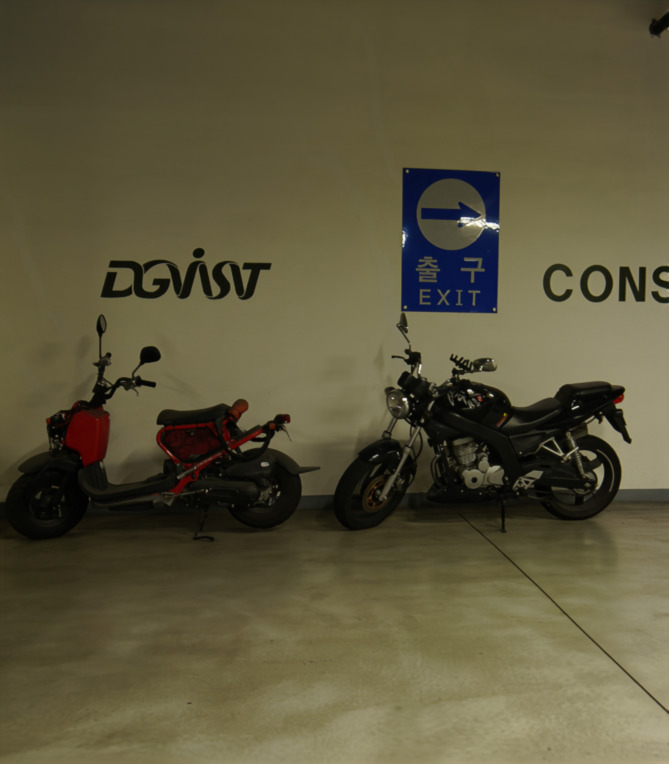}                            &
        \includegraphics[trim=220 300 0 150, clip,width=0.15\textwidth]{imgs/RealBlur/GT/scene011_gt_20.jpg} &
        \includegraphics[width=0.25\textwidth]{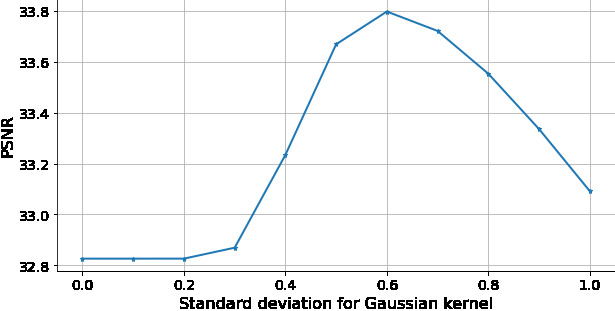} \\
        33.67 dB  &  32.83 dB & 33.80 dB &  & \\
    \includegraphics[trim=20 200 350 320, clip, width=0.15\textwidth]{imgs/RealBlur/Blurry/scene044_blur_06.jpg}   &
      \includegraphics[trim=20 200 350 320, clip,width=0.15\textwidth]{imgs/RealBlur/COCO900_restL2_aug_all_loss_gf1_80k/restored/scene044_blur_06.jpg} &
        \includegraphics[trim=20 200 350 320, clip,width=0.15\textwidth]{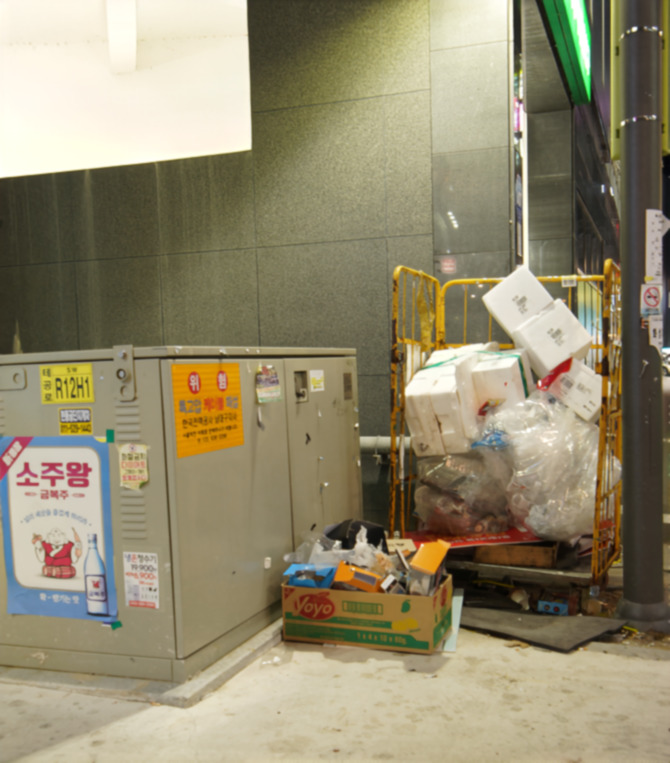}                      &
        \includegraphics[trim=20 200 350 320, clip,width=0.15\textwidth]{imgs/RealBlur/GT/scene044_gt_06.jpg} & 
        \includegraphics[width=0.25\textwidth]{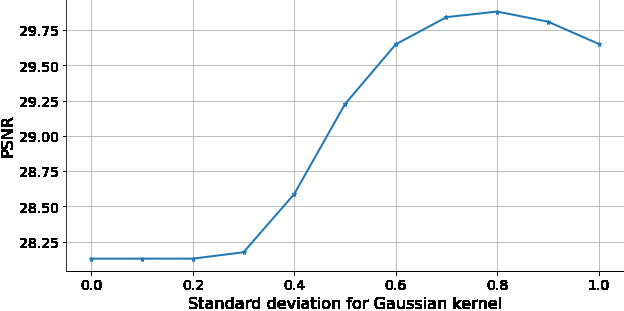} \\
        29.64 dB &   28.13 dB & 29.88 dB &  &   \\
    \includegraphics[trim=0 150 350 320, clip, width=0.15\textwidth]{imgs/RealBlur/Blurry/scene050_blur_17.jpg}   &
      \includegraphics[trim=0 150 350 320, clip,width=0.15\textwidth]{imgs/RealBlur/COCO900_restL2_aug_all_loss_gf1_80k/restored/scene050_blur_17.jpg} &  
       \includegraphics[trim=0 150 350 320, clip,width=0.15\textwidth]{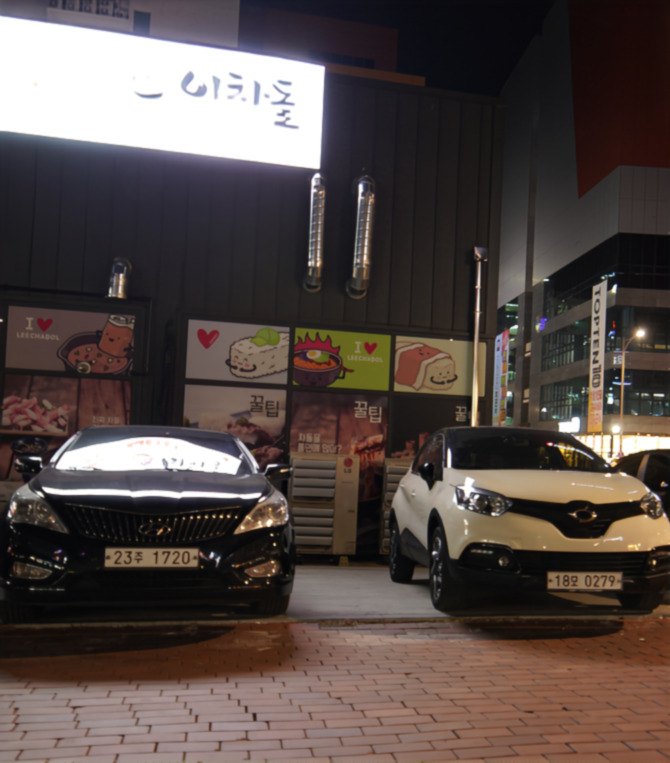}                      &
        \includegraphics[trim=0 150 350 320, clip,width=0.15\textwidth]{imgs/RealBlur/GT/scene050_gt_17.jpg} &   
            \includegraphics[width=0.25\textwidth]{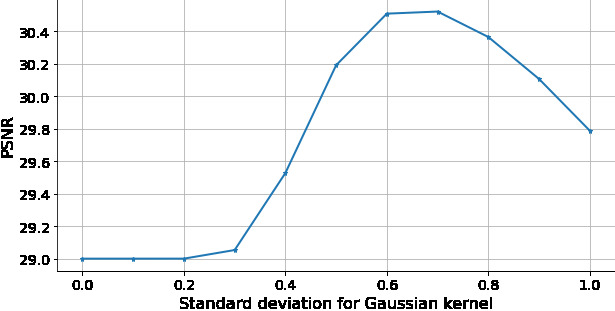} \\
            30.25 dB & 29.0 dB & 30.52 dB &   &    \\
    \includegraphics[trim=200 350 100 120, clip, width=0.15\textwidth]{imgs/RealBlur/Blurry/scene103_blur_04.jpg}   &
      \includegraphics[trim=200 350 100 120, clip,width=0.15\textwidth]{imgs/RealBlur/COCO900_restL2_aug_all_loss_gf1_80k/restored/scene103_blur_04.jpg} &  
       \includegraphics[trim=200 350 100 120, clip,width=0.15\textwidth]{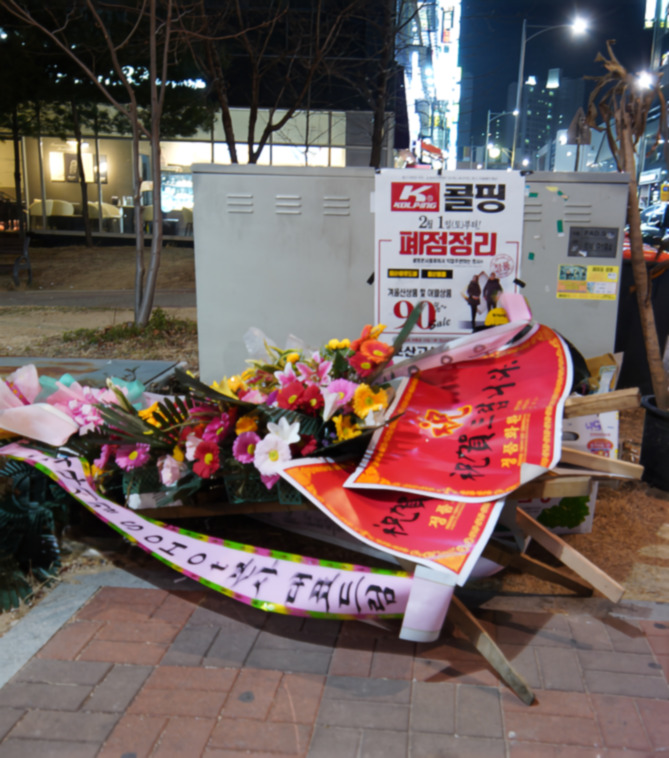}                         &
        \includegraphics[trim=200 350 100 120, clip,width=0.15\textwidth]{imgs/RealBlur/GT/scene103_gt_04.jpg} &
        \includegraphics[width=0.25\textwidth]{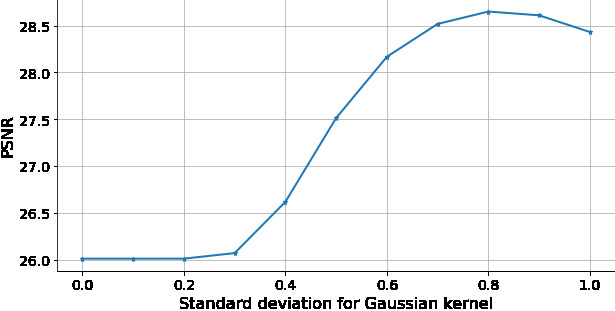} \\
         28.47 dB &  26.01 dB & 28.65 dB &  & \\
    \includegraphics[trim=200 200 50 200, clip, width=0.15\textwidth]{imgs/RealBlur/Blurry/scene107_blur_07.jpg}   &
      \includegraphics[trim=200 200 50 200, clip,width=0.15\textwidth]{imgs/RealBlur/COCO900_restL2_aug_all_loss_gf1_80k/restored/scene107_blur_07.jpg} &
         \includegraphics[trim=200 200 50 200, clip,width=0.15\textwidth]{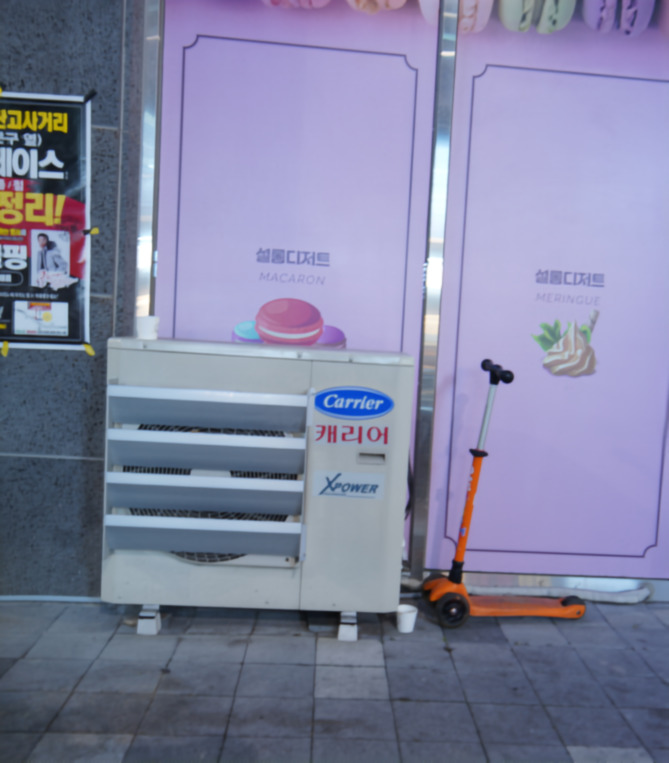}                            &
        \includegraphics[trim=200 200 50 200, clip,width=0.15\textwidth]{imgs/RealBlur/GT/scene107_gt_07.jpg}  & 
        \includegraphics[width=0.25\textwidth]{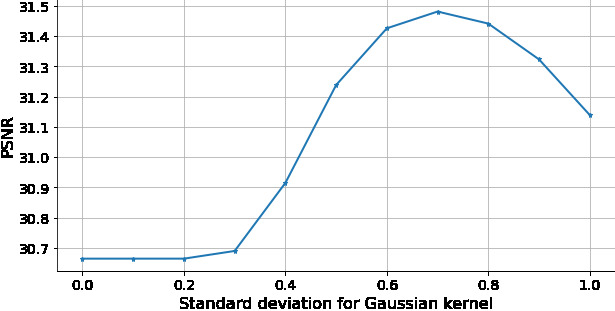} \\
              30.08   & 30.67 dB & 31.48 dB &  &    \\
    \includegraphics[trim=200 200 50 250, clip, width=0.15\textwidth]{imgs/RealBlur/Blurry/scene130_blur_07.jpg}   &
      \includegraphics[trim=200 200 50 250, clip,width=0.15\textwidth]{imgs/RealBlur/COCO900_restL2_aug_all_loss_gf1_80k/restored/scene130_blur_07.jpg} &  
               \includegraphics[trim=200 200 50 250, clip,width=0.15\textwidth]{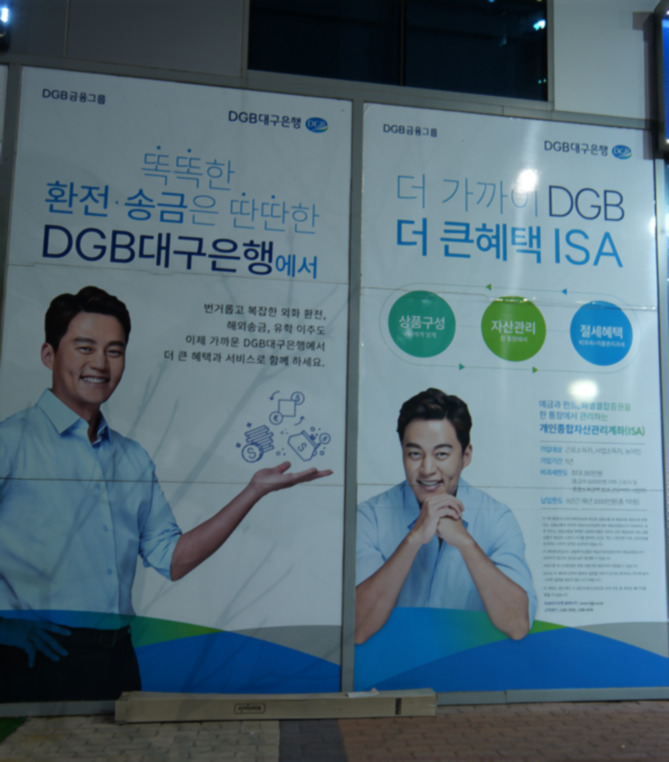}                      &
        \includegraphics[trim=200 200 50 250, clip,width=0.15\textwidth]{imgs/RealBlur/GT/scene130_gt_07.jpg} & 
        \includegraphics[width=0.25\textwidth]{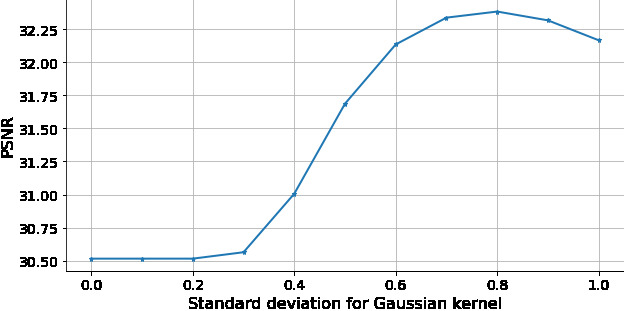} \\
         29.20 & 30.52 dB & 32.38 dB  &   &  \\
    \includegraphics[trim=250 150 50 300, clip, width=0.15\textwidth]{imgs/RealBlur/Blurry/scene208_blur_03.jpg}   &
      \includegraphics[trim=250 150 50 300, clip,width=0.15\textwidth]{imgs/RealBlur/COCO900_restL2_aug_all_loss_gf1_80k/restored/scene208_blur_03.jpg} & 
              \includegraphics[trim=250 150 50 300, clip,width=0.15\textwidth]{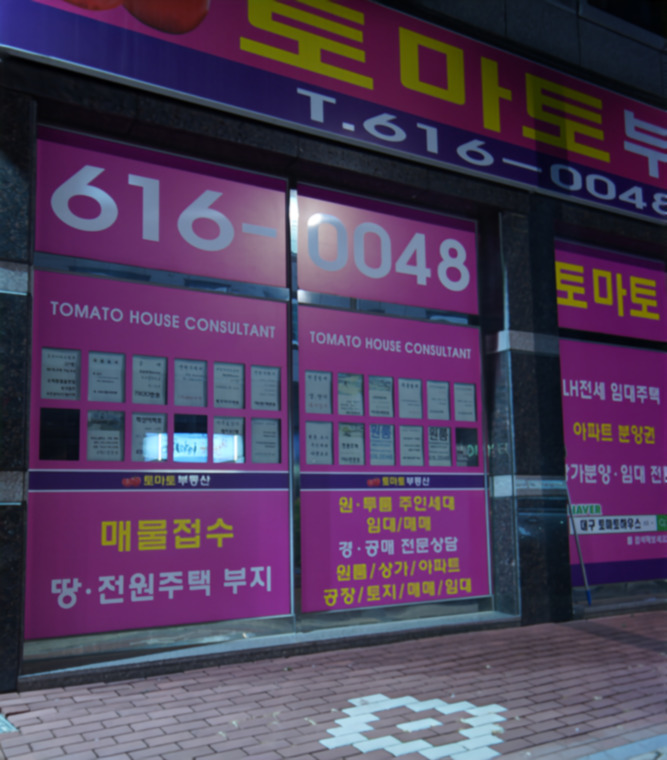}                       &
        \includegraphics[trim=250 150 50 300, clip,width=0.15\textwidth]{imgs/RealBlur/GT/scene208_gt_03.jpg} & 
                \includegraphics[width=0.25\textwidth]{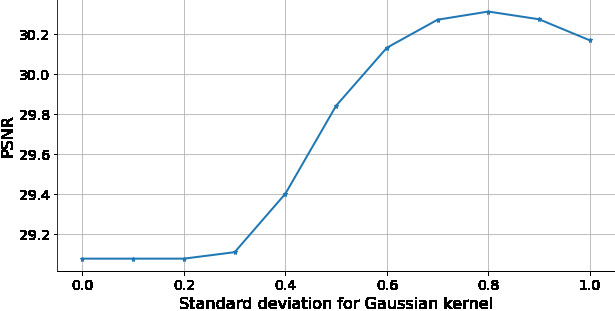} \\
                       29.24 &   29.07 dB & 30.31 dB &  &    \\
  \end{tabular}
  \caption{\textbf{Effect of blurring our restored images on PSNR}. If we blur our restored images with a Gaussian kernel of increasing standard deviation, the PSNR increases, gets to a peak, and then decreases. Since the ground-truth images are a bit blurry, PSNR prefers smooth restorations.   Best viewed in electronic format. }
  \label{fig:blurring_restored}
\end{figure*}

%% file: figure_kernels_Kohler_sup_mat.tex
\begin{figure*}[ht]
  \centering
\setlength{\tabcolsep}{2pt}
  \begin{tabular}{ccc}
         Blurry & Gong \cite{gong2017motion} & Sun \cite{sun2015learning} \\
     \includegraphics[trim=100 170 100 50, clip,width=0.31\textwidth]{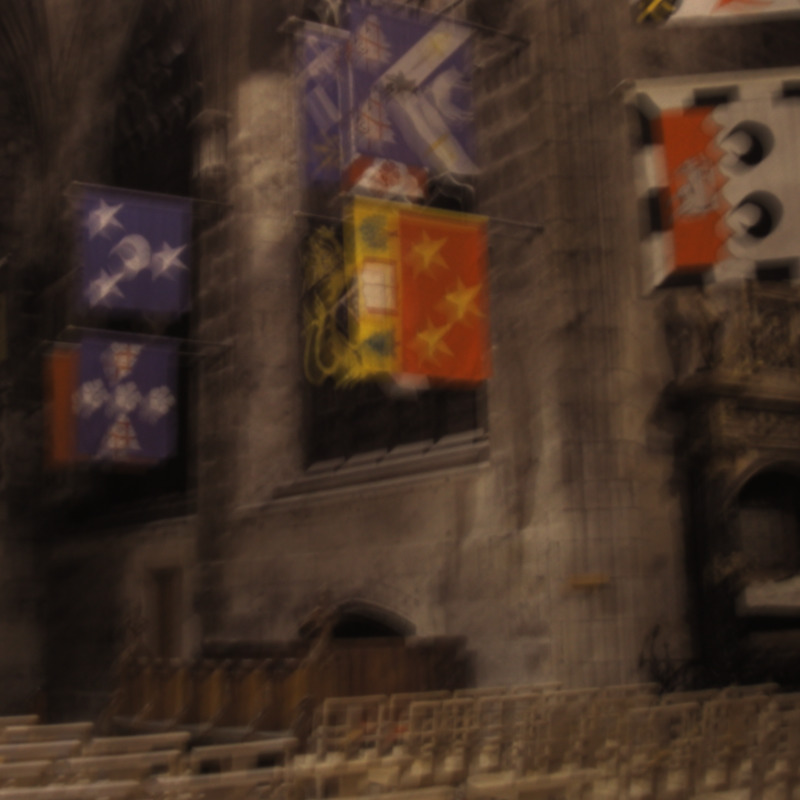}  &
    \includegraphics[trim=100 170 100 50, clip,width=0.31\textwidth]{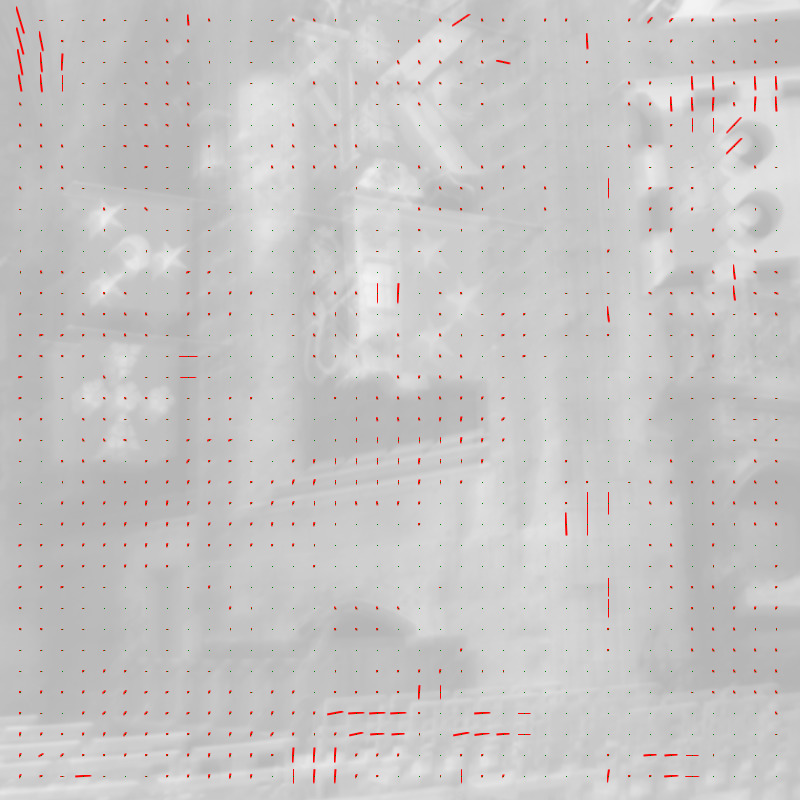}   &
      \includegraphics[trim=100 170 100 50, clip,width=0.31\textwidth]{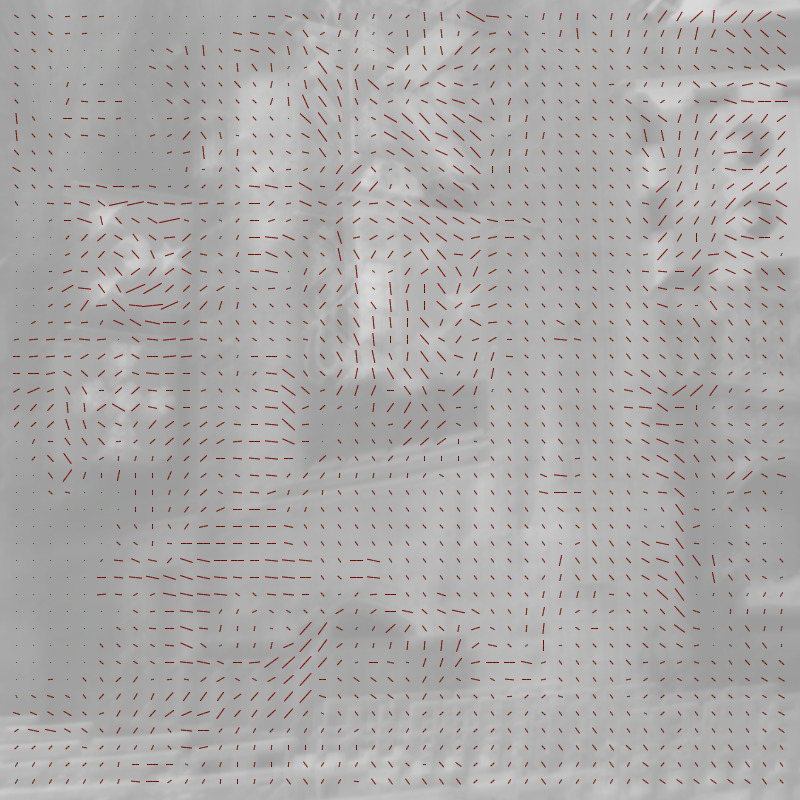}   \\  
      Zhang \cite{zhang2021exposure} & J-MKPD (ours) & GT \\
   \includegraphics[trim=100 170 100 50, clip,width=0.31\textwidth]{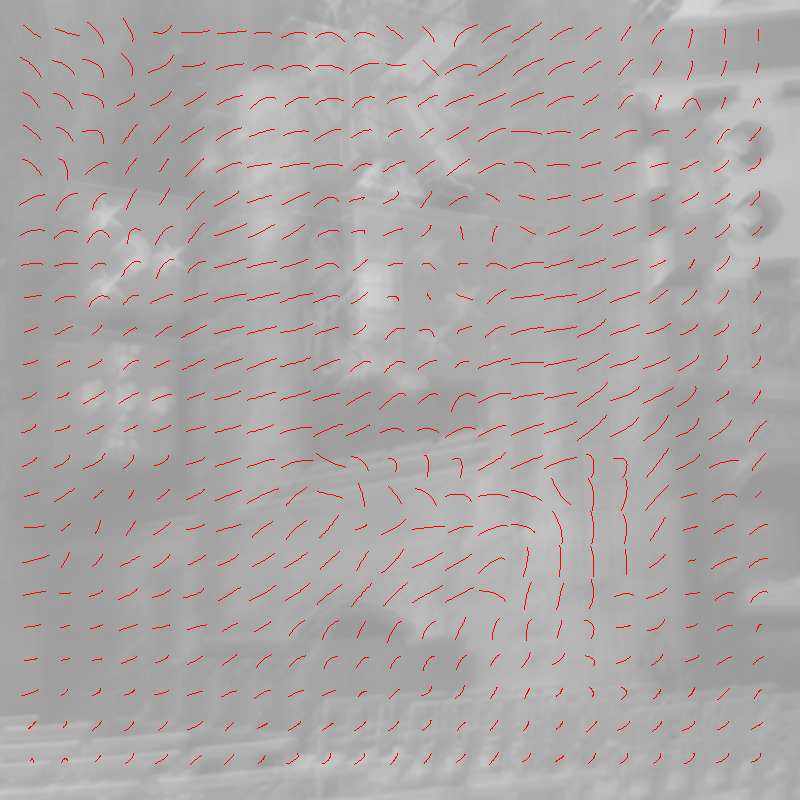}  &
     \includegraphics[trim=100 170 100 50, clip,width=0.31\textwidth]{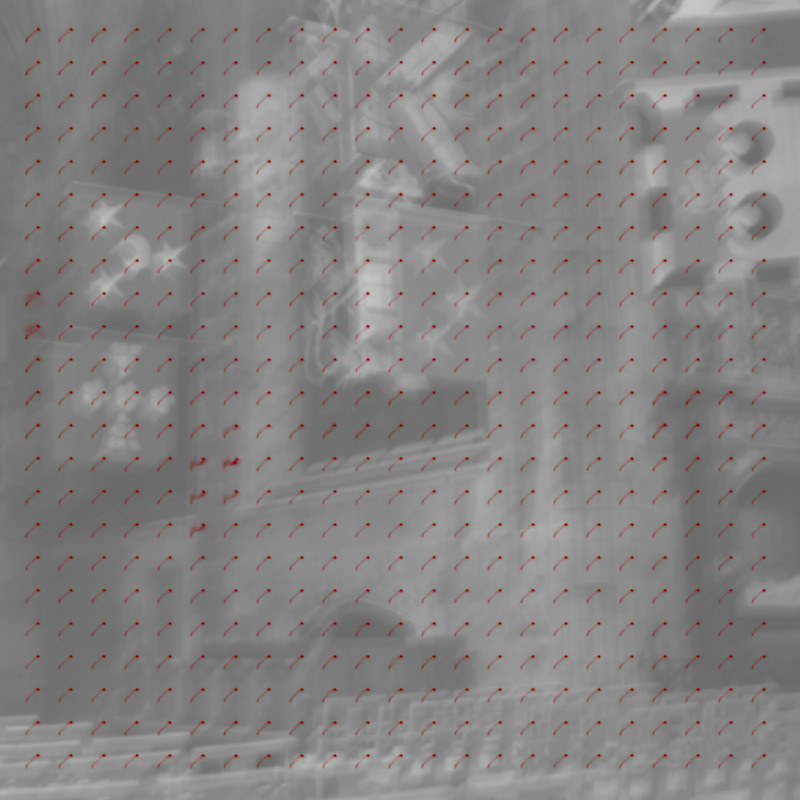}  & 
     {\setlength{\fboxsep}{0pt}%
       \setlength{\fboxrule}{0.05pt}%
        \framebox{%
      \includegraphics[trim=100 170 100 50, clip,width=0.31\textwidth]{imgs/Kohler/kernels_alpha/Blurry1_1_gt_kernel_alpha}}}    \\
      \hline
             Blurry & Gong \cite{gong2017motion} & Sun \cite{sun2015learning} \\
    \includegraphics[trim=100 170 100 50, clip,width=0.31\textwidth]{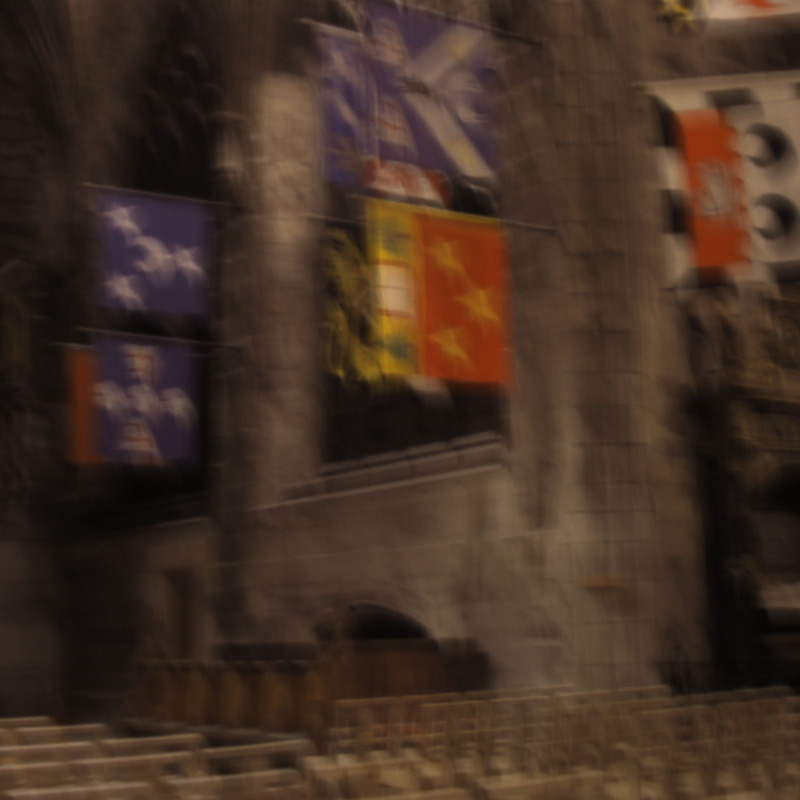}  &
    \includegraphics[trim=100 170 100 50, clip,width=0.31\textwidth]{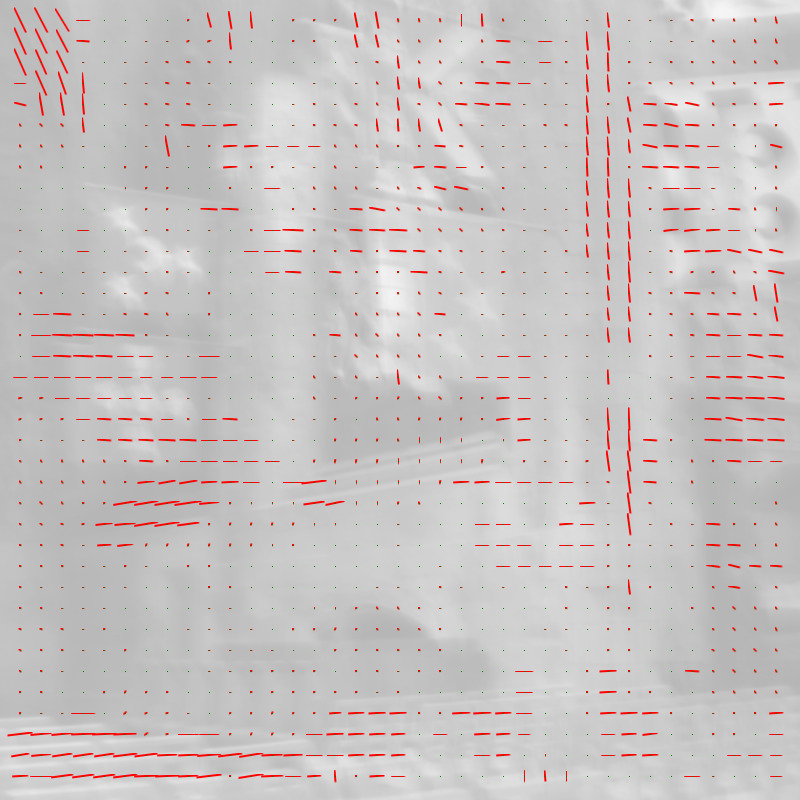}   &
      \includegraphics[trim=100 170 100 50, clip,width=0.31\textwidth]{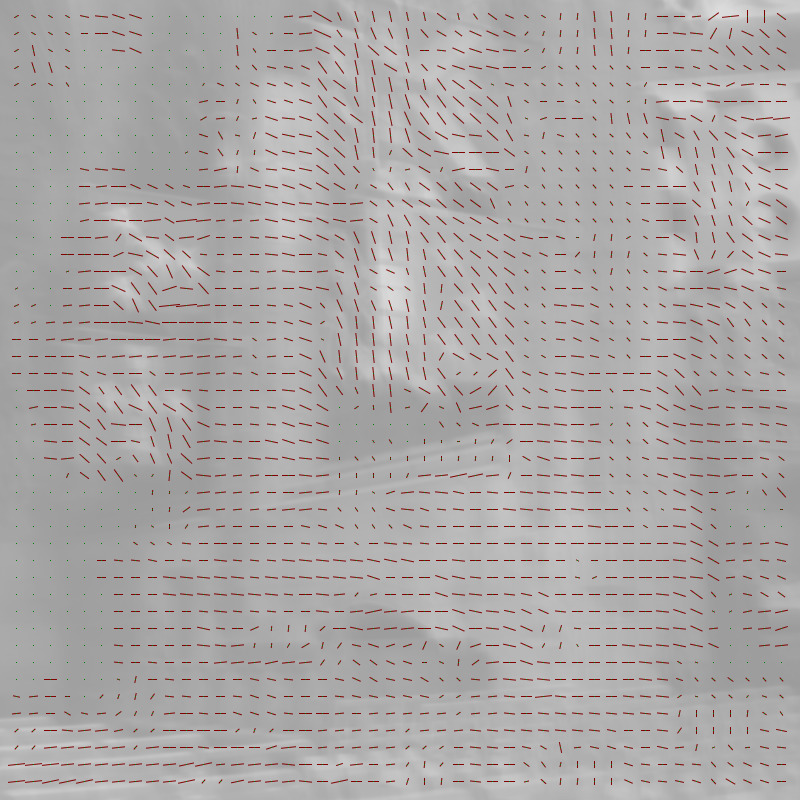}   \\ 
      Zhang \cite{zhang2021exposure} & J-MKPD (ours) & GT \\
   \includegraphics[trim=100 170 100 50, clip,width=0.31\textwidth]{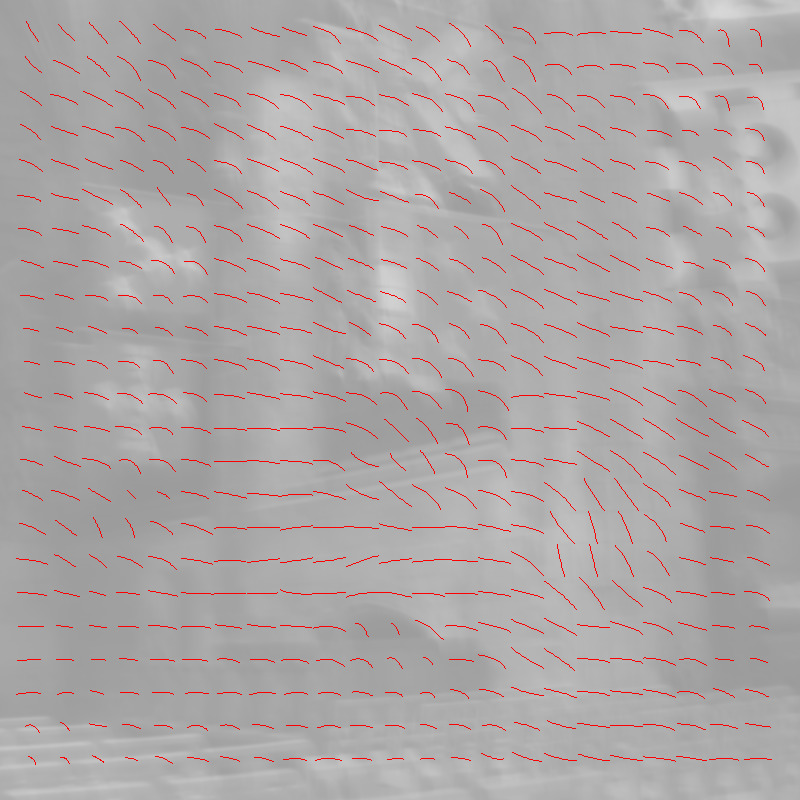}  &
     \includegraphics[trim=100 170 100 50, clip,width=0.31\textwidth]{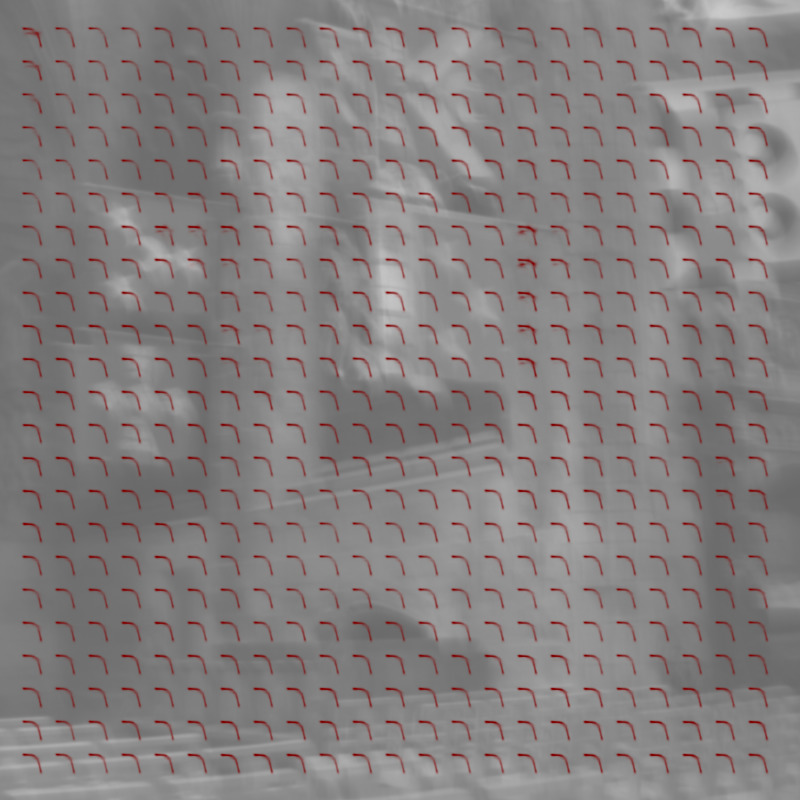}  & 
     {\setlength{\fboxsep}{0pt}%
       \setlength{\fboxrule}{0.05pt}%
        \framebox{%
      \includegraphics[trim=100 170 100 50, clip,width=0.31\textwidth]{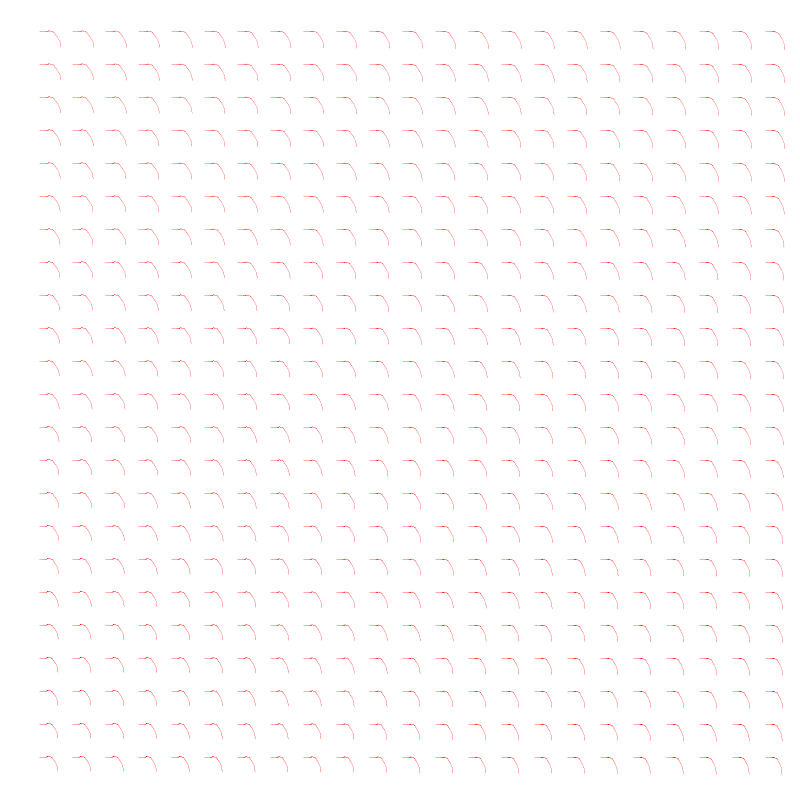}}}    
  \end{tabular}
  \caption{\textbf{Visual comparison of per-pixel kernel estimation on a real image with known ground-truth (GT).} %
  Motion fields generated by other methods are correlated with the image structure, suffer from the aperture problem, or predict deltas in low-variance regions. Best viewed in electronic format. }
  \label{fig:KernelEstimationKohler_1}
\end{figure*}

\begin{figure*}[ht]
  \centering
\setlength{\tabcolsep}{2pt}
  \begin{tabular}{ccc}
  Blurry & Gong \cite{gong2017motion} & Sun \cite{sun2015learning} \\
    \includegraphics[trim=100 170 100 50, clip,width=0.31\textwidth]{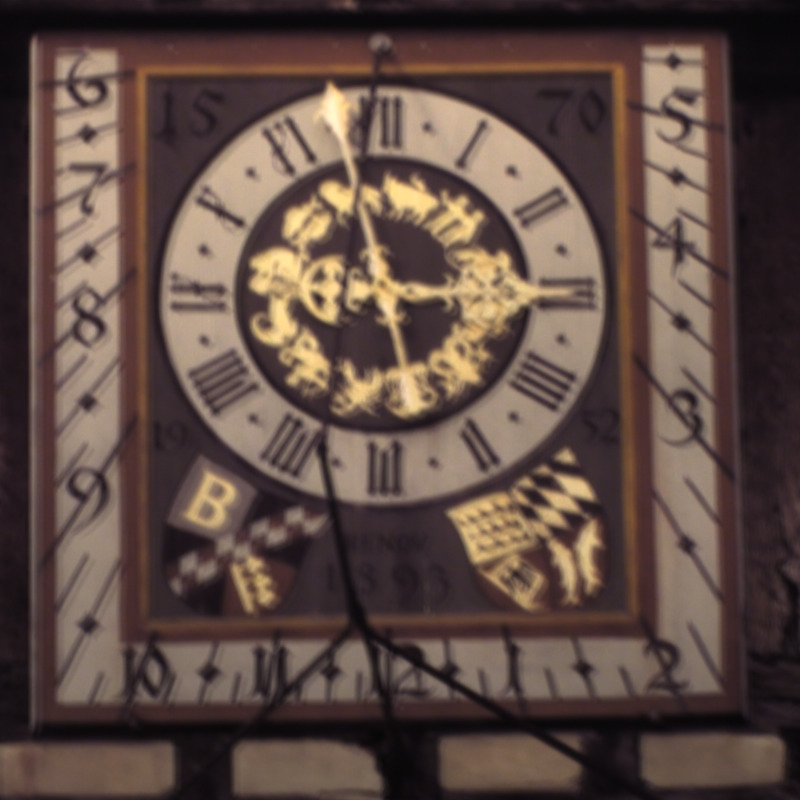}  &
    \includegraphics[trim=100 170 100 50, clip,width=0.31\textwidth]{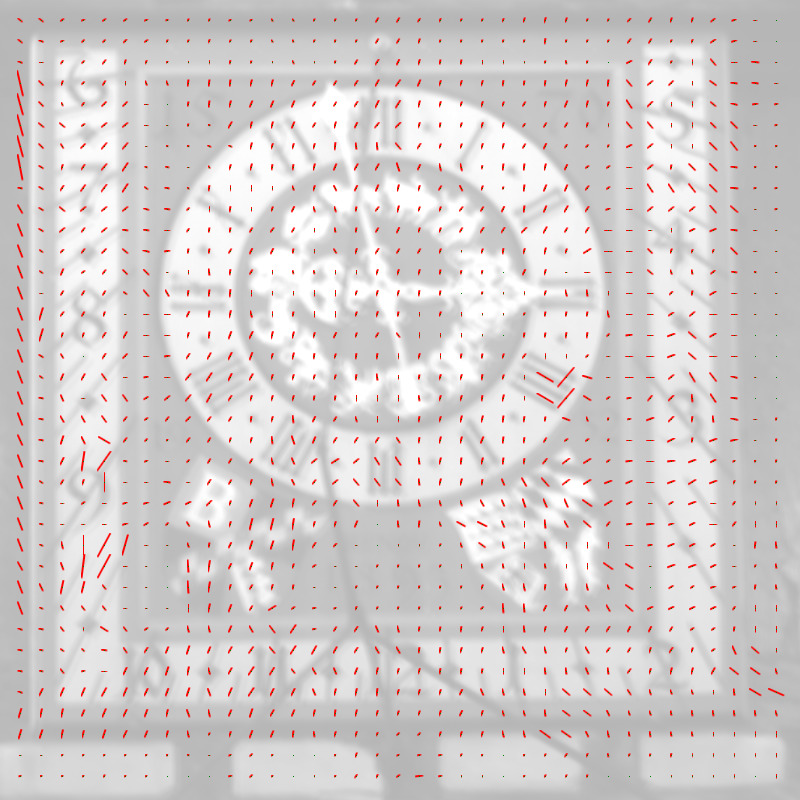}   &
      \includegraphics[trim=100 170 100 50, clip,width=0.31\textwidth]{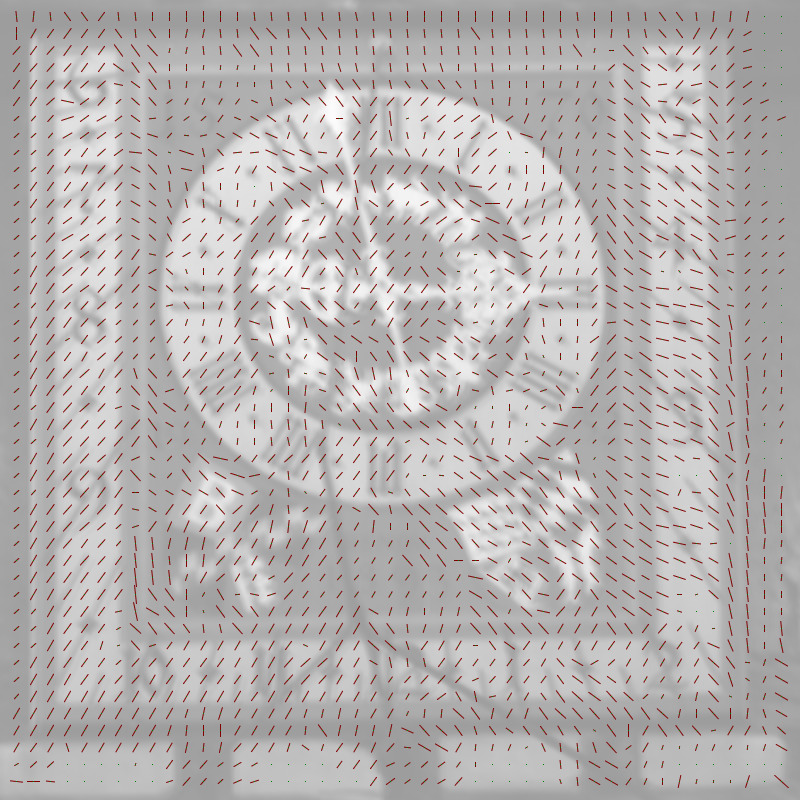}   \\ 
      Zhang \cite{zhang2021exposure} & J-MKPD (ours) & GT \\
   \includegraphics[trim=100 170 100 50, clip,width=0.31\textwidth]{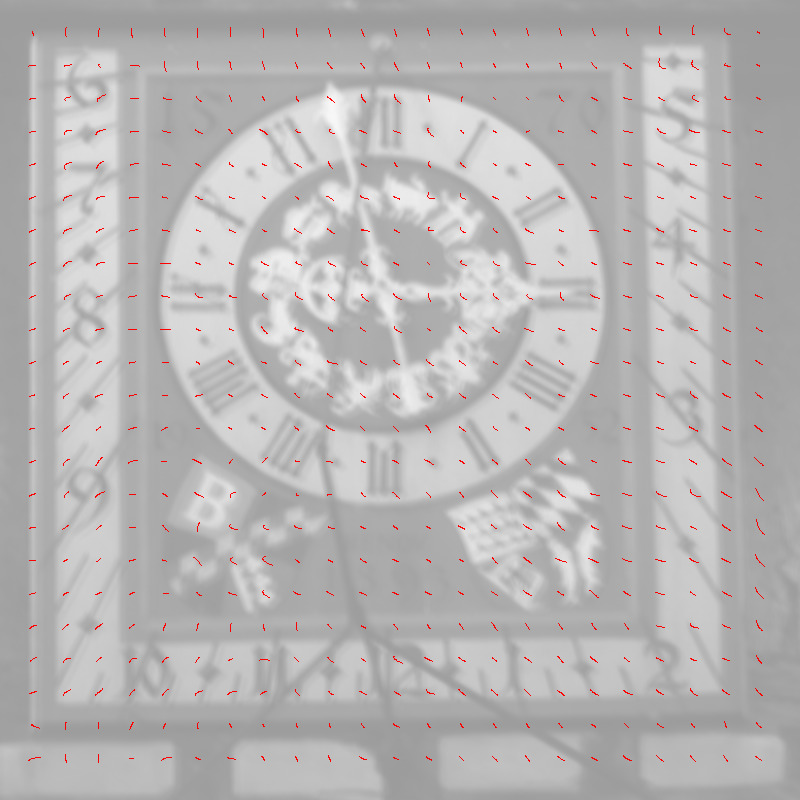}  &
     \includegraphics[trim=100 170 100 50, clip,width=0.31\textwidth]{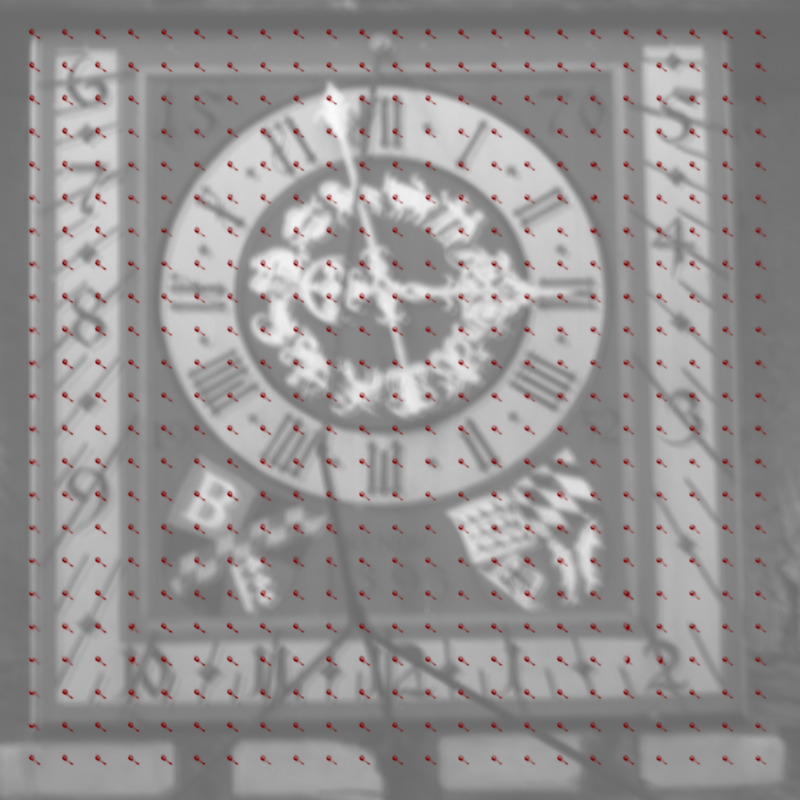}  & 
     {\setlength{\fboxsep}{0pt}%
       \setlength{\fboxrule}{0.05pt}%
        \framebox{%
      \includegraphics[trim=100 170 100 50, clip,width=0.31\textwidth]{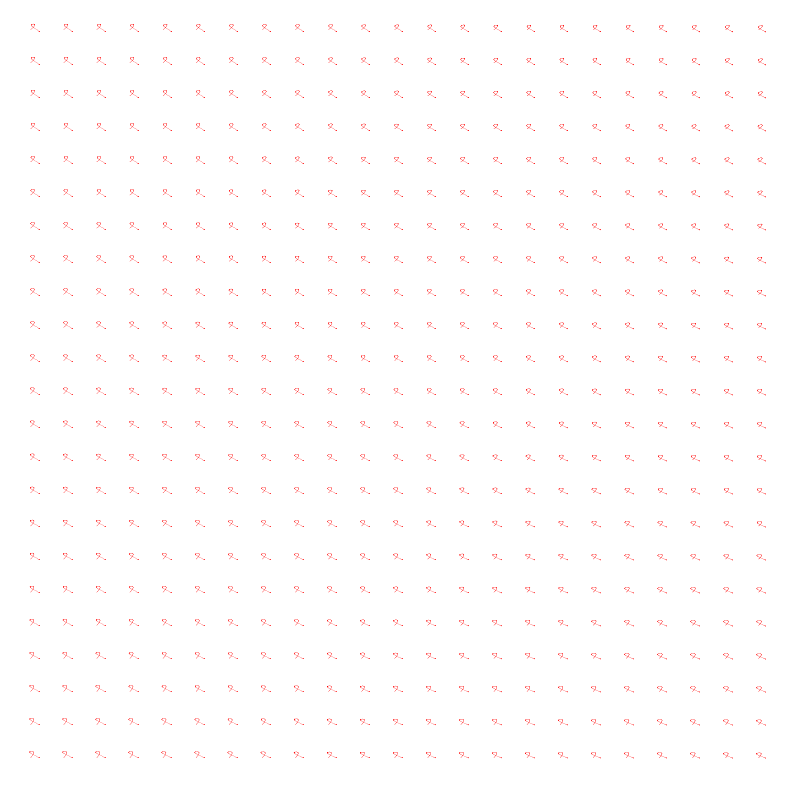}}}   \\ 
      \hline
      Blurry & Gong \cite{gong2017motion} & Sun \cite{sun2015learning} \\
    \includegraphics[trim=100 170 100 50, clip,width=0.31\textwidth]{imgs/Kohler/Blurry/Blurry2_4.jpg}  &
    \includegraphics[trim=100 170 100 50, clip,width=0.31\textwidth]{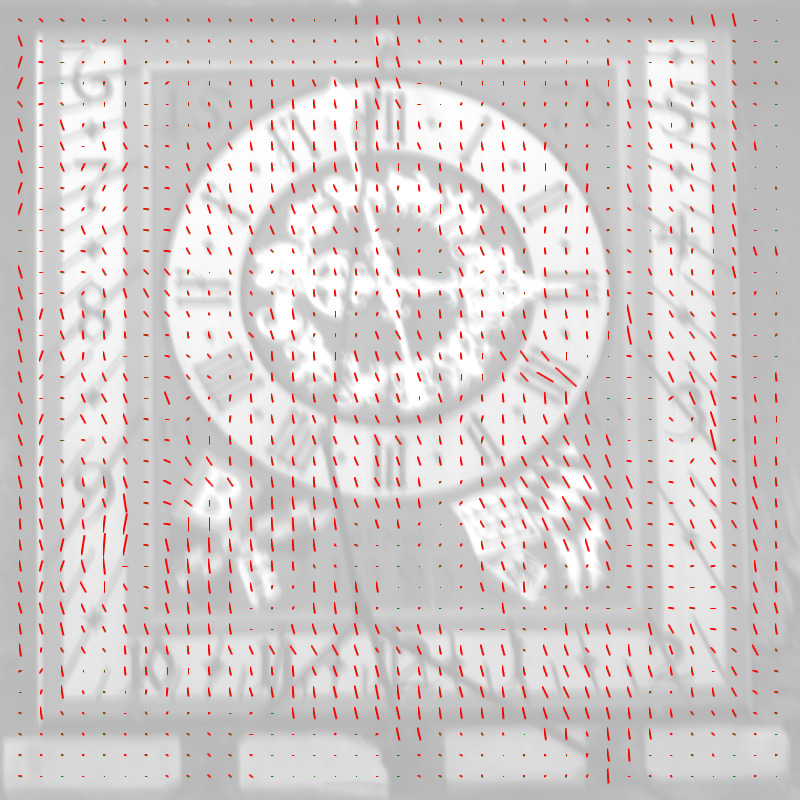}   &
      \includegraphics[trim=100 170 100 50, clip,width=0.31\textwidth]{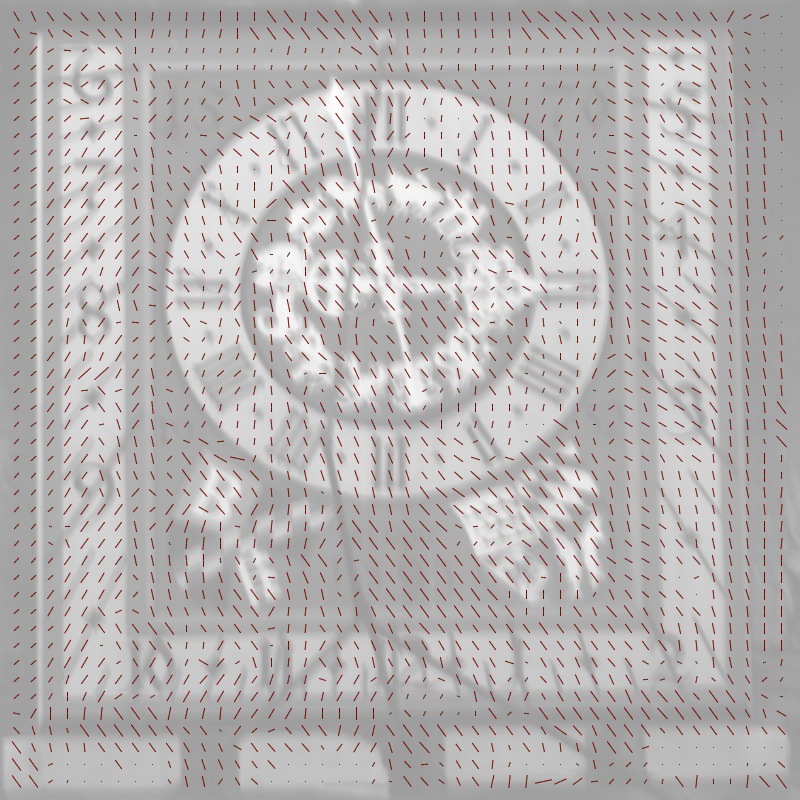}   \\ 
      Zhang \cite{zhang2021exposure} &  J-MKPD (ours) & GT \\
   \includegraphics[trim=100 170 100 50, clip,width=0.31\textwidth]{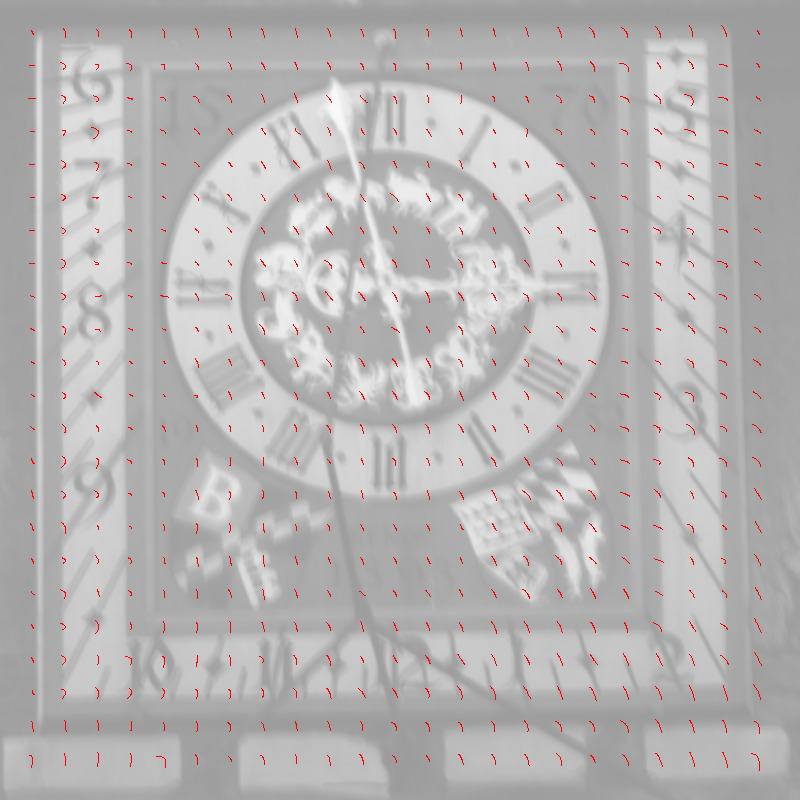}  &
     \includegraphics[trim=100 170 100 50, clip,width=0.31\textwidth]{imgs/Kohler/COCO900_restL2_aug_all_loss_gf1_80k/kernels/Blurry2_4_kernels.jpg}  & 
     {\setlength{\fboxsep}{0pt}%
       \setlength{\fboxrule}{0.05pt}%
        \framebox{%
      \includegraphics[trim=100 170 100 50, clip,width=0.31\textwidth]{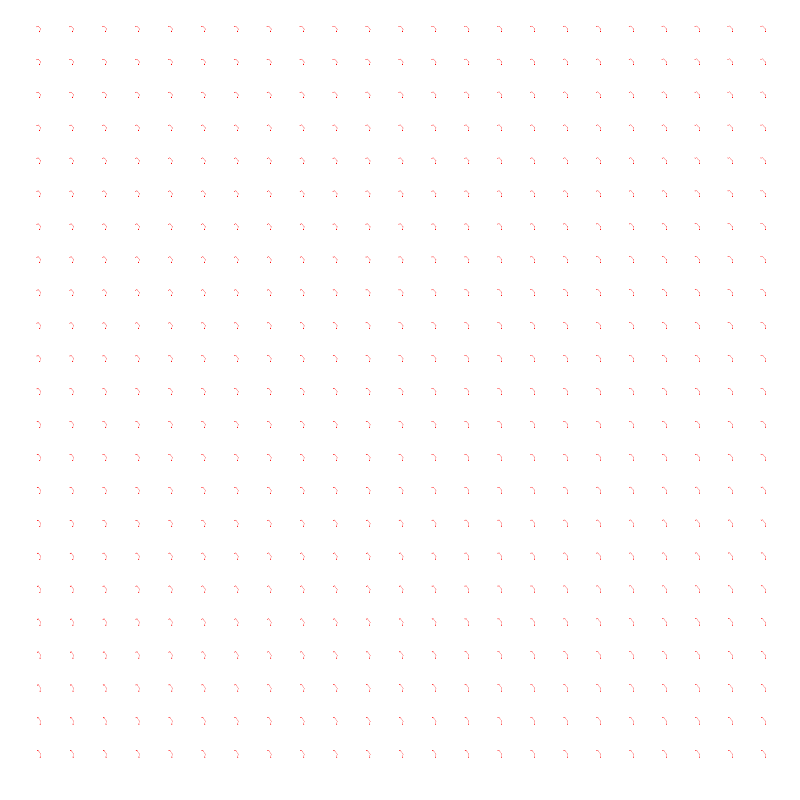}}}    
  \end{tabular}
  \caption{\textbf{Visual comparison of per-pixel kernel estimation on a real image with known ground-truth (GT).} %
  Motion fields generated by other methods are correlated with the image structure, suffer from the aperture problem, or predict deltas in low-variance regions. Best viewed in electronic format. }
  \label{fig:KernelEstimationKohler_2}
\end{figure*}

\begin{figure*}[ht]
  \centering
\setlength{\tabcolsep}{2pt}
  \begin{tabular}{ccc}
         Blurry & Gong \cite{gong2017motion} & Sun \cite{sun2015learning} \\ 
     \includegraphics[trim=100 170 100 50, clip,width=0.31\textwidth]{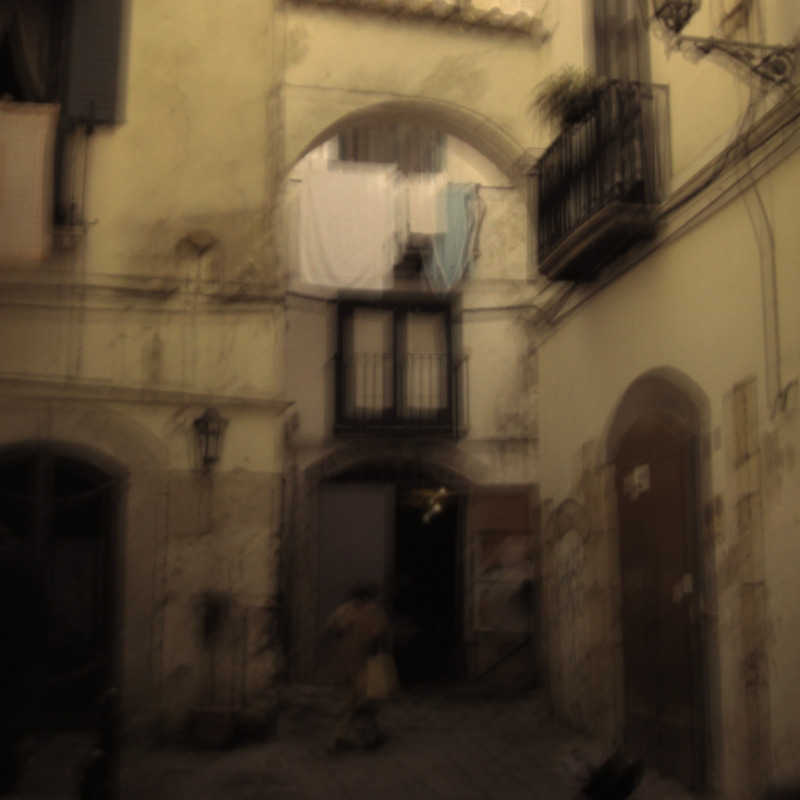}  &
    \includegraphics[trim=100 170 100 50, clip,width=0.31\textwidth]{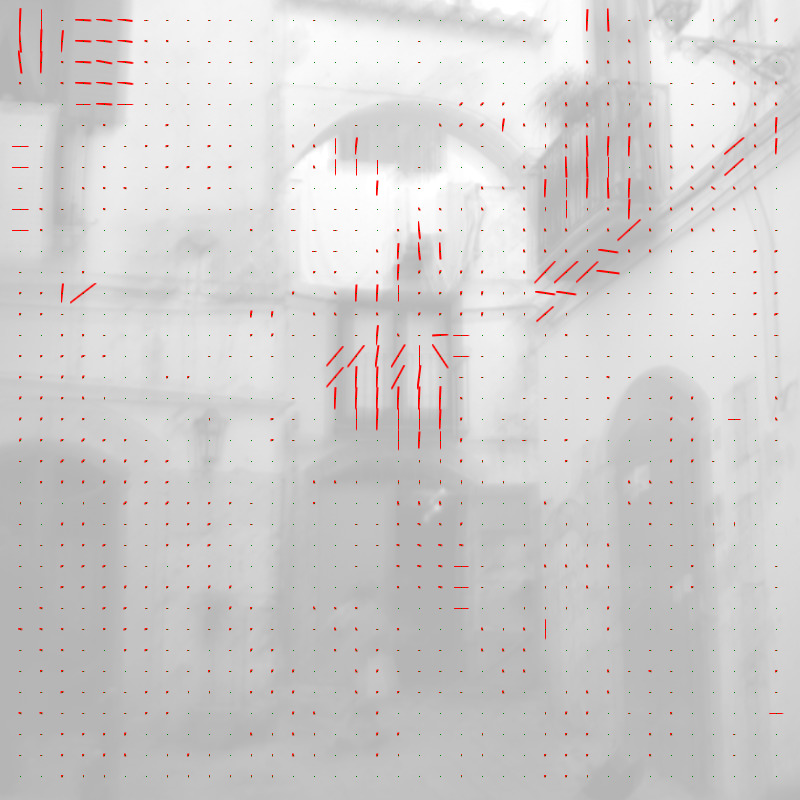}   &
      \includegraphics[trim=100 170 100 50, clip,width=0.31\textwidth]{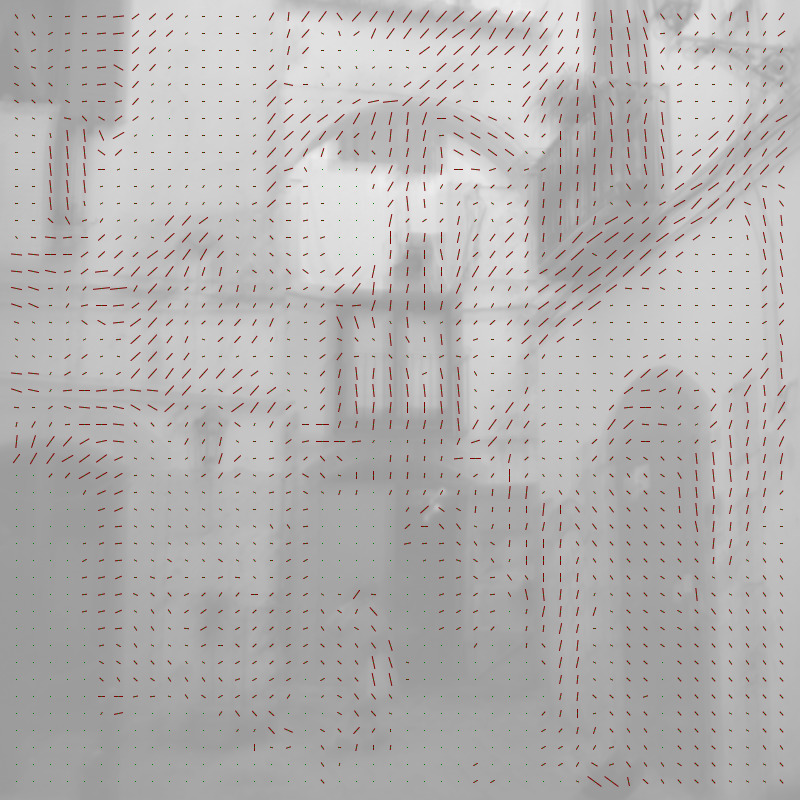}   \\       
      Zhang \cite{zhang2021exposure} &  J-MKPD (ours) & GT \\
   \includegraphics[trim=100 170 100 50, clip,width=0.31\textwidth]{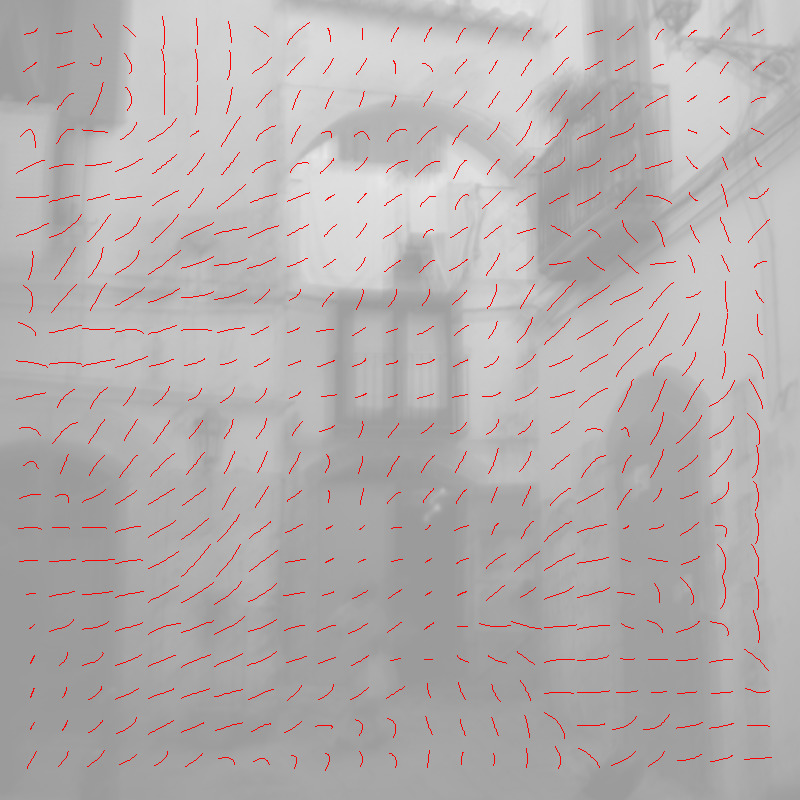}  &
     \includegraphics[trim=100 170 100 50, clip,width=0.31\textwidth]{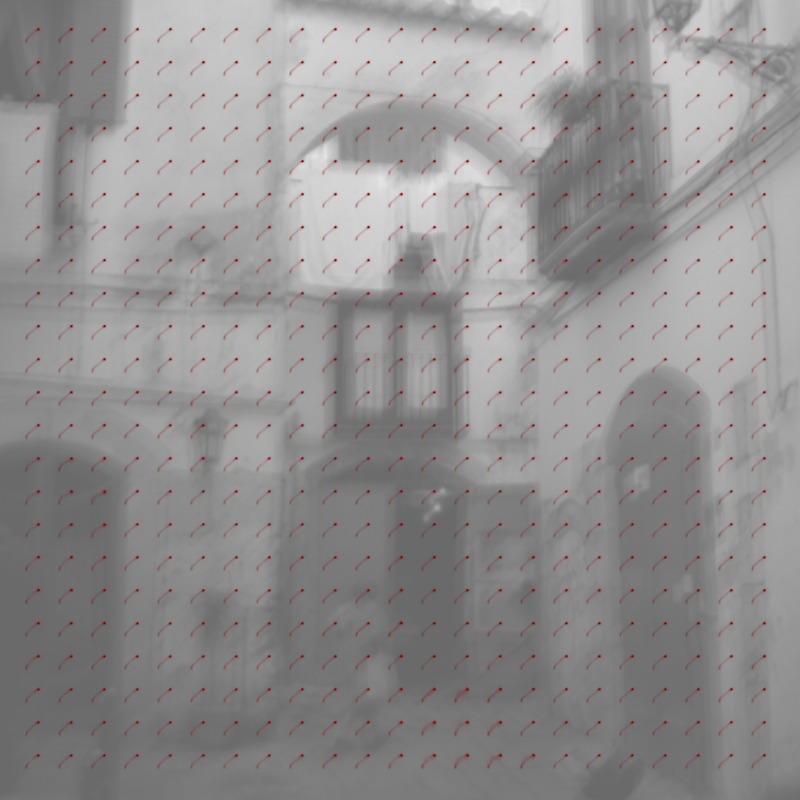}  & 
     {\setlength{\fboxsep}{0pt}%
       \setlength{\fboxrule}{0.05pt}%
        \framebox{%
      \includegraphics[trim=100 170 100 50, clip,width=0.31\textwidth]{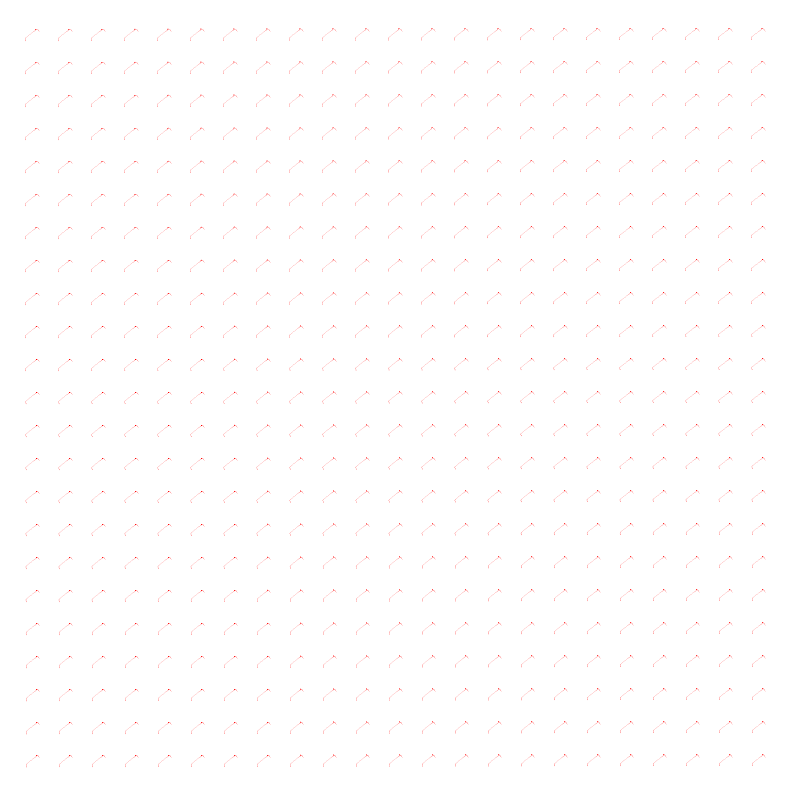}}}    \\
      \hline
             Blurry & Gong \cite{gong2017motion} & Sun \cite{sun2015learning} \\ 
     \includegraphics[trim=100 170 100 50, clip,width=0.31\textwidth]{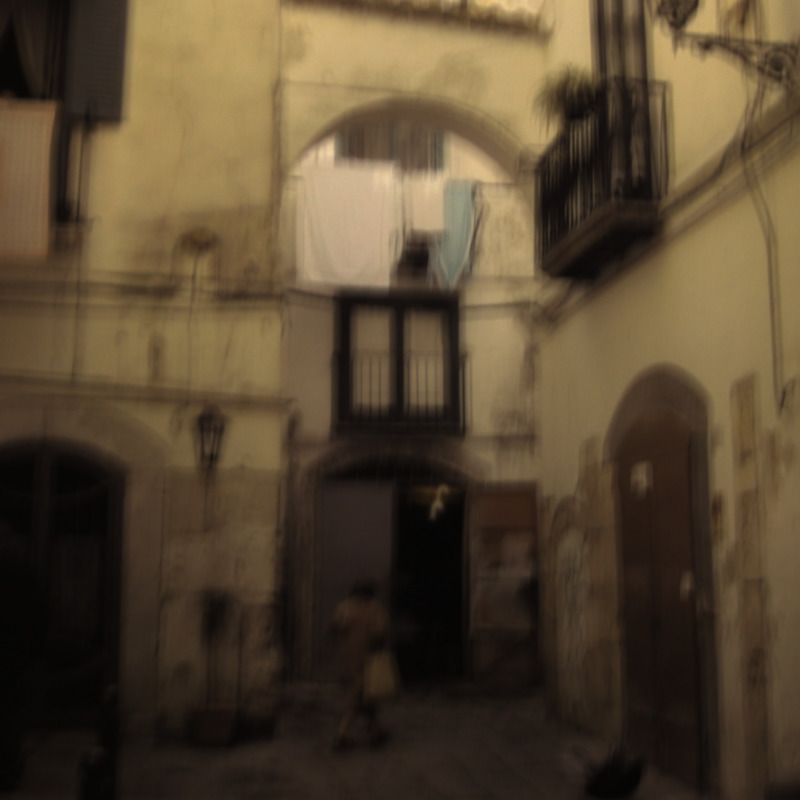}  &
    \includegraphics[trim=100 170 100 50, clip,width=0.31\textwidth]{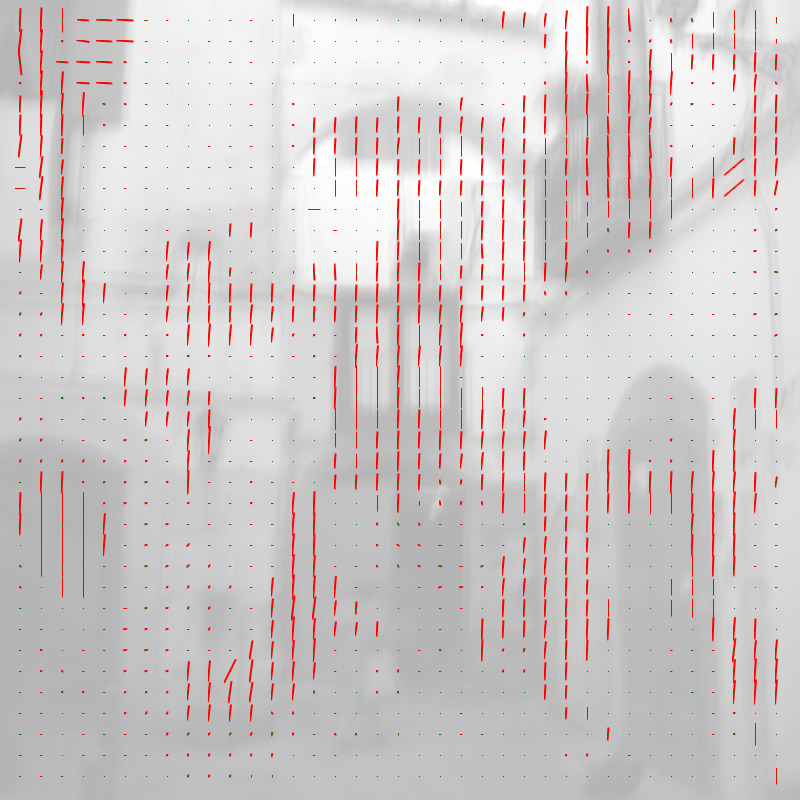}   &
      \includegraphics[trim=100 170 100 50, clip,width=0.31\textwidth]{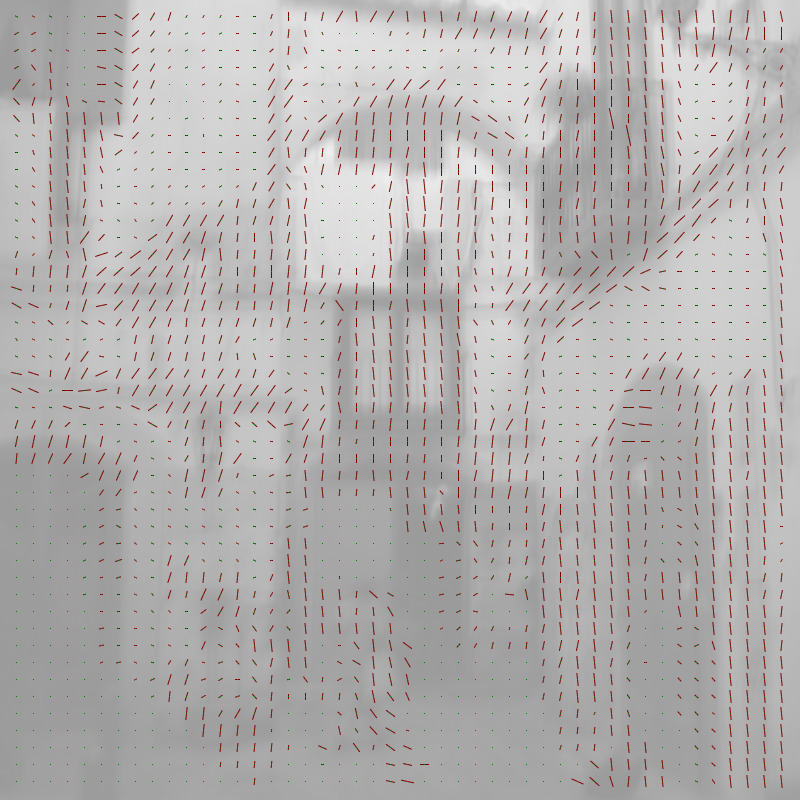}   \\ 
    Zhang \cite{zhang2021exposure} &  J-MKPD (ours) & GT \\
   \includegraphics[trim=100 170 100 50, clip,width=0.31\textwidth]{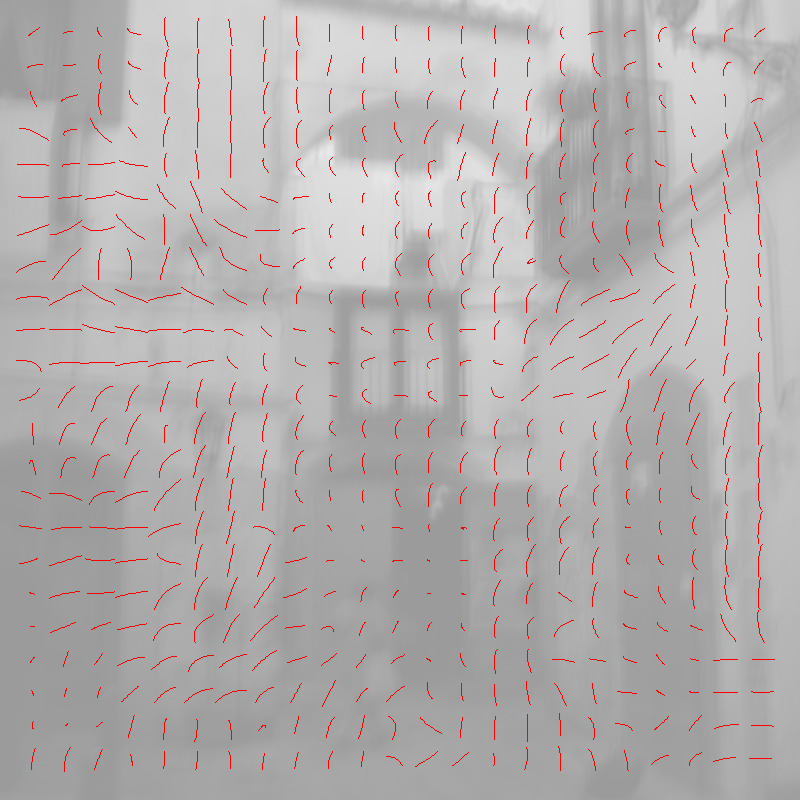}  &
     \includegraphics[trim=100 170 100 50, clip,width=0.31\textwidth]{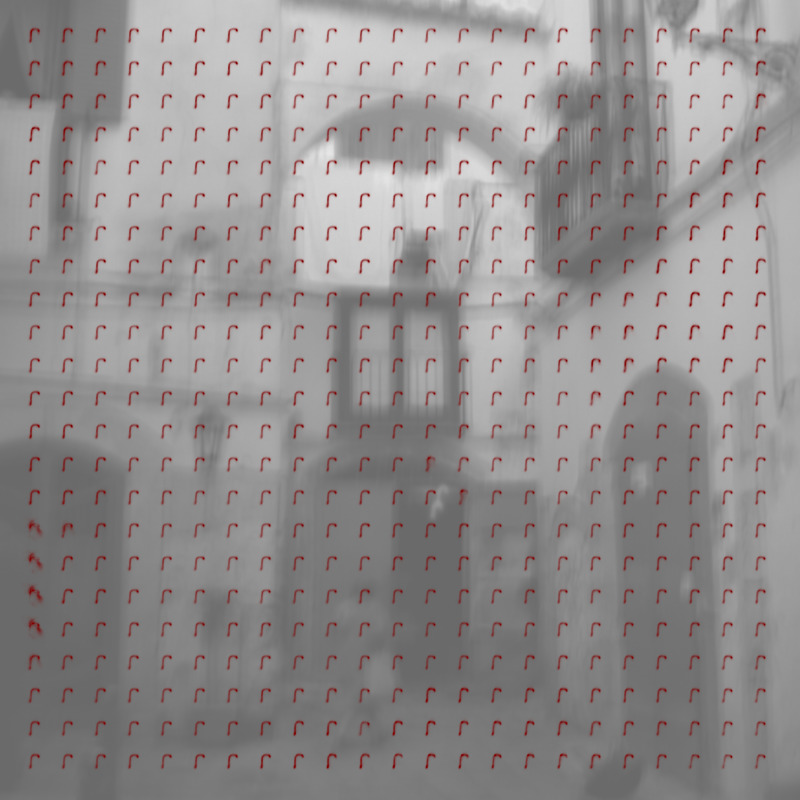}  & 
     {\setlength{\fboxsep}{0pt}%
       \setlength{\fboxrule}{0.05pt}%
        \framebox{%
      \includegraphics[trim=100 170 100 50, clip,width=0.31\textwidth]{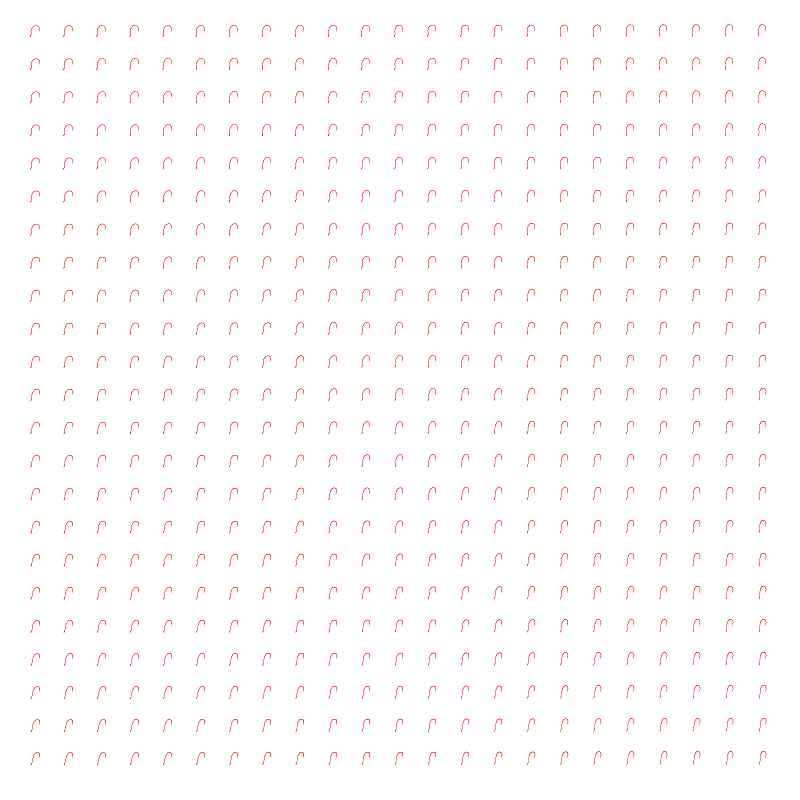}}}    
  \end{tabular}
  \caption{\textbf{Visual comparison of per-pixel kernel estimation on a real image with known ground-truth (GT).} %
  Motion fields generated by other methods are correlated with the image structure, suffer from the aperture problem, or predict deltas in low-variance regions. Best viewed in electronic format. }
  \label{fig:KernelEstimationKohler_3}
\end{figure*}

\begin{figure*}[ht]
  \centering
\setlength{\tabcolsep}{2pt}

  \begin{tabular}{ccc}
       Blurry & Gong \cite{gong2017motion} & Sun \cite{sun2015learning} \\ 
       \includegraphics[trim=100 170 100 50, clip,width=0.31\textwidth]{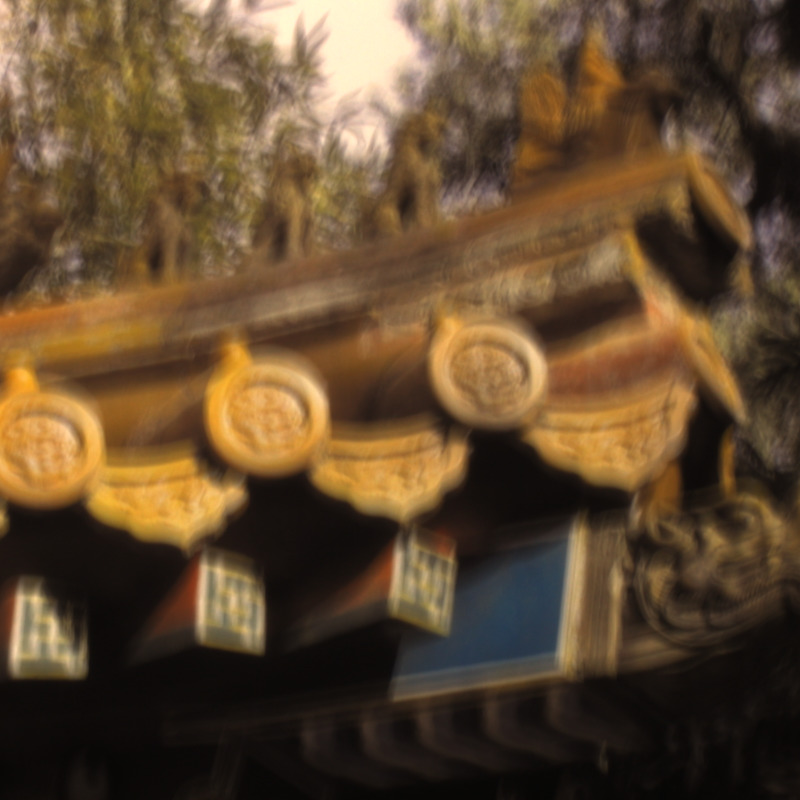}  &
    \includegraphics[trim=100 170 100 50, clip,width=0.31\textwidth]{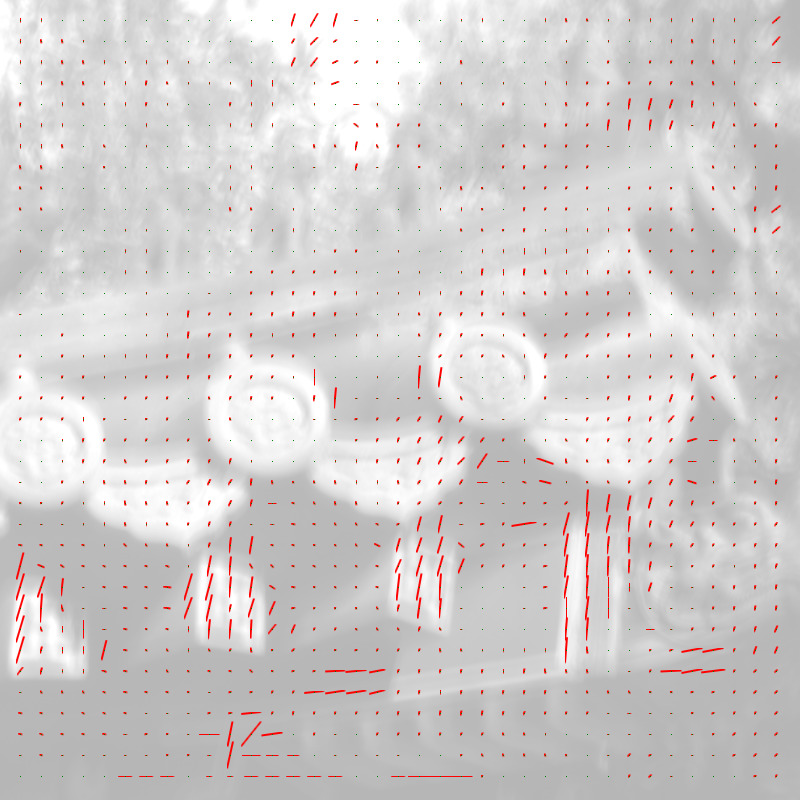}   &
      \includegraphics[trim=100 170 100 50, clip,width=0.31\textwidth]{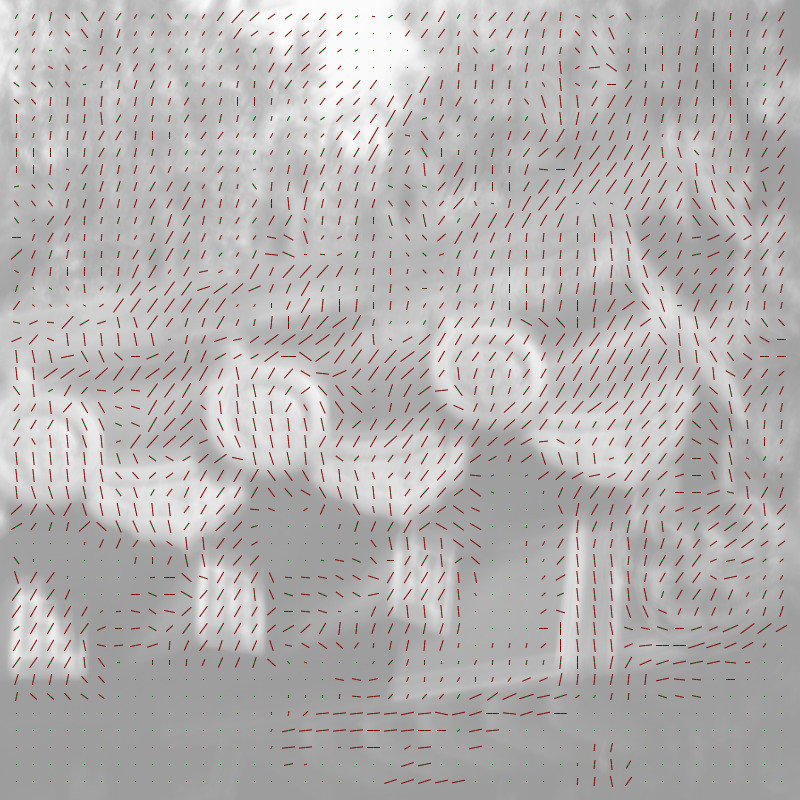}   \\
      Zhang \cite{zhang2021exposure} &  J-MKPD (ours) & GT \\
   \includegraphics[trim=100 170 100 50, clip,width=0.31\textwidth]{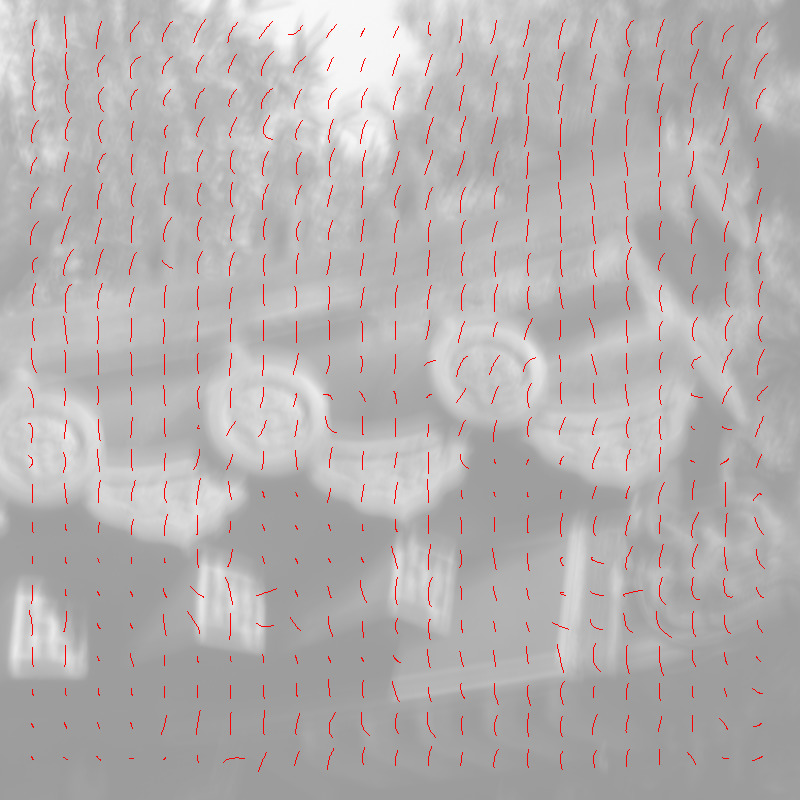}  &
     \includegraphics[trim=100 170 100 50, clip,width=0.31\textwidth]{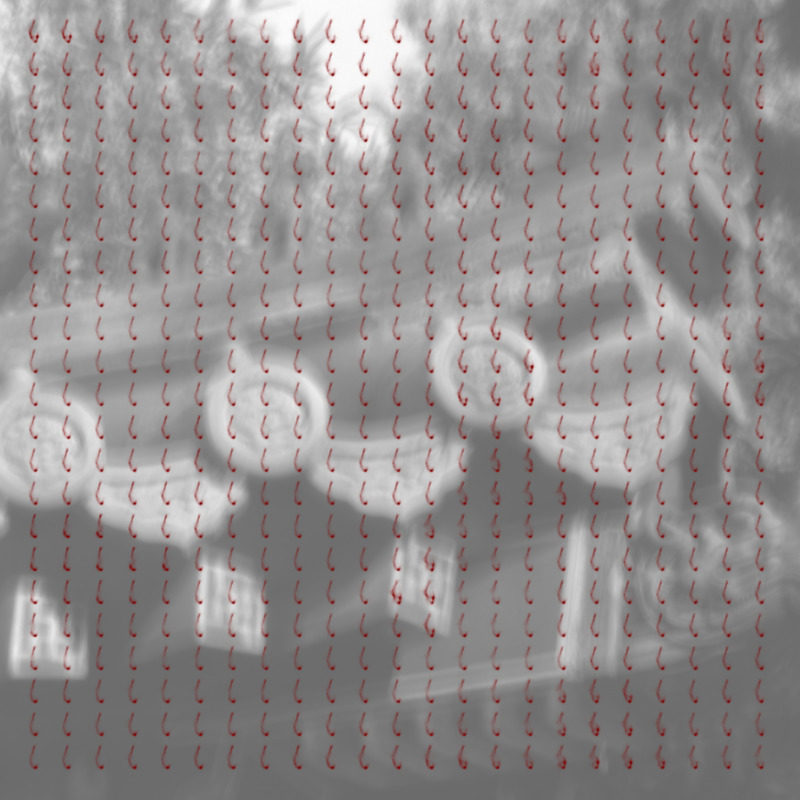}  & 
     {\setlength{\fboxsep}{0pt}%
       \setlength{\fboxrule}{0.05pt}%
        \framebox{%
      \includegraphics[trim=100 170 100 50, clip,width=0.31\textwidth]{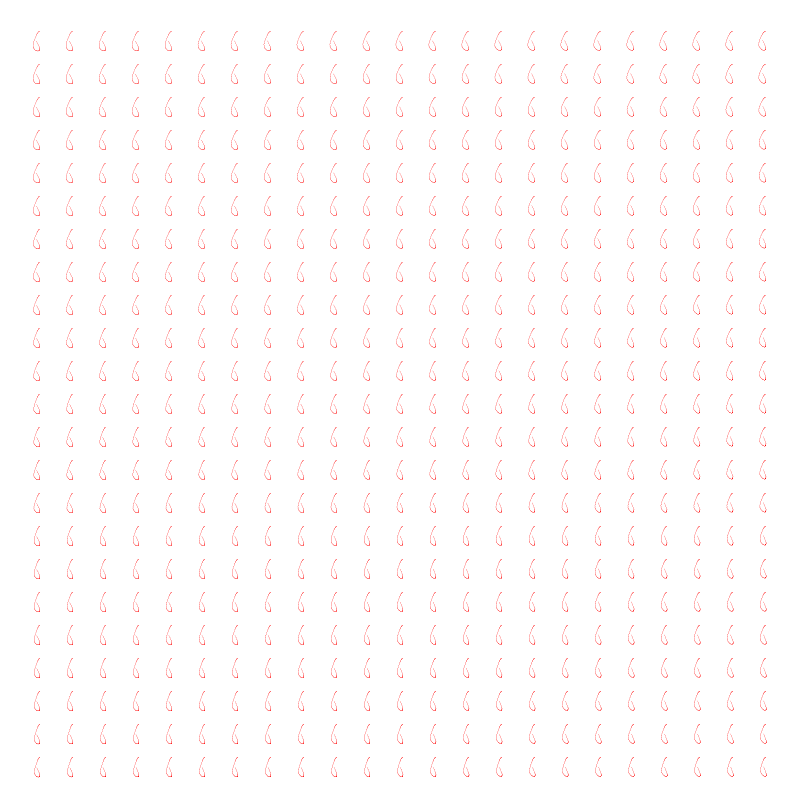}}}   \\
      \hline
       Blurry & Gong \cite{gong2017motion} & Sun \cite{sun2015learning} \\
     \includegraphics[trim=100 170 100 50, clip,width=0.31\textwidth]{imgs/Kohler/Blurry/Blurry4_6.jpg}  &
    \includegraphics[trim=100 170 100 50, clip,width=0.31\textwidth]{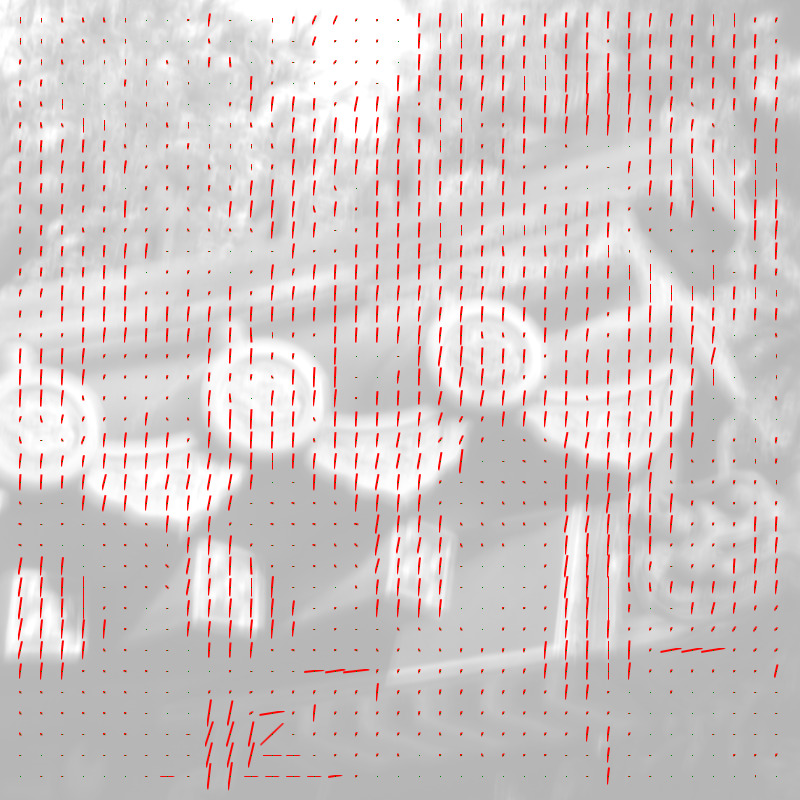}   &
      \includegraphics[trim=100 170 100 50, clip,width=0.31\textwidth]{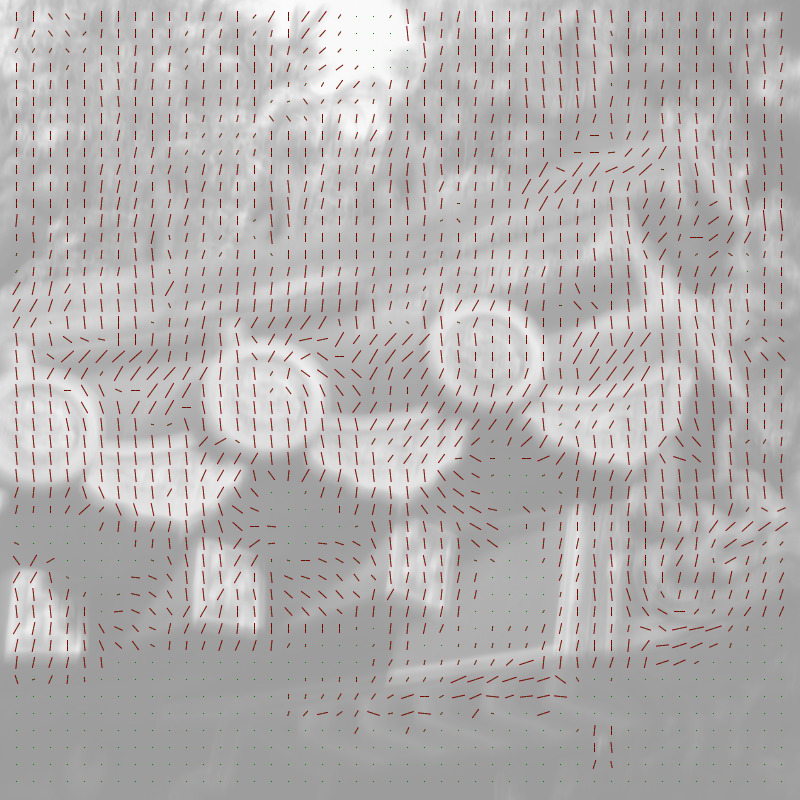}   \\ 
      Zhang \cite{zhang2021exposure} &  J-MKPD (ours) & GT \\
   \includegraphics[trim=100 170 100 50, clip,width=0.31\textwidth]{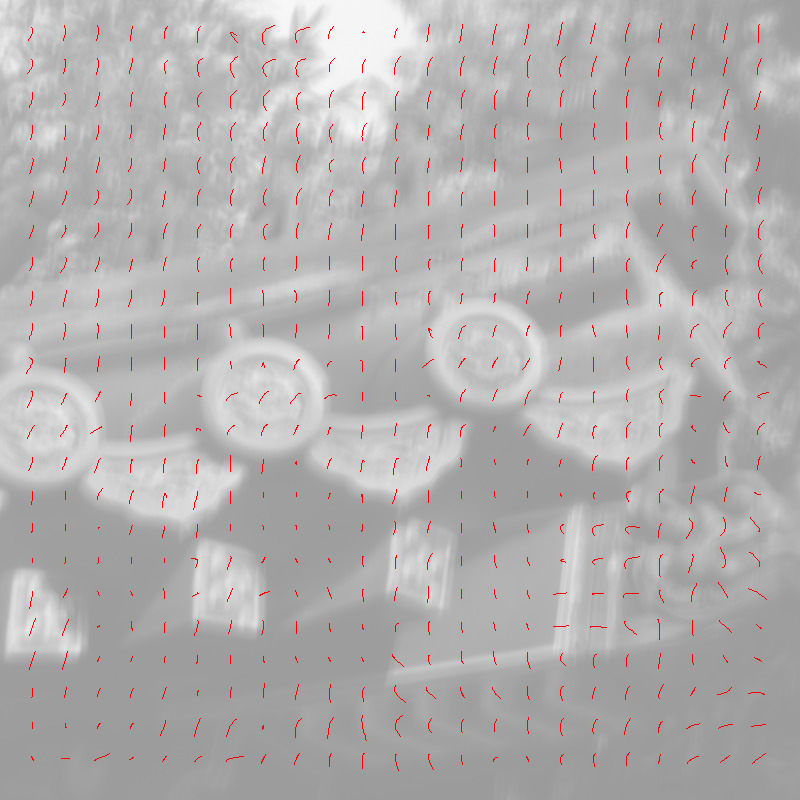}  &
     \includegraphics[trim=100 170 100 50, clip,width=0.31\textwidth]{imgs/Kohler/COCO900_restL2_aug_all_loss_gf1_80k/kernels/Blurry4_6_kernels.jpg}  & 
     {\setlength{\fboxsep}{0pt}%
       \setlength{\fboxrule}{0.05pt}%
        \framebox{%
      \includegraphics[trim=100 170 100 50, clip,width=0.31\textwidth]{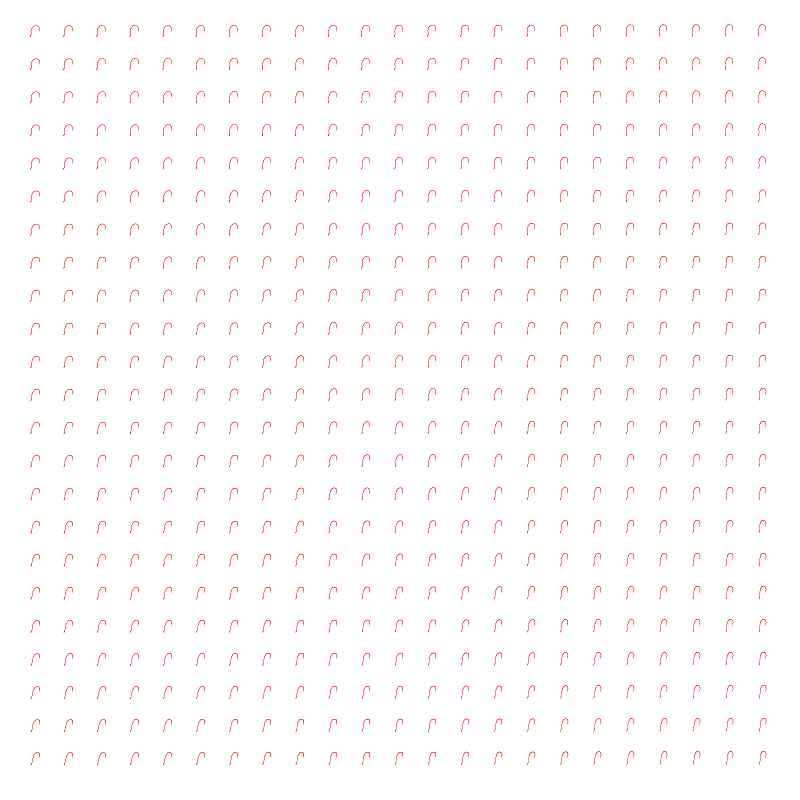}}}    
  \end{tabular}
  \caption{\textbf{Visual comparison of per-pixel kernel estimation on a real image with known ground-truth (GT).} %
  Motion fields generated by other methods are correlated with the image structure, suffer from the aperture problem, or predict deltas in low-variance regions. Best viewed in electronic format. }
  \label{fig:KernelEstimationKohler_4}
\end{figure*}

%% file: figure_kernels_Zhang_vs_ours.tex
\begin{figure}
    \centering
    \setlength{\tabcolsep}{2pt}
    \begin{tabular}{cccc}
    a) & b) & c) & d) \\
    \includegraphics[trim=100 170 100 50, clip, width=0.24\textwidth]{imgs/Kohler/Zhang/Blurry1_12_flow_map.jpg} &
    \includegraphics[trim=100 170 100 50, clip, width=0.24\textwidth]{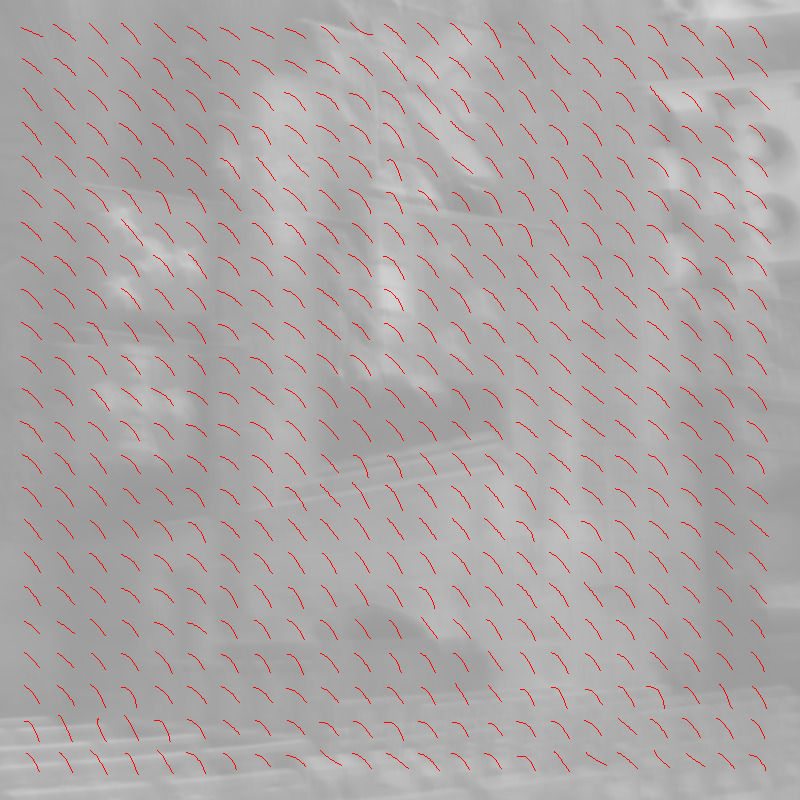} &
    \includegraphics[trim=100 170 100 50, clip, width=0.24\textwidth]{imgs/Kohler/COCO900_restL2_aug_all_loss_gf1_80k/kernels/Blurry1_12_kernels.jpg}  &
        {\setlength{\fboxsep}{0pt}%
       \setlength{\fboxrule}{0.05pt}%
        \framebox{%
        \includegraphics[trim=100 170 100 50, clip, width=0.24\textwidth]{imgs/Kohler/kernels_alpha/Blurry1_12_gt_kernel_alpha} }}   \\
    \includegraphics[trim=100 170 100 50, clip, width=0.24\textwidth]{imgs/Kohler/Zhang/Blurry2_4_flow_map.jpg} &
    \includegraphics[trim=100 170 100 50, clip, width=0.24\textwidth]{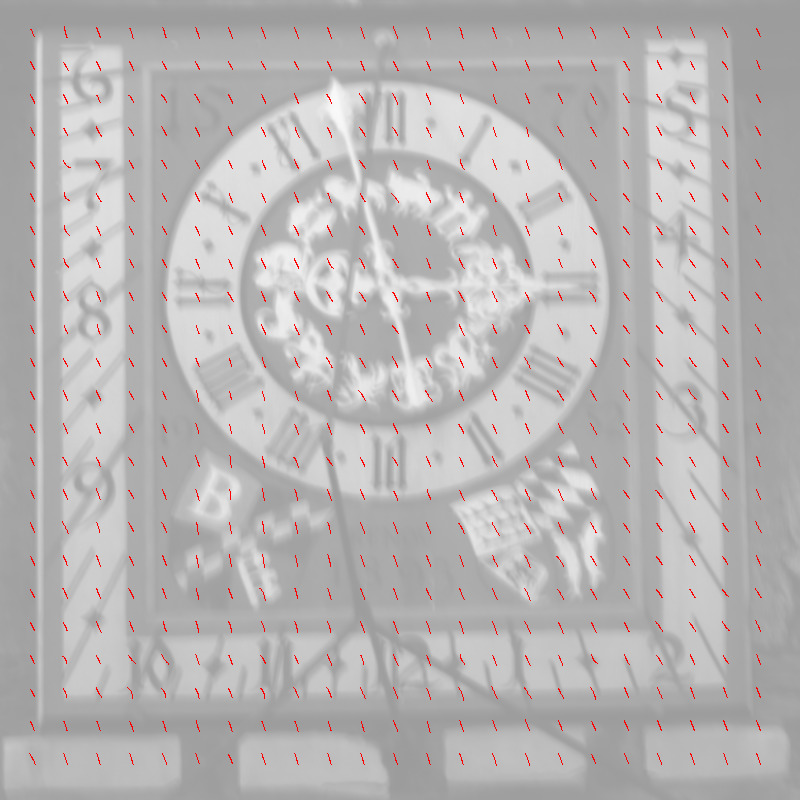} &
    \includegraphics[trim=100 170 100 50, clip, width=0.24\textwidth]{imgs/Kohler/COCO900_restL2_aug_all_loss_gf1_80k/kernels/Blurry2_4_kernels.jpg}  &
    {\setlength{\fboxsep}{0pt}%
       \setlength{\fboxrule}{0.05pt}%
        \framebox{%
    \includegraphics[trim=100 170 100 50, clip, width=0.24\textwidth]{imgs/Kohler/kernels_alpha/Blurry2_4_gt_kernel_alpha} }} \\
    \includegraphics[trim=100 170 100 50, clip, width=0.24\textwidth]{imgs/Kohler/Zhang/Blurry3_1_flow_map.jpg} &
    \includegraphics[trim=100 170 100 50, clip, width=0.24\textwidth]{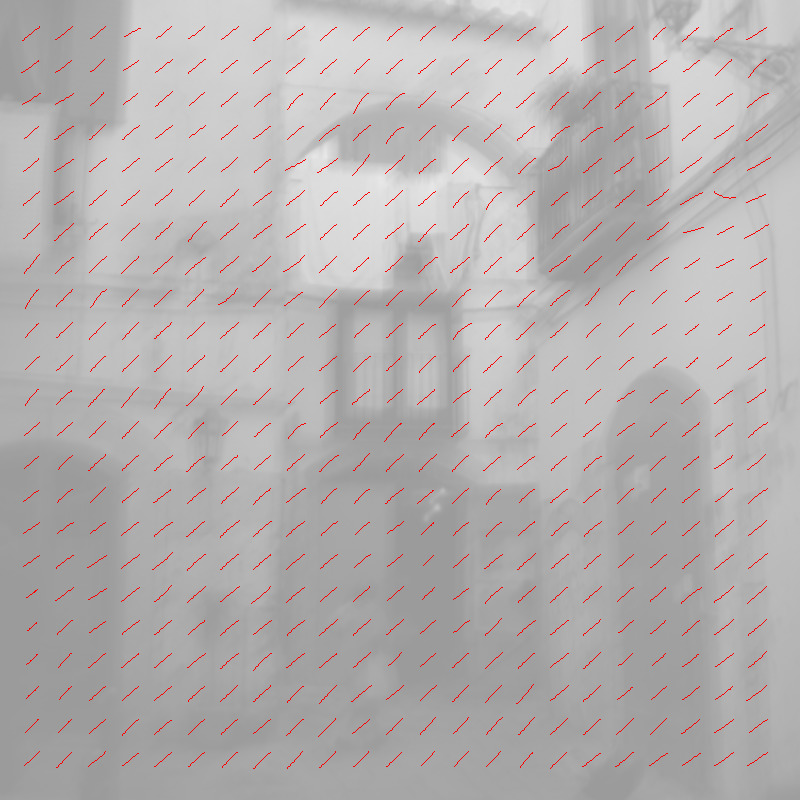} &
    \includegraphics[trim=100 170 100 50, clip, width=0.24\textwidth]{imgs/Kohler/COCO900_restL2_aug_all_loss_gf1_80k/kernels/Blurry3_1_kernels.jpg}  &
    {\setlength{\fboxsep}{0pt}%
       \setlength{\fboxrule}{0.05pt}%
        \framebox{%
    \includegraphics[trim=100 170 100 50, clip, width=0.24\textwidth]{imgs/Kohler/kernels_alpha/Blurry3_1_gt_kernel_alpha}}}  \\
    \includegraphics[trim=100 170 100 50, clip, width=0.24\textwidth]{imgs/Kohler/Zhang/Blurry4_6_flow_map.jpg} &
    \includegraphics[trim=100 170 100 50, clip, width=0.24\textwidth]{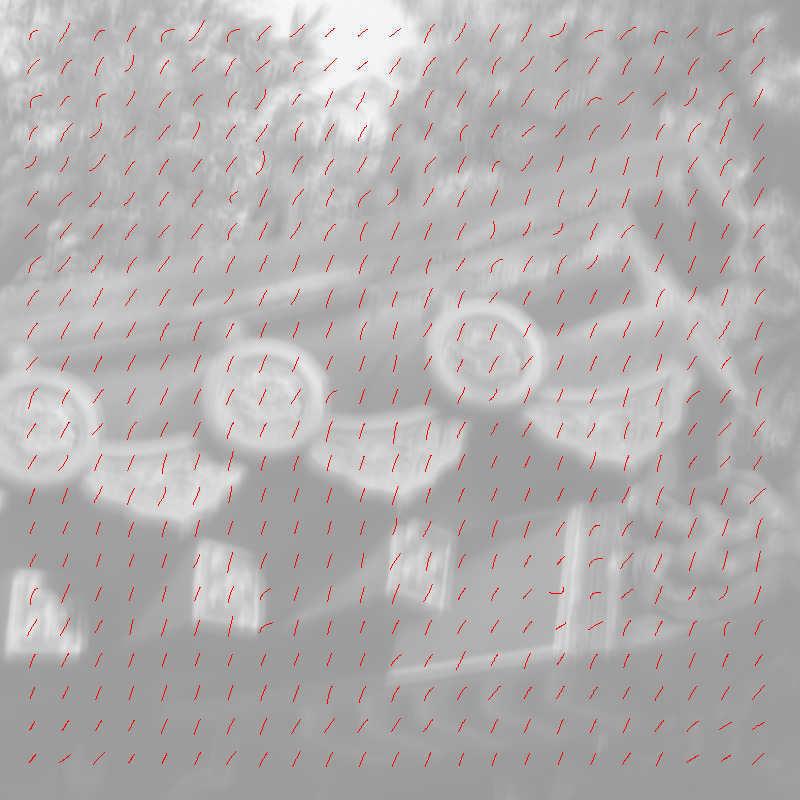} &
    \includegraphics[trim=100 170 100 50, clip, width=0.24\textwidth]{imgs/Kohler/COCO900_restL2_aug_all_loss_gf1_80k/kernels/Blurry4_6_kernels.jpg}  &
    {\setlength{\fboxsep}{0pt}%
       \setlength{\fboxrule}{0.05pt}%
        \framebox{%
        \includegraphics[width=0.24\textwidth]{imgs/Kohler/kernels_alpha/Blurry4_6_gt_kernel_alpha}}}   \\
    \end{tabular}
    \caption{\textbf{a)} Kernels obtained with \cite{zhang2021exposure} trained on the GoPro dataset, \textbf{b)} Kernels obtained with \cite{zhang2021exposure} trained on {\em SBDD}, \textbf{c) } J-MKPD (ours), \textbf{d) } ground truth kernels}
    \label{fig:kernels_Zhang_vs_ours}
\end{figure}

%% file: figure_Lai_suppmat.tex
\begin{figure*}[htb]

    \setlength\tabcolsep{1pt} %
    \centering
    \scriptsize
\begin{tabular}{cccc}

Blurry & Ana-Syn \cite{kaufman2020deblurring} & SRN \cite{tao2018scale} & MIMO-UNet  \cite{cho2021rethinking} \\ 
\includegraphics[width=0.24\textwidth]{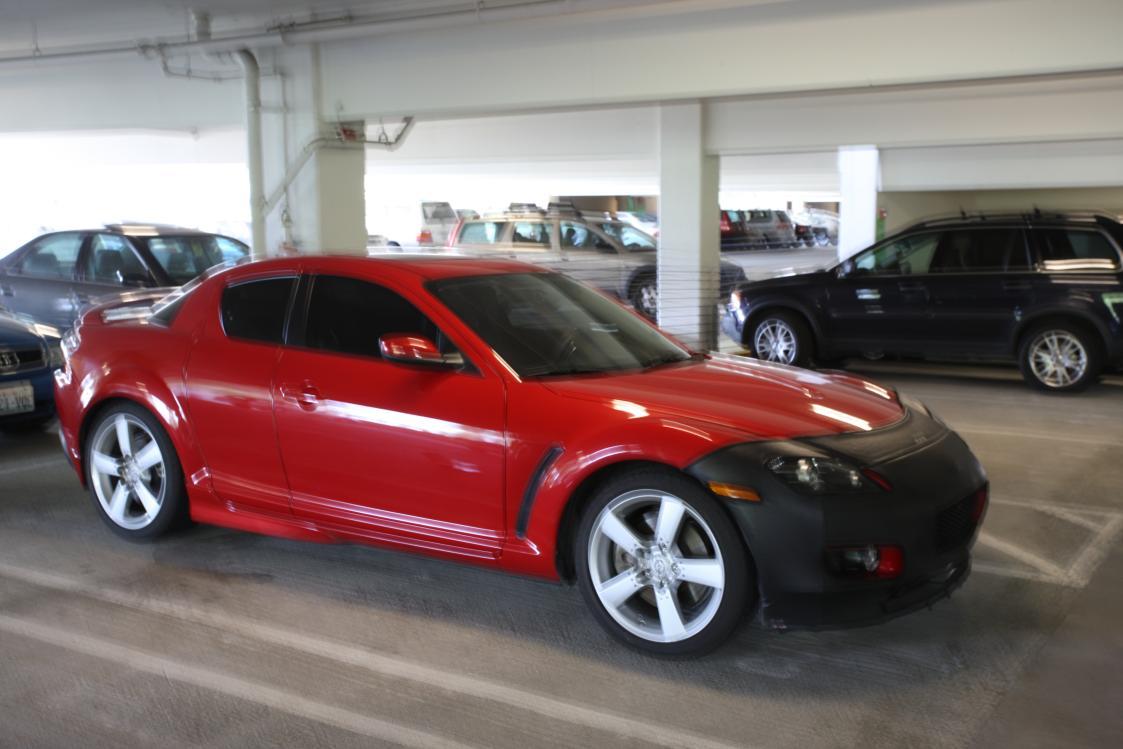}   &
\includegraphics[width=0.24\textwidth]{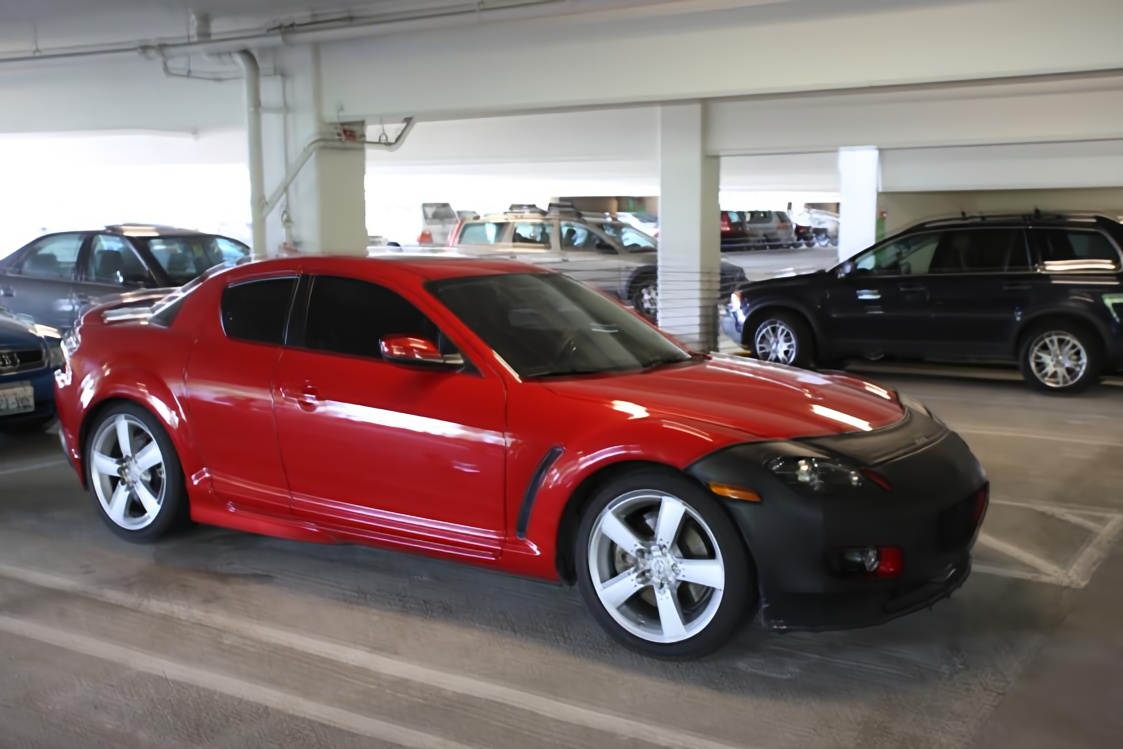} & 
\includegraphics[width=0.24\textwidth]{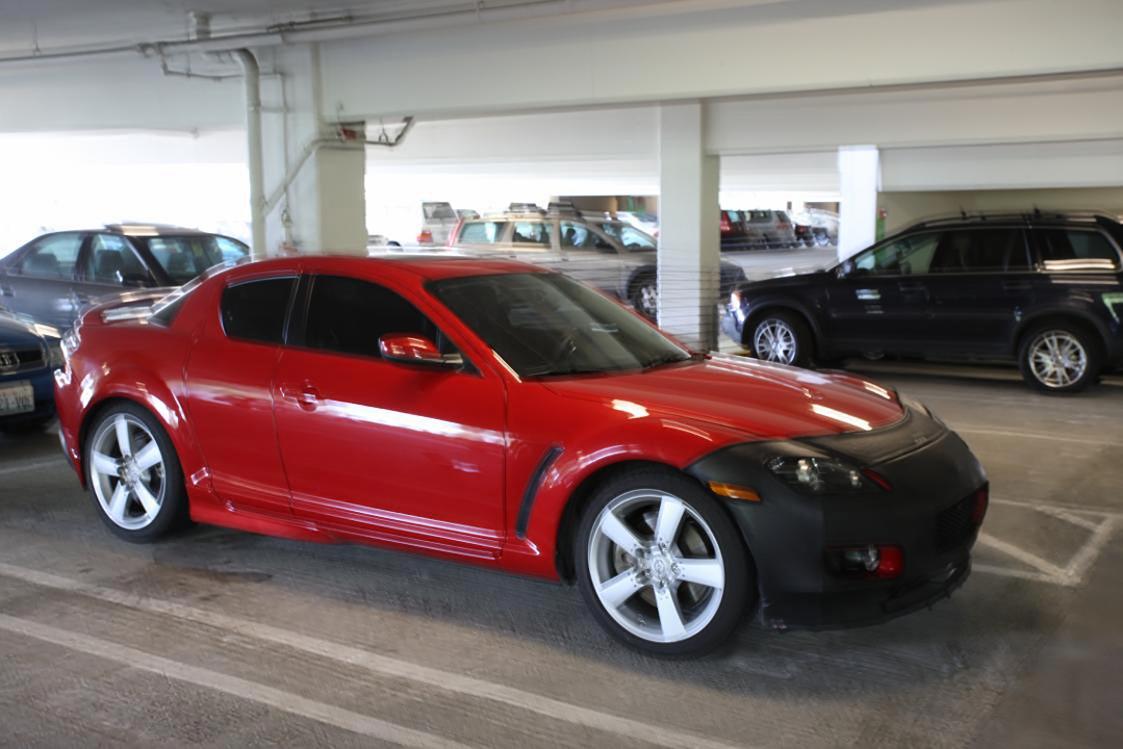}    & 
\includegraphics[width=0.24\textwidth]{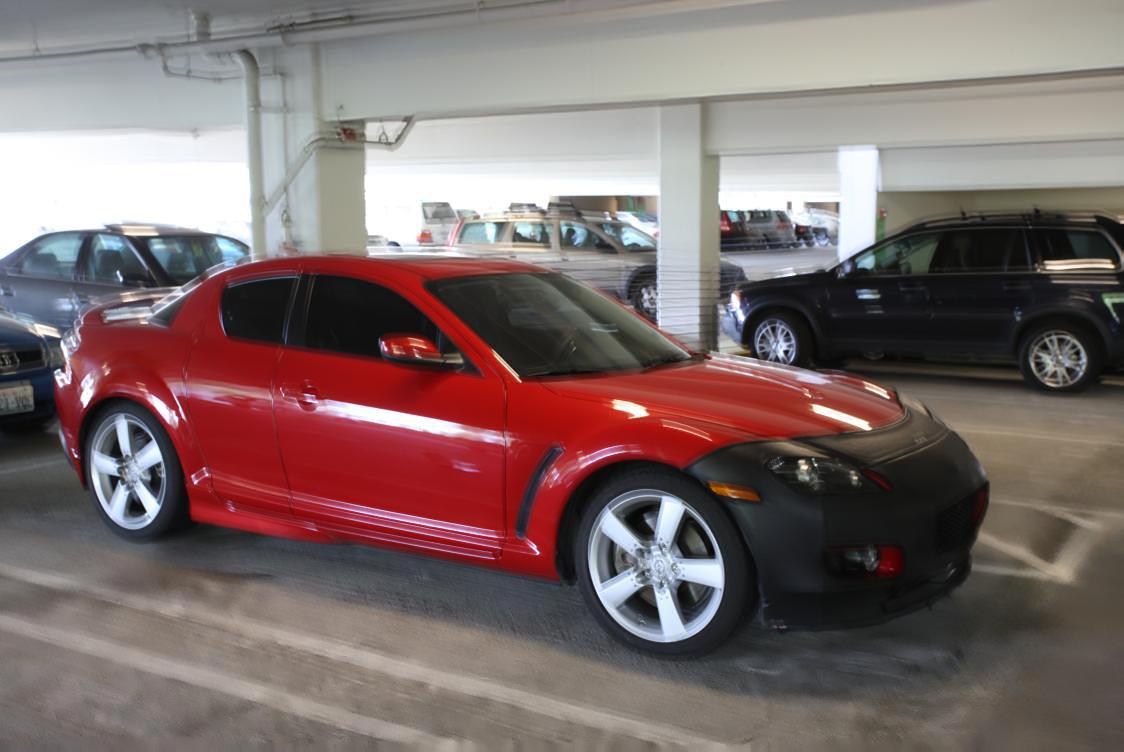} 
\\

\includegraphics[trim=0 280 1000 340, clip,width=0.12\textwidth]{imgs/Lai/Blurry/car1.jpg} 
\includegraphics[trim=485 425 520 200, clip,width=0.12\textwidth]{imgs/Lai/Blurry/car1.jpg} &
\includegraphics[trim=0 280 1000 340, clip,width=0.12\textwidth]{imgs/Lai/AnaSyn/car1_deblurred.jpg} 
\includegraphics[trim=485 425 520 200, clip,width=0.12\textwidth]{imgs/Lai/AnaSyn/car1_deblurred.jpg} &
\includegraphics[trim=0 280 1000 340, clip,width=0.12\textwidth]{imgs/Lai/SRN/car1.jpg} 
\includegraphics[trim=485 425 520 200, clip,width=0.12\textwidth]{imgs/Lai/SRN/car1.jpg} &
\includegraphics[trim=0 280 1000 340, clip,width=0.12\textwidth]{imgs/Lai/MIMO-UNet/car1.jpg} 
\includegraphics[trim=485 425 520 200, clip,width=0.12\textwidth]{imgs/Lai/MIMO-UNet/car1.jpg} \\

DGAN-v2 \cite{kupyn2019deblurgan} & MPRNet \cite{Zamir2021MPRNet} & NAFNet \cite{NAFNet}   &  J-MKPD (ours) \\
\includegraphics[width=0.24\textwidth]{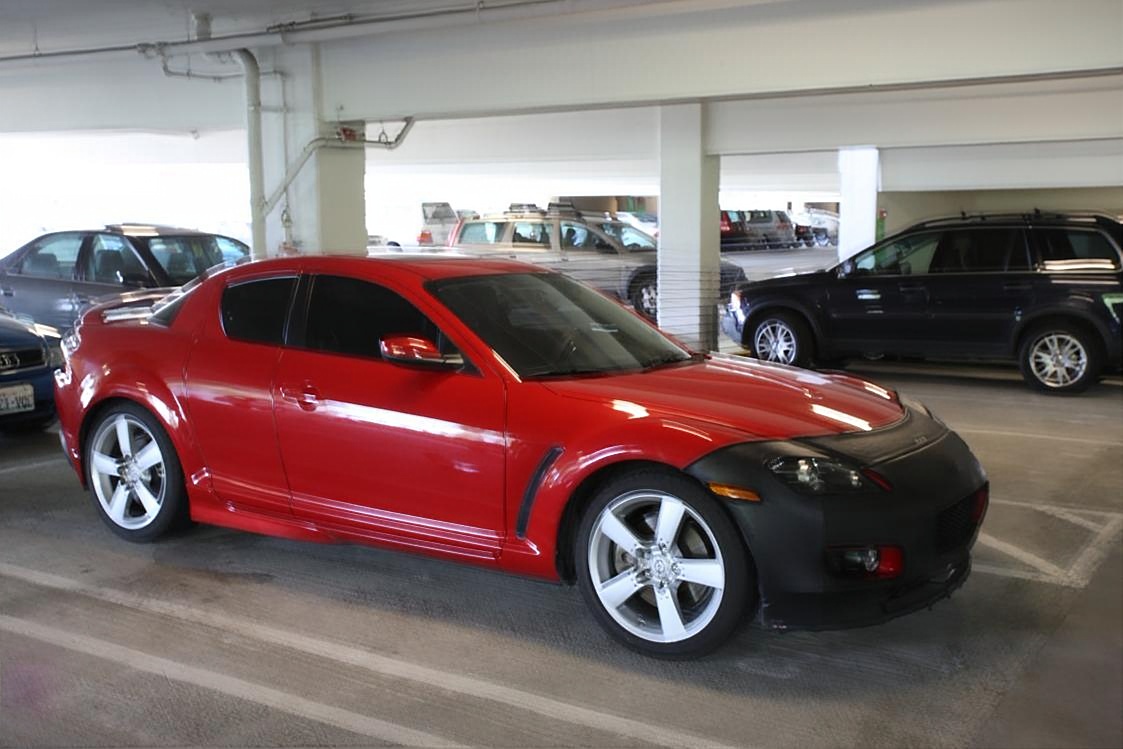}  &
\includegraphics[width=0.24\textwidth]{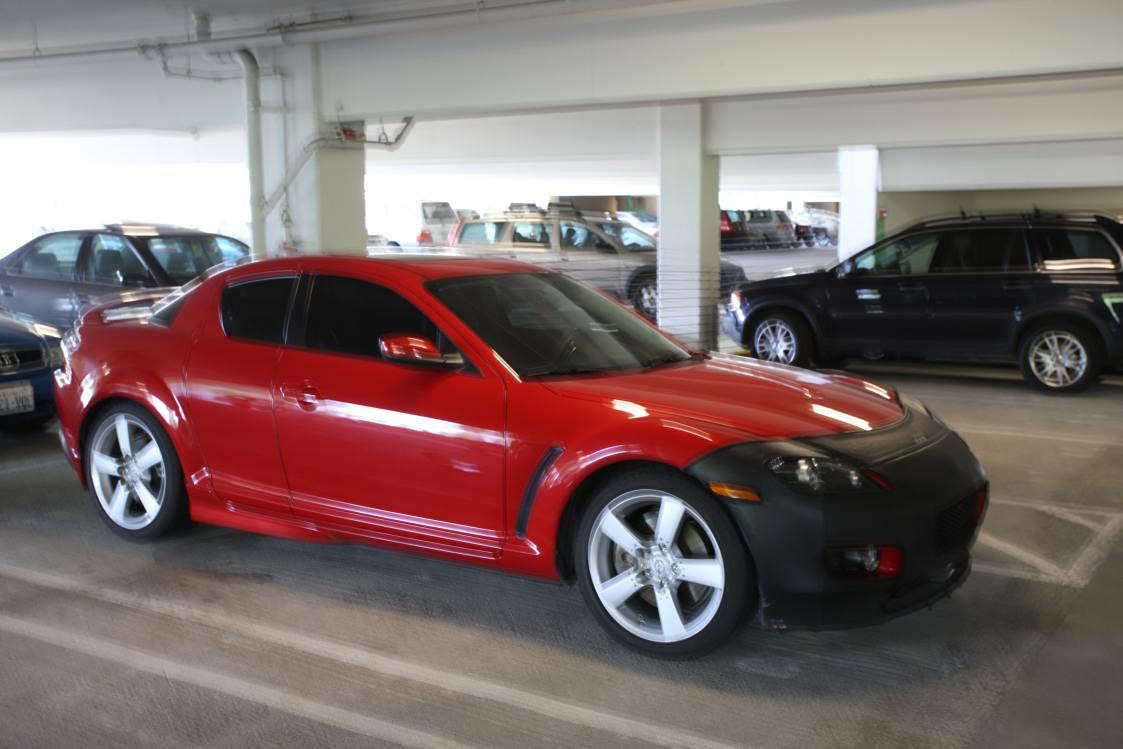}  & \includegraphics[width=0.24\textwidth]{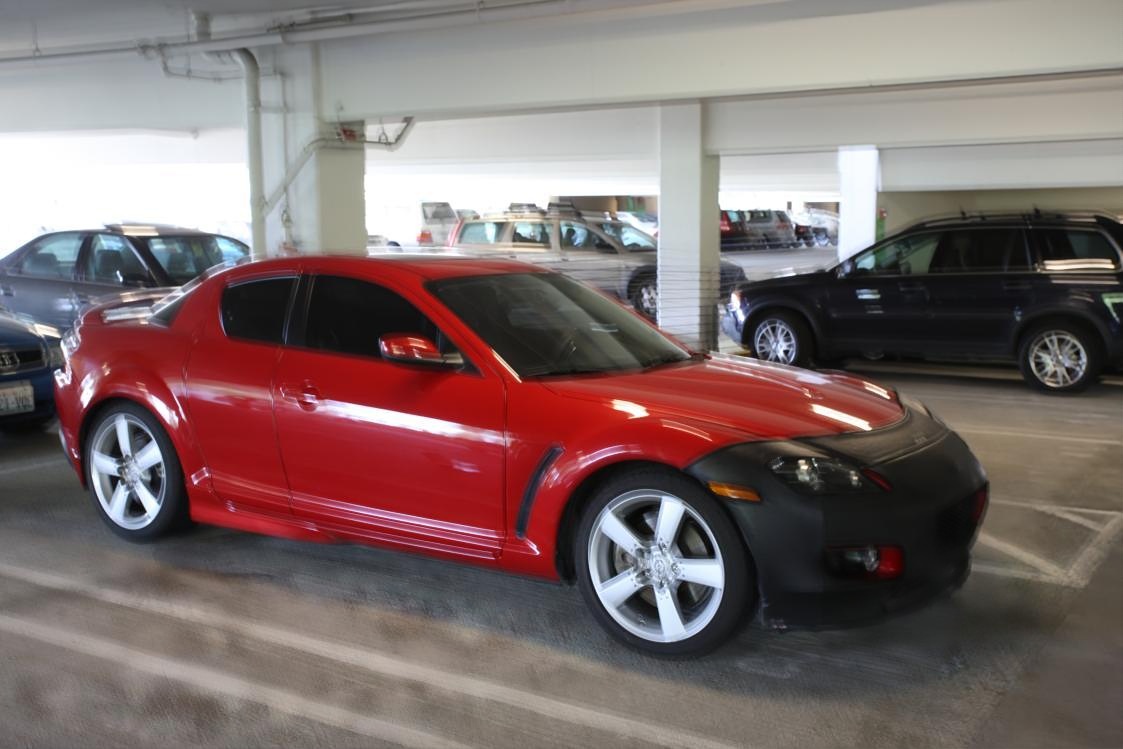}  &
\includegraphics[width=0.24\textwidth]{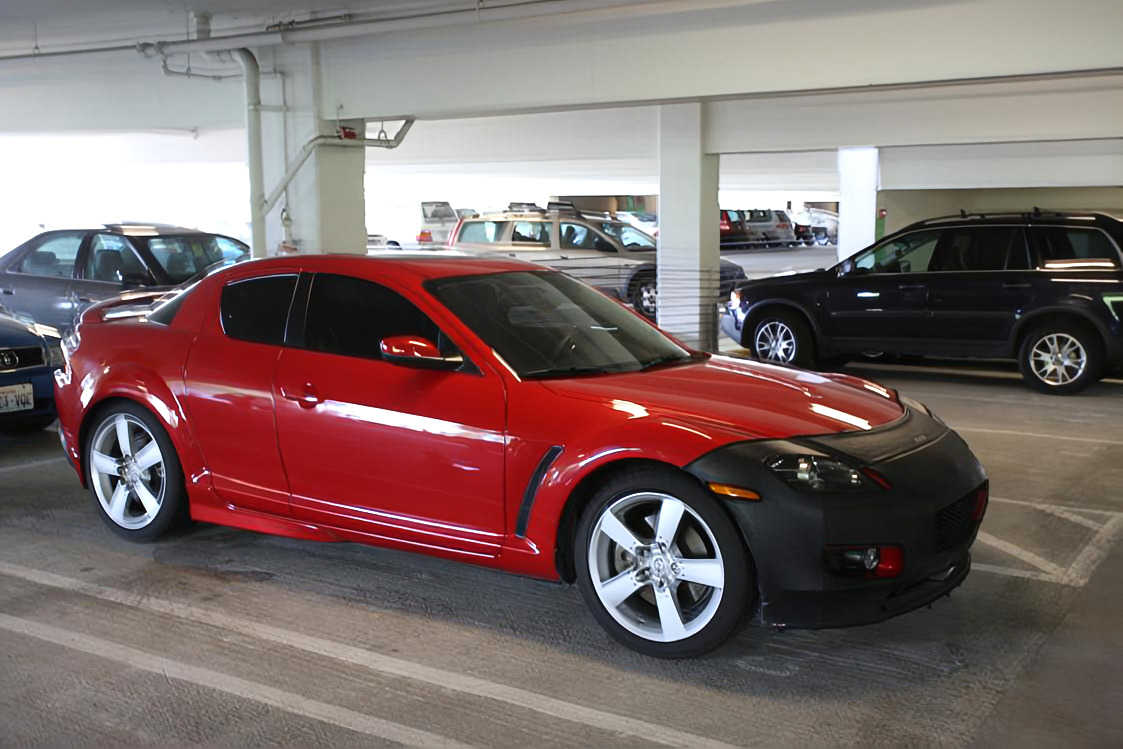}  \\

\includegraphics[trim=0 280 1000 340, clip,width=0.12\textwidth]{imgs/Lai/DeblurGAN_V2/car1.jpg} 
\includegraphics[trim=485 425 520 200, clip,width=0.12\textwidth]{imgs/Lai/DeblurGAN_V2/car1.jpg} &
\includegraphics[trim=0 280 1000 340, clip,width=0.12\textwidth]{imgs/Lai/MPRNet/car1.jpg} 
\includegraphics[trim=485 425 520 200, clip,width=0.12\textwidth]{imgs/Lai/MPRNet/car1.jpg} &
\includegraphics[trim=0 280 1000 340, clip,width=0.12\textwidth]{imgs/Lai/NAFNet/car1.jpg} 
\includegraphics[trim=485 425 520 200, clip,width=0.12\textwidth]{imgs/Lai/NAFNet/car1.jpg} &
\includegraphics[trim=0 280 1000 340, clip,width=0.12\textwidth]{imgs/Lai/COCO900_restL2_aug_all_loss_gf1_80k/car1_NIMBUSR.jpg} 
\includegraphics[trim=485 425 520 200, clip,width=0.12\textwidth]{imgs/Lai/COCO900_restL2_aug_all_loss_gf1_80k/car1_NIMBUSR.jpg} \\

Blurry & Ana-Syn \cite{kaufman2020deblurring} & SRN \cite{tao2018scale} & MIMO-UNet  \cite{cho2021rethinking} \\ 
\includegraphics[width=0.240\textwidth]{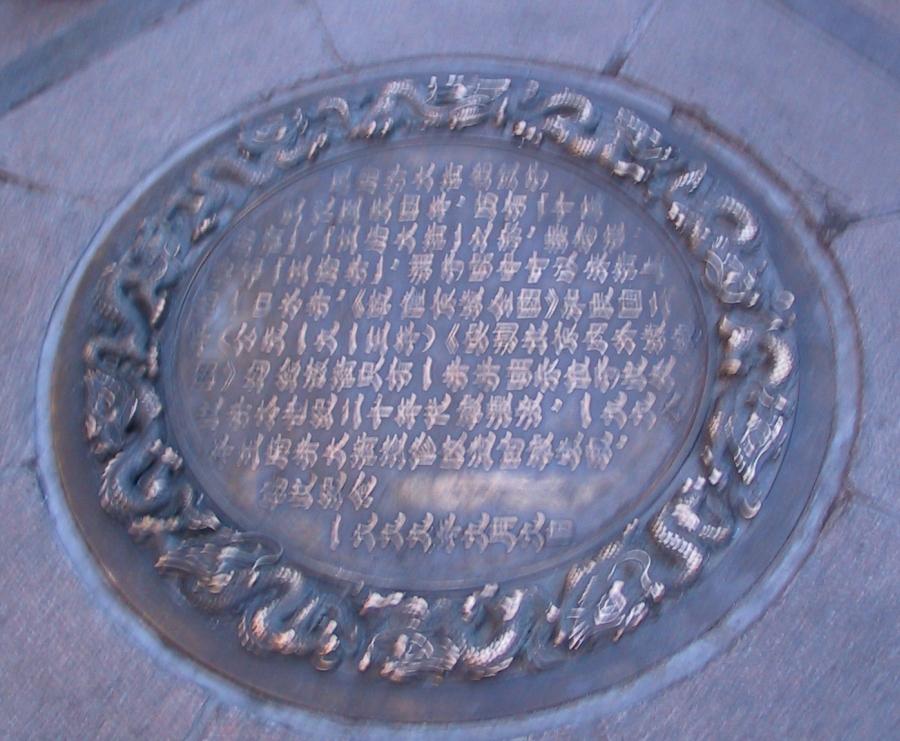}   &
\includegraphics[width=0.240\textwidth]{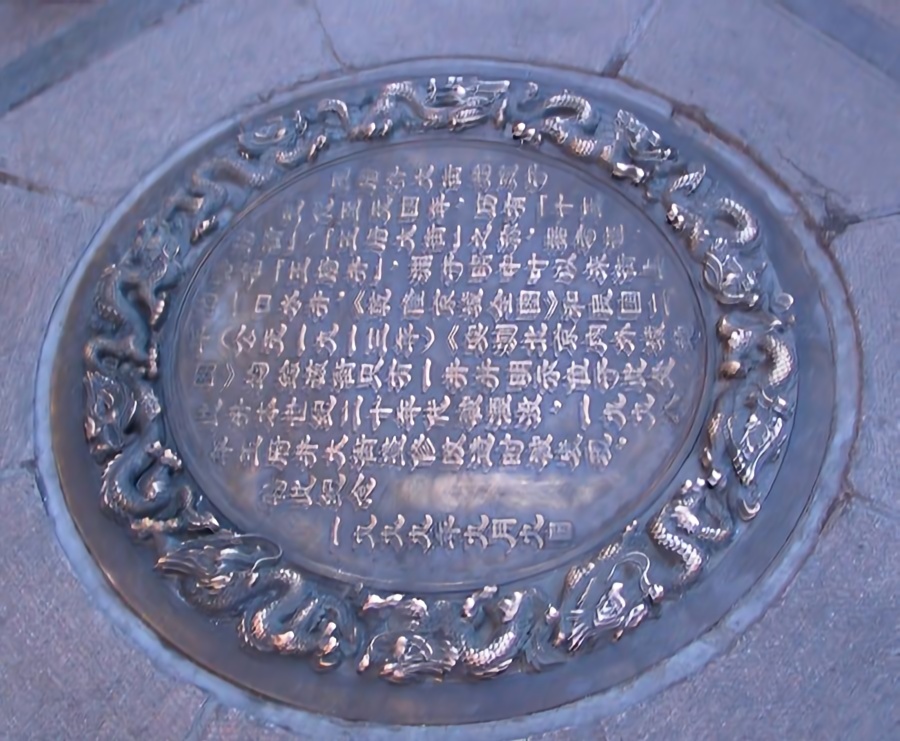} & 
\includegraphics[width=0.240\textwidth]{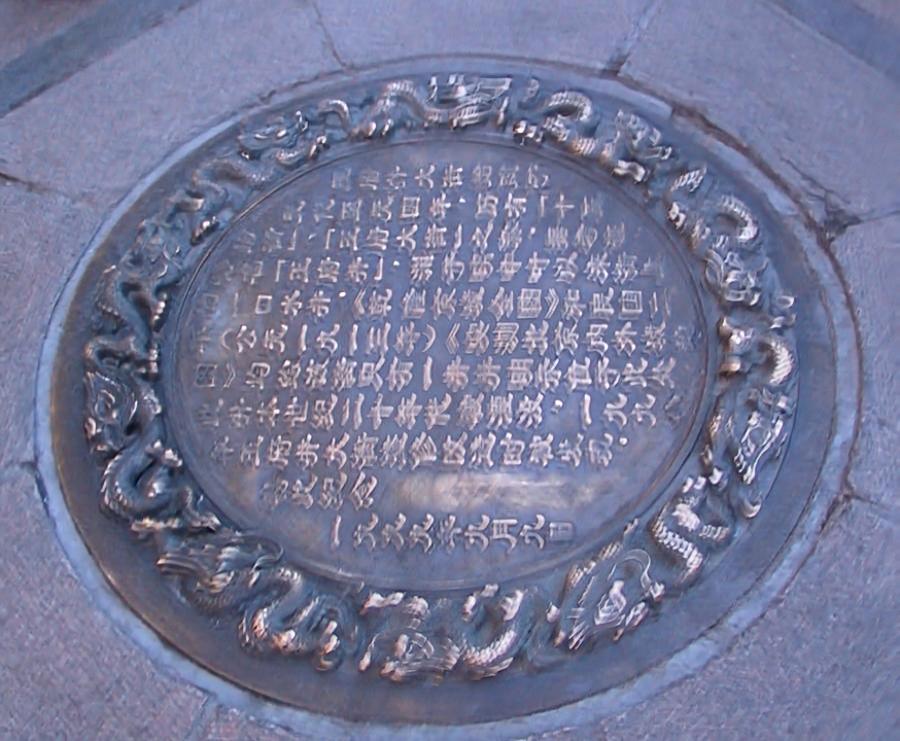}  &  
\includegraphics[width=0.240\textwidth]{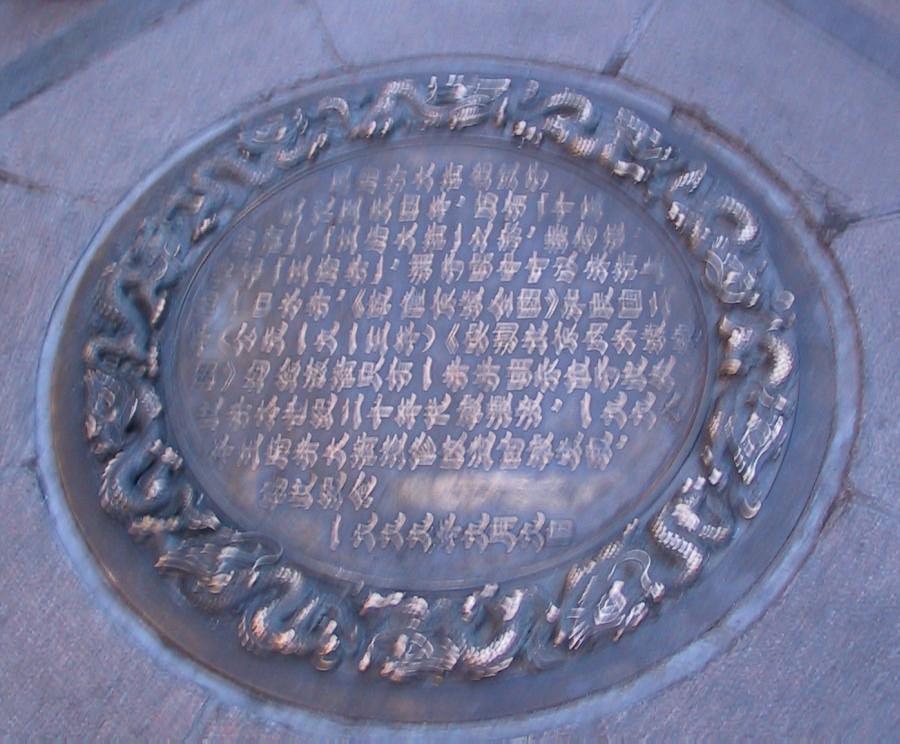}\\

\includegraphics[trim=260 310 500 320, clip,width=0.12\textwidth]{imgs/Lai/Blurry/hanzi.jpg} 
\includegraphics[trim=460 390 313 250, clip,width=0.12\textwidth]{imgs/Lai/Blurry/hanzi.jpg} &
\includegraphics[trim=260 310 500 320, clip,width=0.12\textwidth]{imgs/Lai/AnaSyn/hanzi_deblurred.jpg} 
\includegraphics[trim=460 390 313 250, clip,width=0.12\textwidth]{imgs/Lai/AnaSyn/hanzi_deblurred.jpg} &
\includegraphics[trim=260 310 500 320, clip,width=0.12\textwidth]{imgs/Lai/SRN/hanzi.jpg} 
\includegraphics[trim=460 390 313 250, clip,width=0.12\textwidth]{imgs/Lai/SRN/hanzi.jpg} &
\includegraphics[trim=260 310 500 320, clip,width=0.12\textwidth]{imgs/Lai/MIMO-UNet/hanzi.jpg} 
\includegraphics[trim=460 390 313 250, clip,width=0.12\textwidth]{imgs/Lai/MIMO-UNet/hanzi.jpg} \\

DGAN-v2 \cite{kupyn2019deblurgan} & MPRNet \cite{Zamir2021MPRNet} & NAFNet \cite{NAFNet}    &   J-MKPD (ours) \\
\includegraphics[width=0.240\textwidth]{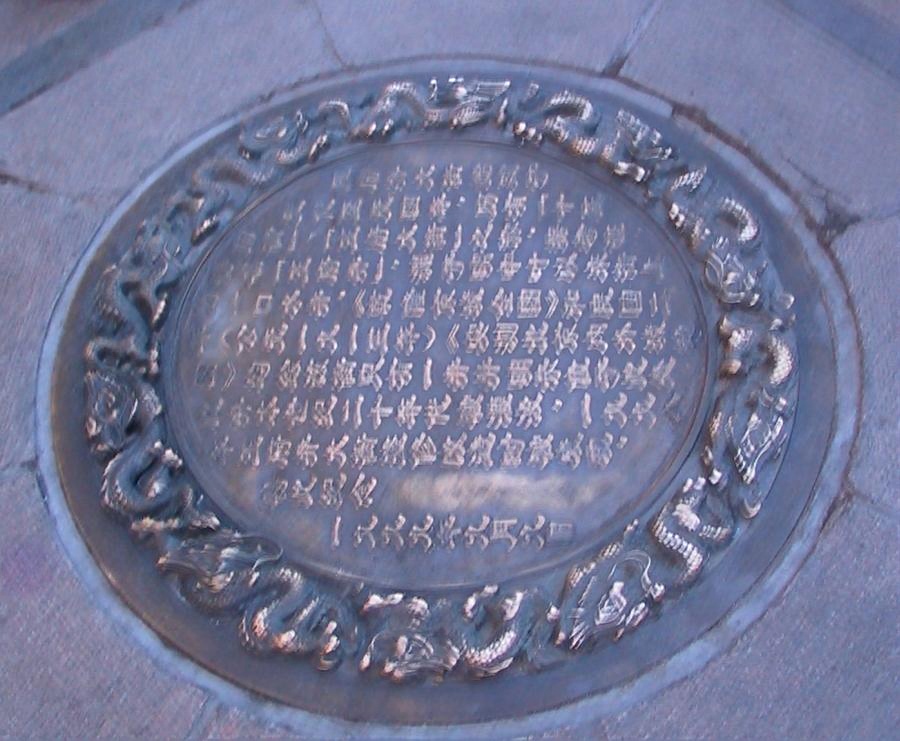}  &
\includegraphics[width=0.240\textwidth]{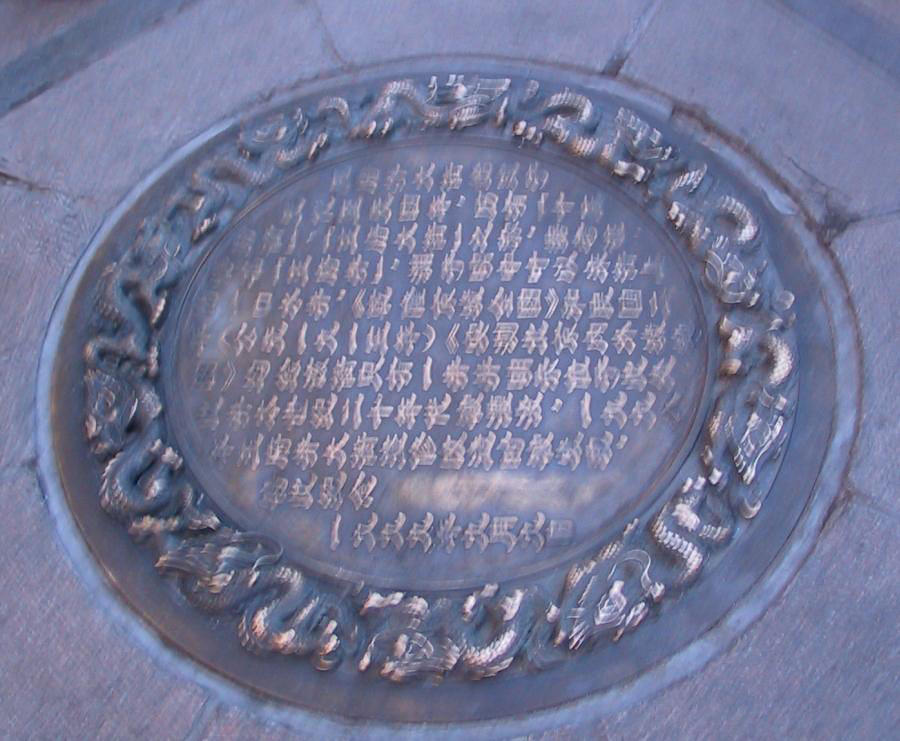}  &  
\includegraphics[width=0.240\textwidth]{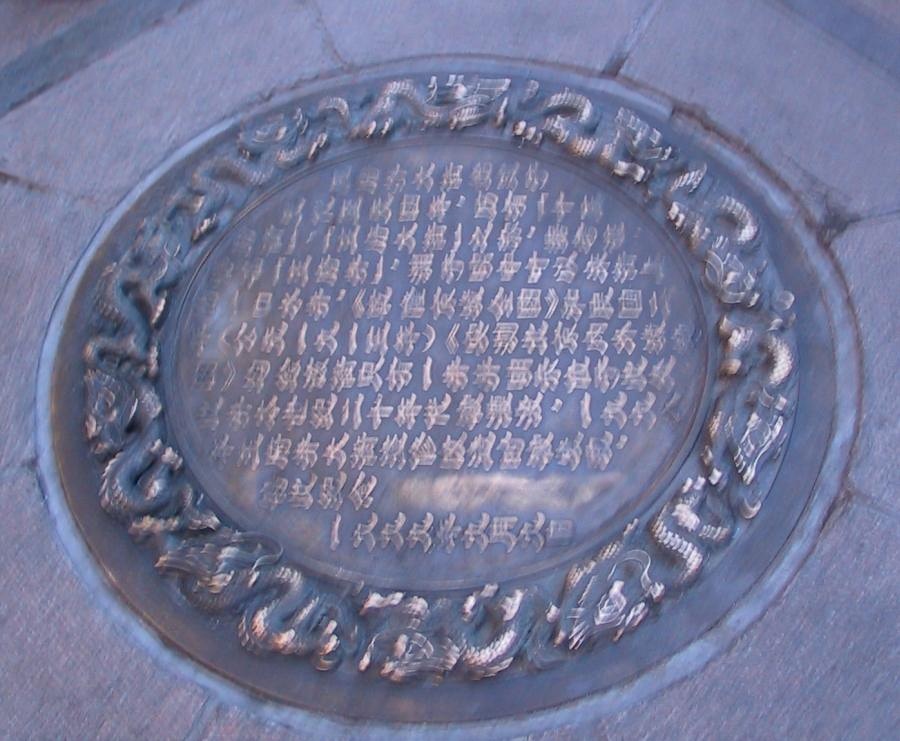} &
\includegraphics[width=0.240\textwidth]{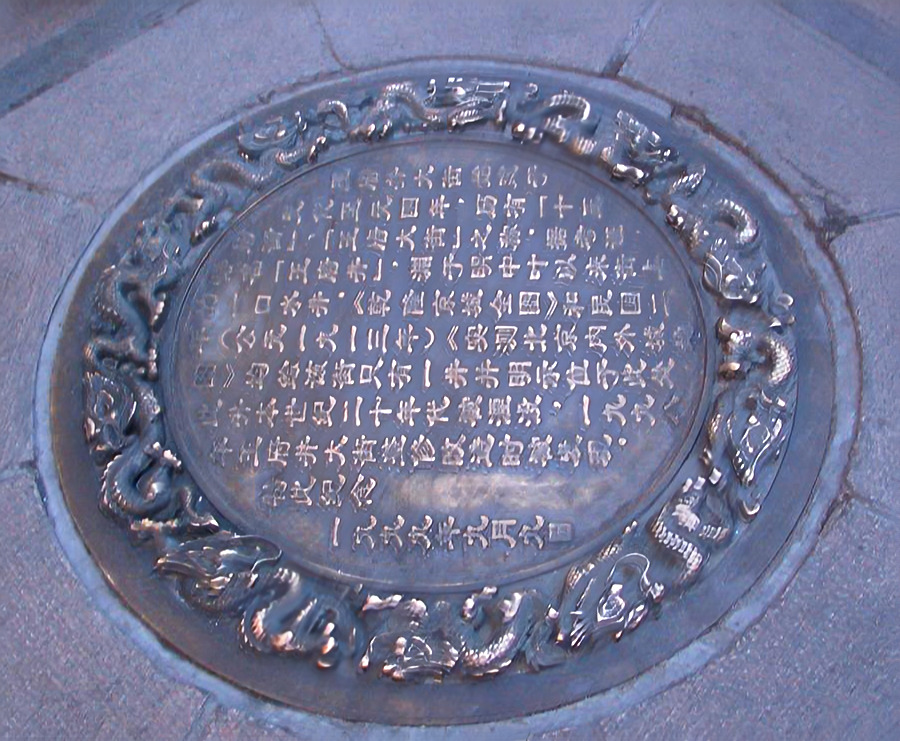}  \\
\includegraphics[trim=260 310 500 320, clip,width=0.12\textwidth]{imgs/Lai/DeblurGAN_V2/hanzi.jpg} 
\includegraphics[trim=460 390 313 250, clip,width=0.12\textwidth]{imgs/Lai/DeblurGAN_V2/hanzi.jpg} &
\includegraphics[trim=260 310 500 320, clip,width=0.12\textwidth]{imgs/Lai/MPRNet/hanzi.jpg} 
\includegraphics[trim=460 390 313 250, clip,width=0.12\textwidth]{imgs/Lai/MPRNet/hanzi.jpg} &
\includegraphics[trim=260 310 500 320, clip,width=0.12\textwidth]{imgs/Lai/NAFNet/hanzi.jpg} 
\includegraphics[trim=460 390 313 250, clip,width=0.12\textwidth]{imgs/Lai/NAFNet/hanzi.jpg} &
\includegraphics[trim=260 310 500 320, clip,width=0.12\textwidth]{imgs/Lai/COCO900_restL2_aug_all_loss_gf1_80k/hanzi_NIMBUSR.jpg} 
\includegraphics[trim=460 390 313 250, clip,width=0.12\textwidth]{imgs/Lai/COCO900_restL2_aug_all_loss_gf1_80k/hanzi_NIMBUSR.jpg} \\
\end{tabular} 
    \caption{\textbf{Deblurring examples on real blurry images from Lai dataset  \cite{lai2016comparative}.} Our model-based approach is competitive with state-of-the-art data-driven methods. Best viewed in electronic format.}
    \label{fig:Lai_suppmat}
\end{figure*}

\begin{figure*}[h]

    \setlength\tabcolsep{1pt} %
    \centering
    \scriptsize
\begin{tabular}{cccc}
Blurry & Ana-Syn \cite{kaufman2020deblurring} & SRN \cite{tao2018scale} & MIMO-UNet  \cite{cho2021rethinking}  \\ 
\includegraphics[trim=0 0 0 50, clip, width=0.240\textwidth]{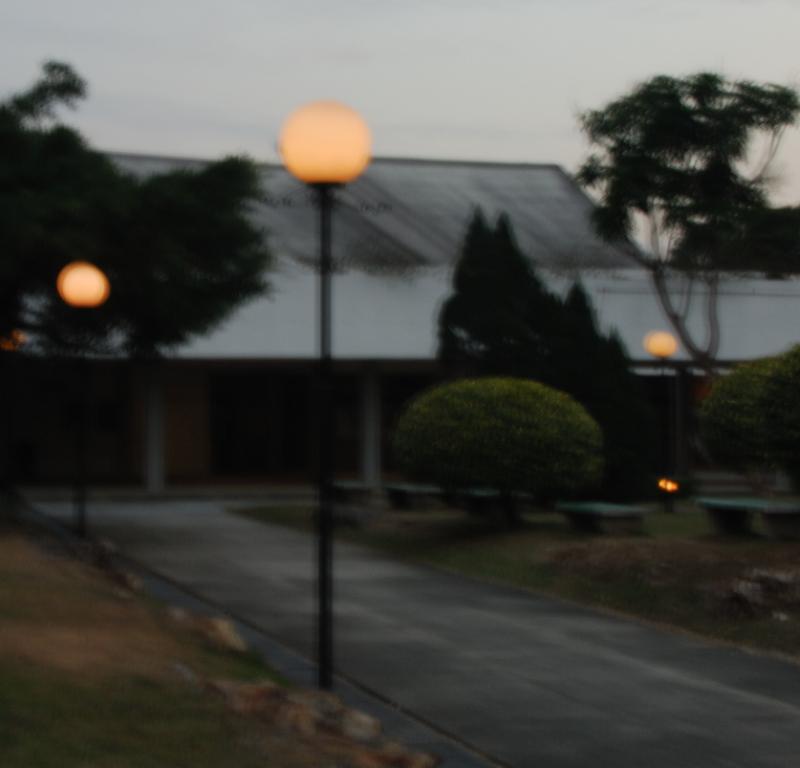}   &
\includegraphics[trim=0 0 0 50, clip, width=0.240\textwidth]{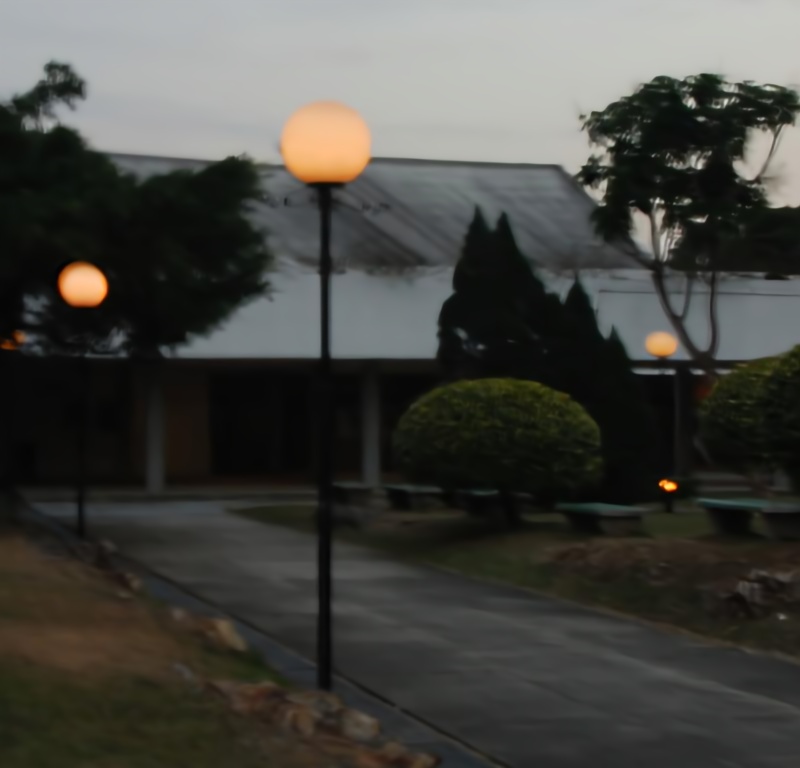} & 
\includegraphics[trim=0 0 0 50, clip, width=0.240\textwidth]{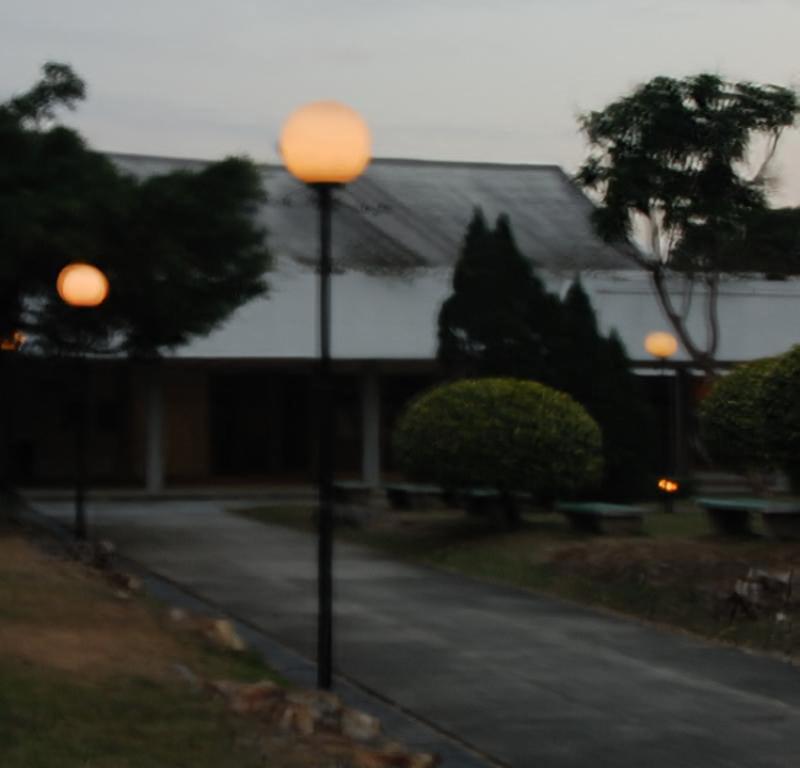}  &    
\includegraphics[trim=0 0 0 50, clip, width=0.240\textwidth]{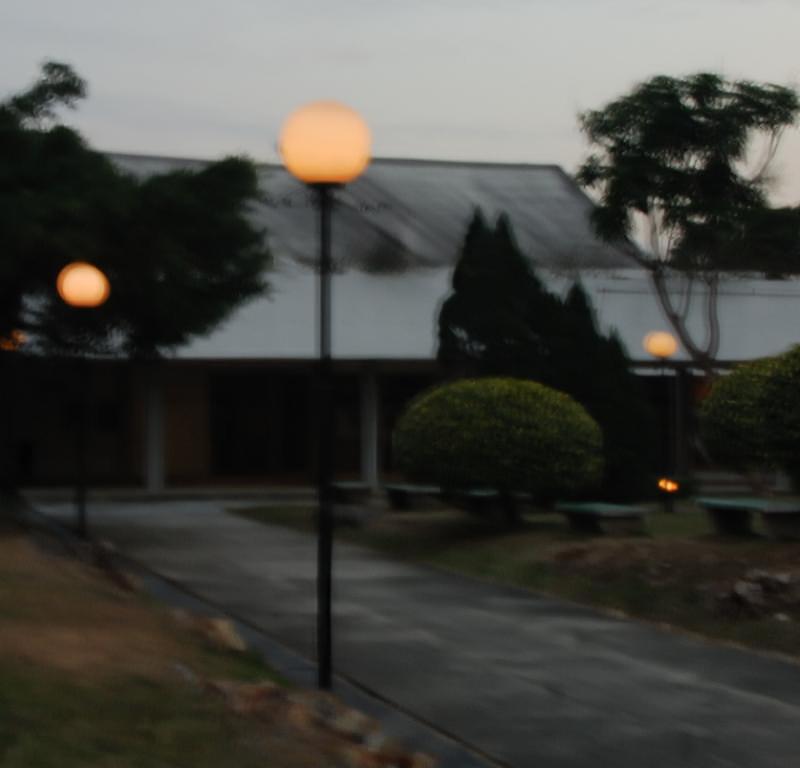} \\

\includegraphics[trim=200 520 380 81, clip,width=0.12\textwidth]{imgs/Lai/Blurry/nv.jpg}
\includegraphics[trim=600 410 70 260, clip,width=0.12\textwidth]{imgs/Lai/Blurry/nv.jpg} &
\includegraphics[trim=200 520 380 81, clip,width=0.12\textwidth]{imgs/Lai/AnaSyn/nv_deblurred.jpg} 
\includegraphics[trim=600 410 70 260, clip,width=0.12\textwidth]{imgs/Lai/AnaSyn/nv_deblurred.jpg} &
\includegraphics[trim=200 520 380 81, clip,width=0.12\textwidth]{imgs/Lai/SRN/nv.jpg}
\includegraphics[trim=600 410 70 260, clip,width=0.12\textwidth]{imgs/Lai/SRN/nv.jpg} &
\includegraphics[trim=200 520 380 81, clip,width=0.12\textwidth]{imgs/Lai/MIMO-UNet/nv.jpg} 
\includegraphics[trim=600 410 70 260, clip,width=0.12\textwidth]{imgs/Lai/MIMO-UNet/nv.jpg} \\

DGAN-v2 \cite{kupyn2019deblurgan} & MPRNet 
\cite{Zamir2021MPRNet} & NAFNet \cite{NAFNet}    &   J-MKPD (ours) \\
\includegraphics[trim=0 0 0 50, clip, width=0.240\textwidth]{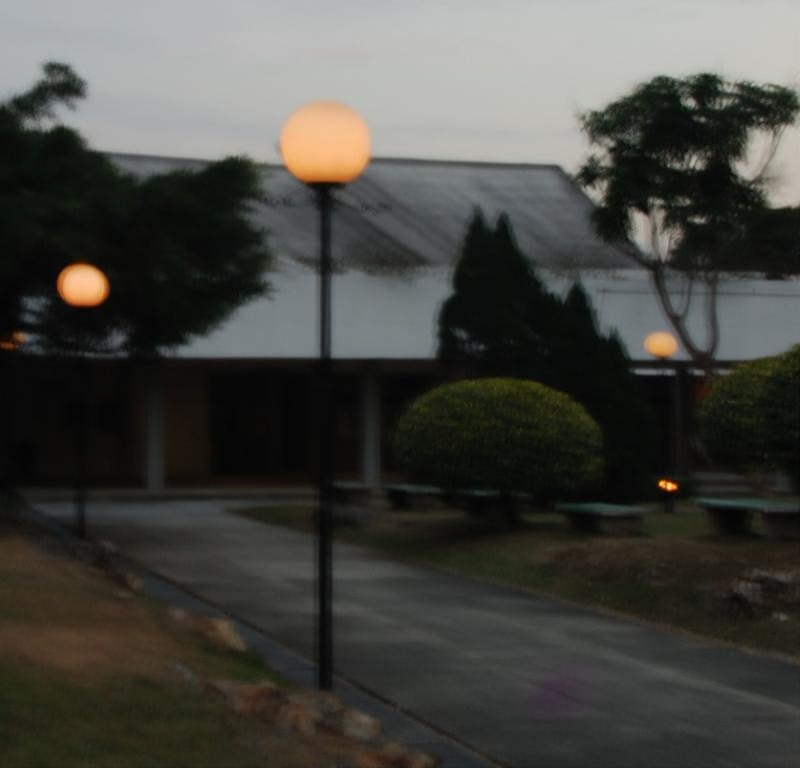}  &
\includegraphics[trim=0 0 0 50, clip, width=0.240\textwidth]{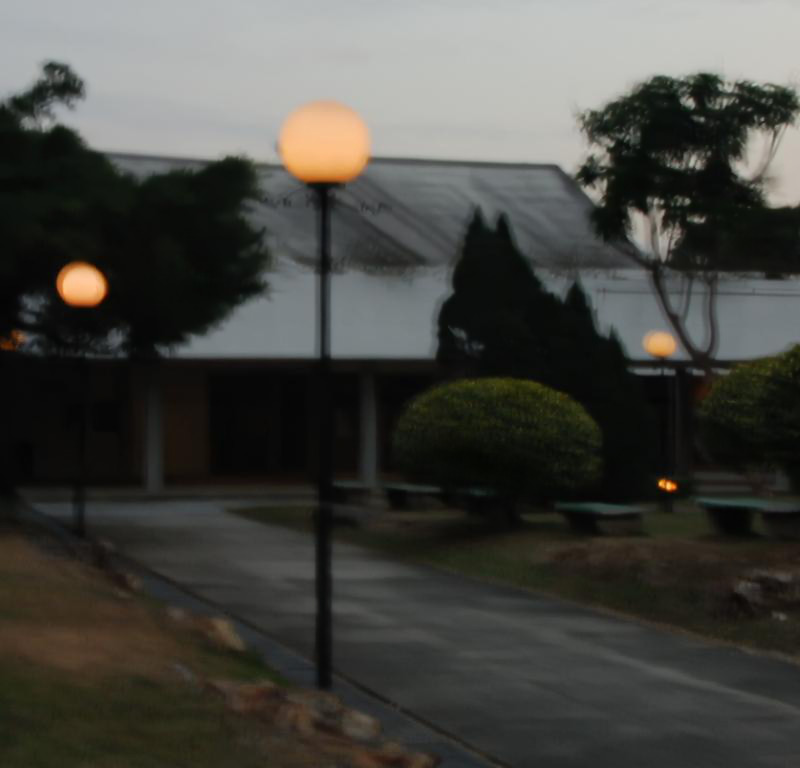}  & 
\includegraphics[trim=0 0 0 50, clip, width=0.240\textwidth]{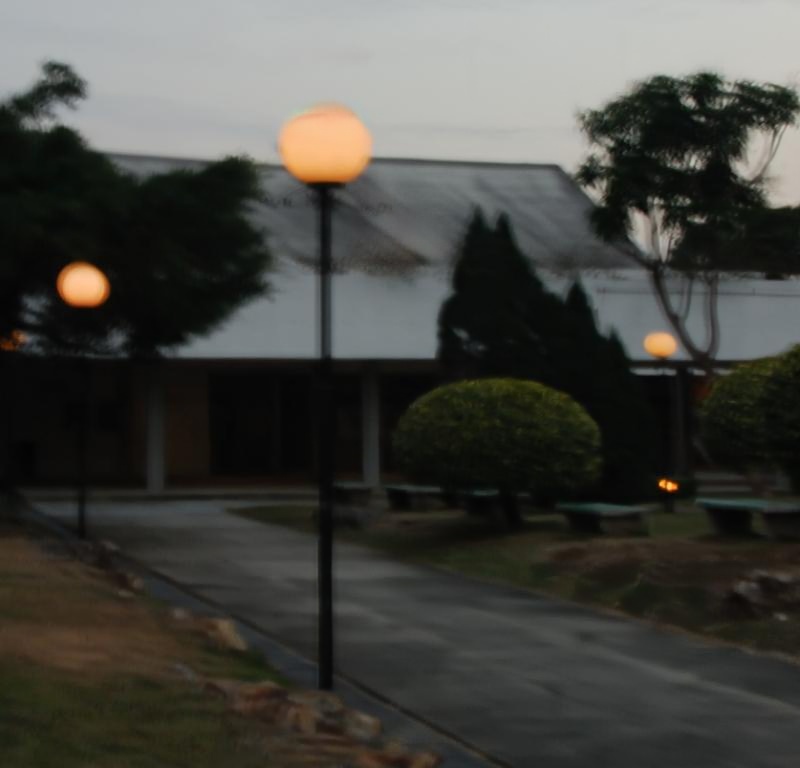} &
\includegraphics[trim=0 0 0 50, clip, width=0.240\textwidth]{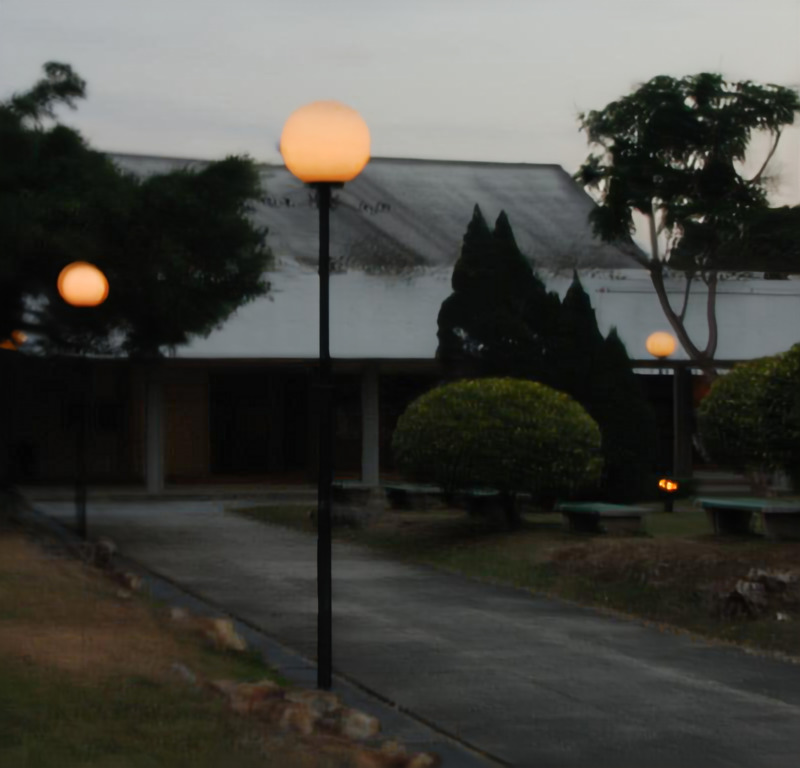}  \\

\includegraphics[trim=200 520 380 81, clip,width=0.12\textwidth]{imgs/Lai/DeblurGAN_V2/nv.jpg}
\includegraphics[trim=600 410 70 260, clip,width=0.12\textwidth]{imgs/Lai/DeblurGAN_V2/nv.jpg} &
\includegraphics[trim=200 520 380 81, clip,width=0.12\textwidth]{imgs/Lai/MPRNet/nv.jpg} 
\includegraphics[trim=600 410 70 260, clip,width=0.12\textwidth]{imgs/Lai/MPRNet/nv.jpg} &
\includegraphics[trim=200 520 380 81, clip,width=0.12\textwidth]{imgs/Lai/NAFNet/nv.jpg}
\includegraphics[trim=600 410 70 260, clip,width=0.12\textwidth]{imgs/Lai/NAFNet/nv.jpg} &
\includegraphics[trim=200 520 380 81, clip,width=0.12\textwidth]{imgs/Lai/COCO900_restL2_aug_all_loss_gf1_80k/nv_NIMBUSR.jpg} 
\includegraphics[trim=600 410 70 260, clip,width=0.12\textwidth]{imgs/Lai/COCO900_restL2_aug_all_loss_gf1_80k/nv_NIMBUSR.jpg} \\

Blurry & Ana-Syn \cite{kaufman2020deblurring} & SRN \cite{tao2018scale} & MIMO-UNet  \cite{cho2021rethinking}  \\ 
\includegraphics[trim=0 0 0 260, clip, width=0.240\textwidth]{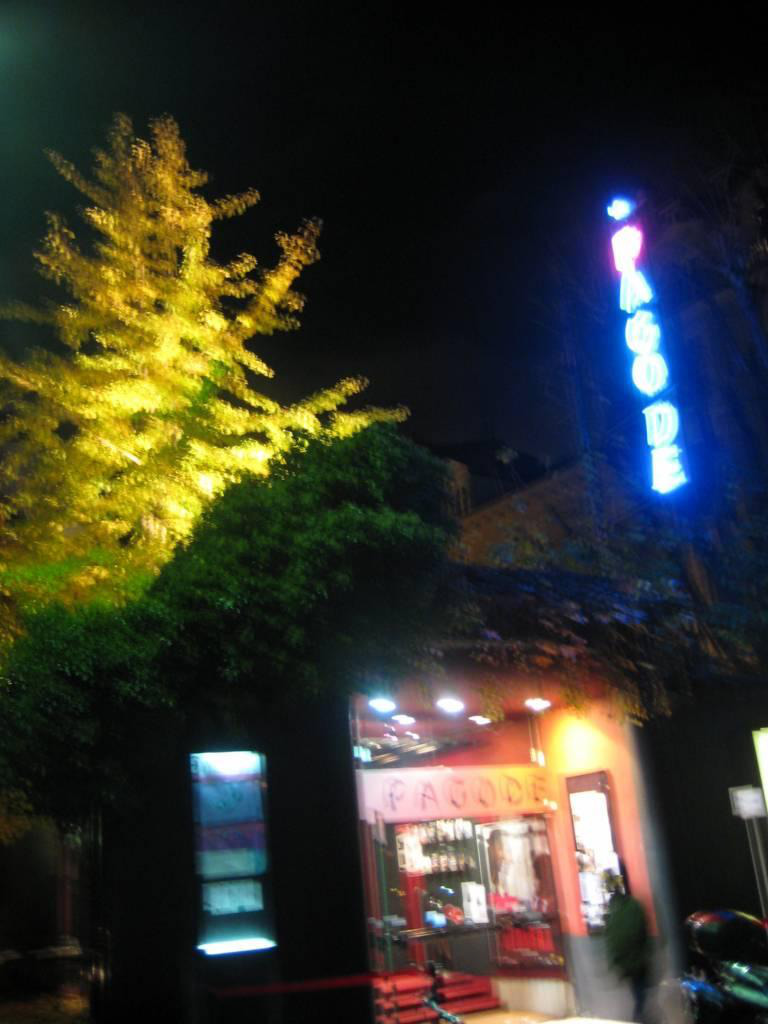}   &
\includegraphics[trim=0 0 0 260, clip, width=0.240\textwidth]{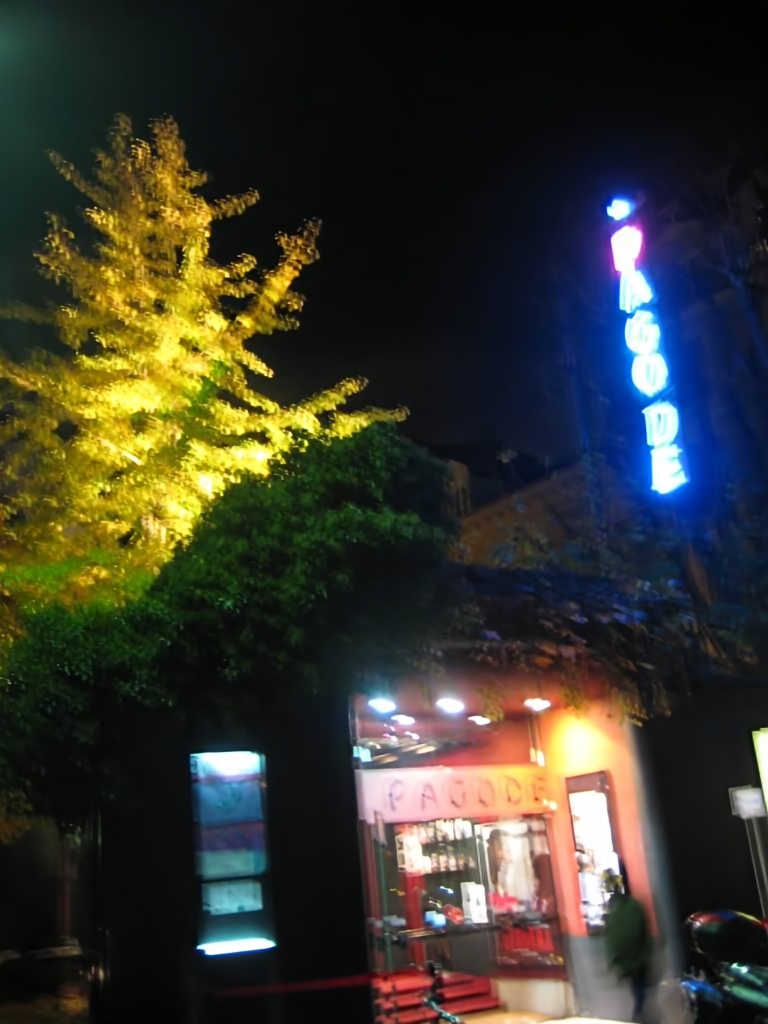} & 
\includegraphics[trim=0 0 0 260, clip, width=0.240\textwidth]{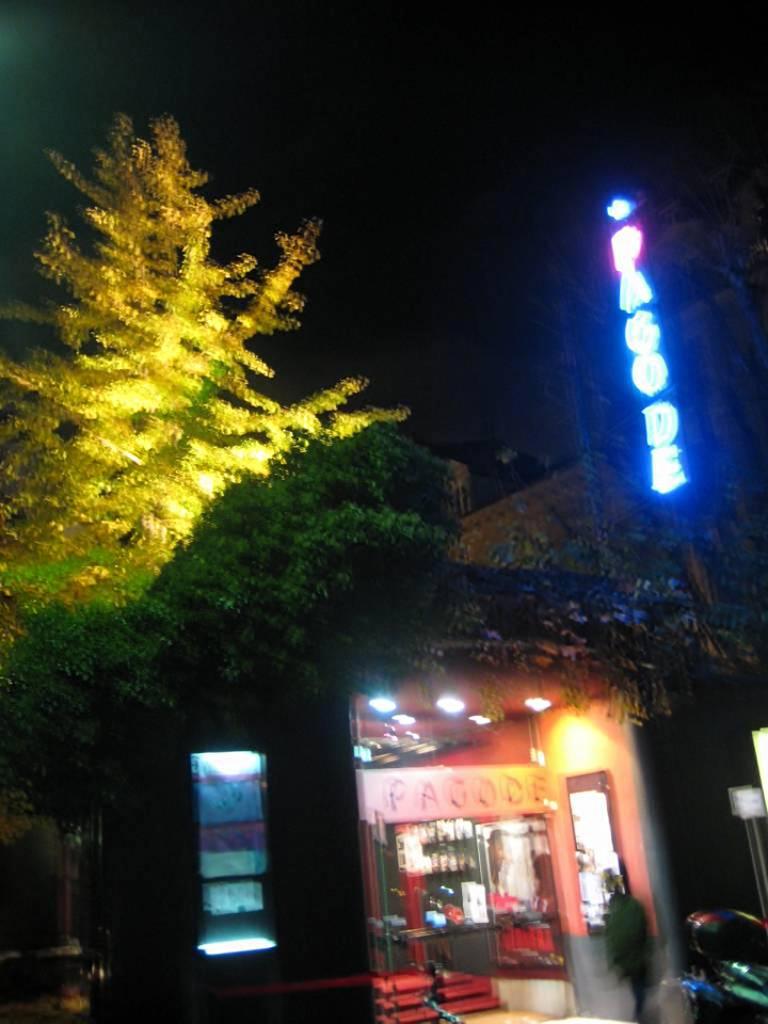}  &    
\includegraphics[trim=0 0 0 260, clip, width=0.240\textwidth]{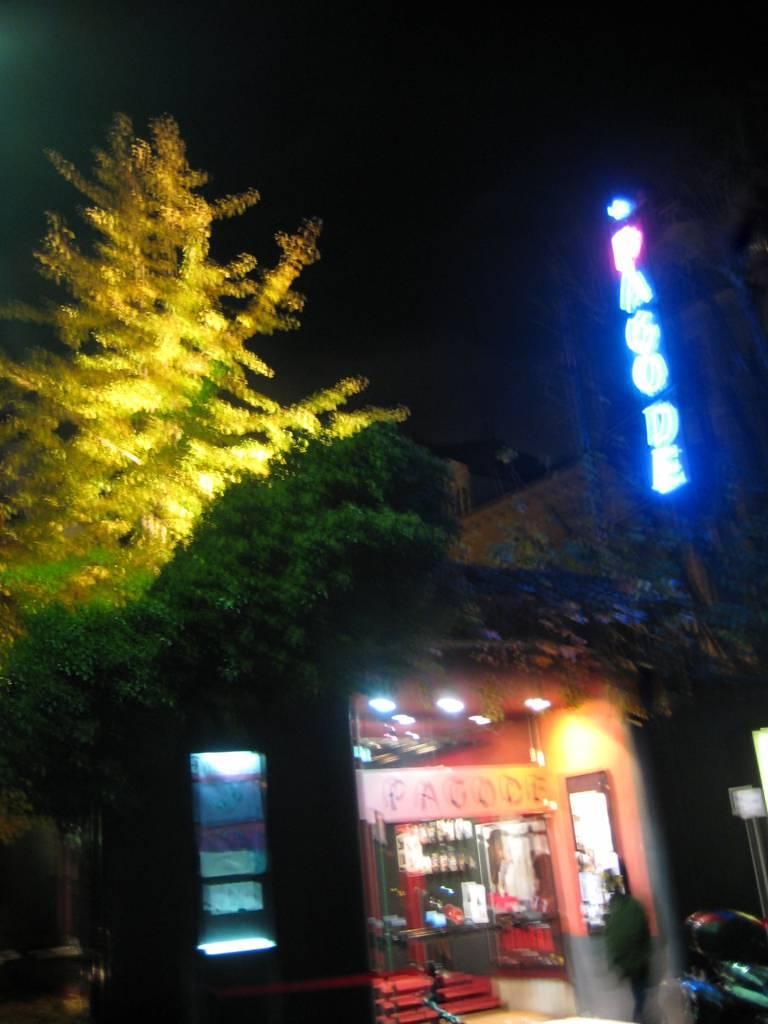} \\

\includegraphics[trim=360 70 300 891, clip,width=0.12\textwidth]{imgs/Lai/Blurry/pagode.jpg}
\includegraphics[trim=340 180 260 745, clip,width=0.12\textwidth]{imgs/Lai/Blurry/pagode.jpg} &
\includegraphics[trim=360 70 300 891, clip,width=0.12\textwidth]{imgs/Lai/AnaSyn/pagode_deblurred.jpg} 
\includegraphics[trim=340 180 260 745, clip,width=0.12\textwidth]{imgs/Lai/AnaSyn/pagode_deblurred.jpg} &
\includegraphics[trim=360 70 300 891, clip,width=0.12\textwidth]{imgs/Lai/SRN/pagode.jpg}
\includegraphics[trim=340 180 260 745, clip,width=0.12\textwidth]{imgs/Lai/SRN/pagode.jpg} &
\includegraphics[trim=360 70 300 891, clip,width=0.12\textwidth]{imgs/Lai/MIMO-UNet/pagode.jpg} 
\includegraphics[trim=340 180 260 745, clip,width=0.12\textwidth]{imgs/Lai/MIMO-UNet/pagode.jpg} \\

DGAN-v2 \cite{kupyn2019deblurgan} & MPRNet 
\cite{Zamir2021MPRNet} & NAFNet \cite{NAFNet}    &   J-MKPD (ours) \\
\includegraphics[trim=0 0 0 260, clip, width=0.240\textwidth]{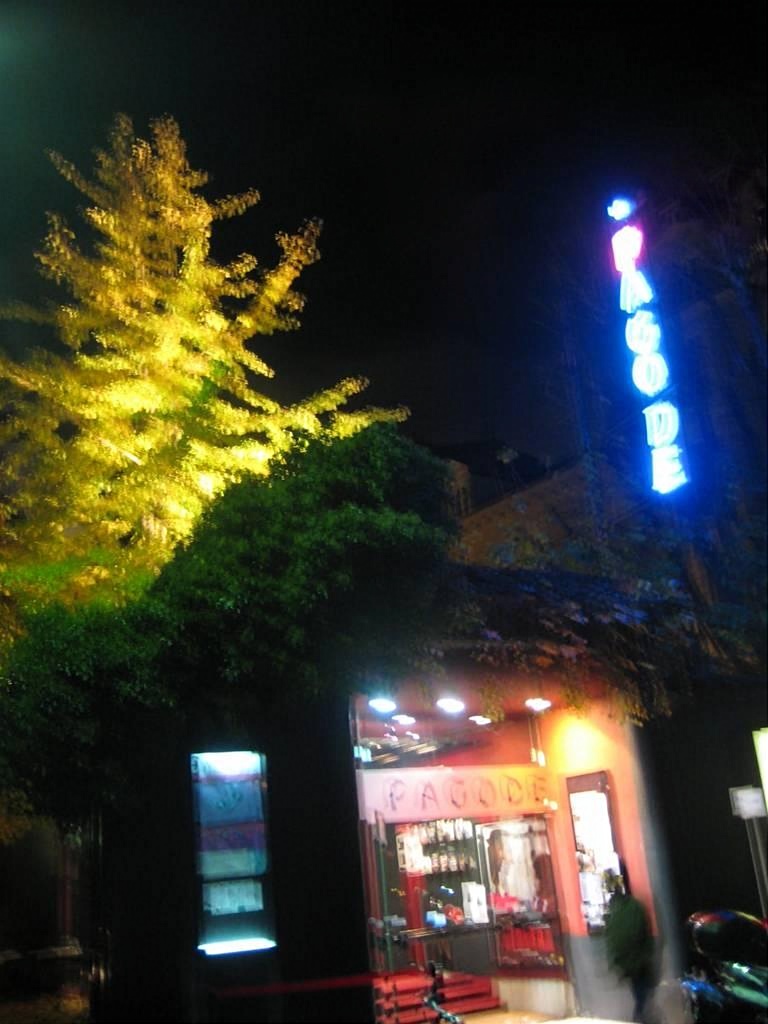}  &
\includegraphics[trim=0 0 0 260, clip, width=0.240\textwidth]{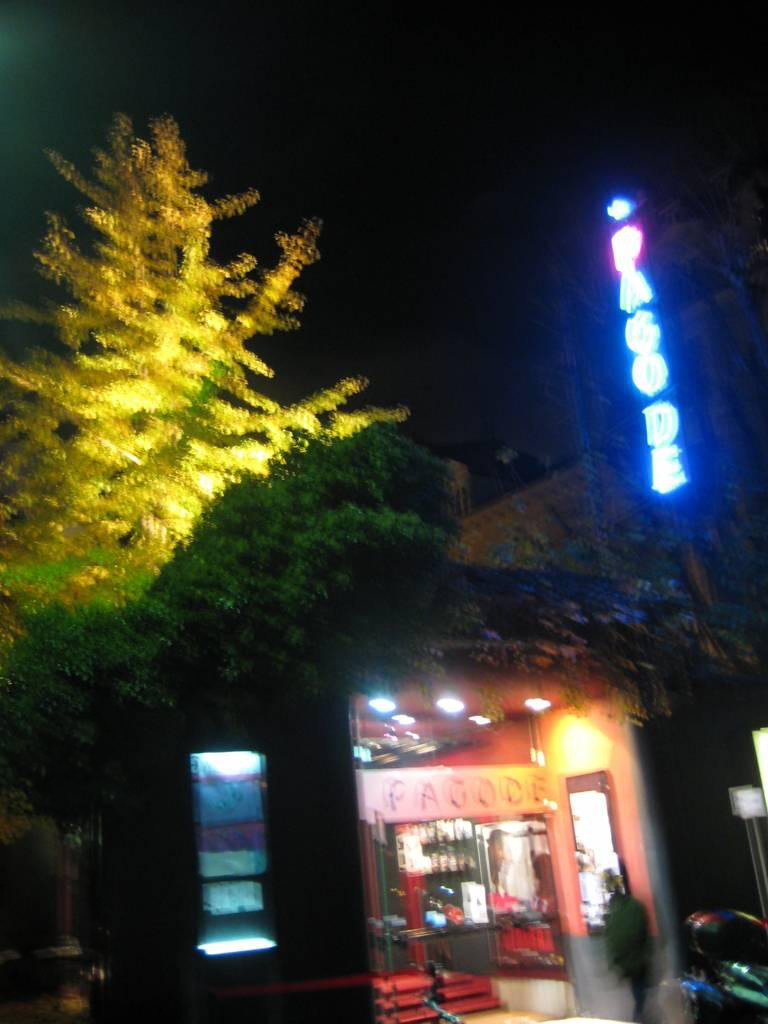}  & 
\includegraphics[trim=0 0 0 260, clip, width=0.240\textwidth]{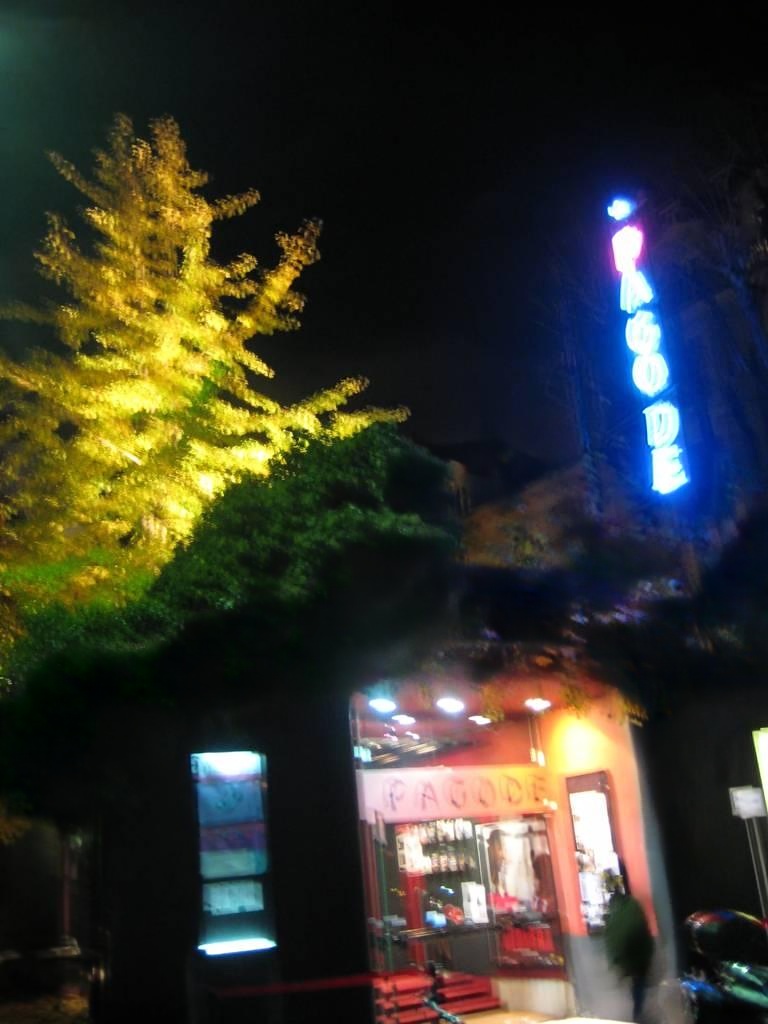} &
\includegraphics[trim=0 0 0 260, clip, width=0.240\textwidth]{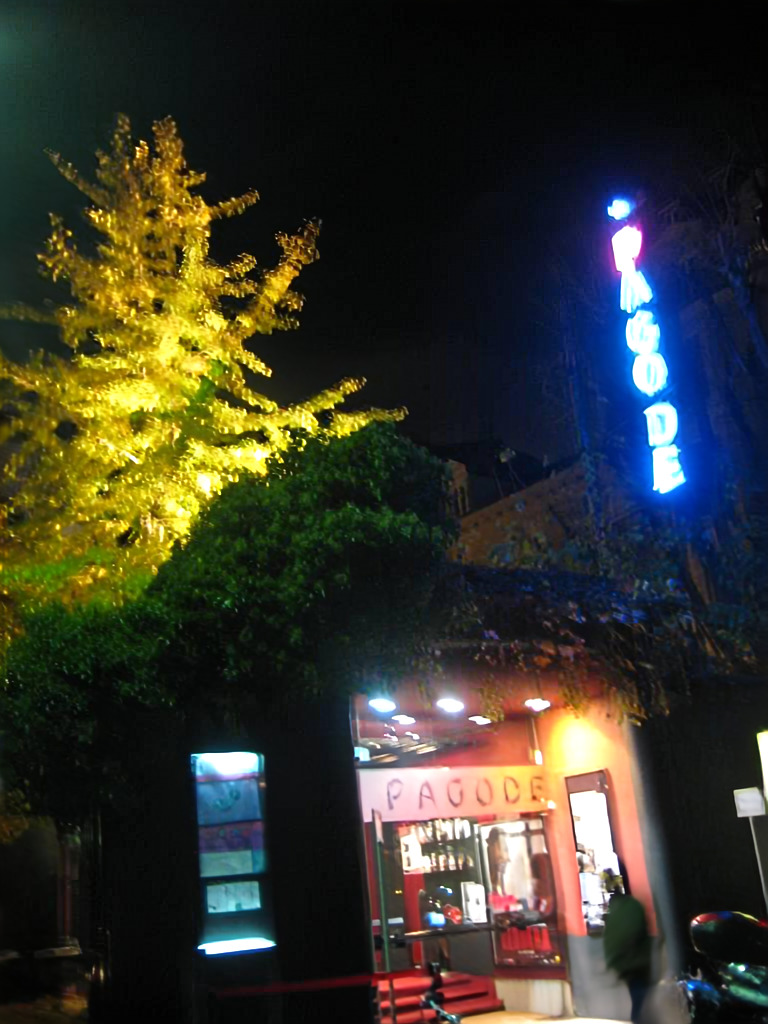}  \\

\includegraphics[trim=360 70 300 891, clip,width=0.12\textwidth]{imgs/Lai/DeblurGAN_V2/pagode.jpg}
\includegraphics[trim=340 180 260 745, clip,width=0.12\textwidth]{imgs/Lai/DeblurGAN_V2/pagode.jpg} &
\includegraphics[trim=360 70 300 891, clip,width=0.12\textwidth]{imgs/Lai/MPRNet/pagode.jpg} 
\includegraphics[trim=340 180 260 745, clip,width=0.12\textwidth]{imgs/Lai/MPRNet/pagode.jpg} &
\includegraphics[trim=360 70 300 891, clip,width=0.12\textwidth]{imgs/Lai/NAFNet/pagode.jpg}
\includegraphics[trim=340 180 260 745, clip,width=0.12\textwidth]{imgs/Lai/NAFNet/pagode.jpg} &
\includegraphics[trim=360 70 300 891, clip,width=0.12\textwidth]{imgs/Lai/COCO900_restL2_aug_all_loss_gf1_80k/pagode_NIMBUSR.jpg} 
\includegraphics[trim=340 180 260 745, clip,width=0.12\textwidth]{imgs/Lai/COCO900_restL2_aug_all_loss_gf1_80k/pagode_NIMBUSR.jpg} \\

\end{tabular} 
    \caption{\textbf{Deblurring examples on real blurry images  from Lai dataset  \cite{lai2016comparative}.} Our model-based approach is competitive with state-of-the-art data-driven methods. Best viewed in electronic format.}
    \label{fig:Lai2_suppmat}
\end{figure*}

\begin{figure*}[h]

    \setlength\tabcolsep{1pt} %
    \centering
    \scriptsize
\begin{tabular}{cccc}

Blurry & Ana-Syn \cite{kaufman2020deblurring} & SRN \cite{tao2018scale} & MIMO-UNet  \cite{cho2021rethinking}  \\ 
\includegraphics[trim=0 0 0 50, clip, width=0.240\textwidth]{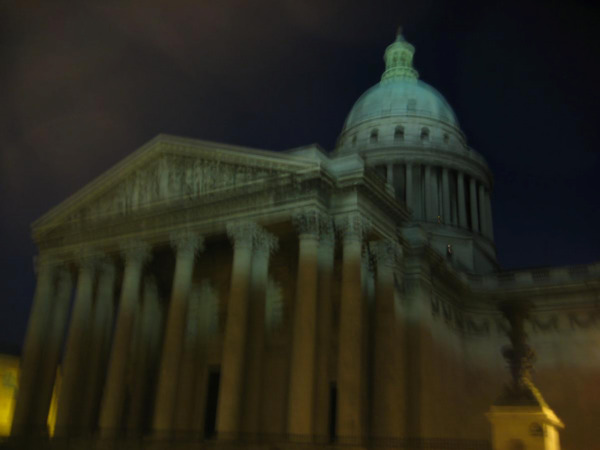}   &
\includegraphics[trim=0 0 0 50, clip, width=0.240\textwidth]{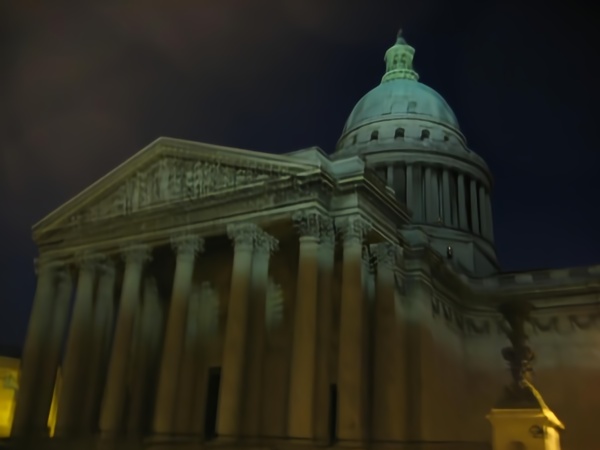} & 
\includegraphics[trim=0 0 0 50, clip, width=0.240\textwidth]{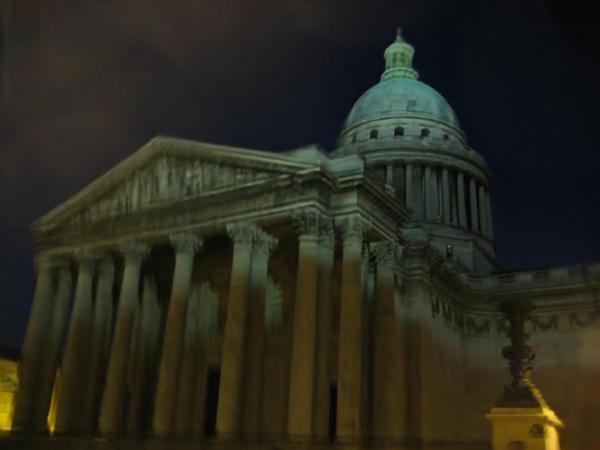}  &    
\includegraphics[trim=0 0 0 50, clip, width=0.240\textwidth]{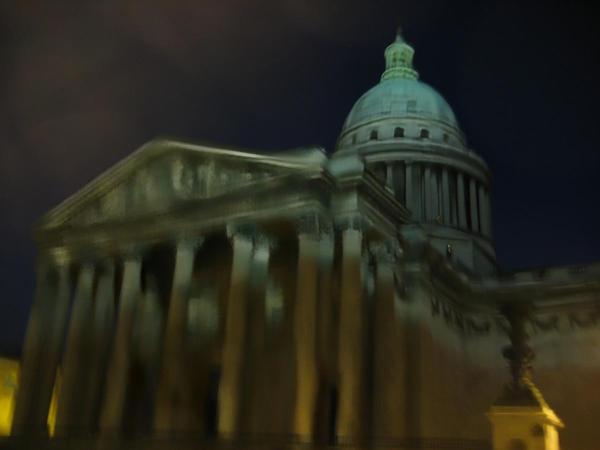} \\

\includegraphics[trim=320 280 120 100, clip,width=0.12\textwidth]{imgs/Lai_HS/Blurry/Pantheon.jpg}
\includegraphics[trim=440 5 0 375, clip,width=0.12\textwidth]{imgs/Lai_HS/Blurry/Pantheon.jpg} &
\includegraphics[trim=320 280 120 100, clip,width=0.12\textwidth]{imgs/Lai_HS/AnaSyn/Pantheon_deblurred.jpg} 
\includegraphics[trim=440 5 0 375, clip,width=0.12\textwidth]{imgs/Lai_HS/AnaSyn/Pantheon_deblurred.jpg} &
\includegraphics[trim=320 280 120 100, clip,width=0.12\textwidth]{imgs/Lai_HS/SRN/Pantheon.jpg}
\includegraphics[trim=440 5 0 375, clip,width=0.12\textwidth]{imgs/Lai_HS/SRN/Pantheon.jpg} &
\includegraphics[trim=320 280 120 100, clip,width=0.12\textwidth]{imgs/Lai_HS/MIMO-UNet/Pantheon.jpg} 
\includegraphics[trim=440 5 0 375, clip,width=0.12\textwidth]{imgs/Lai_HS/MIMO-UNet/Pantheon.jpg} \\

DGAN-v2 \cite{kupyn2019deblurgan} & MPRNet 
\cite{Zamir2021MPRNet} & NAFNet \cite{NAFNet}    &   J-MKPD (ours) \\
\includegraphics[trim=0 0 0 50, clip, width=0.240\textwidth]{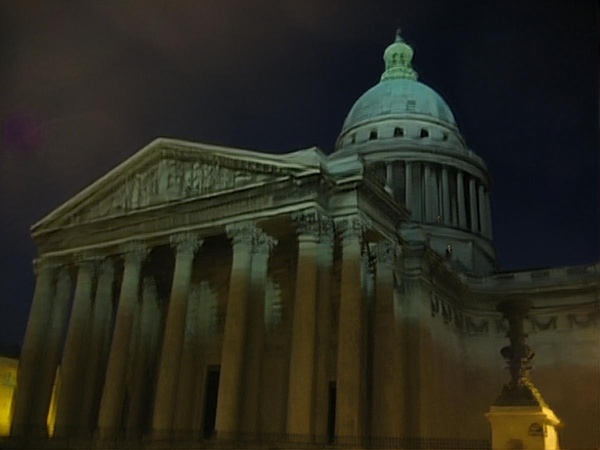}  &
\includegraphics[trim=0 0 0 50, clip, width=0.240\textwidth]{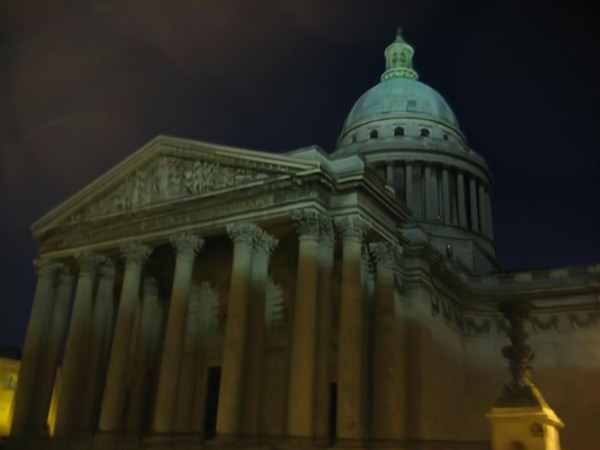}  & 
\includegraphics[trim=0 0 0 50, clip, width=0.240\textwidth]{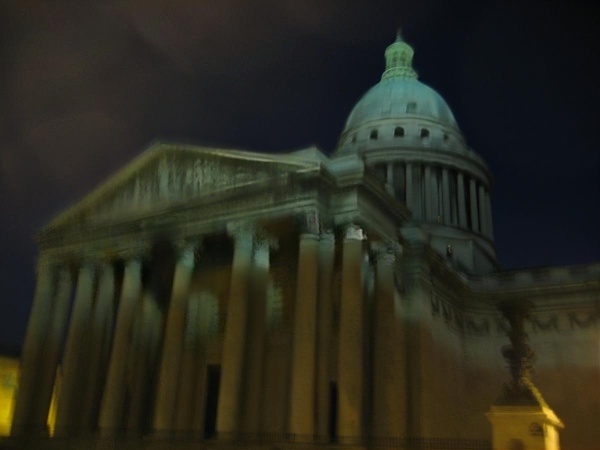} &
\includegraphics[trim=0 0 0 50, clip, width=0.240\textwidth]{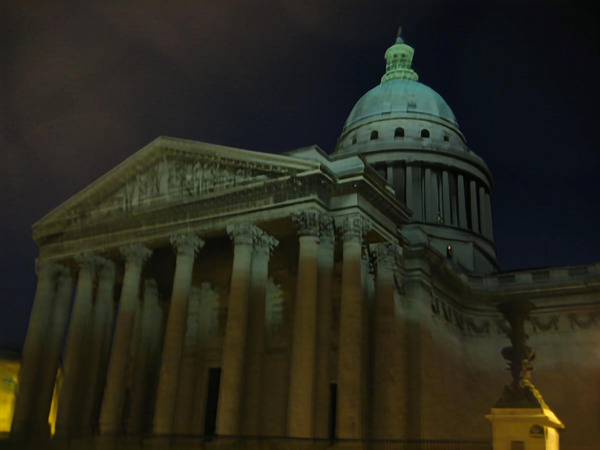}  \\

\includegraphics[trim=320 280 120 100, clip,width=0.12\textwidth]{imgs/Lai_HS/DeblurGAN-v2/Pantheon.jpg}
\includegraphics[trim=440 5 0 375, clip,width=0.12\textwidth]{imgs/Lai_HS/DeblurGAN-v2/Pantheon.jpg} &
\includegraphics[trim=320 280 120 100, clip,width=0.12\textwidth]{imgs/Lai_HS/MPRNet/Pantheon.jpg} 
\includegraphics[trim=440 5 0 375, clip,width=0.12\textwidth]{imgs/Lai_HS/MPRNet/Pantheon.jpg} &
\includegraphics[trim=320 280 120 100, clip,width=0.12\textwidth]{imgs/Lai_HS/NAFNet/Pantheon.jpg}
\includegraphics[trim=440 5 0 375, clip,width=0.12\textwidth]{imgs/Lai_HS/NAFNet/Pantheon.jpg} &
\includegraphics[trim=320 280 120 100, clip,width=0.12\textwidth]{imgs/Lai_HS/COCO900_restL2_aug_all_loss_gf1_80k/Pantheon_NIMBUSR.jpg} 
\includegraphics[trim=440 5 0 375, clip,width=0.12\textwidth]{imgs/Lai_HS/COCO900_restL2_aug_all_loss_gf1_80k/Pantheon_NIMBUSR.jpg} \\

Blurry & Ana-Syn \cite{kaufman2020deblurring} & SRN \cite{tao2018scale} & MIMO-UNet  \cite{cho2021rethinking}  \\ 
\includegraphics[trim=0 0 0 0, clip, width=0.240\textwidth]{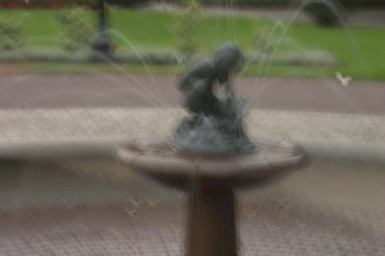}   &
\includegraphics[trim=0 0 0 0, clip, width=0.240\textwidth]{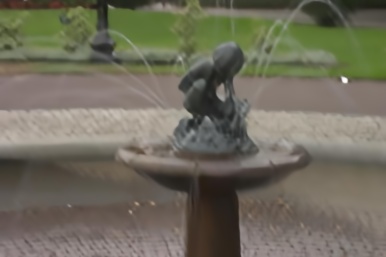} & 
\includegraphics[trim=0 0 0 0, clip, width=0.240\textwidth]{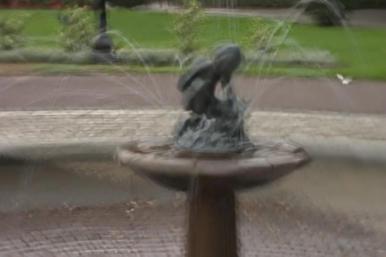}  &    
\includegraphics[trim=0 0 0 0, clip, width=0.240\textwidth]{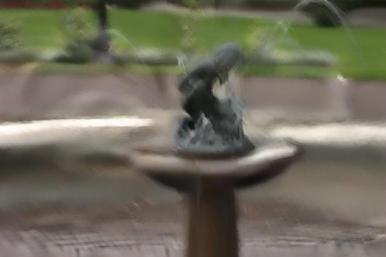} \\

\includegraphics[trim=120 30 191 160, clip,width=0.12\textwidth]{imgs/Lai_HS/Blurry/fountain1.jpg}
\includegraphics[trim=210 160 80 10, clip,width=0.12\textwidth]{imgs/Lai_HS/Blurry/fountain1.jpg} &
\includegraphics[trim=120 30 191 160, clip,width=0.12\textwidth]{imgs/Lai_HS/AnaSyn/fountain1_deblurred.jpg} 
\includegraphics[trim=210 160 80 10, clip,width=0.12\textwidth]{imgs/Lai_HS/AnaSyn/fountain1_deblurred.jpg} &
\includegraphics[trim=120 30 191 160, clip,width=0.12\textwidth]{imgs/Lai_HS/SRN/fountain1.jpg}
\includegraphics[trim=210 160 80 10, clip,width=0.12\textwidth]{imgs/Lai_HS/SRN/fountain1.jpg} &
\includegraphics[trim=120 30 191 160, clip,width=0.12\textwidth]{imgs/Lai_HS/MIMO-UNet/fountain1.jpg} 
\includegraphics[trim=210 160 80 10, clip,width=0.12\textwidth]{imgs/Lai_HS/MIMO-UNet/fountain1.jpg} \\

DGAN-v2 \cite{kupyn2019deblurgan} & MPRNet 
\cite{Zamir2021MPRNet} & NAFNet \cite{NAFNet}    &   J-MKPD (ours) \\
\includegraphics[trim=0 0 0 0, clip, width=0.240\textwidth]{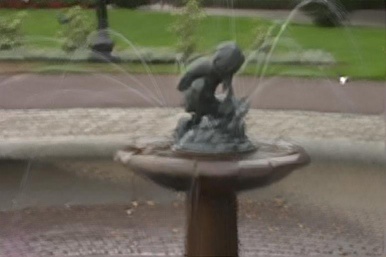}  &
\includegraphics[trim=0 0 0 0, clip, width=0.240\textwidth]{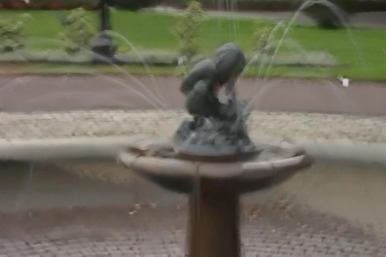}  & 
\includegraphics[trim=0 0 0 0, clip, width=0.240\textwidth]{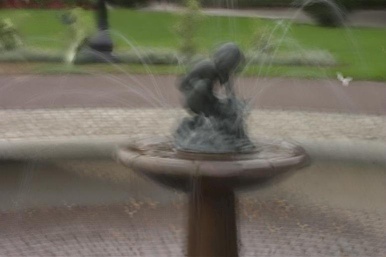} &
\includegraphics[trim=0 0 0 0, clip, width=0.240\textwidth]{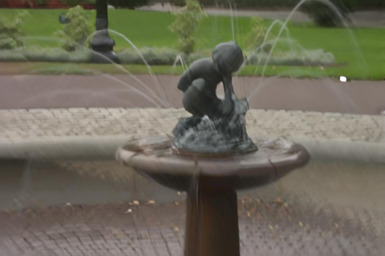}  \\

\includegraphics[trim=120 30 191 160, clip,width=0.12\textwidth]{imgs/Lai_HS/DeblurGAN-v2/fountain1.jpg}
\includegraphics[trim=210 160 80 10, clip,width=0.12\textwidth]{imgs/Lai_HS/DeblurGAN-v2/fountain1.jpg} &
\includegraphics[trim=120 30 191 160, clip,width=0.12\textwidth]{imgs/Lai_HS/MPRNet/fountain1.jpg} 
\includegraphics[trim=210 160 80 10, clip,width=0.12\textwidth]{imgs/Lai_HS/MPRNet/fountain1.jpg} &
\includegraphics[trim=120 30 191 160, clip,width=0.12\textwidth]{imgs/Lai_HS/NAFNet/fountain1.jpg}
\includegraphics[trim=210 160 80 10, clip,width=0.12\textwidth]{imgs/Lai_HS/NAFNet/fountain1.jpg} &
\includegraphics[trim=120 30 191 160, clip,width=0.12\textwidth]{imgs/Lai_HS/COCO900_restL2_aug_all_loss_gf1_80k/fountain1_NIMBUSR.jpg} 
\includegraphics[trim=210 160 80 10, clip,width=0.12\textwidth]{imgs/Lai_HS/COCO900_restL2_aug_all_loss_gf1_80k/fountain1_NIMBUSR.jpg} \\

\end{tabular} 
    \caption{\textbf{Deblurring examples on real blurry images (half resolution) from Lai dataset  \cite{lai2016comparative}.} Our model-based approach is competitive with state-of-the-art data-driven methods. Best viewed in electronic format.}
    \label{fig:LaiHS_suppmat}
\end{figure*}

\begin{figure*}[h]

    \setlength\tabcolsep{1pt} %
    \centering
    \scriptsize
\begin{tabular}{cccc}

Blurry & Ana-Syn \cite{kaufman2020deblurring} & SRN \cite{tao2018scale} & MIMO-UNet  \cite{cho2021rethinking}  \\ 
\includegraphics[trim=0 0 0 100, clip, width=0.240\textwidth]{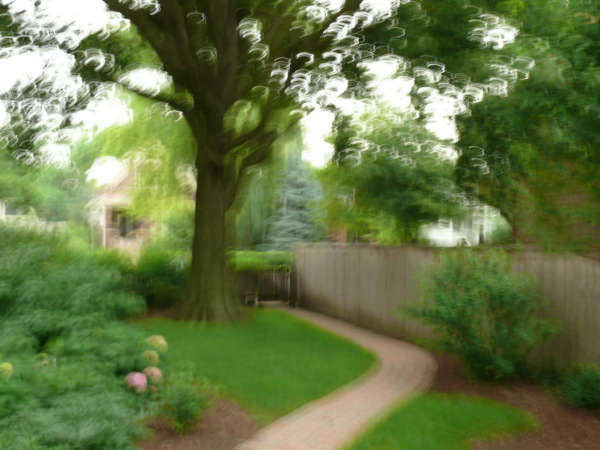}   &
\includegraphics[trim=0 0 0 100, clip, width=0.240\textwidth]{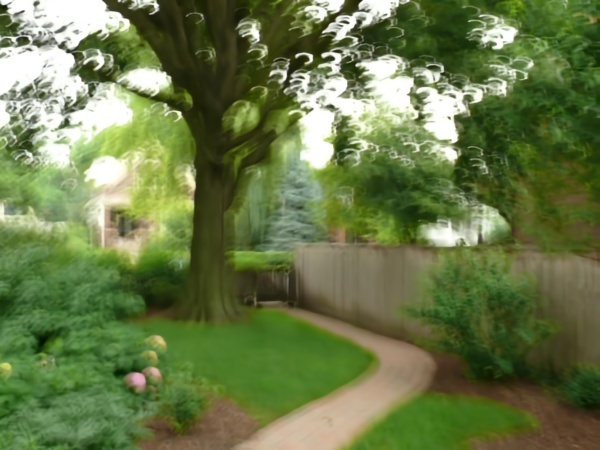} & 
\includegraphics[trim=0 0 0 100, clip, width=0.240\textwidth]{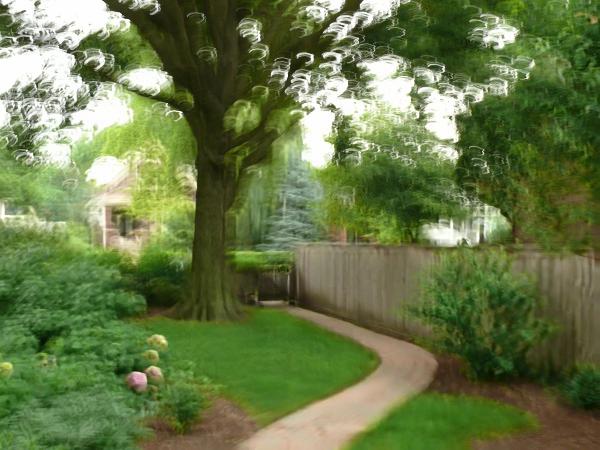}  &    
\includegraphics[trim=0 0 0 100, clip, width=0.240\textwidth]{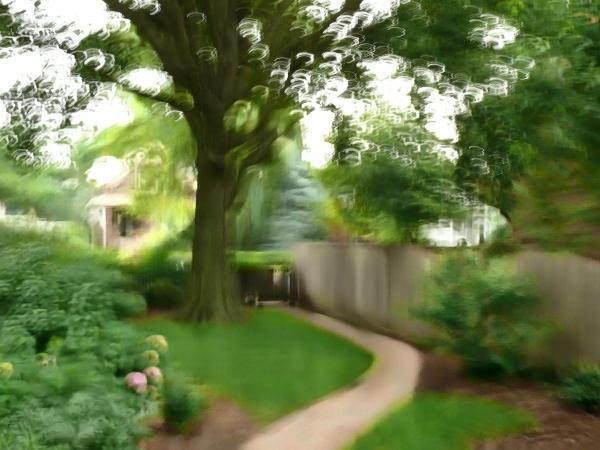} \\

\includegraphics[trim=390 40 120 350, clip,width=0.12\textwidth]{imgs/Lai_HS/Blurry/garden.jpg}
\includegraphics[trim=180 150 310 220, clip,width=0.12\textwidth]{imgs/Lai_HS/Blurry/garden.jpg} &
\includegraphics[trim=390 40 120 350, clip,width=0.12\textwidth]{imgs/Lai_HS/AnaSyn/garden_deblurred.jpg} 
\includegraphics[trim=180 150 310 220, clip,width=0.12\textwidth]{imgs/Lai_HS/AnaSyn/garden_deblurred.jpg} &
\includegraphics[trim=390 40 120 350, clip,width=0.12\textwidth]{imgs/Lai_HS/SRN/garden.jpg}
\includegraphics[trim=180 150 310 220, clip,width=0.12\textwidth]{imgs/Lai_HS/SRN/garden.jpg} &
\includegraphics[trim=390 40 120 350, clip,width=0.12\textwidth]{imgs/Lai_HS/MIMO-UNet/garden.jpg} 
\includegraphics[trim=180 150 310 220, clip,width=0.12\textwidth]{imgs/Lai_HS/MIMO-UNet/garden.jpg} \\

DGAN-v2 \cite{kupyn2019deblurgan} & MPRNet 
\cite{Zamir2021MPRNet} & NAFNet \cite{NAFNet}    &   J-MKPD (ours) \\
\includegraphics[trim=0 0 0 100, clip, width=0.240\textwidth]{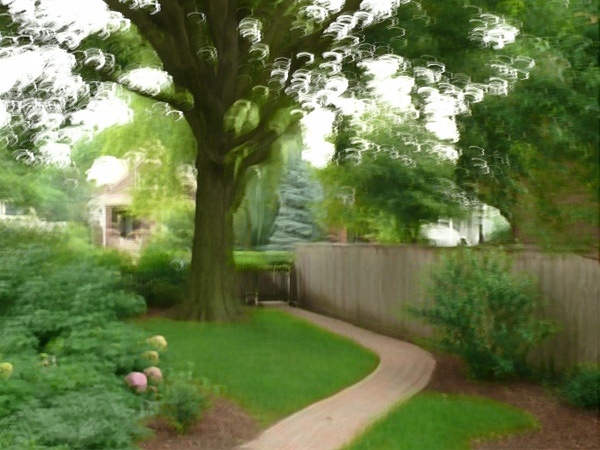}  &
\includegraphics[trim=0 0 0 100, clip, width=0.240\textwidth]{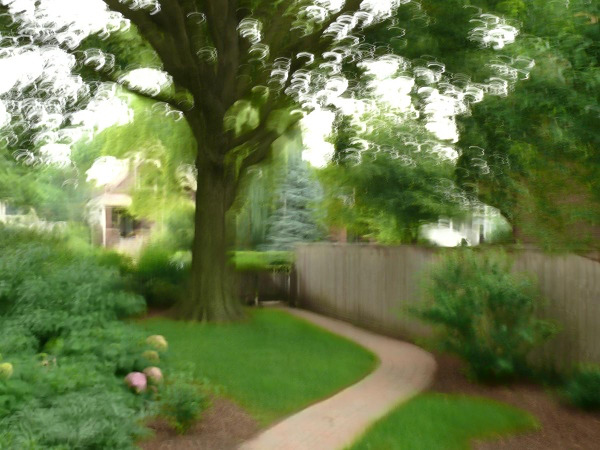}  & 
\includegraphics[trim=0 0 0 100, clip, width=0.240\textwidth]{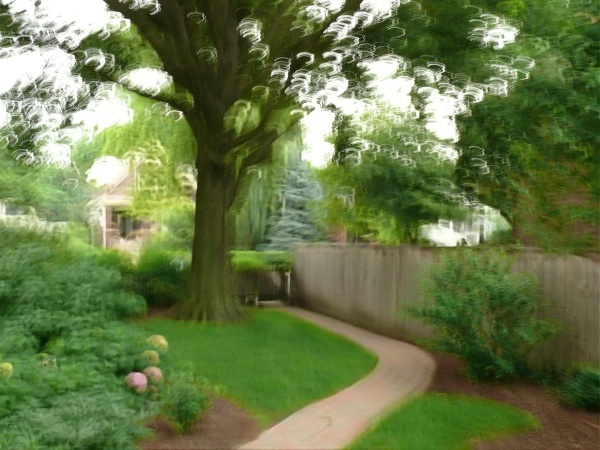} &
\includegraphics[trim=0 0 0 100, clip, width=0.240\textwidth]{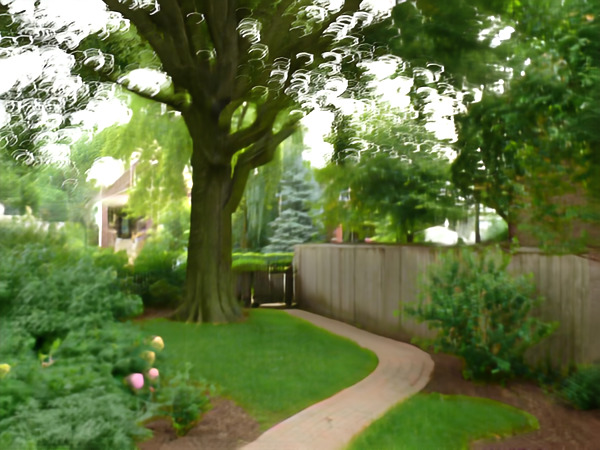}  \\

\includegraphics[trim=390 40 100 330, clip,width=0.12\textwidth]{imgs/Lai_HS/DeblurGAN-v2/garden.jpg}
\includegraphics[trim=180 150 310 220, clip,width=0.12\textwidth]{imgs/Lai_HS/DeblurGAN-v2/garden.jpg} &
\includegraphics[trim=390 40 100 330, clip,width=0.12\textwidth]{imgs/Lai_HS/MPRNet/garden.jpg} 
\includegraphics[trim=180 150 310 220, clip,width=0.12\textwidth]{imgs/Lai_HS/MPRNet/garden.jpg} &
\includegraphics[trim=390 40 100 330, clip,width=0.12\textwidth]{imgs/Lai_HS/NAFNet/garden.jpg}
\includegraphics[trim=180 150 310 220, clip,width=0.12\textwidth]{imgs/Lai_HS/NAFNet/garden.jpg} &
\includegraphics[trim=390 40 100 330, clip,width=0.12\textwidth]{imgs/Lai_HS/COCO900_restL2_aug_all_loss_gf1_80k/garden_NIMBUSR.jpg} 
\includegraphics[trim=180 150 310 220, clip,width=0.12\textwidth]{imgs/Lai_HS/COCO900_restL2_aug_all_loss_gf1_80k/garden_NIMBUSR.jpg} \\

Blurry & Ana-Syn \cite{kaufman2020deblurring} & SRN \cite{tao2018scale} & MIMO-UNet  \cite{cho2021rethinking}  \\ 
\includegraphics[trim=0 0 0 50, clip, width=0.240\textwidth]{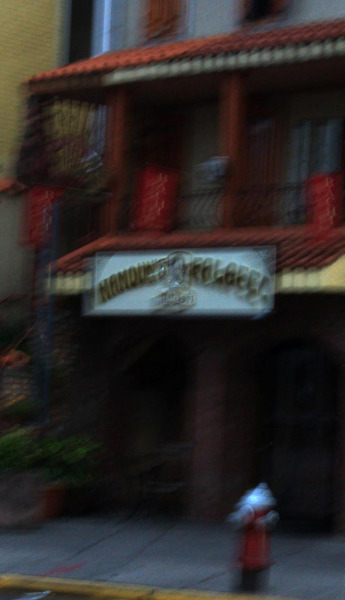}   &
\includegraphics[trim=0 0 0 50, clip, width=0.240\textwidth]{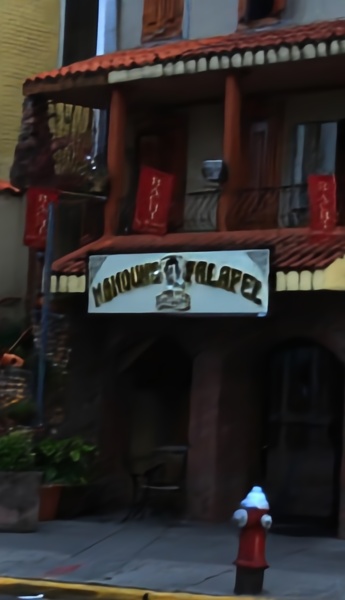} & 
\includegraphics[trim=0 0 0 50, clip, width=0.240\textwidth]{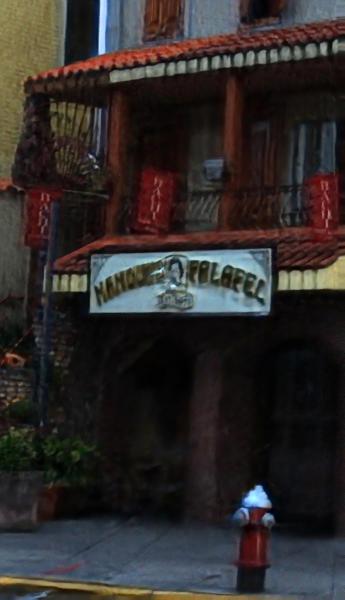}  &    
\includegraphics[trim=0 0 0 50, clip, width=0.240\textwidth]{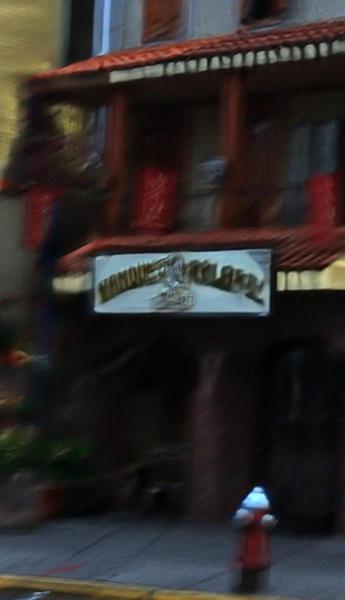} \\

DGAN-v2 \cite{kupyn2019deblurgan} & MPRNet 
\cite{Zamir2021MPRNet} & NAFNet \cite{NAFNet}    &   J-MKPD (ours) \\
\includegraphics[trim=0 0 0 50, clip, width=0.240\textwidth]{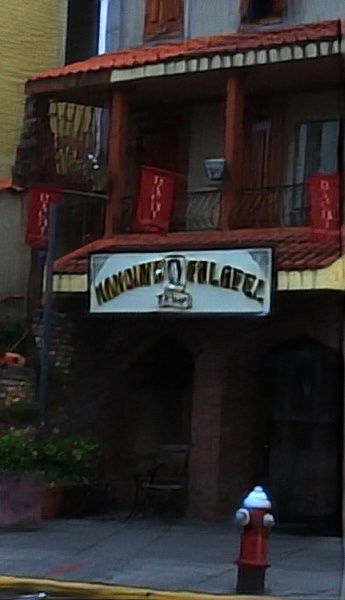}  &
\includegraphics[trim=0 0 0 50, clip, width=0.240\textwidth]{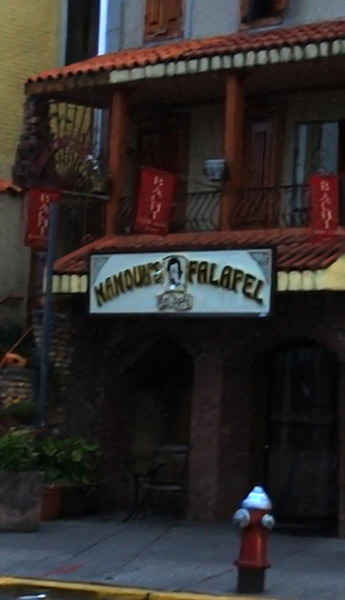}  & 
\includegraphics[trim=0 0 0 50, clip, width=0.240\textwidth]{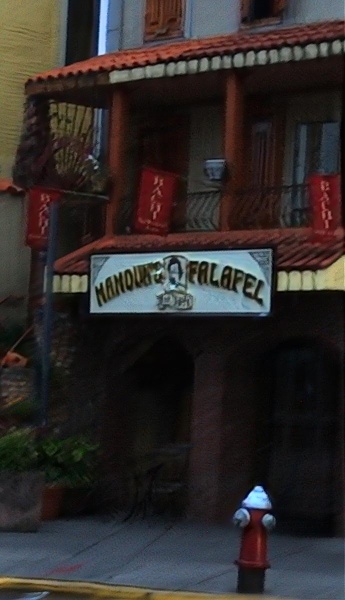} &
\includegraphics[trim=0 0 0 50, clip, width=0.240\textwidth]{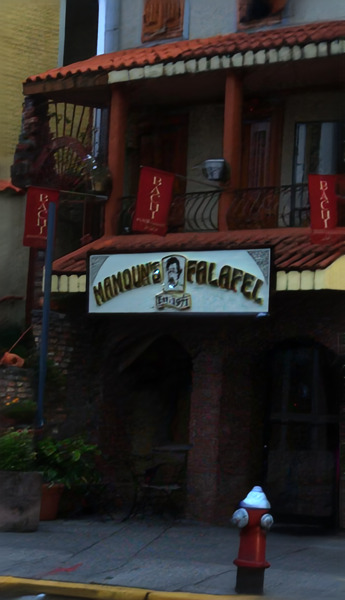}  \\

\end{tabular} 
    \caption{\textbf{Deblurring examples on real blurry images (half resolution) from Lai dataset  \cite{lai2016comparative}.} Our model-based approach is competitive with state-of-the-art data-driven methods. Best viewed in electronic format.}
    \label{fig:LaiHS2_suppmat}
\end{figure*}